%% file: main.tex
\documentclass[11pt]{article}

\usepackage{amsmath,amssymb,amsthm,mathtools}
\usepackage{bbm}
\usepackage{graphicx}
\usepackage{booktabs}
\usepackage{array}
\usepackage{algorithm}
\usepackage{algpseudocode}
\usepackage{xurl}
\usepackage{arxiv}
\usepackage{etoc}
\usepackage{bookmark}
\hypersetup{linktoc=all,bookmarksdepth=2}
\usepackage{subcaption}
\usepackage{float}
\usepackage{placeins}
\usepackage{enumitem}
\usepackage{multirow}

\newcommand{\dd}{\mathrm{d}}

\DeclareMathOperator*{\argmin}{arg\,min}

\theoremstyle{plain}
\newtheorem{theorem}{Theorem}[section]
\newtheorem{corollary}[theorem]{Corollary}
\newtheorem{proposition}[theorem]{Proposition}
\newtheorem{lemma}[theorem]{Lemma}

\theoremstyle{definition}

\theoremstyle{remark}
\newtheorem{remark}[theorem]{Remark}

\makeatletter
\AddToHook{env/theorem/after}{\@afterindentfalse\@afterheading}
\AddToHook{env/corollary/after}{\@afterindentfalse\@afterheading}
\AddToHook{env/proposition/after}{\@afterindentfalse\@afterheading}
\AddToHook{env/lemma/after}{\@afterindentfalse\@afterheading}
\AddToHook{env/definition/after}{\@afterindentfalse\@afterheading}
\AddToHook{env/remark/after}{\@afterindentfalse\@afterheading}
\AddToHook{env/proof/after}{\@afterindentfalse\@afterheading}
\AddToHook{env/enumerate/after}{\@afterindentfalse\@afterheading}
\AddToHook{env/itemize/after}{\@afterindentfalse\@afterheading}
\makeatother

\title{Model Selection for Off-Policy Evaluation
\\ via Bellman Tournaments}
\author{
  Lars van der Laan \\
 Stanford University \\
   \texttt{vdlaan@stanford.edu}
    \And
  Vasilis Syrgkanis \\
 Stanford University \\
  \And
  Nathan Kallus \\
  Netflix and Cornell University
}

\begin{document}
\etocdepthtag.toc{main}

\maketitle

\begin{abstract}
Model selection and hyperparameter tuning pose a central challenge in
off-policy evaluation (OPE) because estimation errors in fitted
$Q$-functions and occupancy ratios cannot be directly evaluated using
held-out transition data. Existing Bellman-residual-based validation methods often require
auxiliary regressions, while many minimax-based methods require
optimization over separate critic classes.
We propose \emph{pairwise Bellman validation (PBV)}, a family of
penalized Bellman tournament methods that reduces validation to
computing held-out sample averages.
For each candidate, PBV constructs critics from pairwise differences
with the other candidates, using differences on the log scale for occupancy ratios.
It then selects the candidate that minimizes the worst-case penalized
score across these pairwise critics. The key insight motivating these pairwise critics is that
$Q$-function errors and log ratios of candidate to true occupancy
ratios serve as oracle critics whose moments against the Bellman
and adjoint Bellman residuals control estimation error.
The tournament criterion avoids the need to identify which pairwise critic
best approximates the oracle critic, while the critic penalties
account for approximation error under candidate-class misspecification. We establish finite-sample oracle inequalities relative to the best
available candidate, even under candidate-class misspecification.
For $Q$-function selection, we first propose unweighted PBV, which
avoids importance weighting but requires a coercivity condition on a
quadratic form induced by the Bellman operator.
We then develop importance-weighted PBV (IW-PBV), which uses marginalized
importance weighting to exploit contraction of the Bellman operator
under the target-policy occupancy measure, thereby avoiding this
condition.
For occupancy-ratio selection, we propose adjoint PBV (A-PBV),
which exploits KL contraction of the adjoint Bellman operator to
obtain oracle guarantees directly under the offline data law.
On Hopper and DeepOPE, unweighted PBV is competitive with
LSTD-Tournament, often achieving higher accuracy on Hopper,
while importance weighting can further improve performance
under controlled overlap.
\end{abstract}


\section{Introduction}
\subsection{Motivation}
The goal of off-policy evaluation (OPE) is to estimate the value of a
target policy using data collected under a different policy in a Markov
decision process (MDP). In the infinite-horizon setting, two central objects in OPE are the target-policy
action-value function, or $Q$-function, and the discounted occupancy ratio
\citep{ueharaShiKallus2022OPEReview}. The former gives the expected
discounted cumulative reward from each state--action pair under the
target policy, while the latter reweights the logged state--action
distribution to the target policy's discounted state--action distribution. These functions underlie direct, importance-weighted, and
doubly robust OPE, as well as semiparametrically efficient inference
\citep{jiangLi2016DoublyRobust,xieEtAl2019MarginalizedIS,
kallusUehara2020DoubleRL,kallusUehara2022BreakingHorizonDRL,
vanDerLaanEtAl2025AutomaticDRL}.
They are characterized, respectively, as fixed points of the Bellman
and adjoint Bellman operators
\citep{ueharaEtAl2021FiniteSampleMinimax}.
These fixed-point equations motivate fitted $Q$-evaluation (FQE) for
learning $Q$-functions and fitted occupancy-ratio evaluation (FORE)
for learning occupancy ratios
\citep{munosSzepesvari2008FVI,van2026fitted}.
They also motivate minimax methods, such as DualDICE and minimax
$Q$-learning, which test Bellman or adjoint Bellman residuals against
critic functions, that is, test functions used to detect violations
of the corresponding equations
\citep{nachumEtAl2019DualDICE,ueharaEtAl2020MWLMQL,
ueharaEtAl2021FiniteSampleMinimax}.

A central challenge in OPE is model selection and hyperparameter tuning.
Different choices of algorithms, function classes, regularization, and hyperparameters
produce competing $Q$-function and occupancy-ratio estimates, but their
estimation errors cannot be directly evaluated on held-out logged data.
For $Q$-functions, squared temporal-difference error is generally unsuitable
for comparing candidates under stochastic transitions because it includes a
candidate-dependent conditional-variance term
\citep{baird1995residual}.
Bellman-residual validation instead estimates
the Bellman residual $Q-\mathcal B_\pi Q$ for each candidate, where
$\mathcal B_\pi$ denotes the target-policy Bellman operator, and
selects the candidate with the smallest estimated mean-squared residual.
However, estimating these residuals typically requires fitting a separate
regression for each candidate \citep{zitovsky2023revisiting},
which can be computationally prohibitive and introduces an additional
tuning problem.
An alternative is minimax Bellman-moment validation, which evaluates
empirical moments of one-step temporal-difference residuals against
a class of critic functions and selects the candidate that minimizes
the largest absolute moment over that class
\citep{ueharaEtAl2021FiniteSampleMinimax,miyaguchi2026duality}.
This avoids separate Bellman-residual regressions but requires
specifying and potentially tuning a critic class, and can involve
computationally demanding and numerically delicate optimization,
particularly with flexible critics.

For $Q$-function selection, LSTD-Tournament \citep{liu2025modelselection} offers a simpler
alternative that uses least-squares temporal-difference Q-learning
(LSTDQ) to compare candidate $Q$-functions pairwise.
Its selection criterion can be viewed as a minimax Bellman-moment
validation procedure that uses the candidate $Q$-functions
themselves as critics, avoiding the need to fit or tune a
separate critic class. However, its theoretical guarantees depend
on the conditioning of the pairwise LSTD matrices, and the
resulting bounds can deteriorate when candidates are nearly collinear,
as may occur when comparing similar hyperparameter settings.  

\subsection{Contributions}

We introduce pairwise Bellman validation (PBV), a family of penalized
Bellman tournament methods for selecting among finite classes of
fitted $Q$-functions and occupancy-ratio estimators using held-out
transitions.
We develop a new validation criterion based on critics constructed
from pairwise differences between candidates, show how importance
weighting relaxes the conditions needed for $Q$-function selection,
and extend the approach to occupancy-ratio selection.
We establish finite-sample oracle inequalities relative to the best
available candidate, even under candidate-class misspecification
and near-collinearity among candidates.

Operationally, PBV evaluates each candidate using empirical moments
of its Bellman or adjoint Bellman residual against critics constructed
from pairwise differences with the other candidates.
For occupancy ratios, these differences are taken on the log scale.
For $Q$-function selection, each moment is the held-out average of a
one-step temporal-difference residual multiplied by a critic.
PBV then uses a penalized tournament, subtracting a critic penalty
from each absolute empirical moment and selecting the candidate
with the smallest worst-case penalized score across all pairwise critics.
Validation thus reduces to computing a finite collection of
sample averages, without fitting auxiliary validation models or
separately optimizing critic functions.

The motivation for these pairwise critics is that the estimation errors
$Q_i-Q^\pi$ and $\log(w_i/w^\pi)$ serve as oracle
critics for the Bellman and adjoint Bellman residuals, respectively.
Under suitable conditions, we show that evaluating the Bellman residual of $Q_i$
against $Q_i-Q^\pi$ controls its squared $L^2$ error, while evaluating
the adjoint Bellman residual of $w_i$ against $\log(w_i/w^\pi)$
controls its generalized-KL error.
The pairwise critics $Q_i-Q_j$ and $\log(w_i/w_j)$ coincide with
their oracle counterparts when the comparator equals the truth,
and approximate them when the comparator is close to the truth.
By taking the worst case over these pairwise critics, the tournament
avoids the need to identify which comparator best approximates the truth,
while the critic penalties account for approximation error under
candidate-class misspecification.

For $Q$-function selection, we first develop unweighted PBV, which
evaluates pairwise Bellman moments under the offline distribution
using a quadratic critic penalty.
Because the target-policy Bellman operator need not
be contractive in the corresponding $L^2$ norm, our guarantees require
a coercivity condition (Condition~\ref{cond:q-pairwise-coercivity})
on a quadratic form induced by the Bellman operator.
This condition plays a role similar to that of the matrix-conditioning
assumptions in LSTD-Tournament and linear TD learning
\citep{bradtkeBarto1996Linear,tsitsiklisVanRoy1997Analysis,
lagoudakisParr2003LSPI,amortila2026unifying}.
Unlike those assumptions, however, coercivity can hold even when
candidate functions are nearly collinear.
We then develop importance-weighted PBV (IW-PBV), which uses marginalized
importance weighting to evaluate the corresponding Bellman moments
under the discounted target-occupancy distribution, where the Bellman
operator is contractive.
Thus, unweighted PBV avoids occupancy-ratio estimation but requires
coercivity, whereas IW-PBV removes this requirement at the cost
of estimating occupancy ratios.

For occupancy-ratio selection, we develop adjoint PBV (A-PBV),
which evaluates adjoint Bellman residuals against log-ratio critics
with a penalty proportional to the squared Hellinger distance
between candidate ratios.
A-PBV requires no importance weighting,
coercivity, or matrix-conditioning assumptions, since the adjoint
Bellman operator is contractive in KL divergence under the
offline distribution \citep{van2026fitted}.

Finally, as contributions of independent interest, we extend
the analysis of LSTD-Tournament to establish finite-sample
oracle inequalities under candidate-class misspecification
(Appendix~\ref{app:lstd-misspecification}) and extend
the method to occupancy-ratio selection
(Appendix~\ref{app:adjoint-lstd-tournament}).
Both extensions require matrix-conditioning assumptions
that IW-PBV and A-PBV, respectively, avoid.

\subsection{Related work}
\label{sec:related-work-main}

\noindent\textbf{Model selection for $Q$-functions and occupancy ratios.}
Existing OPE model-selection methods have focused primarily on $Q$-functions,
using Bellman residuals, candidate-dependent projections, nested classes,
or auxiliary regressions and critics to replace the unavailable supervised loss
\citep{farahmand2011model,lee2022oracle,
zitovsky2023revisiting,miyaguchi2026duality}.
An early influential Bellman tournament method is BVFT
\citep{xie2021batch}, which uses pairwise candidate-dependent
state--action partitions and projected Bellman errors to select the $Q$-function.
BVFT-PE \citep{zhang2021towards} extends this approach to policy evaluation.
These approaches generally introduce additional estimation or structural
requirements beyond the candidate class itself.
PBV instead uses pairwise candidate differences directly as Bellman critics,
avoiding state aggregation and allowing misspecified candidate classes.
A more closely related approach is LSTD-Tournament \citep{liu2025modelselection}, which also constructs
pairwise validation criteria directly from the candidate class.
We discuss this connection separately below.

By contrast, model selection among already-fitted occupancy-ratio
estimators has received substantially less attention, despite
extensive work on occupancy-ratio estimation
\citep{nachumEtAl2019DualDICE,ueharaEtAl2020MWLMQL,
ueharaEtAl2021FiniteSampleMinimax,van2026fitted}.
Our proposed A-PBV extends Bellman tournament methods to occupancy-ratio
selection without coercivity or matrix-conditioning assumptions.
For comparison, we also extend LSTD-Tournament to
occupancy-ratio selection in Appendix~\ref{app:adjoint-lstd-tournament},
but its guarantees require matrix-conditioning assumptions that A-PBV avoids.
We describe the main validation methods in Section~\ref{sec:q-selection},
with further discussion in Appendix~\ref{app:related-work}.

\medskip
\noindent\textbf{Relation to LSTD-Tournament.}
Among our methods, unweighted PBV is most closely related to
LSTD-Tournament \citep{liu2025modelselection}.
Although derived from pairwise LSTDQ, its selection criterion
can be expressed as minimax Bellman-moment validation using
the candidate $Q$-functions themselves as critics.
In contrast, PBV uses pairwise candidate differences as critics
and introduces a quadratic penalty to accommodate misspecification. The theoretical guarantees for LSTD-Tournament require invertible
pairwise LSTD matrices, with bounds that deteriorate as these
matrices approach singularity.

In their numerical experiments, the authors instead use a candidate
function and a pairwise difference as features, each normalized by
its empirical standard deviation.
Although $(Q_i,Q_j)$ and $(Q_i,Q_i-Q_j)$ span the same space,
the latter representation can yield better-conditioned LSTD matrices
after normalization.
However, the authors do not provide finite-sample theoretical guarantees
for this variant.
We analyze the normalized selector in
Appendix~\ref{app:lstd-normalized} and show that it can perform poorly
when empirical critic standard deviations underestimate their population
counterparts, as may occur when critics vary primarily on rare states
missed by the validation sample
(Proposition~\ref{prop:lstd-normalized-empirical-failure}).
This sensitivity is reflected in our finite-sample bound through the
``spikiness'' factor
$\|f-\mathbb E_\nu f\|_\infty^2/
\|f-\mathbb E_\nu f\|_{2,\nu}^2$,
which must be uniformly bounded over the candidate functions and
their pairwise differences; see Appendix~\ref{app:discussionLSTD}
for additional discussion.

Unlike the empirical LSTD-Tournament criterion, unweighted PBV
uses unnormalized pairwise differences as critics, together with
quadratic penalties rather than empirical normalization.
Our choice of pairwise differences is motivated by their role as
approximations of oracle critics, rather than by numerical stability
or matrix-conditioning considerations.
Our guarantees do not require pairwise LSTD matrices to be invertible.
Instead, we impose a coercivity condition on a quadratic form induced
by the Bellman operator along differences between each candidate
and the best candidate.
This condition can hold even when candidates are nearly collinear,
without causing the constants in our error bounds to deteriorate
(Section~\ref{sec:theoryunweighted}).
IW-PBV removes the coercivity requirement altogether through
importance weighting.
In our experiments, unweighted PBV often outperforms LSTD-Tournament
on Hopper and performs comparably on Deep OPE.

\section{OPE setup}
\subsection{MDP and validation data}
\label{sec:offline-data-setup}

Consider a time-homogeneous Markov decision process with state space
$\mathcal S$, action space $\mathcal A$, transition kernel
$P(\dd s'\mid s,a)$, reward $R$, and discount factor $\gamma\in[0,1)$.
Write $X=(S,A)$ and let
$r(x):=\mathbb E[R\mid X=x]$ denote the mean reward function.
Fix a target policy $\pi(\dd a\mid s)$ and an initial state--action
distribution $d_0$. The induced state--action transition kernel is
$P_\pi(\dd s',\dd a'\mid s,a)
:=
P(\dd s'\mid s,a)\pi(\dd a'\mid s')$.
For any state--action function $f$, define
$(\pi f)(s')
:=
\int f(s',a')\,\pi(\dd a'\mid s')$ and
$(P_\pi f)(s,a)
:=
\int (\pi f)(s')\,P(\dd s'\mid s,a)$.
The goal of OPE is to estimate the discounted value
$V^\pi:=\mathbb E_\pi[\sum_{t=0}^{\infty}\gamma^tR_t
\mid X_0\sim d_0]$, where $\mathbb E_\pi$ denotes expectation over
trajectories generated by the MDP under $\pi$ and initialized from $d_0$.

For model selection, we use sample splitting. Candidate $Q$-functions
and occupancy-ratio estimates are fitted on a training sample and
evaluated on an independent validation sample
$\mathcal D_n:=\{(X_t,R_t,S_t')\}_{t=1}^n$.
We condition on the training sample and assume that $(X_t)_{t\geq1}$
is a stationary Markov chain with invariant offline distribution $\nu$.
This setting includes stationary trajectories generated by a stationary
behavior policy $b(\dd a\mid s)$, as well as independently sampled
transitions.
In either case, conditional on the observed history through $X_t$,
the law of $(R_t,S_t')$ is the one-step MDP law at $X_t$.
For any state--action distribution $\rho$, square-integrable
functions $f,g$, and positive integer $m$, write
$\langle f,g\rangle_\rho:=\mathbb E_\rho[f(X)g(X)]$ and
$\|f\|_{2,\rho}:=\langle f,f\rangle_\rho^{1/2}$, and set
$[m]:=\{1,\ldots,m\}$.

\subsection{Bellman and adjoint Bellman equations}

The two nuisance functions we study are the target $Q$-function and the
discounted occupancy ratio. Both can be used to express the policy value
and satisfy the Bellman and adjoint Bellman equations, respectively.
The target $Q$-function is defined by
$Q^\pi(x)
:=
\mathbb E_\pi[
\sum_{t=0}^{\infty}\gamma^t R_t
\,|\,X_0=x
]$.
It gives the policy value $V^\pi=\mathbb E_{d_0}[Q^\pi(X)]$ and
satisfies the Bellman fixed-point equation
\begin{equation}
Q^\pi=\mathcal B_\pi Q^\pi,
\qquad
\mathcal B_\pi Q:=r+\gamma P_\pi Q,
\label{eq:primal-bellman}
\end{equation}
where $\mathcal B_\pi$ is the Bellman operator
\citep{sutton1998reinforcement}.

The normalized discounted occupancy ratio is defined by
\[
w^\pi
:=
\frac{\dd d_{\pi,\gamma}}{\dd\nu},
\qquad
d_{\pi,\gamma}
:=
(1-\gamma)\sum_{t=0}^{\infty}
\gamma^t\mathcal L_\pi(X_t),
\qquad
X_0\sim d_0,
\]
where $\mathcal L_\pi(X_t)$ denotes the marginal law of $X_t$ under the target
policy. We assume $d_{\pi,\gamma}\ll\nu$.
The occupancy ratio also gives the policy value through
$V^\pi=(1-\gamma)^{-1}\mathbb E_{d_{\pi,\gamma}}[r(X)]
=(1-\gamma)^{-1}\mathbb E_\nu[w^\pi(X)r(X)]$.
Moreover, it satisfies the adjoint Bellman fixed-point equation
\citep{van2026fitted}
\begin{equation}
\mathcal B_\pi^\ast w^\pi=w^\pi, \qquad
\mathcal B_\pi^\ast w
:=
(1-\gamma)\frac{\dd d_0}{\dd\nu}
+
\gamma P_\pi^\ast w,
\qquad
P_\pi^\ast w
:=
\frac{\dd\{(w\nu)P_\pi\}}{\dd\nu},
\label{eq:w-adjoint-update}
\end{equation}
which follows from the occupancy identity
$d_{\pi,\gamma}
=
(1-\gamma)d_0+\gamma d_{\pi,\gamma}P_\pi$
\citep{nachumEtAl2019DualDICE}. The term ``adjoint'' refers to the fact that $P_\pi^\ast$ is the adjoint
of $P_\pi$ under the $\nu$-inner product:
$\langle P_\pi^\ast w,Q\rangle_\nu
=\langle w,P_\pi Q\rangle_\nu$
\citep{ueharaEtAl2021FiniteSampleMinimax}.

\section{Motivation and key ideas for pairwise Bellman validation}
\label{sec:q-selection}

We begin by reviewing existing validation approaches for $Q$-function selection, then develop the key
ideas behind PBV. Section~\ref{sec:q-pbv} presents the algorithms,
including occupancy-ratio selection.

\subsection{Existing validation criteria}

Let $Q_1,\ldots,Q_K$ be a finite class of candidate $Q$-functions fitted on
the training sample. A natural approach, which we refer to as
\emph{Bellman-residual validation}, selects a candidate by minimizing the
mean-squared Bellman residual under the offline distribution $\nu$
\citep{antos2008learning,farahmand2011model}:
\[
\mathcal L_{\mathrm{BR}}(Q)
:=
\|Q-\mathcal B_\pi Q\|_{2,\nu}^2.
\]
Because $Q^\pi$ is a Bellman fixed point,
$\mathcal L_{\mathrm{BR}}(Q^\pi)=0$, so the minimum over a class containing
$Q^\pi$ is attained at $Q^\pi$. The criterion is not directly estimable
because $(\mathcal B_\pi Q)(X)$ depends on unknown conditional laws. Without an
MDP model, evaluating it generally requires estimating the Bellman residual
$Q_i-\mathcal B_\pi Q_i$, for example by separately regressing
$Q_i(X)-R-\gamma(\pi Q_i)(S')$ on $X$ for each candidate
\citep{zitovsky2023revisiting}. This introduces additional computation,
tuning, and estimation error.

A tempting alternative is to select candidates by minimizing the
mean-squared one-step temporal-difference (TD) error
\citep{sutton1988learning}:
\[
\mathcal L_{\mathrm{TD}}(Q)
:=
\mathbb E_\nu\!\left[
\{Q(X)-R-\gamma(\pi Q)(S')\}^2
\right].
\]
Unlike the mean-squared Bellman residual, this criterion can be
estimated directly from held-out transitions.
Although the TD residual is conditionally unbiased for the Bellman residual,
$\mathbb E[Q(X)-R-\gamma(\pi Q)(S')\mid X]
=Q(X)-(\mathcal B_\pi Q)(X)$, squaring introduces a conditional-variance
term into the objective:
\[
\mathcal L_{\mathrm{TD}}(Q)
=
\mathcal L_{\mathrm{BR}}(Q)
+
\mathbb E_\nu\!\left[
\operatorname{Var}\!\left(
R+\gamma(\pi Q)(S')
\,\middle|\,X
\right)
\right].
\]
For $\gamma>0$ and stochastic transitions, this term generally depends on $Q$,
so minimizing squared TD error need not minimize the Bellman-residual error
$\mathcal L_{\mathrm{BR}}(Q)$ or select $Q^\pi$, even when it belongs to the
candidate class \citep{zitovsky2023revisiting}. Directly estimating the squared
Bellman residual without this variance term encounters the classical
double-sampling obstruction: unbiased estimation generally requires two
conditionally independent targets at the same state--action pair
\citep{baird1995residual,feng2019kernel}.

A natural way around the double-sampling obstruction is to construct
a minimax criterion based on moments of the Bellman residual against
critic functions
\citep{antos2008learning,feng2019kernel,ueharaEtAl2020MWLMQL,
zanette2022bellman}.
Since $Q^\pi$ is a Bellman fixed point,
$\langle f,Q^\pi-\mathcal B_\pi Q^\pi\rangle_\nu=0$ for every suitable
critic $f$. These moments can be estimated directly from one-step
transitions because
\[
\mathbb E_\nu[f(X)\{Q(X)-R-\gamma(\pi Q)(S')\}]
=
\langle f,Q-\mathcal B_\pi Q\rangle_\nu.
\]
Given a critic class $\mathcal F$, this motivates the minimax population
criterion
\[
\mathcal L_{\mathcal F}(Q)
:=
\sup_{f\in\mathcal F:\ \|f\|_{2,\nu}\leq 1}
\left|
\mathbb E_\nu\!\left[
f(X)\{Q(X)-R-\gamma(\pi Q)(S')\}
\right]
\right|,
\]
which measures the largest violation of the Bellman moment
conditions over the critic class.

We refer to model selection based on an empirical analogue of this
criterion as \emph{minimax Bellman-moment validation}.
The main practical challenge is choosing a critic class $\mathcal F$
for which these moments distinguish among candidate $Q$-functions.
If $\mathcal F$ contains the normalized Bellman residual
$(Q_i-\mathcal B_\pi Q_i)/\|Q_i-\mathcal B_\pi Q_i\|_{2,\nu}$ for each
$i\in[K]$ with nonzero residual, then Cauchy--Schwarz and inclusion of
this critic give
$\mathcal L_{\mathcal F}(Q_i)
=
\|Q_i-\mathcal B_\pi Q_i\|_{2,\nu}$
\citep{ueharaEtAl2021FiniteSampleMinimax}.
Thus, minimax Bellman-moment validation can recover Bellman-residual
validation without directly estimating $\mathcal B_\pi Q_i$.
However, these ideal critics depend on the unknown Bellman residuals.
In practice, restrictive critic classes may fail to distinguish
among candidates, whereas richer classes increase statistical and
computational complexity
\citep{ueharaEtAl2020MWLMQL,zanette2022bellman}.

LSTD-Tournament \citep{liu2025modelselection} offers a simple
alternative that avoids choosing a separate critic class by
comparing candidate $Q$-functions using pairwise LSTDQ.
Its unnormalized selection criterion can be expressed as a
minimax Bellman-moment criterion with the candidate $Q$-functions
themselves as critics.
Specifically, for a finite candidate class $\{Q_1,\ldots,Q_K\}$,
the selector is
\begin{equation}
\label{eqn::tournamentunorm}
\widehat i
=
\arg\min_{i\in[K]}\max_{j\in[K]}
\left|
\frac{1}{n}\sum_{t=1}^n
Q_j(X_t)
\left\{Q_i(X_t)-R_t-\gamma(\pi Q_i)(S_t')\right\}
\right|.
\end{equation}
Its theoretical guarantees require invertibility of the pairwise
LSTD matrices, and the corresponding bounds can deteriorate
under poor conditioning, as may occur when candidates are nearly
collinear.
We discuss these limitations and the normalization variant used
in the authors' experiments in Section~\ref{sec:related-work-main}
and Appendix~\ref{app:discussionLSTD}.

\subsection{Key ideas behind PBV}
\label{sec:sketch}

We develop a penalized Bellman tournament that uses pairwise differences
between candidate $Q$-functions as critics, rather than the candidate
functions themselves.
The following three observations motivate this construction and the
algorithms in Section~\ref{sec:q-pbv}.

The first observation is that the estimation error
$\Delta_i:=Q_i-Q^\pi$ defines an oracle critic whose Bellman
moment is a quadratic form in $\Delta_i$. This leads to the oracle criterion
\begin{align*}
\mathcal L^\star(Q_i)
&:=
\left|\mathbb E_\nu\!\left[
\Delta_i(X)
\left\{Q_i(X)-R-\gamma(\pi Q_i)(S')\right\}
\right]\right| \\
&=
\left|\langle \Delta_i,Q_i-\mathcal B_\pi Q_i\rangle_\nu\right|
=
\left|\langle \Delta_i,(I-\gamma P_\pi)\Delta_i\rangle_\nu\right|,
\end{align*}
where the final equality follows from the fixed-point identity
$\mathcal B_\pi Q^\pi=Q^\pi$. Unlike squared Bellman-residual criteria, $\mathcal L^\star(Q_i)$
is the absolute value of a Bellman moment and, if $\Delta_i$ were
known, could be estimated directly from one-step transitions
without estimating $\mathcal B_\pi Q_i$.

The oracle criterion controls squared estimation error when the
Bellman quadratic form satisfies a coercivity condition.
Specifically, suppose that, for some $\kappa_\nu>0$ and every
$i\in[K]$,
\begin{equation}
\label{eqn:coerex}
\left|\langle \Delta_i,(I-\gamma P_\pi)\Delta_i\rangle_\nu\right|
\geq
\kappa_\nu\|\Delta_i\|_{2,\nu}^2.
\end{equation}
A corresponding upper bound holds under the standard one-step
concentrability condition
$C_{\nu,\pi}:=\|\dd(\nu P_\pi)/\dd\nu\|_\infty<\infty$.
By Cauchy--Schwarz and Jensen's inequality, for every $i\in[K]$,
\[
\left|\langle \Delta_i,(I-\gamma P_\pi)\Delta_i\rangle_\nu\right|
\leq
(1+\gamma\sqrt{C_{\nu,\pi}})\|\Delta_i\|_{2,\nu}^2.
\]
Thus, under \eqref{eqn:coerex} and one-step concentrability,
$\mathcal L^\star(Q_i)$ is comparable to
$\|\Delta_i\|_{2,\nu}^2$ up to constant factors.
Consequently, minimizing the oracle criterion yields an $L^2(\nu)$
error within a factor
$\{(1+\gamma\sqrt{C_{\nu,\pi}})/\kappa_\nu\}^{1/2}$
of the best candidate's error
$\min_{i\in[K]}\|Q_i-Q^\pi\|_{2,\nu}$.

The second observation is that pairwise differences between candidate
$Q$-functions can approximate the unknown oracle critics.
For any $i,j\in[K]$, write $h_{ij}:=Q_i-Q_j=\Delta_i-\Delta_j$.
Under realizability, there exists $i^\star\in[K]$ such that
$Q_{i^\star}=Q^\pi$, so $h_{ii^\star}=\Delta_i$ exactly.
More generally, comparison with a candidate close to $Q^\pi$
provides an accurate approximation to the oracle critic $\Delta_i$.
Because the best comparator is unknown, we evaluate each candidate
against all $K$ pairwise critics and take the largest penalized
Bellman moment as its worst-case score.
The population version of \emph{unweighted PBV} selects
\[
\arg\min_{i\in[K]}\max_{j\in[K]}
\left\{
\left|\langle h_{ij},Q_i-\mathcal B_\pi Q_i\rangle_\nu\right|
-\frac14\left|\langle h_{ij},(I-\gamma P_\pi)h_{ij}\rangle_\nu\right|
\right\}.
\]
Unlike LSTD-Tournament in \eqref{eqn::tournamentunorm},
this criterion uses pairwise differences rather than the candidate
functions themselves as critics.
Under realizability and coercivity, the unpenalized
tournament already recovers $Q^\pi$, since the true candidate has
a worst-case score of zero and comparison with it detects the
error of every other candidate.
Under misspecification, however, even the best candidate has a nonzero
Bellman residual, and poor comparators can inflate its worst-case score.
The quadratic penalty limits this inflation because the Bellman moment
grows at most linearly with critic size, whereas the penalty grows
quadratically, thereby enabling oracle guarantees under misspecification.

The guarantee for unweighted PBV under misspecification, however,
relies on a variant of the coercivity condition in \eqref{eqn:coerex},
requiring a positive lower bound along differences between each
candidate and the best available candidate.
This condition depends on the relationship between the offline
distribution $\nu$ and the target-policy dynamics and is related to,
but distinct from, the matrix-conditioning assumptions of linear TD
learning and LSTD-Tournament \citep{amortila2026unifying}
(see Section~\ref{sec:theoryunweighted} for a comparison).
A simple, albeit overly strong, sufficient condition is
$\|P_\pi h\|_{2,\nu}\leq c\|h\|_{2,\nu}$
for every such difference $h$ and some $c$ satisfying $\gamma c<1$.
By Cauchy--Schwarz, this gives the required lower bound with
$\kappa_\nu=1-\gamma c$.
Under one-step concentrability, one may take
$c=\sqrt{C_{\nu,\pi}}$, so the condition holds whenever
$\gamma^2C_{\nu,\pi}<1$.
In particular, under stationary on-policy sampling,
$C_{\nu,\pi}=1$ and $P_\pi$ is an $L^2(\nu)$ contraction,
so the condition holds automatically with $\kappa_\nu=1-\gamma$.
Under distribution shift, however, coercivity need not hold.

The third observation is that the coercivity requirement under the
offline distribution can be avoided by weighting each Bellman moment
by $w^\pi=\dd d_{\pi,\gamma}/\dd\nu$.
This weighting evaluates Bellman moments under the discounted
target-occupancy distribution, with respect to which
$\gamma P_\pi$ is a $\sqrt{\gamma}$-contraction in
$L^2(d_{\pi,\gamma})$
\citep{vanDerLaanKallus2025StationaryWeightedFQE,geladaBellemare2019CovariateShift}.
Consequently,
\begin{equation}
\label{eqn:contr}
(1-\sqrt{\gamma})\|\Delta_i\|_{2,d_{\pi,\gamma}}^2
\leq
\langle \Delta_i,(I-\gamma P_\pi)\Delta_i\rangle_{d_{\pi,\gamma}}
\leq
(1+\sqrt{\gamma})\|\Delta_i\|_{2,d_{\pi,\gamma}}^2.
\end{equation}
Thus, the weighted oracle criterion controls mean-squared
$Q$-function error under the discounted target occupancy without
requiring \eqref{eqn:coerex}.
Combining importance weighting with pairwise-difference critics and
the same quadratic penalty as unweighted PBV yields the population
version of \emph{importance-weighted PBV} (IW-PBV), which selects
\[
\arg\min_{i\in[K]}\max_{j\in[K]}
\left\{
\left|\langle h_{ij},Q_i-\mathcal B_\pi Q_i
\rangle_{d_{\pi,\gamma}}\right|
-\frac14
\left|\langle h_{ij},(I-\gamma P_\pi)h_{ij}
\rangle_{d_{\pi,\gamma}}\right|
\right\}.
\]
The contraction inequality ensures that the quadratic penalty
controls critic size without an additional coercivity assumption.
The tradeoff is that the occupancy ratio $w^\pi$ must be estimated
sufficiently accurately, introducing an additional model-selection
problem that we address using adjoint PBV in the next section.

\section{Pairwise Bellman validation algorithms}
\label{sec:q-pbv}

We now present the empirical unweighted PBV and IW-PBV selectors
for $Q$-function selection, along with A-PBV for occupancy-ratio
selection.

For both $Q$-function selectors, we also include $Q_0\equiv0$
as an additional comparator, but not as a candidate for selection.
This augmentation adds each $Q_i$ to the set of pairwise-difference
critics, since $Q_i-Q_0=Q_i$, and improved performance in our
experiments (Appendix~\ref{app:zero-comparator}).

\subsection{Unweighted PBV}
\label{sec:unweightedpbv}

For $i\in[K]$ and $0\le j\le K$, let $h_{ij}:=Q_i-Q_j$.
We define the unweighted PBV scores and selector by
\begin{align}
\widehat{\mathcal M}_{ij}^{\mathrm{unw}}
&:=
\left|\frac{1}{n}\sum_{t=1}^n
h_{ij}(X_t)
\{Q_i(X_t)-R_t-\gamma(\pi Q_i)(S_t')\}\right|
\notag\\
&\quad-
\frac14\left|
\frac{1}{n}\sum_{t=1}^n
h_{ij}(X_t)
\{h_{ij}(X_t)-\gamma(\pi h_{ij})(S_t')\}
\right|,
\notag\\
\widehat i_Q^{\mathrm{unw}}
&:=
\min\arg\min_{1\leq i\leq K}\max_{0\leq j\leq K}
\widehat{\mathcal M}_{ij}^{\mathrm{unw}},
\label{eq:q-unweighted-selector}
\end{align}
where both moments are evaluated under the offline distribution,
with absolute values taken after averaging.
By iterated expectation, the population penalty is
$\frac14|\langle h_{ij},(I-\gamma P_\pi)h_{ij}\rangle_\nu|$,
which, under Condition~\ref{cond:q-pairwise-coercivity},
is bounded below by
$\frac14\kappa_\nu\|h_{ik^\star}\|_{2,\nu}^2$
when $j=k^\star$.
Using the absolute quadratic form rather than its lower bound
avoids the need to know $\kappa_\nu$.
More generally, the penalty coefficient can be any
$\lambda\in(0,1)$; the choice $\lambda=1/4$ minimizes the
multiplicative constant in our population oracle bound.

\subsection{Adjoint PBV and importance-weighted PBV}

To implement IW-PBV, we first introduce adjoint PBV (A-PBV), which
applies pairwise validation to the adjoint Bellman equation to select
an occupancy-ratio estimator. We summarize the construction here and
defer further details and intuition to Appendix~\ref{sec:w-selection}.
As in the $Q$-function case, adjoint Bellman moments for fixed critics
can be estimated directly from one-step transitions and independent
draws from $d_0$ \citep{nachumEtAl2019DualDICE}.

For nonnegative integrable functions $p$ and $q$, define the generalized KL
divergence
\[
\mathsf D_\nu(p\|q)
:=
\mathbb E_\nu\!\left[
p(X)\log\frac{p(X)}{q(X)}-p(X)+q(X)
\right].
\]
Unlike the usual KL divergence, $\mathsf D_\nu$ does not require $p$ and $q$
to integrate to one and therefore permits unnormalized ratio candidates.
When the three divergences below are finite, the Bregman three-point
identity \citep{bauschkeCombettes2017}, together
with contraction of the adjoint Bellman operator in generalized KL divergence
\citep{van2026fitted}, gives
\[
\begin{aligned}
\left\langle w-\mathcal B_\pi^\ast w,\log(w/w^\pi)\right\rangle_\nu
&=
\mathsf D_\nu(w\|w^\pi)
-\mathsf D_\nu(\mathcal B_\pi^\ast w\|w^\pi)
+\mathsf D_\nu(\mathcal B_\pi^\ast w\|w)
\\
&\geq
(1-\gamma)\mathsf D_\nu(w\|w^\pi).
\end{aligned}
\]
Thus, $\log(w/w^\pi)$ provides an oracle critic whose adjoint
Bellman moment controls generalized-KL error.
Given $M$ candidate occupancy ratios $w_1,\ldots,w_M$, we
approximate this critic using pairwise log-ratios
$\ell_{ab}:=\log(w_a/w_b)$.

As in unweighted PBV, we penalize pairwise moments to control the
influence of inaccurate comparators under misspecification.
We use Hellinger and Bellman quadratic penalties for A-PBV and
IW-PBV, respectively.
Let $X_{0,1},\ldots,X_{0,n}$ be independent draws from $d_0$.\footnote{
If expectations under $d_0$ are known, the corresponding empirical
averages can be replaced by their population values.}
Define the penalized empirical pairwise scores
\begin{align*}
\widehat{\mathcal S}_{ab}^{\mathrm{adj}}
&:=
\left|\frac{1}{n}\sum_{t=1}^n
w_a(X_t)\{\ell_{ab}(X_t)-\gamma(\pi\ell_{ab})(S_t')\}
-\frac{1-\gamma}{n}\sum_{t=1}^{n}\ell_{ab}(X_{0,t})\right|
\\
&\quad
-\frac{1-\gamma}{4n}\sum_{t=1}^n
\{\sqrt{w_a(X_t)}-\sqrt{w_b(X_t)}\}^2,
\\
\widehat{\mathcal M}_{ij}^{\mathrm{IW}}(w)
&:=
\left|\frac{1}{n}\sum_{t=1}^n
w(X_t)h_{ij}(X_t)
\{Q_i(X_t)-R_t-\gamma(\pi Q_i)(S_t')\}\right|
\\
&\quad-
\frac14\left|
\frac1n\sum_{t=1}^n w(X_t)h_{ij}(X_t)
\{h_{ij}(X_t)-\gamma(\pi h_{ij})(S_t')\}
\right|.
\end{align*}
The IW-PBV penalty uses the same quadratic form as unweighted PBV,
but weighted by the fitted occupancy ratio.
We combine the two stages in Algorithm~\ref{alg:q-pairwise-selection},
which evaluates $M^2+K(K+1)$ pairwise scores without fitting
additional validation models or critics.

\begin{algorithm}[h!]
\caption{Adjoint and importance-weighted pairwise Bellman validation}
\label{alg:q-pairwise-selection}
\begin{algorithmic}[1]
\Require Ratio candidates $w_1,\ldots,w_M$; $Q$-function candidates
$Q_1,\ldots,Q_K$; validation trajectory $\mathcal D_n$; independent draws
$X_{0,1},\ldots,X_{0,n}\sim d_0$ (or exact $d_0$ moments); discount factor
$\gamma$
\State Set $Q_0\equiv 0$ as an additional comparator.
\State Select $\widehat i_w$ as the smallest index in
$\arg\min_{1\leq a\leq M}\max_{1\leq b\leq M}
\widehat{\mathcal S}_{ab}^{\mathrm{adj}}$ and set
$\widehat w:=w_{\widehat i_w}$.
\State Select $\widehat i_Q$ as the smallest index in
$\arg\min_{1\leq i\leq K}\max_{0\leq j\leq K}
\widehat{\mathcal M}_{ij}^{\mathrm{IW}}(\widehat w)$ and set
$\widehat Q:=Q_{\widehat i_Q}$.
\State \Return $(\widehat w,\widehat Q)$.
\end{algorithmic}
\end{algorithm}

\begin{samepage}
To improve finite-sample stability, we recommend screening out
occupancy-ratio candidates with excessively concentrated weights,
using the effective sample size (ESS) diagnostic
\citep{metelli2020importance,zenati2025counterfactual}. For a candidate $w_k$, its empirical ESS is
\[
\widehat{\mathrm{ESS}}(w_k)
:=
n_w\frac{(\mathbb P_{n,w}w_k)^2}{\mathbb P_{n,w}w_k^2},
\]
where $\mathbb P_{n,w}$ averages over the $n_w$ ratio-training
state--action observations.
\end{samepage}
Candidates with $\widehat{\mathrm{ESS}}(w_k)/n_w$ below a prespecified
threshold can be excluded to mitigate instability from extreme
ratio estimates; a practical default is $0.01$ or $0.05$.
We also recommend tempering the selected occupancy ratio
$\widehat w$ for downstream weighting, following
\citet{sinha2022experience}, as
\[
\widetilde w(x)
:=
\frac{\widehat w(x)^{1/T}}
{\mathbb P_{n,w}\widehat w^{1/T}},
\qquad T\ge 1,
\]
where larger $T$ reduces the concentration of the weights.  
As $T\to\infty$, the tempered weights converge to one,
and tempered IW-PBV reduces to unweighted PBV.
Because unweighted PBV does not require estimating an occupancy ratio,
it may be preferable when reliable importance weighting is difficult.
Truncation of importance weights can also be important for downstream policy-value estimation
\citep{mehrabi2024off}.

\section{Finite-sample guarantees}
\label{sec:finite-sample}

We establish finite-sample oracle inequalities for unweighted PBV and
the two-stage procedure combining A-PBV and IW-PBV. Write
$(P_bf)(x):=\mathbb E[f(X_{t+1})\mid X_t=x]$ for the validation chain's
transition operator. To accommodate temporal dependence in the validation
trajectory, we impose the following spectral-gap condition
\citep{jiangSunFan2025Bernstein}.
\begin{enumerate}[label=\textbf{(A\arabic*)}, ref=A\arabic*]
\item \label{cond:validation-spectral-gap}
\textit{Behavior-process spectral gap.}
There exists $g_b\in(0,1]$ such that
\begin{equation}
\|P_bf\|_{2,\nu}
\leq
(1-g_b)\|f\|_{2,\nu}
\qquad
\text{for every real-valued $f\in L^2(\nu)$ with $\mathbb E_\nu f=0$.}
\label{eq:validation-spectral-gap}
\end{equation}
Alternatively, $P_b$ is self-adjoint in $L^2(\nu)$ and
$\langle f,P_bf\rangle_\nu\le(1-g_b)\|f\|_{2,\nu}^2$ for every such $f$.
\end{enumerate}
For independent transitions, $(P_bf)(x)=\mathbb E_\nu[f(X)]$, so
Condition~\ref{cond:validation-spectral-gap} holds with $g_b=1$.

\subsection{Unweighted PBV}
\label{sec:theoryunweighted}
Unweighted PBV avoids occupancy-ratio estimation but requires
coercivity along differences from the best available candidate.
Let $k^\star\in\argmin_{i\in[K]}\|Q_i-Q^\pi\|_{2,\nu}$ index a best candidate.
\begin{enumerate}[label=\textbf{(A\arabic*)}, ref=A\arabic*, start=2]
\item \label{cond:q-pairwise-coercivity}
\textit{Coercivity along oracle-comparison directions.}
There exists a constant $\kappa_\nu\in(0,1]$ such that
\[
\left|\langle h_{ik^\star},(I-\gamma P_\pi)h_{ik^\star}\rangle_\nu\right|
\ge\kappa_\nu\|h_{ik^\star}\|_{2,\nu}^2
\quad\text{for every }i\in\{0,\ldots,K\}.
\]
\end{enumerate}
For comparison, our analysis extending LSTD-Tournament
\citep{liu2025modelselection} to misspecification
(Appendix~\ref{app:lstd-misspecification}) requires
that, for some $s_{\mathrm{LT}}(k^\star)>0$,
\begin{equation}
\label{eqn:lstd-conditioning}
\sum_{\ell\in\{i,k^\star\}}
\left\langle Q_\ell,
(I-\gamma P_\pi)h_{ik^\star}
\right\rangle_\nu^2
\geq 2s_{\mathrm{LT}}(k^\star)^2,
\end{equation}
for every $i$ with $\|h_{ik^\star}\|_{2,\nu}>0$.
The stronger matrix-conditioning condition used in
\citet{liu2025modelselection} is sufficient for
\eqref{eqn:lstd-conditioning}. Specifically, it requires
the pairwise LSTD matrices
\[
A_{ik^\star}
:=
\mathbb E_\nu\left[
\phi_{ik^\star}(X)
\{\phi_{ik^\star}(X)
-\gamma(P_\pi\phi_{ik^\star})(X)\}^\top
\right],
\qquad
\phi_{ik^\star}:=(Q_i,Q_{k^\star})^\top,
\]
to have smallest singular values bounded below by
$s_{\mathrm{LT}}(k^\star)>0$ for every $i\neq k^\star$.
For a fixed finite candidate class, coercivity implies
\eqref{eqn:lstd-conditioning}, but the converse need not hold.
However, this formal implication does not account for the
constants appearing in the respective finite-sample bounds.
In particular, coercivity only guarantees a lower bound of order
$\kappa_\nu^2\|h_{ik^\star}\|_{2,\nu}^4$ on the squared
LSTD moments.
Thus, when candidates are nearly identical, the LSTD conditioning
constant can become arbitrarily small even when $\kappa_\nu$
remains bounded away from zero.
Consequently, the finite-sample bounds for LSTD-Tournament can
deteriorate under near-collinearity even when the corresponding
PBV bounds remain well conditioned.

\begin{theorem}[Unweighted PBV oracle guarantee]
\label{thm:main-unweighted-pbv}
Suppose Conditions~\ref{cond:validation-spectral-gap},
\ref{cond:q-pairwise-coercivity}, \ref{cond:q-bounded-signal},
\ref{cond:q-subgaussian-td}, and~\ref{cond:q-unweighted-domain} hold.
Then there exists a finite constant
$C_{Q,\mathrm{unw}}>0$, depending only on the fixed constants in
Conditions~\ref{cond:q-bounded-signal} and~\ref{cond:q-subgaussian-td}, such that, for
every $\delta\in(0,1)$, with probability at least $1-\delta$,
\[
\|Q_{\widehat i_Q^{\mathrm{unw}}}-Q^\pi\|_{2,\nu}
\leq
\|Q_{k^\star}-Q^\pi\|_{2,\nu}
+\frac{2}{\kappa_\nu}\|Q_{k^\star}-\mathcal B_\pi Q_{k^\star}\|_{2,\nu}
+
\frac{C_{Q,\mathrm{unw}}}{\kappa_\nu}
\sqrt{\frac{\log(12K/\delta)}{ng_b}}.
\]
\end{theorem}
If the Bellman residual of the best candidate satisfies
$\|Q_{k^\star}-\mathcal B_\pi Q_{k^\star}\|_{2,\nu}
\le L_\nu\|Q_{k^\star}-Q^\pi\|_{2,\nu}$
for some $L_\nu\in[0,\infty)$, then the theorem yields
the oracle inequality
\[
\|Q_{\widehat i_Q^{\mathrm{unw}}}-Q^\pi\|_{2,\nu}
\le\left(1+\frac{2L_\nu}{\kappa_\nu}\right)
\min_{i\in[K]}\|Q_i-Q^\pi\|_{2,\nu}
+\frac{C_{Q,\mathrm{unw}}}{\kappa_\nu}
\sqrt{\frac{\log(12K/\delta)}{ng_b}}.
\]
One-step concentrability implies this condition with
$L_\nu=1+\gamma\sqrt{C_{\nu,\pi}}$.
Temporal dependence reduces the effective validation sample size
from $n$ to $ng_b$, while the dependence on the number of candidates
is only logarithmic.

\subsection{Adjoint PBV and importance-weighted PBV}

We next establish guarantees for the two-stage procedure, which estimates
the discounted occupancy ratio to avoid imposing coercivity under
the offline distribution. We measure occupancy-ratio estimation error
using the Hellinger distance
$H_\nu(p,q):=\|\sqrt p-\sqrt q\|_{2,\nu}$,
defined for nonnegative integrable functions $p$ and $q$.
For A-PBV, we assume one-step coverage and bounded ratio candidates,
which ensure that the pairwise log-ratio critics are uniformly bounded.

\begin{enumerate}[label=\textbf{(A\arabic*)}, ref=A\arabic*, start=3]
\item \label{cond:w-candidate-bounds}
\textit{One-step coverage and bounded ratio candidates.}
The measures $d_0$ and $\nu P_\pi$ are absolutely continuous with respect to
$\nu$. There are constants $\overline w<\infty$ and $C_w\ge1$ such that,
$\nu$-almost surely,
\[
0<w_a\le\overline w\quad(a\in[M]),
\qquad C_w^{-1}\le\frac{w_i}{w_j}\le C_w\quad(i,j\in[M]).
\]
\end{enumerate}

\begin{theorem}[A-PBV and IW-PBV oracle guarantee]
\label{thm:sequential-ta-pbv}
Suppose Conditions~\ref{cond:validation-spectral-gap} and
\ref{cond:w-candidate-bounds} hold, together with the technical conditions in
Appendix~\ref{app:joint-ta-pbv-conditions}. Then there exist finite constants
$C_{\mathrm{ratio}},C_{\mathrm{IW}}>0$, independent of
$n,M,K,\delta$, and $\gamma$, such that, for every $\delta\in(0,1)$, with
probability at least $1-\delta$, the output of
Algorithm~\ref{alg:q-pairwise-selection} satisfies
\begin{equation}
H_\nu(\widehat w,w^\pi)
\leq \frac{C_{\mathrm{ratio}}}{1-\gamma}
\left\{
\min_{a\in[M]}H_\nu(w_a,w^\pi)
+\sqrt{\frac{\log(48M/\delta)}{ng_b}}
\right\}.
\label{eq:main-sequential-w-bound}
\end{equation}
\begin{equation}
\begin{aligned}
\|\widehat Q-Q^\pi\|_{2,d_{\pi,\gamma}}
\leq C_{\mathrm{IW}}\Bigg[
&\frac{\min_{i\in[K]}\|Q_i-Q^\pi\|_{2,d_{\pi,\gamma}}}
{\sqrt{1-\gamma}}
+\frac{1}{1-\gamma}\sqrt{\frac{\log(24MK/\delta)}{ng_b}}
\\
&+\frac{1}{(1-\gamma)^{3/2}}
\left\{
\min_{a\in[M]}H_\nu(w_a,w^\pi)
+\sqrt{\frac{\log(48M/\delta)}{ng_b}}
\right\}
\Bigg].
\end{aligned}
\label{eq:main-sequential-q-bound}
\end{equation}
\end{theorem}
Unlike unweighted PBV, IW-PBV provides a $Q$-function error
guarantee without imposing coercivity under the offline distribution.
The bound separates candidate approximation error, validation error,
and error from occupancy-ratio estimation. The horizon dependence is
$(1-\gamma)^{-1/2}$ for $Q$-function approximation,
$(1-\gamma)^{-1}$ for direct validation error, and $(1-\gamma)^{-3/2}$
for occupancy-ratio estimation error.
Alternatively, applying the unweighted PBV analysis under the fitted
measure $\widehat w\nu$, provided the corresponding coercivity and
moment conditions hold, gives an oracle bound in
$L^2(\widehat w\nu)$ without a separate occupancy-ratio
estimation-error term (Appendix~\ref{sec:q-theory:fitted-weight}).

These guarantees also yield bounds for plug-in policy-value estimators.
For $Q$-functions, Cauchy--Schwarz gives
\[
\left|\mathbb E_{d_0}[\widehat Q(X)]-V^\pi\right|
\leq
\left\|\frac{\dd d_0}{\dd d_{\pi,\gamma}}\right\|_{2,d_{\pi,\gamma}}
\|\widehat Q-Q^\pi\|_{2,d_{\pi,\gamma}},
\]
where the $Q$-function error is bounded in
\eqref{eq:main-sequential-q-bound}.
For occupancy ratios, if
$r\in L^2(\nu)\cap L^2(d_{\pi,\gamma})$, the analogous
plug-in estimator satisfies
\[
\begin{aligned}
&\left|
(1-\gamma)^{-1}\mathbb E_\nu[\widehat w(X)r(X)]-V^\pi
\right|\\
&\qquad\leq
\frac{
\{2\overline w\|r\|_{2,\nu}^2
+2\|r\|_{2,d_{\pi,\gamma}}^2\}^{1/2}
}{1-\gamma}
H_\nu(\widehat w,w^\pi),
\end{aligned}
\]
where the Hellinger error is bounded in
\eqref{eq:main-sequential-w-bound}.
A doubly robust bound is given in Appendix~\ref{app:end-to-end-dr}.

\input{sections/experiments_defaults.tex}

\section{Discussion}

Unweighted PBV avoids occupancy-ratio estimation and admits oracle guarantees
under candidate-class misspecification, provided a coercivity condition holds.
IW-PBV removes this requirement through marginalized importance weighting,
at the cost of estimating occupancy ratios.
For occupancy-ratio selection, we also develop A-PBV and Adjoint
LSTD-Tournament, with the former avoiding matrix-conditioning
assumptions.
Our experiments show that unweighted PBV is competitive with
LSTD-Tournament, while the benefits of importance weighting depend on the
setting, the accuracy of the estimated occupancy ratios, and the degree
of tempering.
In particular, more accurate occupancy-ratio estimates do not necessarily
lead to more accurate downstream policy-value estimates, motivating
tempering and truncation to manage the bias--variance tradeoff
associated with importance weighting.

We extend PBV to projected $Q$-functions and occupancy weights in
Appendix~\ref{app:pbv-projections}
\citep{vanDerLaanEtAl2025AutomaticDRL}.
We also extend PBV to soft optimal control in Appendix~\ref{app:soft-pbv}
using occupancy weighting and temperature annealing, building on
\citet{vanDerLaanKallus2025SoftFQI}.

\subsection*{AI use statement}
Generative AI tools were used to assist with manuscript drafting and editing,
as well as developing and checking mathematical proofs.
In particular, these tools assisted with extending the finite-sample
analysis of LSTD-Tournament to candidate-class misspecification
(Appendix~\ref{app:lstd-misspecification}) and developing its extension
to occupancy-ratio selection
(Appendix~\ref{app:adjoint-lstd-tournament}).
The authors independently verified all mathematical arguments and take
full responsibility for the paper.

\bibliographystyle{plainnat}
\bibliography{refs}

\clearpage
\appendix

% Reset supplementary counters and use the S1, S2, ... convention.
\setcounter{equation}{0}
\renewcommand{\theequation}{S\arabic{equation}}
% The theorem-like environments share the theorem counter.  Detaching that
% counter from sections produces one sequence across the supplement.
\makeatletter
\@removefromreset{theorem}{section}
\makeatother
\setcounter{theorem}{0}
\setcounter{figure}{0}
\setcounter{table}{0}
\setcounter{algorithm}{0}
\renewcommand{\thetheorem}{S\arabic{theorem}}
\renewcommand{\thecorollary}{S\arabic{theorem}}
\renewcommand{\theproposition}{S\arabic{theorem}}
\renewcommand{\thelemma}{S\arabic{theorem}}
\renewcommand{\thedefinition}{S\arabic{theorem}}
\renewcommand{\theremark}{S\arabic{theorem}}
\renewcommand{\thefigure}{S\arabic{figure}}
\renewcommand{\thetable}{S\arabic{table}}
\renewcommand{\thealgorithm}{S\arabic{algorithm}}

% Include appendix sections and subsections, with linked titles and page numbers.
\etocdepthtag.toc{appendices}
\begingroup
\small
\renewcommand{\contentsname}{Appendix contents}
\setcounter{tocdepth}{2}
\etocsettagdepth{main}{none}
\etocsettagdepth{appendices}{subsection}
\pdfbookmark[0]{Appendix contents}{appendix-contents}
\tableofcontents
\endgroup
\clearpage

\section{Comparison with LSTD-Tournament}

\label{app:discussionLSTD}

\begin{figure}[H]
\centering
\includegraphics[width=.99\linewidth]{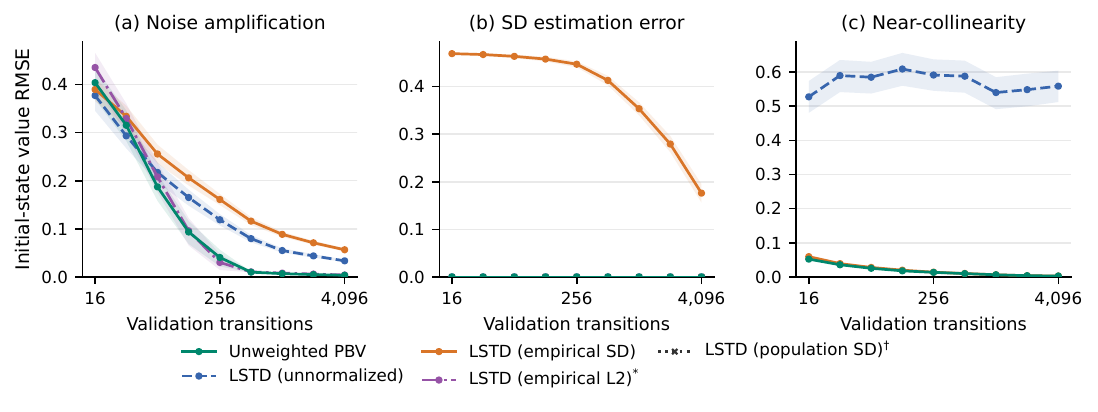}
\caption{Three on-policy examples of LSTD-Tournament:
(a) noise amplification with $Q^\pi=1$, repaired by empirical $L^2$
normalization; (b) SD estimation error with a Gaussian-spike truth,
repaired by oracle population SDs; (c) near-collinearity, addressed by
SD-normalized difference critics. Each library has 200 candidates.
Curves show initial-state value RMSE
over 500 repetitions, with 95\% bootstrap bands. Designs and policy-value
errors appear in Appendix~\ref{app:gaussian-kernel-experiment}.
Three curves coincide in (b).
${}^{*}$ denotes our modification using empirical $L^2$ norms;
${}^{\dagger}$ denotes an oracle control using population SDs unavailable
from validation data alone.}
\label{fig:gaussian-normalization-mechanisms}
\end{figure}

Relative to normalized-difference LSTD-Tournament, unweighted PBV avoids dividing
empirical moments by potentially small or inaccurately estimated critic standard deviations. When a critic has
little variation relative to its magnitude, this normalization can amplify
errors in the empirical moments. In Appendix~\ref{app:lstd-normalized},
we construct a counterexample where underestimating a critic's standard
deviation causes the normalized tournament to select a candidate with error
bounded away from zero, even when the best available candidate has
arbitrarily small error. For bounded candidates,
Appendix~\ref{app:lstd-empirical-normalization}
also establishes a finite-sample oracle inequality for the empirically
normalized tournament under a matrix-conditioning assumption, but with
a bound depending on both the ``spikiness'' measure
$
\|f-\mathbb E_\nu f\|_\infty^2/
\|f-\mathbb E_\nu f\|_{2,\nu}^2$
and the noise amplification factor $\|f\|_{2,\nu}/\|f-\mathbb E_\nu f\|_{2,\nu}$, evaluated for
candidates and pairwise differences
$f\in\{Q_i,Q_i-Q_j:\ i\ne j\}$\footnote{One can show that normalization by the norm $\|f\|_{2,\nu}$
instead of the standard deviation $\|f-\mathbb E_\nu f\|_{2,\nu}$ avoids amplification from the latter
factor but not from the former.} .
These quantities can be arbitrarily large even for uniformly bounded
candidates. For example, if the candidate class
contains Gaussian kernel bumps of bandwidth $h$ added to a nonzero constant,
the spikiness measure can grow at rate $h^{-d}$ and the noise amplification factor at
rate $h^{-d/2}$ as $h\to 0$ in $d$ dimensions. Figure~\ref{fig:gaussian-normalization-mechanisms}
illustrates both failure mechanisms for normalized LSTD-Tournament,
as well as the failure mechanism for unnormalized LSTD-Tournament.

\section{Adjoint PBV for occupancy-ratio selection}
\label{sec:w-selection}

The first stage of Algorithm~\ref{alg:q-pairwise-selection} selects the
discounted occupancy ratio used by its IW-PBV stage.
We consider a finite class of candidates
$w_1,\ldots,w_M$ fitted on the training sample and, conditional on that sample,
treat them as fixed and independent of the validation data. As in the
$Q$-function setting, we use a minimax criterion based on adjoint Bellman
moments and construct the critics directly from the candidate class.

\begin{enumerate}[label=\textbf{(C\arabic*)}, ref=C\arabic*]
\item \label{cond:w-relative-ratios}
\textit{Target-relative support and comparability.}
Let $\mathcal X^\pi:=\{x:w^\pi(x)>0\}$. Each candidate satisfies
$w_i=0$ $\nu$-almost surely on $(\mathcal X^\pi)^c$, and there exists
$\Lambda_w\geq1$ such that
\[
\Lambda_w^{-1}
\leq
\frac{w_i(X)}{w^\pi(X)}
\leq
\Lambda_w
\qquad
\nu\text{-almost surely on }\mathcal X^\pi,
\qquad i=1,\ldots,M.
\]
\end{enumerate}

\subsection{From oracle log-ratio critics to pairwise selection}

For every bounded critic $f$, the adjoint Bellman residual satisfies the
moment identity \citep{nachumEtAl2019DualDICE}
\begin{equation}
\mathbb E_\nu\!\left[
\{w(X)-(\mathcal B_\pi^\ast w)(X)\}f(X)
\right]
=
\mathbb E_\nu\!\left[
w(X)\{f(X)-\gamma(\pi f)(S')\}
\right]
-(1-\gamma)\mathbb E_{d_0}[f(X)].
\label{eq:w-adjoint-moment}
\end{equation}
In particular, this moment vanishes at $w=w^\pi$ for every bounded critic $f$.
We measure occupancy-ratio error using the generalized Kullback--Leibler
divergence, defined for nonnegative integrable functions $p$ and $q$ by
\begin{equation}
\mathsf D_\nu(p\|q)
:=
\mathbb E_\nu\!\left[
p(X)\log\frac{p(X)}{q(X)}-p(X)+q(X)
\right].
\label{eq:w-generalized-kl}
\end{equation}
Unlike the standard KL divergence, $\mathsf D_\nu$ does not require $p$ and
$q$ to integrate to one and therefore accommodates unnormalized candidates.
We use $0\log(0/q)=0$ and $p\log(p/0)=+\infty$ for $p>0$.
The adjoint Bellman operator is a contraction in generalized KL divergence
\citep{van2026fitted}:
\begin{equation}
\mathsf D_\nu(\mathcal B_\pi^\ast w\|w^\pi)
\leq
\gamma\mathsf D_\nu(w\|w^\pi).
\label{eq:w-generalized-kl:contract}
\end{equation}

We therefore seek a critic whose adjoint Bellman moment directly controls
generalized-KL error. In the $Q$-function case, pairing the Bellman
residual with $Q-Q^\pi$ yields a criterion comparable to squared $L^2$ error.
For occupancy ratios, the relevant error is generalized KL, which is the
Bregman divergence generated pointwise by $\phi(u)=u\log u-u$:
\[
\mathsf D_\nu(p\|q)
=
\mathbb E_\nu\!\left[
\phi\{p(X)\}-\phi\{q(X)\}
-\phi'\{q(X)\}\{p(X)-q(X)\}
\right],
\]
Since $\phi'(u)=\log u$, the corresponding critic is
$\phi'\{w(X)\}-\phi'\{w^\pi(X)\}=\log\{w(X)/w^\pi(X)\}$
\citep{amariNagaoka2000}. By the Bregman
three-point identity \citep{bauschkeCombettes2017},
\[
\begin{aligned}
&\mathbb E_\nu\!\left[
\{w(X)-(\mathcal B_\pi^\ast w)(X)\}
\{\phi'(w(X))-\phi'(w^\pi(X))\}
\right]
\\
&\qquad=
\mathsf D_\nu(w\|w^\pi)
-\mathsf D_\nu(\mathcal B_\pi^\ast w\|w^\pi)
+\mathsf D_\nu(\mathcal B_\pi^\ast w\|w)
\\
&\qquad\geq
\mathsf D_\nu(w\|w^\pi)
-\mathsf D_\nu(\mathcal B_\pi^\ast w\|w^\pi).
\end{aligned}
\]
By this inequality and the contraction property of $\mathcal B_\pi^\ast$ in
\eqref{eq:w-generalized-kl:contract},
\[
\mathbb E_\nu\!\left[
\{w(X)-(\mathcal B_\pi^\ast w)(X)\}
\log\{w(X)/w^\pi(X)\}
\right]
\geq
(1-\gamma)\mathsf D_\nu(w\|w^\pi).
\]
This motivates the oracle criterion
\begin{align}
\mathcal S^\star(w)
&:=
\mathbb E_\nu\!\left[
w(X)\left\{
\log\frac{w(X)}{w^\pi(X)}
-\gamma\left(\pi\log\frac{w}{w^\pi}\right)(S')
\right\}
\right]
-(1-\gamma)\mathbb E_{d_0}\!\left[
\log\frac{w(X)}{w^\pi(X)}
\right],
\label{eq:w-oracle-score}
\end{align}
which, by \eqref{eq:w-adjoint-moment}, directly controls generalized-KL error.
Unlike in the $Q$-function case, the adjoint contraction holds under the
offline distribution $\nu$, so no importance weighting or additional pairwise
coercivity condition is required.

Under Condition~\ref{cond:w-relative-ratios}, the pairwise log-ratio critics
satisfy
$\left|\log\{w_i(X)/w_j(X)\}\right|\leq 2\log\Lambda_w$
$\nu$-almost surely for every $i,j\in[M]$. Under this boundedness, the
following two-sided comparison holds.

\begin{lemma}[Oracle log-ratio equivalence]
\label{lem:w-oracle-equivalence}
Suppose Condition~\ref{cond:w-relative-ratios} holds. Then, for every
$i\in\{1,\ldots,M\}$,
\begin{equation}
(1-\gamma)\mathsf D_\nu(w_i\|w^\pi)
\le
\mathcal S^\star(w_i)
\le
2\sqrt{2}\,\Lambda_w(1+\sqrt{\gamma})\,
\mathsf D_\nu(w_i\|w^\pi).
\label{eq:w-oracle-equivalence}
\end{equation}
\end{lemma}
By Lemma~\ref{lem:w-oracle-equivalence}, a minimizer of the oracle score has
generalized-KL error within a fixed factor of the best candidate in the
class. The remaining obstacle is that the oracle critic
$\log(w_i/w^\pi)$ is unknown. As in the $Q$-function construction, we replace
it by pairwise critics from the candidate class. Define $\ell_{ij}:=\log\frac{w_i}{w_j}$ and the corresponding pairwise score
\[
\mathcal S_{ij}
:=
\mathbb E_\nu\!\left[
w_i(X)\{\ell_{ij}(X)-\gamma(\pi\ell_{ij})(S')\}
\right]
-(1-\gamma)\mathbb E_{d_0}[\ell_{ij}(X)].
\]
We then select the candidate with the smallest worst-case score
$\max_{1\leq j\leq M}|\mathcal S_{ij}|$. Under realizability, if
$w_{i^\star}=w^\pi$, then
$\ell_{i i^\star}=\log(w_i/w^\pi)$ is the oracle critic and hence
\[
\max_{1\leq j\leq M}|\mathcal S_{ij}|
\geq
\mathcal S_{i i^\star}
=
\mathcal S^\star(w_i)
\geq
(1-\gamma)\mathsf D_\nu(w_i\|w^\pi).
\]
By contrast, $\mathcal S_{i^\star j}=0$ for every $j$, since
$w_{i^\star}=w^\pi$ satisfies the adjoint Bellman equation. Thus, minimizing
the worst-case pairwise score recovers $w^\pi$ under realizability.

\subsection{Penalized score and selector}

The remaining challenge is misspecification. When $w^\pi$ is not contained in
the class, pairwise log ratios only approximate the oracle critics. We
therefore penalize each pairwise score to account for the error from using a
comparison candidate $w_j$ in place of $w^\pi$. We use Hellinger distance
because it provides a symmetric measure of separation between the two
candidate ratios that can be controlled by their generalized-KL errors
relative to $w^\pi$. For nonnegative $p$ and $q$, define the squared Hellinger
distance
\[
H_\nu^2(p,q)
:=
\mathbb E_\nu\!\left[
\{\sqrt{p(X)}-\sqrt{q(X)}\}^2
\right].
\]
For each ordered pair $(i,j)$, define the penalized population score
\begin{align*}
\mathcal S_{ij}^{\mathrm{adj}}
:=
|\mathcal S_{ij}|
-
\frac{1-\gamma}{4}H_\nu^2(w_i,w_j).
\end{align*}
The factor $1-\gamma$ comes from the lower bound for the oracle score in
\eqref{eq:w-oracle-equivalence}. The coefficient $(1-\gamma)/4$ allows
the error from replacing the oracle critic by a pairwise critic to be
absorbed in the oracle comparison. The A-PBV population
selector is
\[
\widetilde i_w
\in
\arg\min_{1\le i\le M}
\max_{1\le j\le M}
\mathcal S_{ij}^{\mathrm{adj}}.
\]

To implement A-PBV, we evaluate empirical versions of the penalized
pairwise scores using held-out one-step transitions
$\{(X_t,S_t')\}_{t=1}^n$ and independent initial state--action draws
$\{X_{0,t}\}_{t=1}^n$ from $d_0$. Algorithm~\ref{alg:w-pairwise-selection}
summarizes the resulting procedure, which requires only $M^2$ directly
computable pairwise scores and no additional nuisance or critic fitting.

\begin{algorithm}[t!]
\caption{Adjoint PBV selector for occupancy ratios}
\label{alg:w-pairwise-selection}
\begin{algorithmic}[1]
\Require Candidate ratios $w_1,\ldots,w_M$; validation transitions
$\{(X_t,S_t')\}_{t=1}^n$; initial draws $\{X_{0,t}\}_{t=1}^n$; discount
factor $\gamma$

\For{each pair $(i,j)\in\{1,\ldots,M\}^2$}
    \State Set $\ell_{ij}:=\log(w_i/w_j)$ and compute
\[
\begin{aligned}
\widehat{\mathcal S}_{ij}^{\mathrm{adj}}
:={}&
\left|\frac{1}{n}\sum_{t=1}^n
w_i(X_t)
\{\ell_{ij}(X_t)-\gamma(\pi\ell_{ij})(S_t')\} -
\frac{1-\gamma}{n}\sum_{t=1}^n\ell_{ij}(X_{0,t})\right| \\
& -
\frac{1-\gamma}{4n}\sum_{t=1}^n
\{\sqrt{w_i(X_t)}-\sqrt{w_j(X_t)}\}^2
.
\end{aligned}
\]
\EndFor

\State Let $\widehat i_w$ be the smallest index in $\arg\min_{1\le i\le M}
\max_{1\le j\le M}
\widehat{\mathcal S}_{ij}^{\mathrm{adj}}.$
\State \Return $w_{\widehat i_w}$.
\end{algorithmic}
\end{algorithm}

\subsection{Theoretical guarantees for occupancy-ratio selection}
\label{sec:w-theory}

\paragraph{Population oracle inequality.}
The following population oracle inequality compares A-PBV with the best
candidate in generalized KL, allowing candidate-class misspecification.

\begin{theorem}[Population A-PBV oracle inequality]
\label{thm:w-population-oracle}
Suppose Condition~\ref{cond:w-relative-ratios} holds. Then every population
minimizer $\widetilde i_w$ satisfies
\begin{equation}
\sqrt{\mathsf D_\nu(w_{\widetilde i_w}\|w^\pi)}
\le
\left\{
\frac{4\sqrt{2}\,\Lambda_w}{1-\sqrt{\gamma}}
+\frac23
\right\}
\min_{1\le i\le M}
\sqrt{\mathsf D_\nu(w_i\|w^\pi)}.
\label{eq:w-population-oracle}
\end{equation}
\end{theorem}

\paragraph{Finite-sample oracle inequality.}
We allow the initial-distribution sample size to differ from the number
of validation transitions.

\begin{enumerate}[label=\textbf{(C\arabic*)}, ref=C\arabic*, start=2]
\item \label{cond:w-initial-sample}
\textit{Initial-distribution draws.}
Conditional on the training sample, $X_{0,1},\ldots,X_{0,m}$ are independent
and identically distributed according to $d_0$.
\end{enumerate}
In Condition~\ref{cond:w-initial-sample}, we assume access to samples from the
target state--action distribution $d_0$ used to initialize the discounted
occupancy distribution and to evaluate the adjoint Bellman moments. If the
target initial-state law equals the state marginal $\nu_S$ of $\nu$, one may
reuse the validation states and replace the initial-sample average of each
critic $\ell_{ij}$ by $n^{-1}\sum_{t=1}^n(\pi\ell_{ij})(S_t)$, yielding
$d_0(\dd s,\dd a)=\nu_S(\dd s)\pi(\dd a\mid s)$. If expectations under $d_0$
are known exactly, they may replace the empirical initial-sample averages in
Algorithm~\ref{alg:w-pairwise-selection}, and
Condition~\ref{cond:w-initial-sample} is unnecessary.

\begin{theorem}[Hellinger oracle inequality for A-PBV]
\label{thm:w-finite-sample-oracle}
Suppose Conditions~\ref{cond:validation-spectral-gap}
and~\ref{cond:w-candidate-bounds} hold. Use either
Condition~\ref{cond:w-initial-sample}, exact $d_0$ moments, or the
validation-state initial averages described above when
$d_0(\dd s,\dd a)=\nu_S(\dd s)\pi(\dd a\mid s)$. There is a finite constant $C_{w,\mathrm H}>0$,
depending only on $C_w$ and $\overline w$, such that, for every
$\delta\in(0,1)$, conditional on the training sample, with probability
at least $1-\delta$, the output of
Algorithm~\ref{alg:w-pairwise-selection}, with its initial average computed
from $m$ draws, satisfies
\begin{align*}
H_\nu(w_{\widehat i_w},w^\pi)
\le{}&\frac{C_{w,\mathrm H}}{1-\gamma}
\left\{\min_{a\in[M]}H_\nu(w_a,w^\pi)
+\sqrt{\frac{\log(24M/\delta)}{ng_b}}\right\}\\
&+\frac{C_{w,\mathrm H}}{\sqrt{1-\gamma}}
\sqrt{\frac{\log(24M/\delta)}m}.
\end{align*}
The final term is omitted when the initial moments are known exactly.
If $d_0(\dd s,\dd a)=\nu_S(\dd s)\pi(\dd a\mid s)$, the validation-state
average described above satisfies the same bound with the final term omitted.
\end{theorem}

\section{Additional guarantees for \texorpdfstring{\(Q\)}{Q}-function selection}
\label{sec:q-theory}

We establish population and finite-sample guarantees for the importance-weighted
and unweighted PBV selectors.
Write $\mathcal J_Q:=\{0,1,\ldots,K\}$ for the comparator indices,
with $Q_0\equiv0$.

\subsection{Importance-weighted PBV}

\paragraph{Population oracle inequality.}
We justify IW-PBV's quadratic penalty under candidate-class
misspecification, extending the realizable argument in
Section~\ref{sec:sketch}. We require only the following second-moment condition.

\begin{enumerate}[label=\textbf{(B\arabic*)}, ref=B\arabic*]
\item \label{cond:q-population-domain}
\textit{Second moments under the target occupancy.}
The candidate functions and rewards satisfy
\[
\max_{1\leq i\leq K}\|Q_i\|_{2,d_{\pi,\gamma}}^2
+
\mathbb E_\nu[w^\pi(X)R^2]
<\infty.
\]
\end{enumerate}
Condition~\ref{cond:q-population-domain} holds, for example, when the rewards
and candidate functions are uniformly bounded.

Define the penalized population score
$\mathcal M_{ij}^{\mathrm{IW}}
:=|\langle h_{ij},Q_i-\mathcal B_\pi Q_i\rangle_{d_{\pi,\gamma}}|
-\frac14\langle h_{ij},(I-\gamma P_\pi)h_{ij}\rangle_{d_{\pi,\gamma}}$
and let $\widetilde i_Q\in\arg\min_{i\in[K]}\max_{j\in\mathcal J_Q}
\mathcal M_{ij}^{\mathrm{IW}}$.

\begin{theorem}[Population IW-PBV oracle inequality]
\label{thm:q-population-oracle}
Suppose Condition~\ref{cond:q-population-domain} holds. Then every population
minimizer $\widetilde i_Q$ satisfies
\begin{equation}
\|Q_{\widetilde i_Q}-Q^\pi\|_{2,d_{\pi,\gamma}}
\leq
\left\{
1+2\sqrt{\frac{1+\sqrt{\gamma}}{1-\sqrt{\gamma}}}
\right\}
\min_{1\leq i\leq K}
\|Q_i-Q^\pi\|_{2,d_{\pi,\gamma}}.
\label{eq:q-population-oracle}
\end{equation}
\end{theorem}

Thus, every minimizer of the penalized population criterion has error within
a factor depending only on $\gamma$ of that of the best candidate, without
requiring realizability. In particular, if $Q^\pi$ belongs to the candidate
class, every population minimizer equals $Q^\pi$ in
$L^2(d_{\pi,\gamma})$.

\paragraph{Finite-sample oracle inequality.}
We next account for the two additional sources of error in the practical
selector: estimation of $w^\pi$ and sampling variability in the pairwise
scores computed on the validation sample. For each candidate, define the TD
residual
\[
\varepsilon_i
:=
Q_i(X)-R-\gamma(\pi Q_i)(S'),\qquad i\in\mathcal J_Q.
\]
In particular, $\varepsilon_0=-R$. The following conditions are also required for the finite-sample analysis.

\begin{enumerate}[label=\textbf{(B\arabic*)}, ref=B\arabic*, start=2]
\item \label{cond:q-bounded-signal}
\textit{Bounded candidate differences and conditional TD means.}
There are finite constants $H_Q$ and $G_Q$ such that, almost surely,
\begin{align*}
\max_{0\le i,j\le K}|h_{ij}(X)|\le H_Q,
\qquad
\max_{0\le i\le K}
\left|\mathbb E[\varepsilon_i\mid X]\right|
\le G_Q.
\end{align*}

\item \label{cond:q-subgaussian-td}
\textit{Conditional subexponential TD noise.}
There is a finite constant $\sigma_Q>0$ such that, almost surely,
\begin{align*}
\max_{0\le i\le K}
\mathbb E\!\left[
\left.
\exp\left\{
\frac{
\left|\varepsilon_i-\mathbb E[\varepsilon_i\mid X]\right|
}{\sigma_Q}
\right\}
\right|X
\right]
\le 2.
\end{align*}

\item \label{cond:q-fitted-weight-envelope}
\textit{Bounded fitted weights.}
There is a finite constant $W_Q$ such that
$0\le\widehat w(X)\le W_Q$, $\nu$-almost surely.
\end{enumerate}

Under Condition~\ref{cond:validation-spectral-gap}, Bernstein concentration
holds for the stationary validation process with effective sample size of
order $ng_b$. Independent validation transitions correspond to $g_b=1$
\citep{jiangSunFan2025Bernstein}.
Under Condition~\ref{cond:q-bounded-signal}, the candidate differences and
the conditional means of the TD residuals are bounded.
Under Condition~\ref{cond:q-subgaussian-td}, the centered TD residuals are
conditionally subexponential; the residuals themselves need not be bounded. If
$|\varepsilon_i|\le\bar G_Q$ almost surely for every $i$ and some $\bar G_Q>0$,
then the
conditional-mean bound in Condition~\ref{cond:q-bounded-signal} holds with
$G_Q=\bar G_Q$, and Condition~\ref{cond:q-subgaussian-td} holds with
$\sigma_Q=2\bar G_Q/\log 2$.
Finally, the fitted-weight multiplier in the validation scores is bounded
under Condition~\ref{cond:q-fitted-weight-envelope}.

\begin{theorem}[Finite-sample IW-PBV oracle inequality]
\label{thm:q-finite-sample-oracle}
Suppose Condition~\ref{cond:validation-spectral-gap} and
Conditions~\ref{cond:q-population-domain}--\ref{cond:q-fitted-weight-envelope}
hold, and suppose the $Q$-validation fold is independent of all data used to
construct $\widehat w$. There are finite constants
$C_{Q,\mathrm{wt}},C_{Q,\mathrm{val}}>0$,
depending only on $H_Q,G_Q,\sigma_Q$, and $W_Q$, such that, for every
$\delta\in(0,1)$, conditional on the training sample and the data used to
construct $\widehat w$, with probability at least $1-\delta$, the
IW-PBV selector in
Algorithm~\ref{alg:q-pairwise-selection} satisfies
\begin{align}
\|Q_{\widehat i_Q}-Q^\pi\|_{2,d_{\pi,\gamma}}
\le{}&
\left\{
1+2\sqrt{\frac{1+\sqrt{\gamma}}{1-\sqrt{\gamma}}}
\right\}
\min_{1\le i\le K}
\|Q_i-Q^\pi\|_{2,d_{\pi,\gamma}}
\notag\\
&+
\frac{C_{Q,\mathrm{wt}}}{\sqrt{1-\sqrt{\gamma}}}
H_\nu(\widehat w,w^\pi)
\notag\\
&+
\frac{C_{Q,\mathrm{val}}}{1-\sqrt{\gamma}}
\sqrt{\frac{\log(12K/\delta)}{ng_b}}.
\label{eq:q-finite-sample-oracle}
\end{align}
\end{theorem}

Explicit choices for $C_{Q,\mathrm{wt}}$ and $C_{Q,\mathrm{val}}$ are given in
Appendix~\ref{app:q-finite-sample-proof}. The bound in
Theorem~\ref{thm:q-finite-sample-oracle} consists of the best-candidate error,
a Hellinger error for the fitted occupancy ratio, and a validation
term of order
$\sqrt{\log(12K/\delta)/(ng_b)}/(1-\sqrt{\gamma})$. Thus, temporal dependence
enters through the effective sample size $ng_b$, while the class size enters
only logarithmically. For fixed $\gamma$ and envelope constants, the validation
term vanishes whenever $\log K=o(ng_b)$. Consistency follows when the
best-candidate error and Hellinger weight error also vanish. If
$\widehat w=w^\pi$, the weight-error term is identically zero.

A plug-in value-error bound also holds. Since
$d_{\pi,\gamma}\geq(1-\gamma)d_0$, we have
$c_{\mathrm{conc}}
:=\|\dd d_0/\dd d_{\pi,\gamma}\|_\infty
\leq(1-\gamma)^{-1}$, and Cauchy--Schwarz gives
$|\mathbb E_{d_0}[Q_{\widehat i_Q}(X)]-V^\pi|
\leq
\sqrt{c_{\mathrm{conc}}}\,
\|Q_{\widehat i_Q}-Q^\pi\|_{2,d_{\pi,\gamma}}$.

\subsection{Unweighted PBV}
\label{sec:q-theory:unweighted}

The unweighted selector avoids occupancy-ratio estimation but instead requires
coercivity (Condition~\ref{cond:q-pairwise-coercivity}) under the offline distribution $\nu$.

\begin{enumerate}[label=\textbf{(B\arabic*)}, ref=B\arabic*, start=5]
\item \label{cond:q-unweighted-domain}
\textit{Second moments under the offline distribution.}
The true action-value function, candidate functions, rewards, and TD residuals satisfy
\[
\|Q^\pi\|_{2,\nu}^2
+
\max_{1\leq i\leq K}\|Q_i\|_{2,\nu}^2
+
\mathbb E_\nu[R^2]
+\max_{1\le i\le K}\mathbb E_\nu[\varepsilon_i^2]
<\infty.
\]
\end{enumerate}

\paragraph{Population oracle inequality.}
For each $(i,j)$, define
\begin{equation}
\mathcal M_{ij}^{\mathrm{unw}}
:=
\left|\mathbb E_\nu[h_{ij}(X)\varepsilon_i]\right|
-\frac14\left|
\mathbb E_\nu\{h_{ij}(X)(\varepsilon_i-\varepsilon_j)\}
\right|,
\label{eq:app-q-unweighted-population-score}
\end{equation}
This is the population counterpart of
\eqref{eq:q-unweighted-selector}, since
$\varepsilon_i-\varepsilon_j=h_{ij}(X)-\gamma(\pi h_{ij})(S')$.
Condition~\ref{cond:q-unweighted-domain} and Cauchy--Schwarz ensure that
both expectations are finite. The corresponding population selector is
\[
\widetilde i_Q^{\mathrm{unw}}
\in\argmin_{i\in[K]}\max_{j\in\mathcal J_Q}\mathcal M_{ij}^{\mathrm{unw}}.
\]

\begin{theorem}[Population oracle inequality for unweighted PBV]
\label{thm:q-unweighted-population-oracle}
Suppose Condition~\ref{cond:q-unweighted-domain} holds. For any $k\in[K]$ and
$\kappa_\nu\in(0,1]$ for which
Condition~\ref{cond:q-pairwise-coercivity} holds with $k$ in place of
$k^\star$, every population minimizer satisfies
\begin{equation}
\|Q_{\widetilde i_Q^{\mathrm{unw}}}-Q^\pi\|_{2,\nu}
\le\|Q_k-Q^\pi\|_{2,\nu}
+\frac{2}{\kappa_\nu}\|Q_k-\mathcal B_\pi Q_k\|_{2,\nu}.
\label{eq:q-unweighted-population-oracle}
\end{equation}
\end{theorem}

For the best-candidate index $k^\star$, if
$\|Q_{k^\star}-\mathcal B_\pi Q_{k^\star}\|_{2,\nu}
\le L_\nu\|Q_{k^\star}-Q^\pi\|_{2,\nu}$ for some $L_\nu\in[0,\infty)$,
then
\[
\|Q_{\widetilde i_Q^{\mathrm{unw}}}-Q^\pi\|_{2,\nu}
\le\left(1+\frac{2L_\nu}{\kappa_\nu}\right)
\min_{i\in[K]}\|Q_i-Q^\pi\|_{2,\nu}.
\]

\paragraph{Why penalization is needed.}
Without penalization, misspecification can lead to arbitrarily large error
relative to the best available candidate.
\begin{proposition}[Failure of unpenalized pairwise validation under misspecification]
\label{prop:unweighted-unpenalized-counterexample}
For every $\gamma\in[0,1)$ and $\rho>1$, there exist a four-state MDP and
four bounded candidate $Q$-functions satisfying
Conditions~\ref{cond:q-pairwise-coercivity} and~\ref{cond:q-unweighted-domain}
with $\kappa_\nu=1$, such that the unique minimizer
\[
\widetilde i_0\in\argmin_{i\in\{0,1,2,3\}}\max_{j\in\{0,1,2,3\}}
\left|\langle Q_i-Q_j,Q_i-\mathcal B_\pi Q_i\rangle_\nu\right|
\]
satisfies
\[
\|Q_{\widetilde i_0}-Q^\pi\|_{2,\nu}
=\rho\min_{i\in\{0,1,2,3\}}\|Q_i-Q^\pi\|_{2,\nu}.
\]
\end{proposition}

\paragraph{Finite-sample oracle inequality.}
The following bound applies to the empirical selector.

\begin{theorem}[Finite-sample oracle inequality for unweighted PBV]
\label{thm:q-unweighted-finite-sample-oracle}
Suppose Conditions~\ref{cond:validation-spectral-gap},
\ref{cond:q-pairwise-coercivity},
\ref{cond:q-bounded-signal}, \ref{cond:q-subgaussian-td},
and~\ref{cond:q-unweighted-domain} hold.
There is a finite constant $C_{Q,\mathrm{unw}}>0$, depending only on
$H_Q,G_Q$, and $\sigma_Q$, such that, for every $\delta\in(0,1)$,
conditional on the training sample, with probability at least $1-\delta$,
the selector in \eqref{eq:q-unweighted-selector} satisfies
\begin{align}
\|Q_{\widehat i_Q^{\mathrm{unw}}}-Q^\pi\|_{2,\nu}
\leq{}&
\|Q_{k^\star}-Q^\pi\|_{2,\nu}
+\frac{2}{\kappa_\nu}\|Q_{k^\star}-\mathcal B_\pi Q_{k^\star}\|_{2,\nu}
+
\frac{C_{Q,\mathrm{unw}}}{\kappa_\nu}
\sqrt{\frac{\log(12K/\delta)}{ng_b}}.
\label{eq:q-unweighted-finite-sample-oracle}
\end{align}
More generally, the remaining conditions and coercivity relative to
any fixed comparator $k\in[K]$ imply the same bound with $k^\star$ replaced
by $k$.
\end{theorem}

\begin{corollary}[Unweighted PBV under target-policy stationarity]
\label{cor:q-unweighted-stationary}
Suppose $\nu P_\pi=\nu$ and Conditions~\ref{cond:validation-spectral-gap},
\ref{cond:q-bounded-signal}, \ref{cond:q-subgaussian-td},
and~\ref{cond:q-unweighted-domain} hold. Then, for every $\delta\in(0,1)$,
conditional on training, with probability at least $1-\delta$,
\[
\|Q_{\widehat i_Q^{\mathrm{unw}}}-Q^\pi\|_{2,\nu}
\le\left\{1+2\sqrt{\frac{1+\gamma}{1-\gamma}}\right\}
\min_i\|Q_i-Q^\pi\|_{2,\nu}
+\frac{C_{Q,\mathrm{unw}}}{1-\gamma}
\sqrt{\frac{\log(12K/\delta)}{ng_b}}.
\]
\end{corollary}

An explicit choice of $C_{Q,\mathrm{unw}}$ is given in
Appendix~\ref{app:q-unweighted-finite-sample-proof}. The validation term has
the same $\sqrt{\log K/(ng_b)}$ order as for importance-weighted PBV, and there is
no weight-estimation term. The tradeoff is dependence on the constant $\kappa_\nu$ and an oracle comparison in $L^2(\nu)$.
For fixed $\kappa_\nu>0$ and envelope constants, consistency holds when
$\|Q_{k^\star}-Q^\pi\|_{2,\nu}$ and
$\|Q_{k^\star}-\mathcal B_\pi Q_{k^\star}\|_{2,\nu}$ vanish and
$\log K=o(ng_b)$.

A similar plug-in value-error bound holds in the offline norm. If $d_0\ll\nu$
and
$c_{0,\nu}
:=
\|\dd d_0/\dd\nu\|_{2,\nu}
<\infty$, then
$|\mathbb E_{d_0}[Q_{\widehat i_Q^{\mathrm{unw}}}(X)]-V^\pi|
\leq
c_{0,\nu}
\|Q_{\widehat i_Q^{\mathrm{unw}}}-Q^\pi\|_{2,\nu}$.

\input{sections/q_fitted_weight_theory.tex}

\section{Joint PBV selection and doubly robust OPE}
\label{app:end-to-end-dr}
\label{app:joint-ta-pbv-conditions}

This appendix applies the two-stage procedure in
Algorithm~\ref{alg:q-pairwise-selection} to doubly robust OPE. Its first
stage selects an occupancy-ratio candidate, and its second stage uses that
ratio to select a $Q$-function candidate on the same validation fold. The
selected nuisances are then evaluated using a doubly robust estimator on an
evaluation fold.

For the joint guarantee in Theorem~\ref{thm:sequential-ta-pbv}, we additionally
assume Conditions~\ref{cond:q-population-domain}--\ref{cond:q-subgaussian-td}
and either exact $d_0$ moments or
Condition~\ref{cond:w-initial-sample} with $m=n$.

Let the candidates $Q_1,\ldots,Q_K$ and $w_1,\ldots,w_M$ be fitted
on a training sample. Let $\mathcal D_{\mathrm{sel}}$ denote the transition fold shared by the
two selection stages, and let $\mathcal D_V$ denote the evaluation fold,
which may equal $\mathcal D_{\mathrm{sel}}$. Let
$\mathcal D_{0,\mathrm{sel}}$ and $\mathcal D_{0,V}$ denote the corresponding
initial-distribution samples when those moments are evaluated empirically.

\begin{algorithm}[!htbp]
\caption{Doubly robust OPE after PBV selection}
\label{alg:sequential-pbv-dr}
\begin{algorithmic}[1]
\Require Candidates $Q_1,\ldots,Q_K$ and $w_1,\ldots,w_M$ fitted
only on a training sample; transition folds
$\mathcal D_{\mathrm{sel}},\mathcal D_V$, each of size $n$ and independent of
the training sample, with $\mathcal D_V=\mathcal D_{\mathrm{sel}}$ permitted;
when $d_0$ moments are empirical, initial samples
$\mathcal D_{0,\mathrm{sel}},\mathcal D_{0,V}$, each of size $n$ and independent
of training, with $\mathcal D_{0,V}=\mathcal D_{0,\mathrm{sel}}$ permitted;
discount factor $\gamma$
\State Run both stages of Algorithm~\ref{alg:q-pairwise-selection} on
$\mathcal D_{\mathrm{sel}}$, using $\mathcal D_{0,\mathrm{sel}}$ or exact
$d_0$ moments, and set
$(\widehat w,\widehat Q)$ to the returned pair.
\State Evaluate $(\widehat Q,\widehat w)$ on
$\mathcal D_V$ and $\mathcal D_{0,V}$, or using exact $d_0$ moments.
\State \Return $\widehat V_{\mathrm{DR}}$ defined in
\eqref{eq:app-sequential-dr-estimator}.
\end{algorithmic}
\end{algorithm}

Let $\mathbb P_{n,V}$ denote the empirical average over the evaluation transition
fold, and let $\mathbb P_{n,0,V}$ denote the empirical average over
$\mathcal D_{0,V}$. The estimator returned by
Algorithm~\ref{alg:sequential-pbv-dr} is the doubly robust estimator
\citep{jiangLi2016DoublyRobust,kallusUehara2020DoubleRL}:
\begin{equation}
\widehat V_{\mathrm{DR}}
:=
\mathbb P_{n,0,V}\widehat Q
+
\frac{1}{1-\gamma}
\mathbb P_{n,V}\!\left[
\widehat w(X)
\{R+\gamma(\pi\widehat Q)(S')-\widehat Q(X)\}
\right].
\label{eq:app-sequential-dr-estimator}
\end{equation}

The following conditions are imposed for the end-to-end analysis.
\begin{enumerate}[label=\textbf{(D\arabic*)}, ref=D\arabic*]
\item \label{cond:sequential-fold-laws}
\textit{Validation laws.}
Each transition fold follows the stationary validation setup of
Section~\ref{sec:offline-data-setup}, has invariant distribution $\nu$, and
has spectral gap at least $g_b$. Each empirical initial-distribution
sample consists of independent draws from $d_0$.

\item \label{cond:sequential-selector-regularity}
\textit{Selector regularity.}
Conditions~\ref{cond:q-population-domain},
\ref{cond:q-bounded-signal}, \ref{cond:q-subgaussian-td},
and~\ref{cond:w-candidate-bounds} hold on the selection fold, with constants
$H_Q,G_Q,\sigma_Q,C_w$, and $\overline w$ as defined there.

\item \label{cond:sequential-initial-evaluation}
\textit{Initial-value evaluation.}
Either $\mathbb E_{d_0}[Q_i(X)]$ is known exactly for every candidate, or
there is a finite constant $\sigma_{Q,0}>0$ such that
\[
\max_{1\le i\le K}\mathbb E_{d_0}\exp\left\{
\frac{|Q_i(X_0)-\mathbb E_{d_0}Q_i|}{\sigma_{Q,0}}\right\}\le2.
\]
When the initial-state expectations under $d_0$ are evaluated exactly, set $\sigma_{Q,0}:=0$ and interpret
$\mathbb P_{n,0,V}\widehat Q$ in
\eqref{eq:app-sequential-dr-estimator} as
$\mathbb E_{d_0}[\widehat Q(X)]$.
\end{enumerate}

The finite candidate classes permit a uniform analysis of both PBV stages
and the final estimator, including when the selection and evaluation folds
coincide. For the selected ratio,
Condition~\ref{cond:q-fitted-weight-envelope} holds by
Condition~\ref{cond:w-candidate-bounds}.

For the statement below, let
\[
\begin{aligned}
\epsilon_w
&:=\min_{1\le j\le M}H_\nu(w_j,w^\pi),
&
\epsilon_Q
&:=\min_{1\le i\le K}\|Q_i-Q^\pi\|_{2,d_{\pi,\gamma}},
\\
\rho_\delta
&:=\sqrt{\frac{\log\{12(K\vee M)/\delta\}}{ng_b}}.
\end{aligned}
\]

\begin{corollary}[Doubly robust oracle bound after PBV selection]
\label{cor:sequential-pbv-dr}
Run Algorithm~\ref{alg:sequential-pbv-dr} and suppose
Conditions~\ref{cond:sequential-fold-laws}--\ref{cond:sequential-initial-evaluation}
hold. Then, conditional on the training sample, for every
$\delta\in(0,1)$, with probability at least $1-\delta$,
\begin{align}
|\widehat V_{\mathrm{DR}}-V^\pi|
\lesssim{}&
\frac{\epsilon_w\epsilon_Q}{(1-\gamma)^2}
+\frac{\epsilon_w^2}{(1-\gamma)^3}
+\left\{
\frac{\epsilon_Q}{(1-\gamma)^2}
+\frac{\epsilon_w}{(1-\gamma)^3}
+\frac1{1-\gamma}
\right\}\rho_\delta
+\frac{\rho_\delta^2}{(1-\gamma)^3}.
\label{eq:app-sequential-dr-oracle}
\end{align}

Here $\lesssim$ suppresses only universal constants and the fixed structural
constants in Conditions~\ref{cond:sequential-selector-regularity}--\ref{cond:sequential-initial-evaluation}.
\end{corollary}

The first two terms come from the second-order nuisance remainder.
They involve the product $\epsilon_w\epsilon_Q$ and the squared
occupancy-ratio approximation error $\epsilon_w^2$, respectively. The third term collects contributions linear in $\rho_\delta$ from
selection and evaluation, including interactions with approximation error.
The fourth is quadratic in $\rho_\delta$. The selection and evaluation samples may coincide.

\section{Pairwise validation for projected \texorpdfstring{$Q$}{Q}-functions and occupancy weights}
\label{app:pbv-projections}

We extend the selectors in Section~\ref{sec:q-pbv} to the weighted
Bellman projection and Bellman--Riesz problems of
\citet[Section~4 and Appendix~G]{vanDerLaanEtAl2025AutomaticDRL}.
Each candidate consists of the two functions returned by a minimax fit.
Validation uses candidate differences as critics and requires only
held-out moments. The working space and projection weight are fixed
across candidates.

\subsection{Projected Bellman solutions}
\label{app:pbv-projections-targets}

Use the MDP notation of Section~\ref{sec:offline-data-setup}. Write
$T_\pi:=I-\gamma P_\pi$. Fix a nonnegative weight $\omega$ with
$0<\mathbb E_\nu\omega<\infty$, let $\mu:=\omega\nu$, and fix a closed
linear working space $\mathcal H\subseteq L^2(\mu)$. Set
\[
\langle f,h\rangle_{\omega\nu}:=\mathbb E_\nu[\omega fh],
\qquad
\|f\|_{2,\omega\nu}^2:=\langle f,f\rangle_{\omega\nu},
\qquad
\ell_\pi(h):=(1-\gamma)\mathbb E_{d_0}[h].
\]
Assume that $T_\pi$ maps $\mathcal H$ into $L^2(\mu)$,
$r\in L^2(\mu)$, and $\ell_\pi$ is a well-defined finite functional
on $\mathcal H$.
Assume also that the minimizers defined below exist and are unique.
The projected $Q$-function and its Bellman residual are
\begin{equation}
Q_{\mathcal H}
:=\argmin_{Q\in\mathcal H}\frac12\|T_\pi Q-r\|_{2,\omega\nu}^2,
\qquad
e_{\mathcal H}:=T_\pi Q_{\mathcal H}-r.
\label{eq:proj-q-target}
\end{equation}
The Bellman--Riesz representer and projected occupancy weight are
\begin{equation}
\begin{aligned}
\alpha_{\mathcal H}
&:=\argmin_{\alpha\in\mathcal H}
\left\{\frac12\|T_\pi\alpha\|_{2,\omega\nu}^2-\ell_\pi(\alpha)\right\},
\\
\tau_{\mathcal H}&:=T_\pi\alpha_{\mathcal H},
\qquad
w_{\mathcal H}:=\omega\tau_{\mathcal H}.
\end{aligned}
\label{eq:proj-riesz-target}
\end{equation}
These definitions give the normal equations
\begin{equation}
\langle e_{\mathcal H},T_\pi h\rangle_{\omega\nu}=0,
\qquad
\langle\tau_{\mathcal H},T_\pi h\rangle_{\omega\nu}=\ell_\pi(h),
\qquad h\in\mathcal H.
\label{eq:proj-normal-equations}
\end{equation}
Since $T_\pi Q^\pi=r$, $Q_{\mathcal H}$ is the projection of $Q^\pi$
onto $\mathcal H$ under the Bellman seminorm
$f\mapsto\|T_\pi f\|_{2,\mu}$.
For the occupancy weight, let $\mathcal G$ be the closure of
$T_\pi\mathcal H$ in $L^2(\mu)$. If $d_{\pi,\gamma}$ has a density
$g_\pi\in L^2(\mu)$ with respect to $\mu$, the occupancy identity
$\langle g_\pi,T_\pi h\rangle_\mu=\ell_\pi(h)$ and
\eqref{eq:proj-normal-equations} give
$\tau_{\mathcal H}=\Pi_{\mathcal G}g_\pi$, where
$\Pi_{\mathcal G}$ is the $L^2(\mu)$ orthogonal projection.
Equivalently, $w_{\mathcal H}$ is the least-squares projection of
$w^\pi$ onto $\omega\mathcal G$, with squared error
$\int_{\{\omega>0\}}(w-w^\pi)^2/\omega\,\dd\nu$.

For value estimation, $\tau_{\mathcal H}=T_\pi\alpha_{\mathcal H}$
is orthogonal to $e_{\mathcal H}$, so replacing $r$ by
$T_\pi Q_{\mathcal H}$ leaves the weighted reward moment unchanged.
Applying the second normal equation with $h=Q_{\mathcal H}$ gives the
value identity
\begin{equation}
\mathbb E_\nu[w_{\mathcal H}r]
=\langle\tau_{\mathcal H},r\rangle_{\omega\nu}
=\langle\tau_{\mathcal H},T_\pi Q_{\mathcal H}\rangle_{\omega\nu}
=\ell_\pi(Q_{\mathcal H}).
\label{eq:proj-value-identity}
\end{equation}

\subsection{Minimax identification}
\label{app:pbv-projections-minimax}

The target pairs are identified by the population minimax problems
\begin{align}
\min_{e\in L^2(\omega\nu)}\sup_{Q\in\mathcal H}
\mathcal L_Q(e,Q),
\qquad
\mathcal L_Q(e,Q)
&:=\tfrac12\|e\|_{2,\omega\nu}^2
+\langle e,r-T_\pi Q\rangle_{\omega\nu},
\label{eq:proj-q-minimax}\\
\min_{\tau\in L^2(\omega\nu)}\sup_{\alpha\in\mathcal H}
\mathcal L_w(\tau,\alpha),
\qquad
\mathcal L_w(\tau,\alpha)
&:=\tfrac12\|\tau\|_{2,\omega\nu}^2
-\langle\tau,T_\pi\alpha\rangle_{\omega\nu}+\ell_\pi(\alpha).
\label{eq:proj-riesz-minimax}
\end{align}
For fixed $Q$ and $\alpha$, completing the square gives
\[
\inf_e\mathcal L_Q(e,Q)=-\tfrac12\|T_\pi Q-r\|_{2,\omega\nu}^2,
\qquad
\inf_\tau\mathcal L_w(\tau,\alpha)
=-\tfrac12\|T_\pi\alpha\|_{2,\omega\nu}^2+\ell_\pi(\alpha).
\]
The saddle-point equations are
$e=T_\pi Q-r$ and $\langle e,T_\pi h\rangle_{\omega\nu}=0$ for
the first problem, and $\tau=T_\pi\alpha$ and
$\langle\tau,T_\pi h\rangle_{\omega\nu}=\ell_\pi(h)$ for the second,
for every $h\in\mathcal H$. By
\eqref{eq:proj-q-target}--\eqref{eq:proj-normal-equations}, their unique
saddle points are $(e_{\mathcal H},Q_{\mathcal H})$ and
$(\tau_{\mathcal H},\alpha_{\mathcal H})$.

Condition on the training sample. Let $(Q_i,e_i)_{i\in[K]}$ and
$(\alpha_i,\tau_i)_{i\in[M]}$ be fitted pairs, with
$Q_i,\alpha_i\in\mathcal H$ and $e_i,\tau_i\in L^2(\omega\nu)$,
and set $w_i:=\omega\tau_i$.
Candidates may be obtained by varying regularization or optimization
settings in \eqref{eq:proj-q-minimax} and
\eqref{eq:proj-riesz-minimax}.

\subsection{Pairwise validation}
\label{app:pbv-projections-selectors}

We test the relation between the two fitted functions together with
the projection's normal equation, using differences between candidates
as critics. Let $\mathbb P_n$ denote empirical averaging over the
validation sample $\mathcal D_n=\{(X_t,R_t,S_t')\}_{t=1}^n$.
Let $\widehat\ell_\pi(h)$ equal $\ell_\pi(h)$ for exact initial
moments or $(1-\gamma)m^{-1}\sum_{t=1}^m h(X_{0,t})$ for initial
draws $X_{0,1},\ldots,X_{0,m}$ from $d_0$.
For $f\in L^2(\mu)$ and $h\in\mathcal H$, define
\begin{equation}
\begin{aligned}
\widehat{\mathcal R}_i^Q(f)
&:=\mathbb P_n[\omega(X)f(X)
\{e_i(X)-Q_i(X)+R+\gamma(\pi Q_i)(S')\}],\\
\widehat{\mathcal B}_i^Q(h)
&:=\mathbb P_n[\omega(X)e_i(X)\{h(X)-\gamma(\pi h)(S')\}],
\end{aligned}
\label{eq:proj-q-moments}
\end{equation}
and
\begin{equation}
\begin{aligned}
\widehat{\mathcal R}_i^w(f)
&:=\mathbb P_n[\omega(X)f(X)
\{\tau_i(X)-\alpha_i(X)+\gamma(\pi\alpha_i)(S')\}],\\
\widehat{\mathcal B}_i^w(h)
&:=\mathbb P_n[\omega(X)\tau_i(X)\{h(X)-\gamma(\pi h)(S')\}]
-\widehat\ell_\pi(h).
\end{aligned}
\label{eq:proj-riesz-moments}
\end{equation}
For each candidate pair, take
\[
\begin{aligned}
v_{ij}^Q&:=e_i-e_j,& h_{ij}^Q&:=Q_i-Q_j,
&&i,j\in[K],\\
v_{ij}^w&:=\tau_i-\tau_j,& h_{ij}^w&:=\alpha_i-\alpha_j,
&&i,j\in[M].
\end{aligned}
\]
Both selectors use the quadratic penalty
\begin{equation}
\begin{aligned}
\widehat{\mathcal A}(v,h)
&:=\mathbb P_n[\omega(X)v(X)^2]\\
&\quad+|\mathbb P_n[\omega(X)h(X)
\{v(X)-h(X)+\gamma(\pi h)(S')\}]|.
\end{aligned}
\label{eq:proj-empirical-penalty}
\end{equation}
For $s\in\{Q,w\}$, form the pairwise scores
\begin{equation}
\begin{aligned}
\widehat{\mathcal M}_{ij}^s
&:=|\widehat{\mathcal R}_i^s(v_{ij}^s)
+\widehat{\mathcal B}_i^s(h_{ij}^s)|
+|\widehat{\mathcal R}_i^s(h_{ij}^s)|\\
&\quad-\tfrac14\widehat{\mathcal A}(v_{ij}^s,h_{ij}^s),
\end{aligned}
\label{eq:proj-pairwise-scores}
\end{equation}
and select
\begin{equation}
\widehat i_Q:=\min\argmin_{i\in[K]}\max_{j\in[K]}
\widehat{\mathcal M}_{ij}^Q,
\qquad
\widehat i_w:=\min\argmin_{i\in[M]}\max_{j\in[M]}
\widehat{\mathcal M}_{ij}^w.
\label{eq:proj-selectors}
\end{equation}
The selected pairs are $(Q_{\widehat i_Q},e_{\widehat i_Q})$ and
$(\alpha_{\widehat i_w},\tau_{\widehat i_w})$, with occupancy weight
$w_{\widehat i_w}=\omega\tau_{\widehat i_w}$.

\subsection{Finite-sample guarantees}
\label{app:pbv-projections-guarantees}

Measure joint error relative to the projected targets by
\begin{align}
\mathcal E_Q(i)
&:=\|Q_i-Q_{\mathcal H}\|_{2,\omega\nu}^2
+\|e_i-e_{\mathcal H}\|_{2,\omega\nu}^2,
\label{eq:proj-q-error}\\
\mathcal E_w(i)
&:=\|\alpha_i-\alpha_{\mathcal H}\|_{2,\omega\nu}^2
+\|\tau_i-\tau_{\mathcal H}\|_{2,\omega\nu}^2.
\label{eq:proj-riesz-error}
\end{align}
Fix $k^\star\in\argmin_{i\in[K]}\mathcal E_Q(i)$ and
$a^\star\in\argmin_{i\in[M]}\mathcal E_w(i)$.
All conditions and probability statements below are conditional on the
training sample.

\begin{enumerate}[label=\textbf{(P\arabic*)}, ref=P\arabic*]
\item\label{cond:proj-validation}
\textit{Validation process.}
The validation sample is independent of the training sample, follows
the stationary setup of Section~\ref{sec:offline-data-setup}, and satisfies
Condition~\ref{cond:validation-spectral-gap} with spectral gap $g_b\in(0,1]$.

\item\label{cond:proj-envelopes}
\textit{Bounded weight and candidates.}
There are constants $\overline\omega<\infty$ and $B<\infty$ such that
$0\leq\omega\leq\overline\omega$ and every function in
the candidate pairs under consideration is bounded in absolute value
by $B$, $\nu$-almost surely. For $Q$-function selection, assume also
$|R|\leq R_{\max}<\infty$ almost surely.

\item\label{cond:proj-coverage}
\textit{One-step concentrability.}
The measure $\mu P_\pi$ is absolutely continuous with respect to $\mu$,
and $C_{\mu,\pi}:=\|\dd(\mu P_\pi)/\dd\mu\|_\infty<\infty$.

\item\label{cond:proj-initial}
\textit{Initial moments.}
Either $\widehat\ell_\pi=\ell_\pi$, or the initial draws are i.i.d.
from $d_0$, independent of the training and transition samples, and
$C_{0,\mu}:=\|\dd d_0/\dd\mu\|_\infty<\infty$.

\item\label{cond:proj-q-coercivity}
\textit{$Q$-comparison coercivity.}
There is $\kappa_Q>0$ such that
\[
|\langle h_{ik^\star}^Q,T_\pi h_{ik^\star}^Q\rangle_{\omega\nu}|
\geq\kappa_Q\|h_{ik^\star}^Q\|_{2,\omega\nu}^2,
\qquad i\in[K].
\]

\item\label{cond:proj-riesz-coercivity}
\textit{Representer-comparison coercivity.}
There is $\kappa_w>0$ such that
\[
|\langle h_{ia^\star}^w,T_\pi h_{ia^\star}^w\rangle_{\omega\nu}|
\geq\kappa_w\|h_{ia^\star}^w\|_{2,\omega\nu}^2,
\qquad i\in[M].
\]
\end{enumerate}

\begin{theorem}[Projected $Q$-function selection]
\label{thm:proj-q-selection}
Suppose Conditions~\ref{cond:proj-validation}--\ref{cond:proj-coverage}
and~\ref{cond:proj-q-coercivity} hold. There is a finite constant $C_Q$,
depending only on $\overline\omega$, $B$, $R_{\max}$,
$C_{\mu,\pi}$, and $\gamma$, such that, for every $\delta\in(0,1)$,
with probability at least $1-\delta$,
\begin{equation}
\sqrt{\mathcal E_Q(\widehat i_Q)}
\leq C_Q(1+\kappa_Q^{-2})
\left\{\min_{i\in[K]}\sqrt{\mathcal E_Q(i)}
+\sqrt{\frac{\log(2K/\delta)}{ng_b}}\right\}.
\label{eq:proj-q-oracle}
\end{equation}
\end{theorem}

\begin{theorem}[Projected occupancy-weight selection]
\label{thm:proj-riesz-selection}
Suppose Conditions~\ref{cond:proj-validation}--\ref{cond:proj-initial}
and~\ref{cond:proj-riesz-coercivity} hold. There is a finite constant $C_w$,
depending only on $\overline\omega$, $B$,
$C_{\mu,\pi}$, $C_{0,\mu}$, and $\gamma$, such that, for every
$\delta\in(0,1)$, with probability at least $1-\delta$,
\begin{equation}
\sqrt{\mathcal E_w(\widehat i_w)}
\leq C_w(1+\kappa_w^{-2})
\left\{\min_{i\in[M]}\sqrt{\mathcal E_w(i)}
+\sqrt{\frac{\log(2M/\delta)}{ng_b}}
+\sqrt{\frac{\log(2M/\delta)}{m}}\right\}.
\label{eq:proj-riesz-oracle}
\end{equation}
The final term and the dependence on $C_{0,\mu}$ are omitted for exact initial
moments. Moreover,
$\|w_{\widehat i_w}-w_{\mathcal H}\|_{2,\nu/\omega}^2
\leq\mathcal E_w(\widehat i_w)$, where
$\|f\|_{2,\nu/\omega}^2:=\int_{\{\omega>0\}}f^2/\omega\,\dd\nu$.
\end{theorem}

\subsection{Proofs}
\label{app:pbv-projections-proofs}

\begin{proof}[Proof of Theorems~\ref{thm:proj-q-selection}
and~\ref{thm:proj-riesz-selection}]
For the first problem, write $(p_i,f_i)=(e_i,Q_i)$; for the second,
write $(p_i,f_i)=(\tau_i,\alpha_i)$. Let $(p_*,f_*)$ denote
the corresponding target. In the proof, unhatted quantities replace
empirical averages by expectations. Suppress the superscripts $Q$ and
$w$, and set
\[
u_i:=p_i-p_*,\qquad a_i:=f_i-f_*,\qquad
E_i:=\|u_i\|^2+\|a_i\|^2,\qquad
d_{ij}:=(p_i-p_j,f_i-f_j).
\]
Throughout the proof, norms and inner products use $\omega\nu$, and
$\|(v,h)\|^2:=\|v\|^2+\|h\|^2$.
The normal equations~\eqref{eq:proj-normal-equations} imply
\begin{equation}
\mathcal R_i(v)=\langle u_i-T_\pi a_i,v\rangle,
\qquad
\mathcal B_i(h)=\langle u_i,T_\pi h\rangle.
\label{eq:proj-centered-residuals}
\end{equation}
Write
\[
\begin{aligned}
\mathcal F_i(v,h)
&:=|\mathcal R_i(v)+\mathcal B_i(h)|+|\mathcal R_i(h)|,\\
\mathcal A(v,h)
&:=\|v\|^2+|\langle v-T_\pi h,h\rangle|.
\end{aligned}
\]
For $d_{ij}=(v,h)$, subtracting the residual moments gives
\begin{equation}
\begin{aligned}
\{\mathcal R_i(v)+\mathcal B_i(h)\}
-\{\mathcal R_j(v)+\mathcal B_j(h)\}&=\|v\|^2,\\
\mathcal R_i(h)-\mathcal R_j(h)&=\langle v-T_\pi h,h\rangle.
\end{aligned}
\label{eq:proj-pairwise-identities}
\end{equation}
The triangle inequality therefore gives
\begin{equation}
\mathcal F_i(d_{ij})\geq\mathcal A(d_{ij})-\mathcal F_j(d_{ij}).
\label{eq:proj-reverse-comparison}
\end{equation}

Let $k$ index the best fitted pair and write $\kappa$ for its comparison
coercivity constant. For $d_{ik}=(v,h)$, Cauchy--Schwarz and Young's
inequality give
\[
\kappa\|h\|^2
\leq|\langle T_\pi h,h\rangle|
\leq|\langle v-T_\pi h,h\rangle|
+\frac{\|v\|^2}{2\kappa}+\frac\kappa2\|h\|^2.
\]
Since $2/\kappa\leq1+\kappa^{-2}$, rearranging yields
\begin{equation}
\mathcal A(d_{ik})\geq c\|d_{ik}\|^2,
\qquad c:=\frac{\kappa^2}{1+\kappa^2}.
\label{eq:proj-penalty-lower}
\end{equation}
The same bound holds for $d_{ki}=-d_{ik}$.
Jensen's inequality and Condition~\ref{cond:proj-coverage} give
\[
\|P_\pi h\|^2
\leq\mathbb E_{\mu P_\pi}[h^2]
\leq C_{\mu,\pi}\|h\|^2,
\]
so $\|T_\pi h\|\leq\Lambda\|h\|$, where
$\Lambda:=1+\gamma\sqrt{C_{\mu,\pi}}$.
Applying this bound to \eqref{eq:proj-centered-residuals} gives
\begin{align}
\mathcal F_i(v,h)
&\leq(\|u_i\|+\Lambda\|a_i\|)\|v\|
+\{(\Lambda+1)\|u_i\|
+\Lambda\|a_i\|\}\|h\|
\notag\\
&\leq D\sqrt{E_i}\,\|(v,h)\|,
\label{eq:proj-comparison-upper}
\end{align}
where $D:=\sqrt{3\Lambda^2+2\Lambda+2}$ bounds the coefficient
matrix's operator norm by its Frobenius norm.

Next let $N$ be the number of candidates, set
$z:=\log(48N^2/\delta)$ and $\rho_n:=\sqrt{z/(ng_b)}$, and write
$t_{ij}:=\|d_{ij}\|$.
For each pair $(i,j)$, the empirical score involves four transition
averages: the transition parts of
$\widehat{\mathcal R}_i(v)+\widehat{\mathcal B}_i(h)$ and
$\widehat{\mathcal R}_i(h)$, and the two averages in
$\widehat{\mathcal A}(v,h)$, with $(v,h)=d_{ij}$.
Each summand is bounded in absolute value by a fixed constant and by
$C\omega(X)\{|v(X)|+|h(X)|+|(\pi h)(S')|\}$.
Since $\omega^2\leq\overline\omega\,\omega$, Jensen's inequality gives
\[
\mathbb E[\omega(X)^2\{(\pi h)(S')\}^2]
\leq\overline\omega\,\mathbb E_{\mu P_\pi}[h^2]
\leq\overline\omega C_{\mu,\pi}\|h\|^2.
\]
Thus each summand has second moment at most $Ct_{ij}^2$.
Lemma~\ref{lem:trajectory-bernstein}(i) bounds the deviation of each
average by $C\{t_{ij}\rho_n+\rho_n^2\}$, with failure probability
at most $6e^{-z}$.

For empirical initial moments,
Condition~\ref{cond:proj-initial} gives
$\mathbb E_{d_0}[h^2]\leq C_{0,\mu}\|h\|^2$.
Bernstein's inequality therefore bounds each initial-moment deviation
by $C\{t_{ij}\rho_0+\rho_0^2\}$, with failure probability at most
$2e^{-z}$, where $\rho_0:=\sqrt{z/m}$.
Set $\rho_0=0$ in the $Q$-problem and for exact initial moments, and let
$\rho:=\rho_n+\rho_0$.
A union bound over the $N^2$ pairs and
$||x|-|y||\leq|x-y|$ give, simultaneously for all pairs,
\begin{equation}
|\widehat{\mathcal M}_{ij}-\mathcal M_{ij}|
\leq C\{t_{ij}\rho+\rho^2\},
\label{eq:proj-local-score-deviation}
\end{equation}
outside an event of probability at most
$(24+2)N^2e^{-z}\leq\delta$.
Here and below, $C$ depends only on the constants listed in the relevant
theorem and may increase between displays.

Work on this event and put $b:=D\sqrt{E_k}+C\rho$.
By \eqref{eq:proj-penalty-lower}--\eqref{eq:proj-local-score-deviation},
for every $j$,
\[
\widehat{\mathcal M}_{kj}
\leq b t_{kj}-\frac c4t_{kj}^2+C\rho^2
\leq\frac{b^2}{c}+C\rho^2.
\]
For the selected index $\widehat i$, put $t:=t_{\widehat i k}$.
Equations~\eqref{eq:proj-reverse-comparison}--\eqref{eq:proj-local-score-deviation}
also give
\[
\widehat{\mathcal M}_{\widehat i k}
\geq\frac{3c}{4}t^2-bt-C\rho^2.
\]
The selector minimizes the largest pairwise score, so
\[
\frac{3c}{4}t^2-bt\leq\frac{b^2}{c}+2C\rho^2.
\]
At $t=2b/c+s$, with $s\geq0$, the left side is
$b^2/c+2bs+3cs^2/4$. Hence
\[
t\leq\frac{2b}{c}+\sqrt{\frac{8C}{3c}}\,\rho.
\]
The triangle inequality and $0<c\leq1$ now imply
\begin{align*}
\sqrt{E_{\widehat i}}
&\leq\sqrt{E_k}+t
\leq\left(1+\frac{2D}{c}\right)\sqrt{E_k}+\frac{C'}c\rho\\
&\leq C''(1+\kappa^{-2})(\sqrt{E_k}+\rho).
\end{align*}
Since $\log(48N^2/\delta)\leq6\log(2N/\delta)$, substituting
$(N,k,\kappa)=(K,k^\star,\kappa_Q)$ or $(M,a^\star,\kappa_w)$
proves \eqref{eq:proj-q-oracle} and \eqref{eq:proj-riesz-oracle},
respectively. Finally,
\[
\|w_{\widehat i_w}-w_{\mathcal H}\|_{2,\nu/\omega}^2
=\|\tau_{\widehat i_w}-\tau_{\mathcal H}\|_{2,\omega\nu}^2
\leq\mathcal E_w(\widehat i_w).
\]
\end{proof}

\section{Pairwise validation for soft optimal control}
\label{app:soft-pbv}

We extend PBV to KL-regularized optimal control using occupancy weighting
and temperature annealing, building on
\citet{vanDerLaanKallus2025SoftFQI}. We first bound the error of the
selected candidate at a fixed temperature when the initializer and
candidate library lie near the soft optimal $Q$-function. We then lower
the temperature in stages, using the candidate selected by PBV at each
stage to initialize the next. Under the stated conditions on candidate
accuracy, validation sample sizes, and temperature decreases, this
procedure can start from any bounded initializer at a sufficiently high
temperature.

\subsection{The soft Bellman operator}
\label{app:soft-pbv-operator}

Consider a discounted MDP with state space $\mathcal S$, finite action
space $\mathcal A$, transition kernel $P$, discount factor
$\gamma\in(0,1)$, and rewards satisfying $|R|\leq R_{\max}$.
Write $r(s,a):=\mathbb E[R\mid S=s,A=a]$ and $X=(S,A)$.
Fix a reference policy $\pi_{\mathrm{ref}}$ with
$L_{\mathrm{ref}}:=\sup_{s,a}\log\{1/\pi_{\mathrm{ref}}(a\mid s)\}<\infty$.
For temperature $\tau>0$ and bounded $Q$, define the soft state value
and its Gibbs policy by
\begin{equation}
\begin{aligned}
V_{\tau,Q}(s)
&:=\tau\log\sum_{a\in\mathcal A}
\pi_{\mathrm{ref}}(a\mid s)e^{Q(s,a)/\tau},\\
\pi_{\tau,Q}(a\mid s)
&:=\pi_{\mathrm{ref}}(a\mid s)
\exp\{[Q(s,a)-V_{\tau,Q}(s)]/\tau\}.
\end{aligned}
\label{eq:soft-pbv-value-policy}
\end{equation}
The Gibbs policy maximizes expected $Q$ minus $\tau$ times its KL
divergence from $\pi_{\mathrm{ref}}$. The soft Bellman operator and
residual map are
\begin{equation}
(\mathcal B_\tau Q)(s,a)
:=r(s,a)+\gamma\mathbb E[V_{\tau,Q}(S')\mid S=s,A=a],
\qquad F_\tau(Q):=Q-\mathcal B_\tau Q.
\label{eq:soft-pbv-operator}
\end{equation}
The operator satisfies
$\|\mathcal B_\tau Q-\mathcal B_\tau Q'\|_\infty
\leq\gamma\|Q-Q'\|_\infty$ and has a unique bounded fixed point
$Q_\tau^\star=\mathcal B_\tau Q_\tau^\star$, with
$\|Q_\tau^\star\|_\infty\leq R_{\max}/(1-\gamma)$.
For a policy $\pi$, let
$(P_\pi h)(s,a):=\mathbb E[\sum_{a'}\pi(a'\mid S')h(S',a')\mid s,a]$.
The derivative of the soft Bellman operator is
\begin{equation}
D\mathcal B_\tau(Q)[h]=\gamma P_{\pi_{\tau,Q}}h.
\label{eq:soft-pbv-derivative}
\end{equation}
These properties follow from the log-partition representation; see
\citet[Section~2 and Lemma~S3]{vanDerLaanKallus2025SoftFQI} for proofs.

\subsection{A fixed temperature and warm start}
\label{app:soft-pbv-local}

Fix a temperature $\tau$, a bounded initializer $\bar Q$ with
$\|\bar Q-Q_\tau^\star\|_\infty\leq r_{\mathrm{loc}}$ for some
$r_{\mathrm{loc}}>0$, and an initial state--action distribution $d_0$.
Let $\bar\pi:=\pi_{\tau,\bar Q}$ and
let $\mu$ be its normalized discounted occupancy measure:
\begin{equation}
\mu:=d_{\bar\pi,\gamma}
:=(1-\gamma)\sum_{t=0}^\infty\gamma^t d_0P_{\bar\pi}^t,
\qquad
\mu=(1-\gamma)d_0+\gamma\mu P_{\bar\pi}.
\label{eq:soft-pbv-occupancy}
\end{equation}
The adjoint Bellman equation in \eqref{eq:soft-pbv-occupancy} gives
contraction of $\gamma P_{\bar\pi}$ in $L^2(\mu)$; see
\citet[Lemma~S2]{vanDerLaanKallus2025SoftFQI}.

\begin{proposition}[Local contraction and coercivity]
\label{prop:soft-pbv-local}
For bounded $Q,Q'$ satisfying
$\max\{\|Q-\bar Q\|_\infty,\|Q'-\bar Q\|_\infty\}\leq r_{\mathrm{loc}}$,
set $L:=\sqrt\gamma\,e^{r_{\mathrm{loc}}/\tau}$. Then
\begin{align}
\|\mathcal B_\tau Q-\mathcal B_\tau Q'\|_{2,\mu}
&\leq L\|Q-Q'\|_{2,\mu},
\label{eq:soft-pbv-local-contraction}\\
\langle Q-Q',F_\tau(Q)-F_\tau(Q')\rangle_\mu
&\geq(1-L)\|Q-Q'\|_{2,\mu}^2.
\label{eq:soft-pbv-local-coercivity}
\end{align}
In particular, the coercivity constant is positive when
$r_{\mathrm{loc}}<\frac\tau2\log(1/\gamma)$.
\end{proposition}

\subsection{Pairwise validation at a fixed temperature}
\label{app:soft-pbv-selector}

Condition on the training data and use the temperature $\tau$,
initializer $\bar Q$, and occupancy measure $\mu$ from
\eqref{eq:soft-pbv-occupancy}. Assume $\mu\ll\nu$ and set
$\omega:=\dd\mu/\dd\nu$. Let $Q_1,\ldots,Q_K$ be fitted candidates.
Let $\mathbb P_n$ average over held-out transitions
$\{(X_t,R_t,S_t')\}_{t=1}^n$, where $X_t$ has marginal law $\nu$.
For $h_{ij}:=Q_i-Q_j$, define
\begin{equation}
\begin{aligned}
\widehat{\mathcal M}_{ij}^{\tau}
&:=\left|\mathbb P_n\!\left[
\omega(X)h_{ij}(X)
\{Q_i(X)-R-\gamma V_{\tau,Q_i}(S')\}\right]\right|\\
&\quad-\frac14\left|\mathbb P_n\!\left[
\omega(X)h_{ij}(X)
\{h_{ij}(X)-\gamma(V_{\tau,Q_i}-V_{\tau,Q_j})(S')\}\right]\right|,\\
\widehat i_\tau
&:=\min\argmin_{i\in[K]}\max_{j\in[K]}
\widehat{\mathcal M}_{ij}^{\tau}.
\end{aligned}
\label{eq:soft-pbv-selector}
\end{equation}
The first moment tests the soft Bellman equation against the candidate
difference. The penalty uses the difference between the two soft
Bellman residuals, so both terms are computed from the same fitted
functions and held-out transitions.

Let $k^\star\in\argmin_{i\in[K]}\|Q_i-Q_\tau^\star\|_{2,\mu}$.
We impose the following conditions, conditional on training.
\begin{enumerate}[label=\textbf{(SC\arabic*)}, ref=SC\arabic*]
\item\label{cond:soft-pbv-validation}
\textit{Validation process.}
The validation sample is independent of the data used to construct
the candidates, temperature, and weight. It follows the stationary
validation setup of Section~\ref{sec:offline-data-setup} and satisfies
Condition~\ref{cond:validation-spectral-gap} with spectral gap $g_b$.

\item\label{cond:soft-pbv-bounds}
\textit{Bounded candidates and weight.}
There are finite constants $B,W$ such that
$\max_{i\in[K]}\|Q_i\|_\infty\leq B$ and
$0\leq\omega\leq W$, $\nu$-almost surely.

\item\label{cond:soft-pbv-neighborhood}
\textit{Warm start and local library.}
For a radius $0<r_{\mathrm{loc}}<\frac\tau2\log(1/\gamma)$,
\[
\|\bar Q-Q_\tau^\star\|_\infty\leq r_{\mathrm{loc}},
\qquad
\max_{i\in[K]}\|Q_i-\bar Q\|_\infty\leq r_{\mathrm{loc}}.
\]
\end{enumerate}

\begin{theorem}[Oracle bound at a fixed temperature]
\label{thm:soft-pbv-local}
Under Conditions~\ref{cond:soft-pbv-validation}--\ref{cond:soft-pbv-neighborhood},
set $L:=\sqrt\gamma e^{r_{\mathrm{loc}}/\tau}<1$.
There is a finite constant $C$, depending only on $B,W,R_{\max}$ and
$\gamma$, such that, for every $\delta\in(0,1)$, with conditional
probability at least $1-\delta$,
\begin{equation}
\|Q_{\widehat i_\tau}-Q_\tau^\star\|_{2,\mu}
\leq\frac{3+L}{1-L}\min_{i\in[K]}\|Q_i-Q_\tau^\star\|_{2,\mu}
+\frac{C}{1-L}\sqrt{\frac{\log(2K/\delta)}{ng_b}}.
\label{eq:soft-pbv-local-oracle}
\end{equation}
\end{theorem}

\subsection{Stagewise validation and temperature annealing}
\label{app:soft-pbv-annealing}

Fix $\delta\in(0,1)$, temperatures $\tau_0>\cdots>\tau_J>0$, and a constant
$0<c<\frac12\log(1/\gamma)$.
At stage $j$, use an initializer $\bar Q_j$, its policy
$\pi_{\tau_j,\bar Q_j}$, and the occupancy measure
$\mu_j:=d_{\pi_{\tau_j,\bar Q_j},\gamma}$.
Keep this measure and temperature fixed while selecting among
$Q_{j,1},\ldots,Q_{j,K_j}$, and require
\begin{equation}
\max_{i\in[K_j]}\|Q_{j,i}-\bar Q_j\|_\infty\leq c\tau_j.
\label{eq:soft-pbv-stage-library}
\end{equation}
The selected function $\widehat Q_j$ becomes the next initializer,
$\bar Q_{j+1}:=\widehat Q_j$. Use a fresh validation sample at each
stage, independent of the preceding stages and the current fitting
data, with spectral gap at least $g_b$ and $\mu_j=\omega_j\nu$.
Assume the candidate and weight bounds
in Condition~\ref{cond:soft-pbv-bounds} hold uniformly over stages.

Write $L_0:=\sqrt\gamma e^c<1$ and let
\begin{equation}
\varepsilon_j
:=\frac{3+L_0}{1-L_0}
\min_{i\in[K_j]}\|Q_{j,i}-Q_{\tau_j}^\star\|_{2,\mu_j}
+\frac{C}{1-L_0}
\sqrt{\frac{\log\{2K_j(J+1)/\delta\}}{n_jg_b}},
\label{eq:soft-pbv-stage-bound}
\end{equation}
where $n_j$ is the validation sample size at stage $j$ and $C$ is
the constant from Theorem~\ref{thm:soft-pbv-local}.
To transfer these bounds between temperatures, suppose there are
$C_0<\infty$ and $\alpha\in(0,1]$ such that
\begin{equation}
\|Q_{j,i}-Q_{\tau_j}^\star\|_\infty
\leq C_0\|Q_{j,i}-Q_{\tau_j}^\star\|_{2,d_0}^{\alpha}
\qquad (i\in[K_j],\ 0\leq j\leq J).
\label{eq:soft-pbv-interpolation}
\end{equation}
Equation~\eqref{eq:soft-pbv-occupancy} gives
$\mu_j\geq(1-\gamma)d_0$, so a stage
error of $\varepsilon_j$ in $L^2(\mu_j)$ gives a supremum-norm error
of at most $C_0(1-\gamma)^{-\alpha/2}\varepsilon_j^\alpha$.
By \citet[Lemma~S6]{vanDerLaanKallus2025SoftFQI}, the soft optimal
$Q$-functions at two temperatures satisfy
\begin{equation}
\|Q_\tau^\star-Q_{\tau'}^\star\|_\infty
\leq\frac{\gamma L_{\mathrm{ref}}}{1-\gamma}|\tau-\tau'|.
\label{eq:soft-pbv-temperature-path}
\end{equation}

Assume the following bounds hold throughout the stage construction.
\begin{enumerate}[label=\textbf{(AN\arabic*)}, ref=AN\arabic*]
\item\label{cond:soft-pbv-annealing-initial}
\textit{Initial temperature.}
The first temperature satisfies
\[
c\tau_0\geq\|\bar Q_0\|_\infty
+\max\{B,R_{\max}/(1-\gamma)\}.
\]

\item\label{cond:soft-pbv-annealing-transfer}
\textit{Transfer between temperatures.}
For every $0\leq j<J$,
\begin{equation}
C_0(1-\gamma)^{-\alpha/2}\varepsilon_j^\alpha
+\frac{\gamma L_{\mathrm{ref}}}{1-\gamma}(\tau_j-\tau_{j+1})
\leq c\tau_{j+1}.
\label{eq:soft-pbv-annealing-transfer}
\end{equation}
\end{enumerate}

\begin{theorem}[PBV under temperature annealing]
\label{thm:soft-pbv-annealing}
Under Conditions~\ref{cond:soft-pbv-validation}--\ref{cond:soft-pbv-bounds}
at every stage, the library and interpolation bounds
\eqref{eq:soft-pbv-stage-library} and~\eqref{eq:soft-pbv-interpolation}, and
Conditions~\ref{cond:soft-pbv-annealing-initial}--\ref{cond:soft-pbv-annealing-transfer},
with probability at least $1-\delta$, simultaneously for
$0\leq j\leq J$,
\begin{equation}
\|\bar Q_j-Q_{\tau_j}^\star\|_\infty\leq c\tau_j,
\qquad
\|\widehat Q_j-Q_{\tau_j}^\star\|_{2,\mu_j}\leq\varepsilon_j,
\label{eq:soft-pbv-annealing-guarantee}
\end{equation}
where $\varepsilon_j$ is defined in \eqref{eq:soft-pbv-stage-bound}.
\end{theorem}

\subsection{Proofs}
\label{app:soft-pbv-proofs}

\begin{proof}[Proof of Proposition~\ref{prop:soft-pbv-local}]
Let $h:=Q-Q'$ and $Q_t:=Q'+th$ for $0\leq t\leq1$.
The segment satisfies $\|Q_t-\bar Q\|_\infty\leq r_{\mathrm{loc}}$.
The log-partition formula gives
$\|V_{\tau,Q_t}-V_{\tau,\bar Q}\|_\infty
\leq\|Q_t-\bar Q\|_\infty\leq r_{\mathrm{loc}}$.
The Gibbs formula
\eqref{eq:soft-pbv-value-policy} therefore implies
$\pi_{\tau,Q_t}\leq e^{2r_{\mathrm{loc}}/\tau}\bar\pi$
pointwise. Using the adjoint Bellman equation
\eqref{eq:soft-pbv-occupancy}, we obtain
\[
\mu P_{\pi_{\tau,Q_t}}
\leq e^{2r_{\mathrm{loc}}/\tau}\mu P_{\bar\pi}
\leq\gamma^{-1}e^{2r_{\mathrm{loc}}/\tau}\mu.
\]
Conditional Jensen's inequality gives
\[
\|\gamma P_{\pi_{\tau,Q_t}}h\|_{2,\mu}^2
\leq\gamma^2\int h^2\,\dd(\mu P_{\pi_{\tau,Q_t}})
\leq\gamma e^{2r_{\mathrm{loc}}/\tau}\|h\|_{2,\mu}^2.
\]
Integrating the derivative in \eqref{eq:soft-pbv-derivative} along
$Q_t$ now yields
\[
\|\mathcal B_\tau Q-\mathcal B_\tau Q'\|_{2,\mu}
\leq\int_0^1\|\gamma P_{\pi_{\tau,Q_t}}h\|_{2,\mu}\,\dd t
\leq L\|h\|_{2,\mu},
\]
which proves \eqref{eq:soft-pbv-local-contraction}.
Finally, Cauchy--Schwarz gives
\[
\langle h,F_\tau(Q)-F_\tau(Q')\rangle_\mu
=\|h\|_{2,\mu}^2
-\langle h,\mathcal B_\tau Q-\mathcal B_\tau Q'\rangle_\mu
\geq(1-L)\|h\|_{2,\mu}^2,
\]
proving \eqref{eq:soft-pbv-local-coercivity}.
\end{proof}

\begin{proof}[Proof of Theorem~\ref{thm:soft-pbv-local}]
We condition on the training data and use $\mu$ for all norms and
inner products. Let $k:=k^\star$, $\kappa:=1-L$, and
$t_{ij}:=\|Q_i-Q_j\|$. Define the population penalty and score by
\[
a_{ij}:=|\langle h_{ij},F_\tau(Q_i)-F_\tau(Q_j)\rangle|,
\qquad
\mathcal M_{ij}^{\tau}
:=|\langle h_{ij},F_\tau(Q_i)\rangle|-\tfrac14a_{ij}.
\]
Proposition~\ref{prop:soft-pbv-local} and
Condition~\ref{cond:soft-pbv-neighborhood} give
$a_{ij}\geq\kappa t_{ij}^2$ and
$\|F_\tau(Q_k)\|\leq(1+L)\|Q_k-Q_\tau^\star\|$.
Cauchy--Schwarz gives the first bound below. For the second, write
$F_\tau(Q_i)=F_\tau(Q_k)+\{F_\tau(Q_i)-F_\tau(Q_k)\}$
and apply the triangle inequality:
\begin{align}
\mathcal M_{kj}^{\tau}
&\leq\|F_\tau(Q_k)\|t_{kj}-\tfrac\kappa4t_{kj}^2,
\label{eq:soft-pbv-population-score-upper}\\
\mathcal M_{ik}^{\tau}
&\geq\tfrac{3\kappa}4t_{ik}^2-\|F_\tau(Q_k)\|t_{ik}.
\label{eq:soft-pbv-population-score-lower}
\end{align}

Set $z:=\log(24K^2/\delta)$ and $\rho:=\sqrt{z/(ng_b)}$.
Below, $C$ denotes a finite constant depending only on
$B,W,R_{\max},\gamma$, and its value may increase between bounds.
Since $|V_{\tau,Q_i}|\leq B$, each transition summand in
\eqref{eq:soft-pbv-selector} is bounded in absolute value by $C$
and by $C\omega(X)|h_{ij}(X)|$.
Its second moment is at most $Ct_{ij}^2$, because
$\omega^2\leq W\omega$.
Applying Lemma~\ref{lem:trajectory-bernstein}(i) to the two averages
for each of the $K^2$ pairs gives a total failure probability at most
$12K^2e^{-z}\leq\delta$.
Using $||x|-|y||\leq|x-y|$, we therefore obtain, simultaneously for
all pairs,
\begin{equation}
|\widehat{\mathcal M}_{ij}^{\tau}-\mathcal M_{ij}^{\tau}|
\leq C(t_{ij}\rho+\rho^2).
\label{eq:soft-pbv-score-deviation}
\end{equation}

On this event, put $b:=\|F_\tau(Q_k)\|+C\rho$ and
$t:=t_{\widehat i_\tau k}$.
Equations~\eqref{eq:soft-pbv-population-score-upper}
and~\eqref{eq:soft-pbv-score-deviation} imply
\[
\max_j\widehat{\mathcal M}_{kj}^{\tau}
\leq\max_{s\geq0}\{bs-\kappa s^2/4\}+C\rho^2
=\frac{b^2}{\kappa}+C\rho^2.
\]
By \eqref{eq:soft-pbv-population-score-lower}
and~\eqref{eq:soft-pbv-score-deviation},
\[
\widehat{\mathcal M}_{\widehat i_\tau k}^{\tau}
\geq\frac{3\kappa}4t^2-bt-C\rho^2.
\]
The definition of $\widehat i_\tau$ in
\eqref{eq:soft-pbv-selector} gives
\[
\widehat{\mathcal M}_{\widehat i_\tau k}^{\tau}
\leq\max_j\widehat{\mathcal M}_{\widehat i_\tau j}^{\tau}
\leq\max_j\widehat{\mathcal M}_{kj}^{\tau}.
\]
Combining these inequalities yields
\[
\frac{3\kappa}4t^2-bt
\leq\frac{b^2}{\kappa}+2C\rho^2.
\]
For $s\geq0$, substituting $t=2b/\kappa+s$ makes the left side
$b^2/\kappa+2bs+3\kappa s^2/4$. Hence
\[
t\leq\frac{2b}{\kappa}
+\sqrt{\frac{8C}{3\kappa}}\,\rho
\leq\frac{2\|F_\tau(Q_k)\|}{\kappa}
+\frac{C}{\kappa}\rho,
\]
where the last bound uses $0<\kappa\leq1$.
The triangle inequality now gives
\begin{align*}
\|Q_{\widehat i_\tau}-Q_\tau^\star\|
&\leq\|Q_k-Q_\tau^\star\|+t\\
&\leq\frac{3+L}{1-L}\|Q_k-Q_\tau^\star\|
+\frac{C}{1-L}\rho.
\end{align*}
Since $\log(24K^2/\delta)\leq5\log(2K/\delta)$, increasing $C$
proves \eqref{eq:soft-pbv-local-oracle}.
\end{proof}

\begin{proof}[Proof of Theorem~\ref{thm:soft-pbv-annealing}]
For each stage $j$, let $\mathcal G_j$ be the sigma-field generated
by all data available before its validation sample is drawn, and define
\[
E_j:=\{\|\widehat Q_j-Q_{\tau_j}^\star\|_{2,\mu_j}
\leq\varepsilon_j\}.
\]
Condition~\ref{cond:soft-pbv-annealing-initial} gives
\[
\|\bar Q_0-Q_{\tau_0}^\star\|_\infty
\leq\|\bar Q_0\|_\infty+\frac{R_{\max}}{1-\gamma}
\leq c\tau_0.
\]
Whenever $\|\bar Q_j-Q_{\tau_j}^\star\|_\infty\leq c\tau_j$,
\eqref{eq:soft-pbv-stage-library} verifies
Condition~\ref{cond:soft-pbv-neighborhood} with radius $c\tau_j$.
The fresh validation sample is independent of $\mathcal G_j$, so
Theorem~\ref{thm:soft-pbv-local}, applied with failure probability
$\delta/(J+1)$, gives
\[
\mathbb P(E_j^c\mid\mathcal G_j)\leq\frac{\delta}{J+1}
\]
on this event. The error bound in that theorem is
$\varepsilon_j$ as defined in \eqref{eq:soft-pbv-stage-bound}.

For $j<J$, if $E_j$ holds, then
\eqref{eq:soft-pbv-interpolation} and
$\mu_j\geq(1-\gamma)d_0$ imply
\[
\|\widehat Q_j-Q_{\tau_j}^\star\|_\infty
\leq C_0(1-\gamma)^{-\alpha/2}\varepsilon_j^\alpha.
\]
Since $\bar Q_{j+1}=\widehat Q_j$, the temperature-path bound
\eqref{eq:soft-pbv-temperature-path} and
Condition~\ref{cond:soft-pbv-annealing-transfer} give
\begin{align*}
\|\bar Q_{j+1}-Q_{\tau_{j+1}}^\star\|_\infty
&\leq\|\widehat Q_j-Q_{\tau_j}^\star\|_\infty
+\|Q_{\tau_j}^\star-Q_{\tau_{j+1}}^\star\|_\infty\\
&\leq C_0(1-\gamma)^{-\alpha/2}\varepsilon_j^\alpha
+\frac{\gamma L_{\mathrm{ref}}}{1-\gamma}(\tau_j-\tau_{j+1})\\
&\leq c\tau_{j+1}.
\end{align*}
Thus the warm-start bound holds at stage $j$ whenever
$E_0,\ldots,E_{j-1}$ hold. Taking expectations of the conditional
failure bounds yields
\[
\mathbb P\left(\bigcup_{j=0}^J E_j^c\right)
=\mathbb P(E_0^c)
+\sum_{j=1}^J
\mathbb P\left(E_j^c\cap\bigcap_{\ell=0}^{j-1}E_\ell\right)
\leq\delta.
\]
On $\bigcap_{j=0}^J E_j$, both bounds in
\eqref{eq:soft-pbv-annealing-guarantee} hold at every stage.
\end{proof}

\input{sections/experiments_details.tex}

\input{sections/occupancy_experiments.tex}

\input{sections/gaussian_kernel_experiment.tex}

% Keep the contents pages balanced: results first, then proofs and extensions.
\addtocontents{toc}{\protect\newpage\protect\section*{Appendix contents (continued)}}
\section{Shared concentration bounds}
\label{app:trajectory-concentration}

The dependence argument used for all three selectors combines Bernstein
concentration for bounded additive functionals of a Markov chain with
martingale concentration for the centered one-step transition term. Let
$\mathcal F_t:=\sigma(X_1,R_1,S_1',\ldots,X_{t-1},R_{t-1},S_{t-1}',X_t)$.
Under the validation setup, conditional on $\mathcal F_t$, the pair
$(R_t,S_t')$ follows the MDP reward and transition law given $X_t$.

\begin{lemma}[Trajectory Bernstein bounds]
\label{lem:trajectory-bernstein}
Let $U_1,\ldots,U_n$ be a stationary Markov chain with invariant law $\mu$
and transition operator $P_U$. Suppose, for some $g\in(0,1]$,
$\|P_Uf\|_{2,\mu}\le(1-g)\|f\|_{2,\mu}$ for every
centered real-valued $f\in L^2(\mu)$; alternatively, $P_U$ is self-adjoint
and $\langle f,P_Uf\rangle_\mu\le(1-g)\|f\|_{2,\mu}^2$ for every such $f$. Let
$(\mathcal F_t)_{t=1}^{n+1}$ be a filtration such that $U_t$ is
$\mathcal F_t$-measurable and $Y_t$ is $\mathcal F_{t+1}$-measurable. Suppose
there is a Markov kernel $K_Y$, not depending on $t$, such that
$\mathcal L(Y_t\mid\mathcal F_t)=K_Y(U_t,\cdot)$ for every $t$.
For $u\ge1$, set $r_g(u):=\sqrt{u/(ng)}$.

\begin{enumerate}[label=(\roman*)]
\item If $|Y_t|\le B$ almost surely and $\mathbb E(Y_t^2)\le v^2$, then, with
probability at least $1-6e^{-u}$,
\begin{equation}
\left|
\frac1n\sum_{t=1}^nY_t-\mathbb E Y_t
\right|
\le
C_{\mathrm M}\{v r_g(u)+B r_g(u)^2\}.
\label{eq:app-trajectory-bernstein-bounded}
\end{equation}

\item Suppose
$Y_t=a(U_t)\{d(U_t)+\zeta_t\}$, where
$|a|\le B$, $|d|\le D$,
$\mathbb E_\mu(a^2)\le v^2$, $\sigma>0$, and
\begin{equation}
\mathbb E(\zeta_t\mid\mathcal F_t)=0,
\qquad
\mathbb E\!\left[
\left.
\exp\left\{\frac{|\zeta_t|}{\sigma}\right\}
\right|\mathcal F_t
\right]
\le2.
\label{eq:app-trajectory-subgaussian-innovation}
\end{equation}
Then, with probability at least $1-6e^{-u}$,
\begin{equation}
\left|
\frac1n\sum_{t=1}^nY_t-\mathbb E Y_t
\right|
\le
C_{\mathrm M}(D+\sigma)
\{v r_g(u)+B r_g(u)^2\}.
\label{eq:app-trajectory-bernstein-subgaussian}
\end{equation}
\end{enumerate}
Here $C_{\mathrm M}>0$ is a universal constant.
\end{lemma}

\begin{proof}[Proof of Lemma~\ref{lem:trajectory-bernstein}]
We use the following consequence of Theorem~1 of
\citet{jiangSunFan2025Bernstein}, or their Theorem~2 under the
self-adjoint alternative. If $f$ is a bounded, time-independent
function and
$\mathbb E_\mu f=0$, then, for a universal constant $C>0$,
\begin{equation}
\left|
\frac1n\sum_{t=1}^nf(U_t)
\right|
\le
C\left\{
\|f\|_{2,\mu}\sqrt{\frac{u}{ng}}
+
\|f\|_\infty\frac{u}{ng}
\right\}
\label{eq:app-markov-bernstein}
\end{equation}
outside an event of probability $2e^{-u}$.
For the proof, also write $r_1(u):=\sqrt{u/n}$.

For part~(i), let $m(x):=\int y\,K_Y(x,\dd y)$ and
$\eta_t=Y_t-m(U_t)$. The centered function
$m-\mathbb E Y_t$ has variance at most $v^2$ and absolute value at most $2B$.
Thus, by \eqref{eq:app-markov-bernstein}, its empirical average is bounded by
$C\{v r_g(u)+B r_g(u)^2\}$. Next, define
$q(x):=\int\{y-m(x)\}^2K_Y(x,\dd y)$. By conditional variance contraction,
\[
0\le q\le B^2,
\qquad
\mathbb E_\mu q\le v^2,
\qquad
\operatorname{Var}_\mu(q)\le B^2v^2.
\]
By \eqref{eq:app-markov-bernstein}, applied to
$q-\mathbb E_\mu q$, outside an event of probability $2e^{-u}$,
\begin{equation}
\frac1n\sum_{t=1}^nq(U_t)
\le
C\{v+B r_g(u)\}^2.
\label{eq:app-trajectory-variance-process}
\end{equation}
The sequence $\{\eta_t\}$ is a martingale difference with conditional
variance process $\{q(U_t)\}$ and $|\eta_t|\le2B$. By the two-sided form of
Freedman's inequality \citep{freedman1975tail} and
\eqref{eq:app-trajectory-variance-process},
\[
\left|\frac1n\sum_{t=1}^n\eta_t\right|
\le
C\{v r_1(u)+B r_1(u)r_g(u)+B r_1(u)^2\}
\]
outside an event of probability $4e^{-u}$. Combining this bound with the
bound for the predictable component gives
\[
C\{v r_g(u)+B r_g(u)^2
+v r_1(u)+B r_1(u)r_g(u)+B r_1(u)^2\}.
\]
Since $r_1(u)\le r_g(u)$ for $g\le1$,
\eqref{eq:app-trajectory-bernstein-bounded} follows.

For part~(ii), apply \eqref{eq:app-markov-bernstein} first to the centered
function $ad$. Its variance is at most $D^2v^2$ and its absolute value is at
most $2BD$, so its empirical average is bounded by
$CD\{v r_g(u)+B r_g(u)^2\}$. Apply
\eqref{eq:app-markov-bernstein} a second time to
$a^2-\mathbb E_\mu(a^2)$. Since
$\mathbb E_\mu(a^4)\le B^2v^2$,
\begin{equation}
\frac1n\sum_{t=1}^na(U_t)^2
\le
C\{v+B r_g(u)\}^2
\label{eq:app-trajectory-multiplier-process}
\end{equation}
outside an event of probability $2e^{-u}$. Expanding the exponential in
\eqref{eq:app-trajectory-subgaussian-innovation} gives
$\mathbb E(|\zeta_t|^k\mid\mathcal F_t)\le2k!\sigma^k$ for every
integer $k\ge2$. Since $|a|^k\le a^2B^{k-2}$,
\[
\mathbb E\!\left(
|a(U_t)\zeta_t|^k
\,\middle|\,\mathcal F_t
\right)
\le
\frac{k!}{2}
C^2\sigma^2a(U_t)^2(C\sigma B)^{k-2}.
\]
By the martingale Bernstein inequality under this conditional moment condition
\citep{deLaPena1999general} and
\eqref{eq:app-trajectory-multiplier-process},
\[
\left|\frac1n\sum_{t=1}^na(U_t)\zeta_t\right|
\le
C\sigma\{v r_1(u)+B r_1(u)r_g(u)+B r_1(u)^2\}
\]
outside an event of probability $4e^{-u}$. Combining this bound with the
bound for the predictable component gives
\[
C\left[
D\{v r_g(u)+B r_g(u)^2\}
+\sigma\{v r_1(u)+B r_1(u)r_g(u)+B r_1(u)^2\}
\right].
\]
Because $r_1(u)\le r_g(u)$,
\eqref{eq:app-trajectory-bernstein-subgaussian} follows.
\end{proof}

\section{Proofs for A-PBV}
\label{app:w-model-selection-proofs}

\subsection{Generalized-KL contraction and adjoint residual identities}
\label{app:w-generalized-kl}

The following generalized-KL contraction was established in the FORE analysis
of \citet{van2026fitted}.

\begin{lemma}[Generalized-KL contraction of the adjoint update]
\label{lem:w-generalized-kl-contraction}
Let $w$ and $\widetilde w$ be nonnegative integrable functions whose adjoint
updates exist as $\nu$-densities. Then
\begin{equation}
\mathsf D_\nu(
\mathcal B_\pi^\ast w
\|
\mathcal B_\pi^\ast\widetilde w)
\le
\gamma\mathsf D_\nu(w\|\widetilde w),
\label{eq:app-w-generalized-kl-contraction}
\end{equation}
whenever the divergence on the right is finite.
\end{lemma}

\begin{proof}[Proof of Lemma~\ref{lem:w-generalized-kl-contraction}]
Write $\mathsf D$ for generalized KL between finite nonnegative measures.
This divergence is jointly convex and obeys the data-processing inequality
for Markov kernels. Let
$\mu=w\nu$ and $\widetilde\mu=\widetilde w\nu$. By
\eqref{eq:w-adjoint-update}, joint convexity, and data processing,
\begin{align*}
\mathsf D_\nu(\mathcal B_\pi^\ast w\|\mathcal B_\pi^\ast\widetilde w)
&=
\mathsf D\!\left(
(1-\gamma)d_0+\gamma\mu P_\pi
\,\middle\|\,
(1-\gamma)d_0+\gamma\widetilde\mu P_\pi
\right)\\
&\le
(1-\gamma)\mathsf D(d_0\|d_0)
+\gamma\mathsf D(\mu P_\pi\|\widetilde\mu P_\pi)\\
&\le
\gamma\mathsf D(\mu\|\widetilde\mu)
=
\gamma\mathsf D_\nu(w\|\widetilde w),
\end{align*}
which is \eqref{eq:app-w-generalized-kl-contraction}.
\end{proof}

\begin{lemma}[Generalized-KL three-point identity for the adjoint residual]
\label{lem:w-three-point}
Let $w$ and $v$ be nonnegative integrable functions that are positive on a
common measurable support and vanish $\nu$-almost surely off that support.
Set $\log(w/v)=0$ where both functions vanish, and suppose
$w^+:=\mathcal B_\pi^\ast w$ exists as a nonnegative $\nu$-density. If every
divergence in \eqref{eq:app-w-three-point} is finite, then
\begin{equation}
\mathbb E_\nu\!\left[
(w-w^+)\log\frac{w}{v}
\right]
=
\mathsf D_\nu(w\|v)
-
\mathsf D_\nu(w^+\|v)
+
\mathsf D_\nu(w^+\|w).
\label{eq:app-w-three-point}
\end{equation}
In particular, if $v=w^\pi$, then
\begin{equation}
\mathcal S^\star(w)
\ge
(1-\gamma)\mathsf D_\nu(w\|w^\pi).
\label{eq:app-w-oracle-lower}
\end{equation}
\end{lemma}

\begin{proof}[Proof of Lemma~\ref{lem:w-three-point}]
Finiteness of $\mathsf D_\nu(w^+\|v)$ implies that $w^+$ vanishes off the common
support of $w$ and $v$. Every integrand below therefore vanishes there.
Expanding the three generalized-KL terms and canceling their linear parts
gives
\begin{equation}
\begin{aligned}
\mathsf D_\nu(w\|v)
-
\mathsf D_\nu(w^+\|v)
+
\mathsf D_\nu(w^+\|w)
&=
\mathbb E_\nu\!\left[
w\log\frac{w}{v}
-w^+\log\frac{w^+}{v}
+w^+\log\frac{w^+}{w}
\right]\\
&=
\mathbb E_\nu\!\left[
(w-w^+)\log\frac{w}{v}
\right].
\end{aligned}
\label{eq:app-w-three-point-expansion}
\end{equation}
The final expression in \eqref{eq:app-w-three-point-expansion} is the
left-hand side of \eqref{eq:app-w-three-point}, so that identity follows. To
derive \eqref{eq:app-w-oracle-lower}, set $v=w^\pi$. By
\eqref{eq:w-adjoint-moment} and \eqref{eq:w-oracle-score}, this expression is
$\mathcal S^\star(w)$. Now use
$\mathcal B_\pi^\ast w^\pi=w^\pi$ in
Lemma~\ref{lem:w-generalized-kl-contraction}. By that lemma, the middle
divergence in \eqref{eq:app-w-three-point} is at most
$\gamma\mathsf D_\nu(w\|w^\pi)$, while the final divergence is nonnegative.
Thus, \eqref{eq:app-w-oracle-lower} follows.
\end{proof}

\subsection{Target-relative ratio bounds}
\label{app:w-relative-ratio-geometry}

Write
\[
e_i:=\sqrt{\mathsf D_\nu(w_i\|w^\pi)},
\qquad
H_{ij}:=H_\nu(w_i,w_j),
\]
throughout the remainder of this appendix.

\begin{lemma}[Relative generalized-KL inequalities]
\label{lem:w-scalar-generalized-kl}
Let $\Lambda\ge1$. For every
$z\in[\Lambda^{-1},\Lambda]$,
\begin{equation}
(z-1)^2
\le
2\Lambda\{z\log z-z+1\}.
\label{eq:app-w-kl-l2-scalar}
\end{equation}
For $z,\widetilde z\in[\Lambda^{-1},\Lambda]$,
\begin{equation}
|\log z-\log\widetilde z|
\le
2\sqrt{\Lambda}
|\sqrt z-\sqrt{\widetilde z}|.
\label{eq:app-w-log-hellinger-scalar}
\end{equation}
Finally, for $p,q\ge0$,
\begin{equation}
(\sqrt p-\sqrt q)^2
\le
p\log\frac{p}{q}-p+q
\label{eq:app-w-hellinger-kl-scalar}
\end{equation}
whenever the right-hand side is finite.
\end{lemma}

\begin{proof}[Proof of Lemma~\ref{lem:w-scalar-generalized-kl}]
The function $\phi(z)=z\log z-z+1$ satisfies
$\phi''(z)=1/z\ge1/\Lambda$ on
$[\Lambda^{-1},\Lambda]$. By the strong-convexity inequality at $z=1$,
\eqref{eq:app-w-kl-l2-scalar} holds. By the mean-value theorem applied to $\log u$
on $u\in[\Lambda^{-1/2},\Lambda^{1/2}]$, together with
$\log z-\log\widetilde z=2(\log\sqrt z-\log\sqrt{\widetilde z})$,
\eqref{eq:app-w-log-hellinger-scalar} holds.

For \eqref{eq:app-w-hellinger-kl-scalar}, set $t=p/q$ when $q>0$. After
division by $q$, the difference between the right- and left-hand sides is
\[
t\log t-2t+2\sqrt t
=
2\sqrt t
\{\sqrt t\log\sqrt t-\sqrt t+1\}
\ge0.
\]
When $q>0$, the case $p=0$ follows by continuity. When $q=0$, finiteness of
the right-hand side requires $p=0$ under the convention in
\eqref{eq:w-generalized-kl}, and both sides vanish.
\end{proof}

\begin{lemma}[Occupancy-ratio plug-in value bound]
\label{lem:w-plugin-value}
Suppose Condition~\ref{cond:w-relative-ratios} holds and
$r\in L^2(d_{\pi,\gamma})$. Then, for every $i\in\{1,\ldots,M\}$,
\begin{equation}
\left|
\frac{1}{1-\gamma}
\mathbb E_\nu\!\left[w_i(X)r(X)\right]
-
V^\pi
\right|
\le
\frac{\sqrt{2\Lambda_w}}{1-\gamma}
\|r\|_{2,d_{\pi,\gamma}}
\sqrt{\mathsf D_\nu(w_i\|w^\pi)}.
\label{eq:app-w-plugin-value}
\end{equation}
\end{lemma}

\begin{proof}[Proof of Lemma~\ref{lem:w-plugin-value}]
Fix $i\in\{1,\ldots,M\}$ and set $z_i=w_i/w^\pi$ on
$\mathcal X^\pi$. By Condition~\ref{cond:w-relative-ratios}, both
$w_i$ and $w^\pi$ vanish off $\mathcal X^\pi$. By the occupancy
representation of $V^\pi$ and the identity
$d_{\pi,\gamma}=w^\pi\nu$,
\[
\left|
\frac{1}{1-\gamma}
\mathbb E_\nu\!\left[w_i(X)r(X)\right]
-V^\pi
\right|
=
\frac{1}{1-\gamma}
\left|
\mathbb E_{d_{\pi,\gamma}}
\!\left[(z_i(X)-1)r(X)\right]
\right|.
\]
By Cauchy--Schwarz and \eqref{eq:app-w-kl-l2-scalar},
\begin{align*}
\left|
\mathbb E_{d_{\pi,\gamma}}
\!\left[(z_i(X)-1)r(X)\right]
\right|
&\le
\|r\|_{2,d_{\pi,\gamma}}
\|z_i-1\|_{2,d_{\pi,\gamma}},\\
\|z_i-1\|_{2,d_{\pi,\gamma}}^2
&\le
2\Lambda_w
\mathbb E_{d_{\pi,\gamma}}
\!\left[z_i\log z_i-z_i+1\right]\\
&=
2\Lambda_w\mathsf D_\nu(w_i\|w^\pi).
\end{align*}
Combining these inequalities proves \eqref{eq:app-w-plugin-value}.
\end{proof}

\begin{proof}[Proof of Lemma~\ref{lem:w-oracle-equivalence}]
Fix $i\in\{1,\ldots,M\}$ and abbreviate $w=w_i$. The comparison with
$w^\pi$ is preserved by the adjoint update. Indeed,
\begin{equation}
\Lambda_w^{-1}w^\pi\nu
\le
(1-\gamma)d_0+\gamma(w\nu)P_{\pi}
\le
\Lambda_ww^\pi\nu,
\label{eq:app-w-adjoint-relative-comparison}
\end{equation}
because $w^\pi\nu=(1-\gamma)d_0+\gamma(w^\pi\nu)P_{\pi}$ and the corresponding
bounds on $w\nu$ relative to $w^\pi\nu$ hold by
Condition~\ref{cond:w-relative-ratios}. Thus, the middle measure in
\eqref{eq:app-w-adjoint-relative-comparison} admits a $\nu$-density
$w^+:=\mathcal B_\pi^\ast w$. The ratios $w/w^\pi$ and $w^+/w^\pi$ are bounded
above and away from zero on $\mathcal X^\pi$, so the hypotheses of
Lemma~\ref{lem:w-three-point} hold. By the lower bound in that lemma,
\eqref{eq:app-w-oracle-lower} holds.

For the upper bound, define
\[
z:=\frac{w}{w^\pi},
\qquad
\bar z:=\frac{w^+}{w^\pi},
\]
setting both ratios equal to one on $(\mathcal X^\pi)^c$.

Apply \eqref{eq:app-w-kl-l2-scalar} to $z$ and $\bar z$ under the probability
measure $w^\pi\nu=d_{\pi,\gamma}$. By the triangle inequality and
\eqref{eq:app-w-generalized-kl-contraction},
\begin{equation}
\begin{aligned}
\|z-\bar z\|_{2,d_{\pi,\gamma}}
&\le
\|z-1\|_{2,d_{\pi,\gamma}}
+\|\bar z-1\|_{2,d_{\pi,\gamma}}\\
&\le
\sqrt{2\Lambda_w}
\left\{
\sqrt{\mathsf D_\nu(w\|w^\pi)}
+\sqrt{\mathsf D_\nu(w^+\|w^\pi)}
\right\}\\
&\le
\sqrt{2\Lambda_w}(1+\sqrt{\gamma})
\sqrt{\mathsf D_\nu(w\|w^\pi)}.
\end{aligned}
\label{eq:app-w-adjoint-residual-l2}
\end{equation}
Moreover, by the log-Hellinger comparison
\eqref{eq:app-w-log-hellinger-scalar} and the generalized-KL control of
Hellinger distance in \eqref{eq:app-w-hellinger-kl-scalar},
\begin{equation}
\|\log z\|_{2,d_{\pi,\gamma}}
\le
2\sqrt{\Lambda_w}
\sqrt{\mathsf D_\nu(w\|w^\pi)}.
\label{eq:app-w-oracle-log-bound}
\end{equation}
By \eqref{eq:app-w-adjoint-residual-l2},
\eqref{eq:app-w-oracle-log-bound}, \eqref{eq:w-adjoint-moment},
\eqref{eq:w-oracle-score}, and Cauchy--Schwarz,
\begin{equation}
\left|
\mathcal S^\star(w)
\right|
\le
2\sqrt{2}\,\Lambda_w
(1+\sqrt{\gamma})
\mathsf D_\nu(w\|w^\pi).
\label{eq:app-w-oracle-upper}
\end{equation}
The upper bound in \eqref{eq:w-oracle-equivalence} follows from
\eqref{eq:app-w-oracle-upper}. Since $i$ was arbitrary, the result holds for
every candidate.
\end{proof}

\begin{lemma}[Pairwise occupancy-score bounds]
\label{lem:w-pairwise-geometry}
Suppose Condition~\ref{cond:w-relative-ratios} holds. Then, for every ordered
pair $(i,j)$,
\begin{align}
\mathcal S_{ij}
&\ge
(1-\gamma)e_i^2
-2\sqrt{2}\,\Lambda_w(1+\sqrt{\gamma})e_ie_j,
\label{eq:app-w-pairwise-lower}\\
H_{ij}
&\le
e_i+e_j,
\label{eq:app-w-hellinger-triangle}\\
|\mathcal S_{ij}|
&\le
2\sqrt{2}\,\Lambda_w(1+\sqrt{\gamma})e_iH_{ij}.
\label{eq:app-w-pairwise-upper}
\end{align}
\end{lemma}

\begin{proof}[Proof of Lemma~\ref{lem:w-pairwise-geometry}]
Let $w_i^+:=\mathcal B_\pi^\ast w_i$, $z_i=w_i/w^\pi$, and
$\bar z_i=w_i^+/w^\pi$, setting both relative ratios equal to one on
$(\mathcal X^\pi)^c$. By \eqref{eq:app-w-adjoint-residual-l2},
\begin{equation}
\|z_i-\bar z_i\|_{2,d_{\pi,\gamma}}
\le
\sqrt{2\Lambda_w}(1+\sqrt{\gamma})e_i.
\label{eq:app-w-candidate-residual}
\end{equation}
By the log-Hellinger comparison
\eqref{eq:app-w-log-hellinger-scalar} and the generalized-KL control of
Hellinger distance in \eqref{eq:app-w-hellinger-kl-scalar},
\begin{equation}
\|\log z_j\|_{2,d_{\pi,\gamma}}
\le
2\sqrt{\Lambda_w}\,e_j.
\label{eq:app-w-comparator-log}
\end{equation}
By Cauchy--Schwarz, \eqref{eq:app-w-candidate-residual}, and
\eqref{eq:app-w-comparator-log},
\begin{equation}
\left|
\mathbb E_\nu\!\left[
(w_i-w_i^+)\log\frac{w_j}{w^\pi}
\right]
\right|
\le
2\sqrt{2}\,\Lambda_w(1+\sqrt{\gamma})e_ie_j.
\label{eq:app-w-cross-moment}
\end{equation}
By \eqref{eq:w-adjoint-moment}, the definition of $\mathcal S_{ij}$, and
$\ell_{ij}=\log(w_i/w^\pi)-\log(w_j/w^\pi)$,
\begin{align*}
\mathcal S_{ij}
=
\mathcal S^\star(w_i)
-
\mathbb E_\nu\!\left[
(w_i-w_i^+)\log\frac{w_j}{w^\pi}
\right].
\end{align*}
Combining this identity with \eqref{eq:app-w-oracle-lower} and
\eqref{eq:app-w-cross-moment} proves \eqref{eq:app-w-pairwise-lower}.

After integration,
$H_\nu(p,q)\le\sqrt{\mathsf D_\nu(p\|q)}$ by
\eqref{eq:app-w-hellinger-kl-scalar}. Therefore, by the triangle inequality in
$L^2(\nu)$,
\[
H_{ij}
\le
H_\nu(w_i,w^\pi)+H_\nu(w_j,w^\pi)
\le
e_i+e_j,
\]
which proves \eqref{eq:app-w-hellinger-triangle}. Finally, by
\eqref{eq:app-w-log-hellinger-scalar},
\begin{equation}
\|\ell_{ij}\|_{2,d_{\pi,\gamma}}
\le
2\sqrt{\Lambda_w}\,H_{ij}.
\label{eq:app-w-pairwise-log-bound}
\end{equation}
By \eqref{eq:app-w-candidate-residual} and
\eqref{eq:app-w-pairwise-log-bound}, Cauchy--Schwarz applied to
$\mathcal S_{ij}=\mathbb E_\nu[(w_i-w_i^+)\ell_{ij}]$ gives
\eqref{eq:app-w-pairwise-upper}.
\end{proof}

For the penalized scores, write
\[
T_i^w
:=
\max_{1\le j\le M}\mathcal S_{ij}^{\mathrm{adj}},
\]
and define the population oracle constant
\begin{equation}
R_w
:=
\frac23\left[
\frac{2\sqrt{2}\,\Lambda_w}{1-\sqrt{\gamma}}
+\frac12
+\sqrt{
4\left\{\frac{2\sqrt{2}\,\Lambda_w}{1-\sqrt{\gamma}}\right\}^2
+\frac{2\sqrt{2}\,\Lambda_w}{1-\sqrt{\gamma}}
+1
}
\right].
\label{eq:app-w-selector-constants}
\end{equation}

\begin{lemma}[Population A-PBV score bounds]
\label{lem:w-penalized-score-bounds}
Suppose Condition~\ref{cond:w-relative-ratios} holds. Then, for every
$i,k\in\{1,\ldots,M\}$,
\begin{align}
T_i^w
&\ge
\mathcal S_{ik}^{\mathrm{adj}}
\ge
\frac{3(1-\gamma)}{4}e_i^2
-\left\{
2\sqrt{2}\,\Lambda_w(1+\sqrt{\gamma})
+\frac{1-\gamma}{2}
\right\}e_ie_k
-\frac{1-\gamma}{4}e_k^2,
\label{eq:app-w-penalized-lower}\\
T_k^w
&\le
\frac{8\Lambda_w^2(1+\sqrt{\gamma})^2}{1-\gamma}e_k^2.
\label{eq:app-w-penalized-upper}
\end{align}
Moreover,
\begin{equation}
R_w
\le
\frac{4\sqrt{2}\,\Lambda_w}{1-\sqrt{\gamma}}+\frac23.
\label{eq:app-w-population-constant}
\end{equation}
\end{lemma}

\begin{proof}[Proof of Lemma~\ref{lem:w-penalized-score-bounds}]
Since $|\mathcal S_{ik}|\ge\mathcal S_{ik}$, the pairwise lower bound
\eqref{eq:app-w-pairwise-lower} and the triangle bound
\eqref{eq:app-w-hellinger-triangle} give
\[
\mathcal S_{ik}^{\mathrm{adj}}
\ge
(1-\gamma)e_i^2
-2\sqrt{2}\,\Lambda_w(1+\sqrt{\gamma})e_ie_k
-\frac{1-\gamma}{4}(e_i+e_k)^2,
\]
which is \eqref{eq:app-w-penalized-lower}. For every $j$, by the pairwise
upper bound \eqref{eq:app-w-pairwise-upper},
\[
\mathcal S_{kj}^{\mathrm{adj}}
\le
2\sqrt{2}\,\Lambda_w(1+\sqrt{\gamma})e_kH_{kj}
-\frac{1-\gamma}{4}H_{kj}^2
.
\]
Completing the square in $H_{kj}$,
\begin{align*}
&2\sqrt{2}\,\Lambda_w(1+\sqrt{\gamma})e_kH_{kj}
-\frac{1-\gamma}{4}H_{kj}^2\\
&\qquad=
\frac{8\Lambda_w^2(1+\sqrt{\gamma})^2}{1-\gamma}e_k^2
-\frac{1-\gamma}{4}
\left\{
H_{kj}-
\frac{4\sqrt{2}\,\Lambda_w(1+\sqrt{\gamma})}{1-\gamma}e_k
\right\}^2.
\end{align*}
Hence,
$\mathcal S_{kj}^{\mathrm{adj}}
\le
8\Lambda_w^2(1+\sqrt{\gamma})^2e_k^2/(1-\gamma)$.
Taking the maximum over $j$ proves
\eqref{eq:app-w-penalized-upper}.

To bound $R_w$, note that
$2\sqrt{2}\,\Lambda_w/(1-\sqrt\gamma)\ge1$ and that
$\sqrt{4t^2+t+1}\le2t+1/2$ for every $t\ge1$.
By this elementary inequality and
\eqref{eq:app-w-selector-constants},
\eqref{eq:app-w-population-constant} holds.
\end{proof}

\subsection{Population oracle inequality}
\label{app:w-population-proof}

\begin{proof}[Proof of Theorem~\ref{thm:w-population-oracle}]
Choose $k\in\arg\min_{1\le i\le M}e_i$, and write
$x=e_{\widetilde i_w}$ and $y=e_k$. By population optimality,
$T_{\widetilde i_w}^w\leq T_k^w$. Combining this comparison with
Lemma~\ref{lem:w-penalized-score-bounds},
\begin{equation}
\frac{3(1-\gamma)}{4}x^2
\le
\left\{
2\sqrt{2}\,\Lambda_w(1+\sqrt{\gamma})
+\frac{1-\gamma}{2}
\right\}xy
+\left\{
\frac{1-\gamma}{4}
+\frac{8\Lambda_w^2(1+\sqrt{\gamma})^2}{1-\gamma}
\right\}y^2.
\label{eq:app-w-population-quadratic}
\end{equation}
If $y=0$, then $x=0$ by \eqref{eq:app-w-population-quadratic}. If $y>0$, let
$r=x/y$ and $t=2\sqrt{2}\,\Lambda_w/(1-\sqrt\gamma)$. Then
\eqref{eq:app-w-population-quadratic} is equivalent to
\[
\frac34r^2-(t+\tfrac12)r-(t^2+\tfrac14)\leq0,
\]
whose nonnegative root is $R_w$ as defined in
\eqref{eq:app-w-selector-constants}. Thus, $x\leq R_wy$. By
\eqref{eq:app-w-population-constant}, \eqref{eq:w-population-oracle} follows.
\end{proof}

\subsection{Hellinger oracle inequality}
\label{app:w-hellinger-proof}

\begin{proof}[Proof of Theorem~\ref{thm:w-finite-sample-oracle}]
Condition on the training sample. Throughout this proof, $C>0$ denotes
a finite constant depending only on $C_w$ and $\overline w$;
its value may increase between displays. Choose the smallest index
$k\in\argmin_aH_\nu(w_a,w^\pi)$, and write
\[
e:=H_\nu(w_k,w^\pi),\qquad
q:=(w^\pi\vee w_k/2)\wedge2w_k,
\qquad d:=\|q-\mathcal B_\pi^\ast q\|_{1,\nu}.
\]
Condition~\ref{cond:w-candidate-bounds} ensures that the adjoint update
of every bounded nonnegative function exists as a $\nu$-density.

\emph{Comparison with a bounded density.}
On $\{w^\pi<w_k/2\}$,
\[
\frac{w_k}{2}-w^\pi
\le(\sqrt{w_k}-\sqrt{w^\pi})^2,
\]
by expanding $2(\sqrt{w^\pi}-\sqrt{w_k}/2)^2\ge0$.
On $\{w^\pi>2w_k\}$,
\[
w^\pi-2w_k\le2(\sqrt{w^\pi}-\sqrt{w_k})^2,
\]
by expanding $(\sqrt{w^\pi}-2\sqrt{w_k})^2\ge0$.
Moreover, $q$ lies pointwise between $w_k$ and $w^\pi$. Thus
\begin{equation}
\|q-w^\pi\|_{1,\nu}\le2e^2,\qquad
H_\nu(q,w^\pi)\le e,\qquad H_\nu(w_k,q)\le e.
\label{eq:w-hellinger-comparison}
\end{equation}
Since $\mathcal B_\pi^\ast w^\pi=w^\pi$ and a Markov kernel
contracts the total variation norm,
\begin{equation}
d=\|(I-\gamma P_\pi^\ast)(q-w^\pi)\|_{1,\nu}
\le(1+\gamma)\|q-w^\pi\|_{1,\nu}\le4e^2.
\label{eq:w-hellinger-comparison-residual}
\end{equation}

\emph{Adjoint moment bounds.}
For each candidate, set
$D_i:=\mathsf D_\nu(w_i\|q)$ and $x_i:=\sqrt{D_i}$.
The candidate ratio bound and $w_k/2\le q\le2w_k$ imply
\[
\frac1{2C_w}\le\frac{w_i}{q}\le2C_w,
\qquad 0<q\le2\overline w.
\]
Consequently $D_i$ is finite and $\log(w_i/q)$ is bounded. Put $f=w_i/q$.
The generalized-KL variational inequality
\[
\langle p,\log f\rangle_\nu
\le\mathsf D_\nu(p\|v)+\langle v,f-1\rangle_\nu
\]
holds for nonnegative integrable $p,v$ and positive bounded $f$, by the
pointwise inequality $u\log z\le u\log u-u+z$ after scaling by $v$.
Apply it with $p=\mathcal B_\pi^\ast w_i$ and
$v=\mathcal B_\pi^\ast q$. By
Lemma~\ref{lem:w-generalized-kl-contraction},
\[
\langle\mathcal B_\pi^\ast w_i,\log(w_i/q)\rangle_\nu
\le\gamma D_i+
\langle\mathcal B_\pi^\ast q,w_i/q-1\rangle_\nu.
\]
Subtracting from $\langle w_i,\log(w_i/q)\rangle_\nu$ gives
\begin{equation}
\begin{aligned}
\langle w_i-\mathcal B_\pi^\ast w_i,\log(w_i/q)\rangle_\nu
&\ge(1-\gamma)D_i
+\langle q-\mathcal B_\pi^\ast q,w_i/q-1\rangle_\nu\\
&\ge(1-\gamma)x_i^2-Cd.
\end{aligned}
\label{eq:w-hellinger-oracle-lower}
\end{equation}

To control the remaining moments, first note that
\[
\left\|\frac{w_i-q}{\sqrt q}\right\|_{2,\nu}
\le(1+\sqrt{2C_w})H_\nu(w_i,q).
\]
For any bounded critic $g$, Jensen's inequality and
$\gamma P_\pi^\ast q\le\mathcal B_\pi^\ast q$ yield
\[
\begin{aligned}
\gamma^2\|P_\pi g\|_{2,q\nu}^2
&\le\gamma\langle\mathcal B_\pi^\ast q,g^2\rangle_\nu\\
&\le\gamma\{\|g\|_{2,q\nu}^2+d\|g\|_\infty^2\}.
\end{aligned}
\]
Decompose $w_i-\mathcal B_\pi^\ast w_i$
as $(I-\gamma P_\pi^\ast)(w_i-q)+(q-\mathcal B_\pi^\ast q)$.
Weighted Cauchy--Schwarz and the preceding two bounds give
\begin{equation}
|\langle w_i-\mathcal B_\pi^\ast w_i,g\rangle_\nu|
\le C H_\nu(w_i,q)
\{\|g\|_{2,q\nu}+\sqrt d\|g\|_\infty\}
+d\|g\|_\infty.
\label{eq:w-hellinger-cross-moment}
\end{equation}
Here the norm $\|\cdot\|_{2,q\nu}$ is with respect to the finite measure
with density $q$ relative to $\nu$.
For positive $u,v$,
$|\log u-\log v|\le2|\sqrt u-\sqrt v|/\sqrt{\min(u,v)}$.
Since $q/\min(w_i,w_j)\le2C_w$ and the candidate ratios are bounded,
\begin{align*}
\|\ell_{ij}\|_{2,q\nu}&\le\sqrt{8C_w}\,H_\nu(w_i,w_j),
&\|\ell_{ij}\|_\infty&\le\log C_w,\\
\|\log(w_k/q)\|_{2,q\nu}&\le\sqrt8\,H_\nu(w_k,q),
&\|\log(w_k/q)\|_\infty&\le\log2.
\end{align*}

Write $h:=1-\gamma$ and $H_{ij}:=H_\nu(w_i,w_j)$.
Using $\ell_{ik}=\log(w_i/q)-\log(w_k/q)$ in
\eqref{eq:w-hellinger-oracle-lower} and
\eqref{eq:w-hellinger-cross-moment}, and then applying
\eqref{eq:w-hellinger-comparison}--\eqref{eq:w-hellinger-comparison-residual},
we obtain $|\mathcal S_{ik}|\ge\mathcal S_{ik}\ge hx_i^2-Cex_i-Ce^2$.
Moreover, $H_\nu(w_i,q)\le x_i$ by
Lemma~\ref{lem:w-scalar-generalized-kl}, so
$H_{ik}\le x_i+e$. Subtracting the Hellinger penalty gives
\begin{equation}
\mathcal S_{ik}^{\mathrm{adj}}
\ge\frac{3h}{4}x_i^2-Cex_i-Ce^2.
\label{eq:w-hellinger-pair-lower}
\end{equation}
For the comparator $k$, \eqref{eq:w-hellinger-cross-moment} applied to
$g=\ell_{kj}$ instead gives
\begin{equation}
\mathcal S_{kj}^{\mathrm{adj}}
\le CeH_{kj}+Ce^2-\frac h4H_{kj}^2.
\label{eq:w-hellinger-pair-upper}
\end{equation}

\emph{Validation deviation.}
For a validation transition, define
\[
Y_{ij,s}:=s\,w_i(X)\{\ell_{ij}(X)-\gamma(\pi\ell_{ij})(S')\}
-\frac h4\{\sqrt{w_i(X)}-\sqrt{w_j(X)}\}^2,
\qquad s\in\{-1,1\}.
\]
The candidate bounds give $|Y_{ij,s}|\le C$.
The current-state term has second moment at most $CH_{ij}^2$: indeed,
\[
w_i^2\ell_{ij}^2
\le4\frac{w_i^2}{\min(w_i,w_j)}
(\sqrt{w_i}-\sqrt{w_j})^2
\le4C_w\overline w(\sqrt{w_i}-\sqrt{w_j})^2.
\]
For the successor term, use $w_i^2\le2C_w\overline w\,q$ and conditional
Jensen's inequality to obtain
\begin{align*}
\mathbb E[\gamma^2w_i(X)^2\{(\pi\ell_{ij})(S')\}^2]
&\le2C_w\overline w\,\gamma^2
\mathbb E_{q\nu}P_\pi(\ell_{ij}^2)\\
&\le2C_w\overline w\,\gamma
\langle\mathcal B_\pi^\ast q,\ell_{ij}^2\rangle_\nu\\
&\le C(H_{ij}^2+d).
\end{align*}
The last step uses
$\langle\mathcal B_\pi^\ast q,\ell_{ij}^2\rangle_\nu
\le\|\ell_{ij}\|_{2,q\nu}^2+d\|\ell_{ij}\|_\infty^2$.
The penalty has second moment at most $CH_{ij}^2$, since
$\mathbb E_\nu(\sqrt{w_i}-\sqrt{w_j})^4\le CH_{ij}^2$.
By \eqref{eq:w-hellinger-comparison-residual},
\begin{equation}
\mathbb E Y_{ij,s}^2\le C(H_{ij}^2+e^2).
\label{eq:w-hellinger-score-second-moment}
\end{equation}
For the initial contribution, $|h\ell_{ij}(X_0)|\le Ch$ and
$h d_0\le(\mathcal B_\pi^\ast q)\nu$ imply
\[
h^2\mathbb E_{d_0}\ell_{ij}^2
\le h\langle\mathcal B_\pi^\ast q,\ell_{ij}^2\rangle_\nu
\le Ch(H_{ij}^2+e^2).
\]

Let
$\mathcal P_k^w=\{(i,k):i\in[M]\}\cup\{(k,j):j\in[M]\}$,
$u=\log(24M/\delta)$, $r_T=\sqrt{u/(ng_b)}$, and
$r_0=\sqrt{u/m}$. Apply Lemma~\ref{lem:trajectory-bernstein}(i), with
tail parameter $2u$, to each $Y_{ij,s}$ along the validation chain.
Apply the same lemma with $g=1$ and tail parameter $2u$ to
$s h\ell_{ij}(X_{0,t})$ along the iid initial sequence.
The empirical and population scores are maxima over the two signs.
Since the maximum is $1$-Lipschitz in the supremum norm,
a union bound over both signs and at most $2M-1$ ordered pairs gives,
with probability at least $1-\delta$,
\begin{align*}
|\widehat{\mathcal S}_{ij}^{\mathrm{adj}}
-\mathcal S_{ij}^{\mathrm{adj}}|
&\le C\{(H_{ij}+e)r_T+r_T^2\}\\
&\quad+C\{\sqrt h(H_{ij}+e)r_0+hr_0^2\}
\end{align*}
simultaneously for all $(i,j)\in\mathcal P_k^w$.
The two concentration events use only the marginal laws of their respective
samples. Their union bound does not require independence between the
initial sequence and the validation trajectory.
For exact initial moments, set $r_0=0$.
Writing $r=r_T+\sqrt h\,r_0$, we have
$r_T^2+hr_0^2\le r^2$, so the preceding bound implies
\begin{equation}
|\widehat{\mathcal S}_{ij}^{\mathrm{adj}}
-\mathcal S_{ij}^{\mathrm{adj}}|
\le C\{(H_{ij}+e)r+r^2\}.
\label{eq:w-hellinger-validation}
\end{equation}

\emph{Empirical comparison.}
On this event, set $x=x_{\widehat i_w}$ and $z=e+r$.
By \eqref{eq:w-hellinger-pair-lower},
\eqref{eq:w-hellinger-validation}, and
$H_{\widehat i_wk}\le x+e$,
\[
\max_j\widehat{\mathcal S}_{\widehat i_wj}^{\mathrm{adj}}
\ge\frac{3h}{4}x^2-Czx-Cz^2.
\]
For the comparator, \eqref{eq:w-hellinger-pair-upper} and
\eqref{eq:w-hellinger-validation} give
\[
\max_j\widehat{\mathcal S}_{kj}^{\mathrm{adj}}
\le\sup_{u\ge0}\{Czu-hu^2/4+Cz^2\}
\le Cz^2/h.
\]
Empirical optimality therefore implies
$3hx^2/4\le Czx+Cz^2/h$. Multiplication by $h$ and the quadratic formula
yield $x\le Cz/h$. Finally,
\[
H_\nu(w_{\widehat i_w},w^\pi)
\le H_\nu(w_{\widehat i_w},q)+H_\nu(q,w^\pi)
\le x+e
\le\frac C{1-\gamma}(e+r_T)
+\frac C{\sqrt{1-\gamma}}r_0.
\]
For validation-state initial averages, incorporate
$-s h(\pi\ell_{ij})(S)$ into $Y_{ij,s}$. Its expectation is the required
initial moment, and Jensen's inequality gives
\[
h^2\mathbb E_{\nu_S}(\pi\ell_{ij})^2
\le h^2\mathbb E_{d_0}\ell_{ij}^2
\le Ch(H_{ij}^2+e^2).
\]
Thus \eqref{eq:w-hellinger-score-second-moment} and the same trajectory
concentration argument apply with $r=r_T$. This proves both assertions.
\end{proof}

\section{Proofs for \texorpdfstring{\(Q\)}{Q}-function selection}
\label{app:q-selection-proofs}

\subsection{Importance-weighted PBV}
\label{app:q-model-selection-proofs}

\subsubsection{Population score bounds}
\label{app:q-population-proofs}

Recall that
\[
\Delta_i:=Q_i-Q^\pi.
\]

\begin{lemma}[Discounted-occupancy operator bounds]
\label{lem:q-occupancy-operator}
For every $f\in L^2(d_{\pi,\gamma})$,
\begin{align}
\|\gamma P_\pi f\|_{2,d_{\pi,\gamma}}
&\leq
\sqrt{\gamma}\|f\|_{2,d_{\pi,\gamma}},
\label{eq:app-q-operator-contraction}\\
\left\langle f,(I-\gamma P_\pi)f\right\rangle_{d_{\pi,\gamma}}
&\geq
(1-\sqrt{\gamma})\|f\|_{2,d_{\pi,\gamma}}^2,
\label{eq:app-q-residual-coercivity}\\
\|(I-\gamma P_\pi)f\|_{2,d_{\pi,\gamma}}
&\leq
(1+\sqrt{\gamma})\|f\|_{2,d_{\pi,\gamma}}.
\label{eq:app-q-residual-boundedness}
\end{align}
\end{lemma}

\begin{proof}
If $\gamma=0$, then $\gamma P_\pi=0$ and
$I-\gamma P_\pi=I$, so all three bounds hold with equality. Suppose
$\gamma>0$.
By conditional Jensen's inequality,
\[
(P_\pi f)^2\leq P_\pi(f^2).
\]
By the discounted-occupancy identity
$d_{\pi,\gamma}=(1-\gamma)d_0+\gamma d_{\pi,\gamma}P_\pi$,
\begin{equation}
\gamma
\mathbb E_{d_{\pi,\gamma}}\!\left[P_\pi(f^2)\right]
=
\mathbb E_{d_{\pi,\gamma}}(f^2)
-
(1-\gamma)\mathbb E_{d_0}(f^2)
\leq
\mathbb E_{d_{\pi,\gamma}}(f^2).
\label{eq:app-q-occupancy-second-moment}
\end{equation}
Therefore,
\begin{align*}
\|\gamma P_\pi f\|_{2,d_{\pi,\gamma}}^2
&=
\gamma^2
\mathbb E_{d_{\pi,\gamma}}\!\left[(P_\pi f)^2\right]\\
&\leq
\gamma^2
\mathbb E_{d_{\pi,\gamma}}\!\left[P_\pi(f^2)\right]\\
&\leq
\gamma
\mathbb E_{d_{\pi,\gamma}}(f^2)\\
&=
\gamma\|f\|_{2,d_{\pi,\gamma}}^2,
\end{align*}
This proves the contraction bound.

Cauchy--Schwarz and the triangle inequality now give, respectively,
\begin{align*}
\langle f,(I-\gamma P_\pi)f\rangle_{d_{\pi,\gamma}}
&\ge (1-\sqrt\gamma)\|f\|_{2,d_{\pi,\gamma}}^2,\\
\|(I-\gamma P_\pi)f\|_{2,d_{\pi,\gamma}}
&\le (1+\sqrt\gamma)\|f\|_{2,d_{\pi,\gamma}}.
\end{align*}
These are the remaining two bounds.
\end{proof}

For $f\in L^2(d_{\pi,\gamma})$, write
\[
\mathcal E(f):=\langle f,(I-\gamma P_\pi)f\rangle_{d_{\pi,\gamma}}.
\]

\begin{lemma}[Bellman operator bound]
\label{lem:q-refined-residual}
Let $T$ be a bounded linear operator on a real Hilbert space, with
$\|T\|\le c<1$, and let $A=I-T$. Then, for every $f,g$,
\[
\|Af\|^2\vee\|A^*f\|^2\le(1+c)\langle f,Af\rangle,
\qquad
|\langle f,Ag\rangle|\le\sqrt{(1+c)\langle f,Af\rangle}\,\|g\|.
\]
In particular, for every $f,g\in L^2(d_{\pi,\gamma})$,
\begin{equation}
\left|\langle f,(I-\gamma P_\pi)g\rangle_{d_{\pi,\gamma}}\right|
\le \sqrt{(1+\sqrt\gamma)\mathcal E(f)}
\|g\|_{2,d_{\pi,\gamma}}.
\label{eq:app-q-refined-residual-linear}
\end{equation}
\end{lemma}
\begin{proof}
Write $E=\langle f,Af\rangle$. Since $\|T^*\|=\|T\|\le c$ and
$E\le(1+c)\|f\|^2$, expansion of the squared norm gives
\begin{align*}
\|Af\|^2
&=2E-\|f\|^2+\|Tf\|^2\\
&\le2E-(1-c^2)\|f\|^2\le(1+c)E.
\end{align*}
The same expansion with $T^*$ proves the bound for $A^*f$, because
$\langle f,A^*f\rangle=E$. Cauchy--Schwarz applied to
$\langle A^*f,g\rangle$ proves the cross-moment bound. Finally,
Lemma~\ref{lem:q-occupancy-operator} gives $\|\gamma P_\pi\|\le\sqrt\gamma$
in $L^2(d_{\pi,\gamma})$, proving
\eqref{eq:app-q-refined-residual-linear}.
\end{proof}

\begin{lemma}[Second moments and score integrability]
\label{lem:q-target-second-moments}
Suppose Condition~\ref{cond:q-population-domain} holds. Then
$Q^\pi\in L^2(d_{\pi,\gamma})$, every $Q_i-Q^\pi$ belongs to
$L^2(d_{\pi,\gamma})$, and the expectations defining
$\mathcal M_{ij}^{\mathrm{IW}}$ are absolutely integrable for every ordered
pair $(i,j)$.
\end{lemma}

\begin{proof}
Let $r(x):=\mathbb E(R\mid X=x)$ and
$m(x):=\mathbb E(|R|\mid X=x)$. By conditional Jensen's inequality,
$\|r\|_{2,d_{\pi,\gamma}}\le\|m\|_{2,d_{\pi,\gamma}}$ and
$\|m\|_{2,d_{\pi,\gamma}}^2
\le\mathbb E_\nu[w^\pi(X)R^2]<\infty$.
Lemma~\ref{lem:q-occupancy-operator} gives
$\|\gamma P_\pi\|_{L^2(d_{\pi,\gamma})\to L^2(d_{\pi,\gamma})}
\le\sqrt\gamma<1$. Thus the Neumann series
$\sum_{t\ge0}(\gamma P_\pi)^t r$ converges in $L^2(d_{\pi,\gamma})$.
The same bound applied to $m$ gives
\[
\left\|\sum_{t\ge0}(\gamma P_\pi)^t m\right\|_{2,d_{\pi,\gamma}}
\le\frac{\|m\|_{2,d_{\pi,\gamma}}}{1-\sqrt\gamma}<\infty.
\]
By Tonelli's theorem, this nonnegative series equals
$\mathbb E_\pi[\sum_{t\ge0}\gamma^t|R_t|\mid X_0=x]$ and is finite for
$d_{\pi,\gamma}$-almost every $x$. We may therefore interchange expectation
and summation in the definition of $Q^\pi$, obtaining
\[
Q^\pi=\sum_{t\ge0}(\gamma P_\pi)^t r=(I-\gamma P_\pi)^{-1}r,
\qquad
\|Q^\pi\|_{2,d_{\pi,\gamma}}
\le\frac{\|r\|_{2,d_{\pi,\gamma}}}{1-\sqrt\gamma}.
\]
By Condition~\ref{cond:q-population-domain}, every
$\Delta_i=Q_i-Q^\pi$ and
$h_{ij}=Q_i-Q_j=\Delta_i-\Delta_j$ belongs to
$L^2(d_{\pi,\gamma})$.

It remains to verify integrability of the one-step score. For $\gamma>0$,
by conditional Jensen's inequality and
\eqref{eq:app-q-occupancy-second-moment},
$\mathbb E_\nu[w^\pi(X)\{(\pi Q_i)(S')\}^2]
\leq \gamma^{-1}\|Q_i\|_{2,d_{\pi,\gamma}}^2<\infty$;
when $\gamma=0$, the continuation term is identically zero. Hence
$Q_i(X)-R-\gamma(\pi Q_i)(S')$ is square integrable under the
$w^\pi$-weighted transition law. By Cauchy--Schwarz and
$h_{ij}\in L^2(d_{\pi,\gamma})$,
$\mathbb E_\nu[w^\pi(X)|h_{ij}(X)|
|Q_i(X)-R-\gamma(\pi Q_i)(S')|]<\infty$.
Finally, $\varepsilon_i-\varepsilon_j$ is square integrable under the
$w^\pi$-weighted transition law. Hence
$\mathbb E_\nu[w^\pi|h_{ij}(\varepsilon_i-\varepsilon_j)|]<\infty$
by Cauchy--Schwarz, which proves absolute integrability of the penalty as well.
\end{proof}

For each $i\in[K]$, let $T_i:=\max_{j\in\mathcal J_Q}\mathcal M_{ij}^{\mathrm{IW}}$.
\begin{lemma}[Population score comparison]
\label{lem:q-population-score-bounds}
Suppose Condition~\ref{cond:q-population-domain} holds. For every $i,k\in[K]$ and $j\in\mathcal J_Q$,
\begin{align}
T_i\ge\mathcal M_{ik}^{\mathrm{IW}}
&\ge\frac34\mathcal E(h_{ik})
-\sqrt{(1+\sqrt\gamma)\mathcal E(h_{ik})}
\|\Delta_k\|_{2,d_{\pi,\gamma}},
\label{eq:q-population-score-lower}\\
\mathcal M_{kj}^{\mathrm{IW}}
&\le\sqrt{(1+\sqrt\gamma)\mathcal E(h_{kj})}
\|\Delta_k\|_{2,d_{\pi,\gamma}}
-\frac14\mathcal E(h_{kj}),
\label{eq:q-population-pair-upper}\\
T_k&\le(1+\sqrt\gamma)\|\Delta_k\|_{2,d_{\pi,\gamma}}^2.
\label{eq:q-population-score-upper}
\end{align}
\end{lemma}
\begin{proof}
Set $A=I-\gamma P_\pi$. Conditioning on $X$, using the Bellman equation for
$Q^\pi$, and applying Lemma~\ref{lem:q-occupancy-operator} give
\begin{equation}
\mathcal M_{ij}^{\mathrm{IW}}
=|\langle h_{ij},A\Delta_i\rangle_{d_{\pi,\gamma}}|
-\frac14\mathcal E(h_{ij}).
\label{eq:app-q-score-representation}
\end{equation}
The absolute value in the population penalty has no effect because
$\mathcal E(h_{ij})\ge0$. Since $\Delta_i=h_{ik}+\Delta_k$, the reverse
triangle inequality gives
\[
\mathcal M_{ik}^{\mathrm{IW}}
\ge\frac34\mathcal E(h_{ik})-
|\langle h_{ik},A\Delta_k\rangle_{d_{\pi,\gamma}}|.
\]
Lemma~\ref{lem:q-refined-residual} proves
\eqref{eq:q-population-score-lower}. Applying the same lemma directly to
\eqref{eq:app-q-score-representation} with $i=k$ proves
\eqref{eq:q-population-pair-upper}. Finally, for $b\ge0$,
$\sup_{z\ge0}(bz-z^2/4)=b^2$. Maximize the latter bound over
$z=\sqrt{\mathcal E(h_{kj})}$ to obtain
\eqref{eq:q-population-score-upper}.
\end{proof}

\begin{proof}[Proof of Theorem~\ref{thm:q-population-oracle}]
Choose $k\in\arg\min_i\|\Delta_i\|_{2,d_{\pi,\gamma}}$ and write
$y=\|\Delta_k\|_{2,d_{\pi,\gamma}}$, $\beta=1+\sqrt\gamma$, and
$z=\sqrt{\mathcal E(h_{\widetilde i_Q k})}$. By the minimizing property of
$\widetilde i_Q$ and Lemma~\ref{lem:q-population-score-bounds},
\[
\frac34z^2-\sqrt\beta\,yz
\le T_{\widetilde i_Q}\le T_k\le\beta y^2.
\]
Solving this quadratic inequality gives $z\le2\sqrt\beta\,y$.
By \eqref{eq:app-q-residual-coercivity} and the triangle inequality,
\[
\|\Delta_{\widetilde i_Q}\|_{2,d_{\pi,\gamma}}
\le y+\|h_{\widetilde i_Qk}\|_{2,d_{\pi,\gamma}}
\le y+\frac z{\sqrt{1-\sqrt\gamma}}
\le\left\{1+2\sqrt{\frac{1+\sqrt\gamma}{1-\sqrt\gamma}}\right\}y.
\]
This is the asserted population oracle inequality.
\end{proof}

\subsubsection{Concentration of the importance-weighted scores}
\label{app:q-validation-proof}

Define the fitted-weight error by
\begin{equation}
\rho_w:=H_\nu(\widehat w,w^\pi).
\label{eq:q-relative-weight-error}
\end{equation}

\begin{lemma}[Localized validation-score deviation]
\label{lem:q-localized-validation}
Suppose Condition~\ref{cond:validation-spectral-gap} and
Conditions~\ref{cond:q-bounded-signal}--\ref{cond:q-fitted-weight-envelope}
hold, and suppose the $Q$-validation fold is independent of all data used
to construct $\widehat w$. There is a universal constant $C_0>0$ such that,
for every $\delta\in(0,1)$ and every nonempty collection
$\mathcal P\subseteq[K]\times\mathcal J_Q$ fixed conditional on the training sample and the
data used to construct $\widehat w$, conditional on these data, with
probability at least $1-\delta$, simultaneously for all $(i,j)\in\mathcal P$,
\begin{equation}
\left|\widehat{\mathcal M}_{ij}^{\mathrm{IW}}-\mathcal M_{ij}^{\mathrm{IW}}\right|
\le\eta\|h_{ij}\|_{2,d_{\pi,\gamma}}+v,
\label{eq:q-localized-validation}
\end{equation}
where, with $r=\{\log(6|\mathcal P|/\delta)/(ng_b)\}^{1/2}$,
\begin{align*}
\eta&=3G_Q\rho_w+C_0\sqrt{W_Q}(G_Q+\sigma_Q)r,\\
v&=\frac32H_QG_Q\rho_w^2+C_0H_Q(G_Q+\sigma_Q)
\{\sqrt{W_Q}\rho_wr+W_Qr^2\}.
\end{align*}
\end{lemma}
\begin{proof}
Condition on the training sample and the data used to construct
$\widehat w$. Let $\mathbb P_{n,Q}$ denote the empirical average over the
$Q$-validation transitions. For $i\in\mathcal J_Q$, put
$\mu_i(X)=\mathbb E(\varepsilon_i\mid X)$ and $\xi_i=\varepsilon_i-\mu_i(X)$.
For an ordered pair $(i,j)$ and $\tau\in\{3/4,5/4\}$, define
\begin{align*}
Z_{ij,\tau}&=\widehat w(X)h_{ij}(X)
\{\tau\varepsilon_i+(1-\tau)\varepsilon_j\},\\
m_{ij,\tau}&=\tau\mu_i+(1-\tau)\mu_j,
\qquad
\zeta_{ij,\tau}=\tau\xi_i+(1-\tau)\xi_j.
\end{align*}
For $s\in\{-1,1\}$ and $\tau\in\{3/4,5/4\}$, the identity
\[
|a|-|a-b|/4=\max_s\min_\tau s\{\tau a+(1-\tau)b\}
\]
gives
\begin{align}
\widehat{\mathcal M}_{ij}^{\mathrm{IW}}
&=\max_s\min_{\tau}s\,\mathbb P_{n,Q}Z_{ij,\tau},
\notag\\
\mathcal M_{ij}^{\mathrm{IW}}
&=\max_s\min_{\tau}
s\,\mathbb E_\nu[w^\pi h_{ij}m_{ij,\tau}].
\label{eq:app-q-score-branch-representation}
\end{align}
Since maxima and minima are $1$-Lipschitz in the supremum norm, it
suffices to control the difference between the empirical and target-weighted
expectations for both values of $\tau$.

The conditional means satisfy $|m_{ij,\tau}|\le3G_Q/2$ by
Condition~\ref{cond:q-bounded-signal}, because
$|\tau|+|1-\tau|\le3/2$. Moreover,
$\mathbb E(\zeta_{ij,\tau}\mid X)=0$ and, by convexity and
Condition~\ref{cond:q-subgaussian-td},
\[
\mathbb E\left[\left.
\exp\left\{\frac{|\zeta_{ij,\tau}|}{3\sigma_Q/2}\right\}
\right|X\right]\le2.
\]
The validation setup makes these centered residuals martingale differences
conditional on the past.

Write $A_{ij}=\widehat w h_{ij}$ and
$s_{ij}=\|h_{ij}\|_{2,d_{\pi,\gamma}}$. Under
Condition~\ref{cond:q-fitted-weight-envelope},
\begin{equation}
|A_{ij}|\le W_QH_Q,
\qquad
\mathbb E_\nu(A_{ij}^2)\le W_Q(s_{ij}+H_Q\rho_w)^2.
\label{eq:app-q-multiplier-second-moment}
\end{equation}
Indeed, $\widehat w^2\le W_Q\widehat w$ and the triangle inequality give
\[
\|\sqrt{\widehat w}\,h_{ij}\|_{2,\nu}
\le\|\sqrt{w^\pi}\,h_{ij}\|_{2,\nu}
+\|(\sqrt{\widehat w}-\sqrt{w^\pi})h_{ij}\|_{2,\nu}
\le s_{ij}+H_Q\rho_w.
\]
Set $z=\sqrt{\widehat w}-\sqrt{w^\pi}$. Since
$\widehat w-w^\pi=2\sqrt{w^\pi}z+z^2$, Cauchy--Schwarz yields, for either value of $\tau$,
\begin{align}
\left|\mathbb EZ_{ij,\tau}
-\mathbb E_\nu[w^\pi h_{ij}m_{ij,\tau}]\right|
&\le3G_Q\rho_ws_{ij}+\frac32H_QG_Q\rho_w^2.
\label{eq:app-q-weight-bias}
\end{align}

Cauchy--Schwarz also gives
\begin{equation}
\left|\mathbb EZ_{ij,\tau}
-\mathbb E_\nu[w^\pi h_{ij}m_{ij,\tau}]\right|
\le 2H_Q\rho_w\|m_{ij,\tau}\|_{2,d_{\pi,\gamma}}
+\frac32H_QG_Q\rho_w^2.
\label{eq:app-q-weight-bias-residual}
\end{equation}
Indeed, the term linear in $z=\sqrt{\widehat w}-\sqrt{w^\pi}$ is at most
$2\|z\|_{2,\nu}\|h_{ij}m_{ij,\tau}\|_{2,d_{\pi,\gamma}}$, and
$|h_{ij}|\le H_Q$; the term quadratic in $z$ is bounded as in
\eqref{eq:app-q-weight-bias}.

Apply Lemma~\ref{lem:trajectory-bernstein}(ii) with multiplier $A_{ij}$,
conditional mean $m_{ij,\tau}$, and centered residual $\zeta_{ij,\tau}$.
The bounds required by that lemma follow from
Conditions~\ref{cond:q-bounded-signal}--\ref{cond:q-fitted-weight-envelope}
and \eqref{eq:app-q-multiplier-second-moment}. Taking a
union bound over the $2|\mathcal P|$ choices of $(i,j,\tau)$ with
$u=\log(12|\mathcal P|/\delta)$ gives, simultaneously,
\begin{align}
|\mathbb P_{n,Q}Z_{ij,\tau}-\mathbb EZ_{ij,\tau}|
&\le C_0(G_Q+\sigma_Q)
\{\sqrt{W_Q}(s_{ij}+H_Q\rho_w)r+W_QH_Qr^2\},
\label{eq:app-q-empirical-deviation}
\end{align}
after increasing $C_0$, because
$\log(12|\mathcal P|/\delta)\le2\log(6|\mathcal P|/\delta)$.
Combining \eqref{eq:app-q-score-branch-representation}, the bounds for maxima and
minima, \eqref{eq:app-q-weight-bias}, and
\eqref{eq:app-q-empirical-deviation} proves
\eqref{eq:q-localized-validation}.
\end{proof}

\subsubsection{Finite-sample oracle inequality}
\label{app:q-finite-sample-proof}

Let $C_0$ be the universal constant in
Lemma~\ref{lem:q-localized-validation}, and define
\begin{align*}
V_Q&:=H_Q\left\{\frac32G_Q+C_0(G_Q+\sigma_Q)
\left(W_Q+\frac{\sqrt{W_Q}}4\right)\right\},\\
C_{Q,\mathrm{wt}}&:=6\sqrt2H_Q+\frac{2\sqrt6}{3}\sqrt{V_Q},\\
C_{Q,\mathrm{val}}&:=2C_0\sqrt{W_Q}(G_Q+\sigma_Q)
+\frac{2\sqrt6}{3}\sqrt{V_Q}.
\end{align*}
These constants depend only on the envelopes in
Conditions~\ref{cond:q-bounded-signal}--\ref{cond:q-fitted-weight-envelope}.

\begin{proof}[Proof of Theorem~\ref{thm:q-finite-sample-oracle}]
Let $k$ be the smallest index minimizing
$\|\Delta_i\|_{2,d_{\pi,\gamma}}$, and apply the concentration
bound \eqref{eq:app-q-empirical-deviation} from the proof of
Lemma~\ref{lem:q-localized-validation} to
\[
\mathcal P_k=\{(i,k):i\in[K]\}\cup\{(k,j):j\in\mathcal J_Q\}.
\]
This collection is fixed conditional on training and has at most $2K$
elements. Enlarge the logarithmic factor to $\log(12K/\delta)$, and set
\begin{align*}
r&=\sqrt{\frac{\log(12K/\delta)}{ng_b}},
\qquad \alpha=1-\sqrt\gamma,\quad\beta=1+\sqrt\gamma,\\
y&=\|\Delta_k\|_{2,d_{\pi,\gamma}},
\qquad \eta_s=C_0\sqrt{W_Q}(G_Q+\sigma_Q)r,\\
v_s&=C_0H_Q(G_Q+\sigma_Q)
\{\sqrt{W_Q}\rho_wr+W_Qr^2\}.
\end{align*}
On the resulting event, each empirical average in
\eqref{eq:app-q-score-branch-representation} differs from its fitted-weight
expectation by at most
$\eta_s\|h_{ij}\|_{2,d_{\pi,\gamma}}+v_s$.

Let $A=I-\gamma P_\pi$. The conditional residual identity
$\mu_i=A\Delta_i$ holds $d_{\pi,\gamma}$-almost surely.
Lemma~\ref{lem:q-refined-residual} gives
$\|Af\|_{2,d_{\pi,\gamma}}^2\le\beta\mathcal E(f)$.

For the selected pair, put
$z=\sqrt{\mathcal E(h_{\widehat i_Qk})}$.
Since $\Delta_{\widehat i_Q}=h_{\widehat i_Qk}+\Delta_k$, for either
$\tau\in\{3/4,5/4\}$,
\[
\|m_{\widehat i_Qk,\tau}\|_{2,d_{\pi,\gamma}}
=\|A(\Delta_k+\tau h_{\widehat i_Qk})\|_{2,d_{\pi,\gamma}}
\le\beta y+\frac54\sqrt\beta\,z.
\]
Combine \eqref{eq:app-q-weight-bias-residual}, the concentration
event, and the bounds for maxima and minima in
\eqref{eq:app-q-score-branch-representation}. Using
$\|h_{\widehat i_Qk}\|_{2,d_{\pi,\gamma}}\le z/\sqrt\alpha$ and
\eqref{eq:q-population-score-lower}, we obtain
\begin{align}
\max_j\widehat{\mathcal M}_{\widehat i_Qj}^{\mathrm{IW}}
\ge{}&\frac34z^2-
\left(\sqrt\beta\,y+\frac52H_Q\sqrt\beta\,\rho_w
+\frac{\eta_s}{\sqrt\alpha}\right)z
-2H_Q\beta\rho_wy-\frac32H_QG_Q\rho_w^2-v_s.
\label{eq:app-q-selected-score-lower}
\end{align}

For the comparator pair $(k,j)$, put
$u=\sqrt{\mathcal E(h_{kj})}$. For each $s\in\{-1,1\}$, use
$\tau=1-s/4$ in the empirical minimum. For this value,
\[
s\,m_{kj,\tau}=A(s\Delta_k-h_{kj}/4),\qquad
\|m_{kj,\tau}\|_{2,d_{\pi,\gamma}}
\le\beta y+\frac{\sqrt\beta}{4}u.
\]
Thus \eqref{eq:app-q-weight-bias-residual}, the same concentration event,
and \eqref{eq:app-q-refined-residual-linear} applied to both signs give
\[
\widehat{\mathcal M}_{kj}^{\mathrm{IW}}
\le\left(\sqrt\beta\,y+\frac{H_Q\sqrt\beta}{2}\rho_w
+\frac{\eta_s}{\sqrt\alpha}\right)u-\frac14u^2
+2H_Q\beta\rho_wy+\frac32H_QG_Q\rho_w^2+v_s.
\]
Maximizing this quadratic in $u\ge0$, and writing
\[
b_{\mathrm{up}}=\sqrt\beta\,y+\frac{H_Q\sqrt\beta}{2}\rho_w
+\frac{\eta_s}{\sqrt\alpha},
\qquad
b=\sqrt\beta\,y+3H_Q\sqrt\beta\,\rho_w
+\frac{\eta_s}{\sqrt\alpha},
\]
we obtain
\[
\max_j\widehat{\mathcal M}_{kj}^{\mathrm{IW}}
\le b_{\mathrm{up}}^2+2H_Q\beta\rho_wy+\frac32H_QG_Q\rho_w^2+v_s.
\]
The coefficient of $z$ in \eqref{eq:app-q-selected-score-lower} is at
most $b$. Moreover,
\[
b^2-b_{\mathrm{up}}^2
=\frac52H_Q\sqrt\beta\,\rho_w(b+b_{\mathrm{up}})
\ge4H_Q\beta\rho_wy.
\]
Empirical optimality consequently implies
\[
\frac34z^2\le bz+b^2+2\{\tfrac32H_QG_Q\rho_w^2+v_s\}.
\]
Solving the quadratic and using $\sqrt{a+b}\le\sqrt a+\sqrt b$ yields
\[
z\le2b+\frac{2\sqrt6}{3}\sqrt{\frac32H_QG_Q\rho_w^2+v_s}.
\]
The triangle inequality and coercivity now give
\begin{align*}
\|\Delta_{\widehat i_Q}\|_{2,d_{\pi,\gamma}}
&\le y+\frac z{\sqrt\alpha}\\
&\le\left(1+2\sqrt{\frac\beta\alpha}\right)y
+\frac{6H_Q\sqrt\beta}{\sqrt\alpha}\rho_w
+\frac{2\eta_s}{\alpha}
+\frac{2\sqrt6}{3}
\sqrt{\frac{\frac32H_QG_Q\rho_w^2+v_s}{\alpha}}.
\end{align*}
Since $\rho_wr\le(\rho_w+r)^2/4$ and
$\rho_w^2\vee r^2\le(\rho_w+r)^2$, we have
$\frac32H_QG_Q\rho_w^2+v_s\le V_Q(\rho_w+r)^2$.
Finally, $\beta\le2$ and $\alpha^{-1/2}\le\alpha^{-1}$ give
\[
\|\Delta_{\widehat i_Q}\|_{2,d_{\pi,\gamma}}
\le\left\{1+2\sqrt{\frac{1+\sqrt\gamma}{1-\sqrt\gamma}}\right\}y
+\frac{C_{Q,\mathrm{wt}}}{\sqrt{1-\sqrt\gamma}}\rho_w
+\frac{C_{Q,\mathrm{val}}}{1-\sqrt\gamma}r,
\]
The symmetric part of $A=I-\gamma P_\pi$ is positive definite by
\eqref{eq:app-q-residual-coercivity}, so $\sqrt{\mathcal E(\cdot)}$ is a norm.
The triangle inequality, $\mathcal E(\Delta_k)\le\beta y^2$, and the bound
for $z$ therefore also give
\begin{align}
\sqrt{\mathcal E(\Delta_{\widehat i_Q})}
&\le z+\sqrt\beta\,y\notag\\
&\le3\sqrt\beta\,y+6H_Q\sqrt\beta\,\rho_w
+\frac{2\eta_s}{\sqrt\alpha}
+\frac{2\sqrt6}{3}\sqrt{V_Q}(\rho_w+r)\notag\\
&\le C_E\left\{y+\rho_w+\frac r{\sqrt\alpha}\right\},
\label{eq:app-q-selected-energy}
\end{align}
where one may take
\[
C_E=\max\left\{
3\sqrt2,\quad
6\sqrt2H_Q+\frac{2\sqrt6}{3}\sqrt{V_Q},\quad
2C_0\sqrt{W_Q}(G_Q+\sigma_Q)+\frac{2\sqrt6}{3}\sqrt{V_Q}
\right\}.
\]
This constant depends only on the stated envelopes.
These bounds prove Theorem~\ref{thm:q-finite-sample-oracle}.
\end{proof}

\subsection{Unweighted PBV}
\label{app:q-unweighted-proofs}

\subsubsection{Population score bounds}
\label{app:q-unweighted-population-proofs}

Write $\Delta_i=Q_i-Q^\pi$, $\mu_i=\mathbb E(\varepsilon_i\mid X)$, and
$T_i^{\mathrm{unw}}=\max_{j\in\mathcal J_Q}\mathcal M_{ij}^{\mathrm{unw}}$.
For pairwise differences, define
$\mathcal B_\nu(f,g)=\langle f,(I-\gamma P_\pi)g\rangle_\nu$.
Then $\mu_i-\mu_j=(I-\gamma P_\pi)h_{ij}$, so
\begin{equation}
\mathcal M_{ij}^{\mathrm{unw}}
=|\langle h_{ij},\mu_i\rangle_\nu|
-\frac14|\mathcal B_\nu(h_{ij},h_{ij})|.
\label{eq:app-q-unweighted-score-representation}
\end{equation}

One-step concentrability controls the oracle candidate's mean Bellman
residual by its approximation error. Indeed, if $\nu P_\pi\ll\nu$ and
$C_{\nu,\pi}=\|\dd(\nu P_\pi)/\dd\nu\|_\infty<\infty$, conditional
Jensen's inequality gives
$\|P_\pi f\|_{2,\nu}\le\sqrt{C_{\nu,\pi}}\|f\|_{2,\nu}$.
The Bellman equation then yields
$\|\mu_k\|_{2,\nu}\le(1+\gamma\sqrt{C_{\nu,\pi}})\|\Delta_k\|_{2,\nu}$.

\begin{lemma}[Deterministic unweighted score bounds]
\label{lem:q-unweighted-score-bounds}
Suppose Condition~\ref{cond:q-unweighted-domain} holds. Fix $k\in[K]$
and $\kappa_\nu\in(0,1]$ such that
$|\mathcal B_\nu(h_{ik},h_{ik})|\ge\kappa_\nu\|h_{ik}\|_{2,\nu}^2$
for every $i\in\mathcal J_Q$. Set $b_k=\|\mu_k\|_{2,\nu}$. Then
\begin{equation}
T_i^{\mathrm{unw}}
\ge\mathcal M_{ik}^{\mathrm{unw}}
\ge\frac{3\kappa_\nu}{4}\|h_{ik}\|_{2,\nu}^2
-b_k\|h_{ik}\|_{2,\nu},
\label{eq:app-q-unweighted-score-lower}
\end{equation}
whereas, for every $j\in\mathcal J_Q$,
\begin{equation}
\mathcal M_{kj}^{\mathrm{unw}}
\le b_k\|h_{kj}\|_{2,\nu}
-\frac{\kappa_\nu}{4}\|h_{kj}\|_{2,\nu}^2,
\label{eq:app-q-unweighted-pair-upper}
\end{equation}
and consequently
\begin{equation}
T_k^{\mathrm{unw}}\le\frac{b_k^2}{\kappa_\nu}.
\label{eq:app-q-unweighted-score-upper}
\end{equation}
\end{lemma}

\begin{proof}
Using $\mu_i=\mu_k+(I-\gamma P_\pi)h_{ik}$ in
\eqref{eq:app-q-unweighted-score-representation}, the reverse triangle
inequality gives
\[
\mathcal M_{ik}^{\mathrm{unw}}
\ge\frac34|\mathcal B_\nu(h_{ik},h_{ik})|
-|\langle h_{ik},\mu_k\rangle_\nu|.
\]
Coercivity and Cauchy--Schwarz give
\eqref{eq:app-q-unweighted-score-lower}. With first index $k$,
\[
\mathcal M_{kj}^{\mathrm{unw}}
=|\langle h_{kj},\mu_k\rangle_\nu|
-\frac14|\mathcal B_\nu(h_{kj},h_{kj})|.
\]
Since $h_{kj}=-h_{jk}$, the same inequalities prove
\eqref{eq:app-q-unweighted-pair-upper}.
Finally, $b_ku-\kappa_\nu u^2/4\le b_k^2/\kappa_\nu$ for every $u\ge0$,
which proves \eqref{eq:app-q-unweighted-score-upper}.
\end{proof}

\begin{lemma}[Population bound for a general penalty coefficient]
\label{lem:q-unweighted-penalty-coefficient}
Suppose Condition~\ref{cond:q-unweighted-domain} holds, and let $k$ and
$\kappa_\nu$ satisfy the inequality in
Lemma~\ref{lem:q-unweighted-score-bounds}. For $\lambda\in(0,1)$, define
\[
\mathcal M_{ij}^{\mathrm{unw},\lambda}
:=|\langle h_{ij},\mu_i\rangle_\nu|
-\lambda|\mathcal B_\nu(h_{ij},h_{ij})|,
\qquad
\widetilde i_\lambda\in\argmin_{i\in[K]}
\max_{j\in\mathcal J_Q}\mathcal M_{ij}^{\mathrm{unw},\lambda}.
\]
Then
\begin{equation}
\|Q_{\widetilde i_\lambda}-Q^\pi\|_{2,\nu}
\le\|\Delta_k\|_{2,\nu}
+\frac{\|\mu_k\|_{2,\nu}}
{2\kappa_\nu\sqrt\lambda(1-\sqrt\lambda)}.
\label{eq:app-q-unweighted-general-penalty-bound}
\end{equation}
The coefficient of $\|\mu_k\|_{2,\nu}$ is uniquely minimized at
$\lambda=1/4$.
\end{lemma}

\begin{proof}
Write $u=\|h_{\widetilde i_\lambda,k}\|_{2,\nu}$ and
$b=\|\mu_k\|_{2,\nu}$. As in the proof of
Lemma~\ref{lem:q-unweighted-score-bounds},
\[
\mathcal M_{ik}^{\mathrm{unw},\lambda}
\ge(1-\lambda)\kappa_\nu\|h_{ik}\|_{2,\nu}^2
-b\|h_{ik}\|_{2,\nu},
\qquad
\mathcal M_{kj}^{\mathrm{unw},\lambda}
\le\frac{b^2}{4\lambda\kappa_\nu}.
\]
The defining optimality of $\widetilde i_\lambda$ therefore implies
$(1-\lambda)\kappa_\nu u^2-bu\le b^2/(4\lambda\kappa_\nu)$.
Solving this quadratic inequality gives
$u\le b/\{2\kappa_\nu\sqrt\lambda(1-\sqrt\lambda)\}$.
The triangle inequality proves
\eqref{eq:app-q-unweighted-general-penalty-bound}.
Since $\sqrt\lambda(1-\sqrt\lambda)
=1/4-(\sqrt\lambda-1/2)^2$, the coefficient is uniquely minimized at
$\lambda=1/4$.
\end{proof}

\begin{proof}[Proof of Theorem~\ref{thm:q-unweighted-population-oracle}]
Apply Lemma~\ref{lem:q-unweighted-penalty-coefficient} with $\lambda=1/4$
and use $\mu_k=Q_k-\mathcal B_\pi Q_k$.
\end{proof}

\begin{proof}[Proof of Proposition~\ref{prop:unweighted-unpenalized-counterexample}]
Take a single-action MDP on $\{-1,1\}^2$ whose transitions draw the next
state uniformly, independently of the current state. Let $\nu$ be uniform
and define $f_1(s)=s_1$, $f_2(s)=s_2$, and $f_3(s)=s_1s_2$.
These functions are orthonormal in $L^2(\nu)$ and satisfy $P_\pi f_l=0$.
With reward $r=f_3$, the true action-value function is $Q^\pi=f_3$.
Set $\epsilon=1/(2\rho^2)$ and $u=1/(2\rho)$, and choose
\[
Q_0=0,\qquad Q_1=f_3+\epsilon f_1,\qquad
Q_2=f_3+u f_2,\qquad Q_3=f_3+f_1.
\]
The candidates are bounded by $2$, and
$(I-\gamma P_\pi)h_{ij}=h_{ij}$, verifying the stated conditions.
Their errors are $1,\epsilon,u,1$, while their unpenalized absolute-moment
row maxima are
\[
T_0=1,\qquad T_1=\epsilon(1-\epsilon),\qquad T_2=u^2=\epsilon/2,\qquad T_3=1.
\]
Since $0<\epsilon<1/2$, the unique minimizer is $\widetilde i_0=2$.
Its error ratio is $u/\epsilon=\rho$, proving the claim.
\end{proof}

\begin{remark}[$L^2$-normalized pairwise critics]
\label{rem:normalized-pairwise-critics}

Proposition~\ref{prop:unweighted-unpenalized-counterexample} shows that
unnormalized pairwise-difference critics without penalization can fail
under candidate-class misspecification. A population oracle inequality
can also be obtained using $L^2(\nu)$-normalized pairwise differences,
but finite-sample guarantees require additional control of empirical
normalization.

To see this, define
$\psi_{ij}:=h_{ij}/\|h_{ij}\|_{2,\nu}$ when
$\|h_{ij}\|_{2,\nu}>0$, and $\psi_{ij}:=0$ otherwise.
Under Conditions~\ref{cond:q-pairwise-coercivity}
and~\ref{cond:q-unweighted-domain}, any minimizer
$\widetilde i_{\mathrm{norm}}$ of
$\max_j|\langle\psi_{ij},Q_i-\mathcal B_\pi Q_i\rangle_\nu|$
satisfies the population oracle bound in
Theorem~\ref{thm:q-unweighted-population-oracle}.
Indeed, take $k=k^\star$ and write
$y:=\|\Delta_k\|_{2,\nu}$,
$u:=\|h_{\widetilde i_{\mathrm{norm}},k}\|_{2,\nu}$, and
$b:=\|Q_k-\mathcal B_\pi Q_k\|_{2,\nu}$.
The score of candidate $k$ is at most $b$, whereas the score
of the selected candidate is at least $\kappa_\nu u-b$.
Optimality therefore gives $u\le 2b/\kappa_\nu$, and the
triangle inequality yields the stated bound.

Finite-sample guarantees with empirical normalization require control of
empirical norms relative to their population counterparts. Rescaling a
difference alone rescales its moment and norm equally; the difficulty is
concentration on rarely observed states. Sup-norm--$L^2$ interpolation
bounds quantify this concentration. For empirical standard-deviation
normalization, these bounds apply to centered critics.
Appendix~\ref{app:lstd-empirical-normalization} gives a concentration lemma,
examples for linear and smooth function classes, and an oracle inequality
for normalized-difference LSTD-Tournament.
\end{remark}

\subsubsection{Concentration of the unweighted scores}
\label{app:q-unweighted-validation-proof}

\begin{lemma}[Localized unweighted-score deviation]
\label{lem:q-unweighted-localized-validation}
Suppose Conditions~\ref{cond:validation-spectral-gap},
\ref{cond:q-bounded-signal}
and~\ref{cond:q-subgaussian-td} hold.
There is a universal constant
$C_0>0$ such that, for every $\delta\in(0,1)$ and every nonempty collection
$\mathcal P\subseteq[K]\times\mathcal J_Q$ fixed conditional on the training
sample, with probability at least $1-\delta$, simultaneously for all
$(i,j)\in\mathcal P$,
\begin{equation}
\left|
\widehat{\mathcal M}_{ij}^{\mathrm{unw}}
-
\mathcal M_{ij}^{\mathrm{unw}}
\right|
\le
C_0\left(G_Q+\sigma_Q\right)
\left\{
\|h_{ij}\|_{2,\nu}
\sqrt{\frac{\log(6|\mathcal P|/\delta)}{ng_b}}
+
H_Q\frac{\log(6|\mathcal P|/\delta)}{ng_b}
\right\}.
\label{eq:q-unweighted-localized-validation}
\end{equation}
\end{lemma}

\begin{proof}
Condition on the training sample, and let $\mathbb P_{n,Q}$ denote the empirical
average over the $Q$-validation transitions. For a fixed ordered pair $(i,j)$,
write
\begin{align*}
\widehat L_{ij}
&:=
\mathbb P_{n,Q}\!\left[h_{ij}(X)\varepsilon_i\right],
&
L_{ij}
&:=
\mathbb E_\nu\!\left[h_{ij}(X)\varepsilon_i\right],
\\
\widehat A_{ij}
&:=
\mathbb P_{n,Q}\!\left[
h_{ij}(X)\{h_{ij}(X)-\gamma(\pi h_{ij})(S')\}
\right],
&
A_{ij}
&:=
\mathbb E_\nu\!\left[
h_{ij}(X)\{h_{ij}(X)-\gamma(\pi h_{ij})(S')\}
\right].
\end{align*}
\[
\widehat{\mathcal M}_{ij}^{\mathrm{unw}}
=|\widehat L_{ij}|-\frac14|\widehat A_{ij}|,
\qquad
\mathcal M_{ij}^{\mathrm{unw}}
=|L_{ij}|-\frac14|A_{ij}|.
\]
Since the absolute-value map is $1$-Lipschitz,
\begin{equation}
\left|
\widehat{\mathcal M}_{ij}^{\mathrm{unw}}
-\mathcal M_{ij}^{\mathrm{unw}}
\right|
\le
|\widehat L_{ij}-L_{ij}|
+\frac14|\widehat A_{ij}-A_{ij}|.
\label{eq:app-q-unweighted-clipped-decomposition}
\end{equation}

Set
\[
\mu_i(X):=\mathbb E[\varepsilon_i\mid X],
\qquad
\xi_i:=\varepsilon_i-\mu_i(X).
\]
By the validation setup in Section~\ref{sec:offline-data-setup}, the
conditional law of $(R_t,S_t')$ given $\mathcal F_t$ is the MDP kernel given
$X_t$. Conditions~\ref{cond:q-bounded-signal}
and~\ref{cond:q-subgaussian-td} therefore verify the conditions of
Lemma~\ref{lem:trajectory-bernstein}(ii) for
$h_{ij}(X)\{\mu_i(X)+\xi_i\}$ with
\[
B=H_Q,
\qquad
D=G_Q,
\qquad
\sigma=\sigma_Q,
\qquad
v=\|h_{ij}\|_{2,\nu}.
\]
Thus, outside an event of probability $6e^{-u}$,
\begin{equation}
|\widehat L_{ij}-L_{ij}|
\le
C_{\mathrm M}(G_Q+\sigma_Q)
\left\{
\|h_{ij}\|_{2,\nu}r_{g_b}(u)
+H_Qr_{g_b}(u)^2
\right\}.
\label{eq:app-q-unweighted-moment-deviation}
\end{equation}

The penalty integrand can be written as
\[
h_{ij}(X)\{h_{ij}(X)-\gamma(\pi h_{ij})(S')\}
=h_{ij}(X)(\varepsilon_i-\varepsilon_j).
\]
Its conditional mean factor is $\mu_i-\mu_j$, bounded by $2G_Q$.
Its centered TD residual is $\xi_i-\xi_j$. Conditional Cauchy--Schwarz gives
\[
\mathbb E\!\left[
\left.\exp\left\{\frac{|\xi_i-\xi_j|}{2\sigma_Q}\right\}
\right|X\right]
\le
\left\{\mathbb E\!\left[
\left.e^{|\xi_i|/\sigma_Q}\right|X\right]
\mathbb E\!\left[
\left.e^{|\xi_j|/\sigma_Q}\right|X\right]\right\}^{1/2}
\le2.
\]
Lemma~\ref{lem:trajectory-bernstein}(ii), with
$a=h_{ij}$, $D=2G_Q$, and $\sigma=2\sigma_Q$, therefore gives,
outside an event of probability $6e^{-u}$,
\begin{equation}
|\widehat A_{ij}-A_{ij}|
\le
2C_{\mathrm M}(G_Q+\sigma_Q)
\left\{
\|h_{ij}\|_{2,\nu}r_{g_b}(u)
+H_Qr_{g_b}(u)^2
\right\}.
\label{eq:app-q-unweighted-penalty-deviation}
\end{equation}

Apply \eqref{eq:app-q-unweighted-moment-deviation}
and~\eqref{eq:app-q-unweighted-penalty-deviation} with
$u=\log(12|\mathcal P|/\delta)$ and take a union bound over the two deviations
and all $(i,j)\in\mathcal P$. Combining the resulting bounds with
\eqref{eq:app-q-unweighted-clipped-decomposition} proves
\eqref{eq:q-unweighted-localized-validation} after increasing the universal
constant $C_0$ to replace $\log(12|\mathcal P|/\delta)$ by
$\log(6|\mathcal P|/\delta)$; their ratio is bounded by
$\log(12)/\log(6)$ because $|\mathcal P|/\delta\ge1$.
\end{proof}

\subsubsection{Finite-sample oracle inequality}
\label{app:q-unweighted-finite-sample-proof}

Let $C_0$ be the universal constant in
Lemma~\ref{lem:q-unweighted-localized-validation}, and define
\begin{align*}
C_{Q,\mathrm{unw}}
:=
2C_0\left(G_Q+\sigma_Q\right)
+
\frac{2}{3}
\sqrt{
6C_0H_Q\left(G_Q+\sigma_Q\right)
}.
\end{align*}

\begin{proof}[Proof of Theorems~\ref{thm:main-unweighted-pbv}
and~\ref{thm:q-unweighted-finite-sample-oracle}]
\emph{Localized event.}
Fix a comparator $k$ satisfying the inequality in Lemma~\ref{lem:q-unweighted-score-bounds}, and define
\[
\mathcal P_k^{\mathrm{unw}}
:=
\{(i,k):1\le i\le K\}
\cup
\{(k,j):j\in\mathcal J_Q\}.
\]
The collection $\mathcal P_k^{\mathrm{unw}}$ is fixed conditional on the
training sample and has at most $2K$ elements. By
Lemma~\ref{lem:q-unweighted-localized-validation}, with conditional
probability at least $1-\delta$, the localized bound holds on
$\mathcal P_k^{\mathrm{unw}}$ with its logarithmic factor enlarged to
$\log(12K/\delta)$. On this event, abbreviate
\[
u:=\|h_{\widehat i_Q^{\mathrm{unw}},k}\|_{2,\nu},
\qquad
y:=\|\Delta_k\|_{2,\nu},
\qquad
\mathfrak r_Q:=\sqrt{\frac{\log(12K/\delta)}{ng_b}}.
\]
Also set
\[
\eta:=C_0\left(G_Q+\sigma_Q\right)\mathfrak r_Q,
\qquad
v:=C_0H_Q\left(G_Q+\sigma_Q\right)\mathfrak r_Q^2.
\]

\emph{Empirical comparison.}
Set $b_k=\|\mu_k\|_{2,\nu}$. For the selected candidate, by
\eqref{eq:app-q-unweighted-score-lower} and the localized deviation bound,
\begin{equation}
\max_j\widehat{\mathcal M}_{\widehat i_Q^{\mathrm{unw}},j}^{\mathrm{unw}}
\ge
\frac{3\kappa_\nu}{4}u^2
-
\{b_k+\eta\}u
-v.
\label{eq:app-q-unweighted-empirical-lower}
\end{equation}
For the comparator $k$, each population score is bounded by
\eqref{eq:app-q-unweighted-pair-upper}, and the corresponding empirical
deviation is bounded on the localized event. Thus, for every $j$,
\[
\widehat{\mathcal M}_{kj}^{\mathrm{unw}}
\le
\{b_k+\eta\}
\|h_{kj}\|_{2,\nu}
-
\frac{\kappa_\nu}{4}\|h_{kj}\|_{2,\nu}^2
+v.
\]
Maximizing the right-hand side over the nonnegative value of
$\|h_{kj}\|_{2,\nu}$ yields
\begin{equation}
\max_j\widehat{\mathcal M}_{kj}^{\mathrm{unw}}
\le
\frac{\{b_k+\eta\}^2}{\kappa_\nu}
+v.
\label{eq:app-q-unweighted-empirical-upper}
\end{equation}

\emph{Solving the quadratic inequality.}
By the defining optimality of $\widehat i_Q^{\mathrm{unw}}$ and
\eqref{eq:app-q-unweighted-empirical-lower}--\eqref{eq:app-q-unweighted-empirical-upper},
\begin{equation}
\frac{3\kappa_\nu}{4}u^2
-
\{b_k+\eta\}u
-
\frac{\{b_k+\eta\}^2}{\kappa_\nu}
-2v
\le0.
\label{eq:app-q-unweighted-empirical-quadratic}
\end{equation}
Solving \eqref{eq:app-q-unweighted-empirical-quadratic} and using
$\sqrt{a+b}\le\sqrt a+\sqrt b$, we obtain
\begin{align*}
u
\le
\frac{2\{b_k+\eta\}}{\kappa_\nu}
+
\frac{2\sqrt6}{3}\sqrt{\frac{v}{\kappa_\nu}}.
\end{align*}
Finally, the triangle inequality, $\kappa_\nu\le1$, and the definitions of
$\eta$, $v$, and $C_{Q,\mathrm{unw}}$ give
\[
\|\Delta_{\widehat i_Q^{\mathrm{unw}}}\|_{2,\nu}
\le y+\frac{2b_k}{\kappa_\nu}
+\frac{C_{Q,\mathrm{unw}}}{\kappa_\nu}\mathfrak r_Q.
\]
This proves the bound for any fixed comparator satisfying the assumed
inequality, since $b_k=\|Q_k-\mathcal B_\pi Q_k\|_{2,\nu}$.
Taking $k=k^\star$ proves \eqref{eq:q-unweighted-finite-sample-oracle} and the main-text guarantee.
\end{proof}

\begin{proof}[Proof of Corollary~\ref{cor:q-unweighted-stationary}]
All norms and inner products in this proof are under $\nu$.
Stationarity and Jensen's inequality give $\|P_\pi f\|\le\|f\|$.
Let $A=I-\gamma P_\pi$ and $E_\nu(f)=\langle f,Af\rangle$.
Then $E_\nu(f)\ge\alpha\|f\|^2$, where $\alpha=1-\gamma$.
Lemma~\ref{lem:q-refined-residual}, with $c=\gamma$, gives
$|\langle f,Ag\rangle|\le\sqrt{\beta E_\nu(f)}\|g\|$, where
$\beta=1+\gamma$.

Let $k$ minimize $\|Q_i-Q^\pi\|$, and write $y=\|Q_k-Q^\pi\|$.
Use the localized event from the proof of
Theorem~\ref{thm:q-unweighted-finite-sample-oracle}.
With $r=\{\log(12K/\delta)/(ng_b)\}^{1/2}$, write
$\eta=C_0(G_Q+\sigma_Q)r$ and $v=C_0H_Q(G_Q+\sigma_Q)r^2$.
The Bellman equation gives $\mu_i=A\Delta_i$. Since every population
energy is nonnegative, the population score equals
$|\langle h_{ij},A\Delta_i\rangle|-E_\nu(h_{ij})/4$.
The reverse triangle inequality and the operator bound yield the same
population score bounds as in Lemma~\ref{lem:q-population-score-bounds}.
For $z=\sqrt{E_\nu(h_{\widehat i_Q^{\mathrm{unw}},k})}$ and
$b=\sqrt\beta\,y+\eta/\sqrt\alpha$, the operator bound and localized
event therefore imply
\[
\max_j\widehat{\mathcal M}_{\widehat i_Q^{\mathrm{unw}},j}^{\mathrm{unw}}
\ge\frac34z^2-bz-v,
\qquad
\max_j\widehat{\mathcal M}_{kj}^{\mathrm{unw}}
\le\sup_{u\ge0}\{bu-u^2/4+v\}=b^2+v.
\]
Empirical optimality and the quadratic formula give
$z\le2b+(2\sqrt6/3)\sqrt v$.
By the triangle inequality and coercivity,
\[
\|\Delta_{\widehat i_Q^{\mathrm{unw}}}\|
\le y+\frac z{\sqrt\alpha}
\le\left(1+2\sqrt{\frac\beta\alpha}\right)y
+\frac{2\eta}{\alpha}
+\frac{2\sqrt6}{3}\sqrt{\frac v\alpha}.
\]
Substituting $\eta,v$ and using $\sqrt\alpha\ge\alpha$ yields the
asserted bound with the stated $C_{Q,\mathrm{unw}}$.
\end{proof}

\input{sections/q_fitted_weight_proofs.tex}

\section{Proofs for joint selection and doubly robust OPE}
\label{app:joint-dr-proofs}

\subsection{Proof of the A-PBV and IW-PBV guarantee}
\label{app:joint-ta-pbv-proof}

\begin{proof}[Proof of Theorem~\ref{thm:sequential-ta-pbv}]
Apply Theorem~\ref{thm:w-finite-sample-oracle} on the shared validation
trajectory with failure probability $\delta/2$. When the initial moments
are estimated empirically, $m=n$, $g_b\le1$, and
$(1-\gamma)^{-1/2}\le(1-\gamma)^{-1}$ allow its initial-sample term
to be absorbed into the validation term. When these moments are known
exactly, the initial-sample term is absent. This gives
\eqref{eq:main-sequential-w-bound}, where $C_{\mathrm{ratio}}$ depends
only on $C_w$ and $\overline w$.

For each $a\in[M]$, let $\widehat i_Q(a)$ be the $Q$-index selected when
the second-stage score uses $w_a$. Conditional on training, every $w_a$
is fixed and independent of the validation trajectory, and
Condition~\ref{cond:q-fitted-weight-envelope} holds with
$W_Q=\overline w$. Apply
Theorem~\ref{thm:q-finite-sample-oracle} to each $w_a$, with failure
probability $\delta/(2M)$. A union bound gives an event of probability
at least $1-\delta/2$ on which all $M$ conclusions hold. Their validation
rate is
\[
\sqrt{\frac{\log\{12K/(\delta/(2M))\}}{ng_b}}
=\sqrt{\frac{\log(24MK/\delta)}{ng_b}}.
\]
On this event, the bound also holds for $a=\widehat i_w$, whose selected
$Q$-function is exactly $\widehat Q$. Substitute
\eqref{eq:main-sequential-w-bound} into its Hellinger weight-error term
and use
\[
1-\sqrt\gamma\ge\frac{1-\gamma}{2},
\qquad
1+2\sqrt{\frac{1+\sqrt\gamma}{1-\sqrt\gamma}}
\le\frac5{\sqrt{1-\gamma}}.
\]
Absorbing the resulting universal factors into $C_{\mathrm{IW}}$ gives
\eqref{eq:main-sequential-q-bound}, with $C_{\mathrm{IW}}$ depending only
on $C_w,\overline w,H_Q,G_Q$, and $\sigma_Q$.
The two events intersect with probability at least $1-\delta$.
\end{proof}

\subsection{Proof of the doubly robust oracle bound}
\label{app:sequential-pbv-dr-proof}

\begin{proof}[Proof of Corollary~\ref{cor:sequential-pbv-dr}]
For this proof, set $\ell=\log\{12(K\vee M)/\delta\}$, so that
$\rho_\delta=\sqrt{\ell/(ng_b)}$.

\emph{Selection events.}
Conditions~\ref{cond:sequential-fold-laws}
and~\ref{cond:sequential-selector-regularity} permit applying
Theorem~\ref{thm:w-finite-sample-oracle} to the shared selection fold with
failure probability $\delta/3$. Since
$\log(72M/\delta)\le2\ell$, there is an event $\mathcal E_w$ of
conditional probability at least $1-\delta/3$ on which
\begin{equation}
H_\nu(\widehat w,w^\pi)
\lesssim\frac{\epsilon_w+\rho_\delta}{1-\gamma}.
\label{eq:app-sequential-w-event}
\end{equation}

For each $a\in[M]$, let $\widehat i_Q(a)$ be the index selected on
$\mathcal D_{\mathrm{sel}}$ when the second-stage score uses $w_a$.
Conditional on training, each $w_a$ is fixed. Condition~\ref{cond:w-candidate-bounds}
gives $0\le w_a\le\overline w$.
Apply the concentration event used in the proof of
Theorem~\ref{thm:q-finite-sample-oracle} to each of these $M$ fixed weights,
with failure probability $\delta/(3M)$ each. Since
$\log(36MK/\delta)\le2\ell$, a union bound gives an event
$\mathcal E_Q$ of conditional probability at least $1-\delta/3$ on which
\eqref{eq:app-q-selected-energy} holds simultaneously for all $a$.
It therefore holds for the selected index $a=\widehat i_w$:
\begin{equation}
\sqrt{\mathcal E(\widehat Q-Q^\pi)}
\lesssim \epsilon_Q+H_\nu(\widehat w,w^\pi)
+\frac{\rho_\delta}{\sqrt{1-\gamma}}.
\label{eq:app-sequential-q-energy-event}
\end{equation}
Here $1-\sqrt\gamma\ge(1-\gamma)/2$ is used only to express the validation
term in $1-\gamma$.

\emph{Doubly robust remainder.}
We next express the population value error as a product of the two nuisance
errors.
For fixed candidates $Q\in\{Q_1,\ldots,Q_K\}$ and
$w\in\{w_1,\ldots,w_M\}$, define the population functional
\[
\Psi(Q,w)
:=
\mathbb E_{d_0}[Q(X)]
+
\frac{1}{1-\gamma}
\mathbb E_\nu\!\left[
w(X)\{R+\gamma(\pi Q)(S')-Q(X)\}
\right].
\]
Let $\Delta_Q=Q-Q^\pi$ and
$\mu_Q(X):=\mathbb E[Q(X)-R-\gamma(\pi Q)(S')\mid X]$.
Condition~\ref{cond:q-bounded-signal} gives $|\mu_Q|\le G_Q$,
while the Bellman equation \eqref{eq:primal-bellman} gives
$\mu_Q=(I-\gamma P_\pi)\Delta_Q$ under $d_{\pi,\gamma}$.
By Lemma~\ref{lem:q-target-second-moments}, the occupancy identity applies
to $\Delta_Q$. Hence
\begin{align*}
\Psi(Q,w)-V^\pi
&=\mathbb E_{d_0}\Delta_Q
-\frac{1}{1-\gamma}\mathbb E_\nu[w\mu_Q]\\
&=\frac{1}{1-\gamma}\mathbb E_\nu[(w^\pi-w)\mu_Q].
\end{align*}
The expectations are finite because $\mu_Q$ and $w$ are bounded,
$\mathbb E_\nu w^\pi=1$, and $\Delta_Q\in L^2(d_0)$.
In particular, the bias of $\Psi$ vanishes if either $Q=Q^\pi$ or
$w=w^\pi$; hence, $\Psi$ is doubly robust.

Put $z=\sqrt w-\sqrt{w^\pi}$. The identity
$w-w^\pi=2\sqrt{w^\pi}z+z^2$, weighted Cauchy--Schwarz, and
$|\mu_Q|\le G_Q$ give
\begin{align}
|\Psi(Q,w)-V^\pi|
&\le\frac{2H_\nu(w,w^\pi)}{1-\gamma}
\|\mu_Q\|_{2,d_{\pi,\gamma}}
+\frac{G_Q}{1-\gamma}H_\nu^2(w,w^\pi)\notag\\
&\le\frac{2\sqrt{1+\sqrt\gamma}}{1-\gamma}
H_\nu(w,w^\pi)\sqrt{\mathcal E(Q-Q^\pi)}
+\frac{G_Q}{1-\gamma}H_\nu^2(w,w^\pi),
\label{eq:app-sequential-dr-product-bound}
\end{align}
where $A=I-\gamma P_\pi$, and the last step uses
$\mu_Q=A(Q-Q^\pi)$ under $d_{\pi,\gamma}$ and
$\|Af\|_{2,d_{\pi,\gamma}}^2\le(1+\sqrt\gamma)\mathcal E(f)$.
For $(Q,w)=(\widehat Q,\widehat w)$, set $h=1-\gamma$ and
$\rho_w=H_\nu(\widehat w,w^\pi)$. On
$\mathcal E_w\cap\mathcal E_Q$, equations
\eqref{eq:app-sequential-w-event},
\eqref{eq:app-sequential-q-energy-event}, and
\eqref{eq:app-sequential-dr-product-bound} imply
\begin{align}
|\Psi(\widehat Q,\widehat w)-V^\pi|
&\lesssim\frac1h
\left\{\rho_w\epsilon_Q+\rho_w^2
+\frac{\rho_w\rho_\delta}{\sqrt h}\right\}\notag\\
&\lesssim
\frac{(\epsilon_w+\rho_\delta)\epsilon_Q}{h^2}
+\frac{(\epsilon_w+\rho_\delta)^2}{h^3}
+\frac{(\epsilon_w+\rho_\delta)\rho_\delta}{h^{5/2}}\notag\\
&\lesssim
\frac{\epsilon_w\epsilon_Q}{h^2}
+\frac{\epsilon_w^2}{h^3}
+\left\{\frac{\epsilon_Q}{h^2}
+\frac{\epsilon_w}{h^3}\right\}\rho_\delta
+\frac{\rho_\delta^2}{h^3}.
\label{eq:app-sequential-dr-expanded-remainder}
\end{align}
The last inequality uses $h\le1$ and
$(\epsilon_w+\rho_\delta)\rho_\delta\le(\epsilon_w+\rho_\delta)^2$.

\emph{Evaluation error.}
Condition on the training sample, so that all $MK$ candidate pairs are fixed.
When the initial-state expectations are estimated empirically,
Bernstein's inequality for subexponential variables and a union bound over
$i\in[K]$ give, with
conditional probability at least $1-\delta/6$,
\begin{equation}
\max_{i\in[K]}
|\mathbb P_{n,0,V}Q_i-\mathbb E_{d_0}Q_i|
\lesssim
\rho_\delta+\rho_\delta^2.
\label{eq:app-sequential-initial-deviation}
\end{equation}
Here Condition~\ref{cond:sequential-initial-evaluation} supplies the
subexponential moment bound,
and $\log(12K/\delta)\le\ell$. When the initial-state expectations are
known exactly, the left-hand side is zero.

For each $(a,i)\in[M]\times[K]$, set
\[
Y_{ai,t}:=
 w_a(X_t^V)
\{R_t^V+\gamma(\pi Q_i)(S_t^{\prime,V})-Q_i(X_t^V)\}.
\]
Writing $\mu_i(X)=\mathbb E(\varepsilon_i\mid X)$ and
$\xi_i=\varepsilon_i-\mu_i(X)$, apply
Lemma~\ref{lem:trajectory-bernstein}(ii) with
$a=w_a$, $d=-\mu_i$, $\zeta=-\xi_i$, $B=\overline w$,
$D=G_Q$, $\sigma=\sigma_Q$, and $v=\overline w$.
Conditions~\ref{cond:sequential-fold-laws}--\ref{cond:sequential-selector-regularity}
verify the spectral-gap, envelope, and conditional noise requirements.
Taking $u=\log(36MK/\delta)$ and a union bound over the $MK$ pairs gives,
with conditional probability at least $1-\delta/6$,
\begin{equation}
\max_{a\in[M],\,i\in[K]}
|\mathbb P_{n,V}Y_{ai}-\mathbb E Y_{ai}|
\lesssim
\rho_\delta+\rho_\delta^2,
\label{eq:app-sequential-transition-deviation}
\end{equation}
because $\log(36MK/\delta)\le2\ell$. Both uniform bounds therefore apply
to the selected pair, even when the selection and evaluation data coincide.
On the two corresponding events, by
\eqref{eq:app-sequential-initial-deviation} and
\eqref{eq:app-sequential-transition-deviation},
\[
|\widehat V_{\mathrm{DR}}-\Psi(\widehat Q,\widehat w)|
\lesssim
\frac{\rho_\delta+\rho_\delta^2}{1-\gamma}.
\]
Adding this evaluation error to the population remainder in
\eqref{eq:app-sequential-dr-expanded-remainder}, we obtain
\eqref{eq:app-sequential-dr-oracle}. Here $1-\gamma\le1$ is used only to
absorb $\rho_\delta^2/(1-\gamma)$ into
$\rho_\delta^2/(1-\gamma)^3$. By a union bound over
$\mathcal E_w$, $\mathcal E_Q$, and the two final evaluation events, the total
failure probability is at most
$\delta/3+\delta/3+\delta/6+\delta/6=\delta$, which proves the corollary.
\end{proof}

\clearpage
\section{Unnormalized LSTD-Tournament under misspecification}
\label{app:lstd-misspecification}

We describe LSTDQ and establish a misspecification oracle inequality for
the LSTD-Tournament selector of \citet[Equation~(10)]{liu2025modelselection}.
Their realizable guarantee bounds policy-value error through a uniform
value-function bound; here we study $L^2(\nu)$ error relative to the best candidate.

\subsection{Algorithm and oracle guarantee}

For a feature map $\phi:\mathcal S\times\mathcal A\to\mathbb R^d$,
least-squares temporal-difference $Q$-learning (LSTDQ) estimates the linear
approximation $Q_\theta=\phi^\top\theta$ by solving empirical Bellman
moment equations \citep{lagoudakisParr2003LSPI}:
\[
\widehat A
:=\mathbb P_{n,Q}\!\left[
\phi(X)\{\phi(X)-\gamma(\pi\phi)(S')\}^\top
\right],
\qquad
\widehat b:=\mathbb P_{n,Q}[\phi(X)R],
\qquad
\widehat\theta=\widehat A^{-1}\widehat b,
\]
when $\widehat A$ is invertible. Here $\mathbb P_{n,Q}$ denotes the
validation-sample average, and $\pi$ acts on each coordinate of $\phi$.
LSTD-Tournament uses $\phi_{ij}:=(Q_i,Q_j)^\top$ for each candidate pair.
For candidate $i$, the corresponding LSTD moment score is
$\max_{\ell\in\{i,j\}}|\mathbb P_{n,Q}[Q_\ell(X)\varepsilon_i]|$,
where $\varepsilon_i:=Q_i(X)-R-\gamma(\pi Q_i)(S')$.
Taking the largest pairwise score gives the unnormalized selector
\begin{equation}
\widehat T_i
:=\max_{j\in[K]}|\widehat m_{ij}|,
\qquad
\widehat m_{ij}:=\mathbb P_{n,Q}[Q_j(X)\varepsilon_i],
\qquad
\widehat i_{\mathrm{LT}}\in\argmin_{i\in[K]}\widehat T_i.
\label{eq:lstd-selector}
\end{equation}
The criterion uses candidate functions as critics, absolute moments, and
no critic penalty. It requires only empirical moments, without inverting
the pairwise empirical matrices.

Condition on the training sample and retain the validation setup of
Section~\ref{sec:offline-data-setup} and
Condition~\ref{cond:validation-spectral-gap}. Assume the conditional TD
mean bound $G_Q$ in Condition~\ref{cond:q-bounded-signal} and the TD noise
bound in Condition~\ref{cond:q-subgaussian-td}, both for $i\in[K]$.
Assume also $Q^\pi\in L^2(\nu)$ and $\mathbb E_\nu|R|<\infty$.
Set
\[
\mu_i:=Q_i-\mathcal B_\pi Q_i,
\qquad
m_{ij}:=\langle Q_j,\mu_i\rangle_\nu,
\qquad
T_i:=\max_{j\in[K]}|m_{ij}|.
\]
Let $K\ge2$ and let
$k^\star\in\argmin_{i\in[K]}\|Q_i-Q^\pi\|_{2,\nu}$ index a best candidate.
In addition to the existing conditions, impose the following:
\begin{enumerate}[label=\textbf{(LT\arabic*)}, ref=LT\arabic*]
\item \label{cond:lstd-candidate-envelope}
\textit{Bounded candidate functions.}
There is a finite $B_Q>0$ such that
$\max_{i\in[K]}|Q_i(X)|\le B_Q$, $\nu$-almost surely.

\item \label{cond:lstd-invertibility}
\textit{LSTD moments along oracle-comparison directions.}
For $\phi_{ik^\star}:=(Q_i,Q_{k^\star})^\top$, define
\[
A_{ik^\star}
:=\mathbb E_\nu\!\left[
\phi_{ik^\star}(X)
\{\phi_{ik^\star}(X)-\gamma(P_\pi\phi_{ik^\star})(X)\}^\top
\right].
\]
There is a constant $s_{\mathrm{LT}}(k^\star)>0$ such that
\begin{equation}
\sum_{\ell\in\{i,k^\star\}}
\left\langle Q_\ell,(I-\gamma P_\pi)h_{ik^\star}\right\rangle_\nu^2
\ge 2s_{\mathrm{LT}}(k^\star)^2,
\qquad
\kappa_{\mathrm{LT}}(k^\star)
:=\frac{s_{\mathrm{LT}}(k^\star)}{2B_Q^2}.
\label{eq:lstd-conditioning}
\end{equation}
The inequality is required for every $i$ with
$\|h_{ik^\star}\|_{2,\nu}>0$.
\end{enumerate}
The full singular-value bound
$\min_{i\ne k^\star}\sigma_{\min}(A_{ik^\star})
\ge s_{\mathrm{LT}}(k^\star)>0$, analogous to that assumed by
\citet{liu2025modelselection}, is sufficient for
Condition~\ref{cond:lstd-invertibility}. Indeed, for
$v=(1,-1)^\top$, the left-hand side of
\eqref{eq:lstd-conditioning} equals
\[
\|A_{ik^\star}v\|_2^2
\ge \sigma_{\min}(A_{ik^\star})^2\|v\|_2^2
\ge 2s_{\mathrm{LT}}(k^\star)^2.
\]
Our condition is weaker, requiring only that the Bellman
moments along the candidate--oracle difference do not both vanish.
In particular, it allows the pairwise LSTD matrices to be singular.

\begin{theorem}[LSTD-Tournament oracle inequality under misspecification]
\label{thm:lstd-misspecification}
Under the conditions above, for every $\delta\in(0,1)$, conditional on
training, with probability at least $1-\delta$, the selector in
\eqref{eq:lstd-selector} satisfies
\begin{align}
\|Q_{\widehat i_{\mathrm{LT}}}-Q^\pi\|_{2,\nu}
\le{}&
\|Q_{k^\star}-Q^\pi\|_{2,\nu}
+\frac{2}{\kappa_{\mathrm{LT}}(k^\star)}
\|Q_{k^\star}-\mathcal B_\pi Q_{k^\star}\|_{2,\nu}
\nonumber\\
&+\frac{2C_{\mathrm M}(G_Q+\sigma_Q)}
{\kappa_{\mathrm{LT}}(k^\star)}
\{r_{n,\delta}+r_{n,\delta}^2\},
\qquad
r_{n,\delta}:=\sqrt{\frac{\log(18K/\delta)}{ng_b}},
\label{eq:lstd-misspecification}
\end{align}
where $C_{\mathrm M}$ is the universal constant in
Lemma~\ref{lem:trajectory-bernstein}. Every population minimizer
$\widetilde i_{\mathrm{LT}}\in\argmin_i T_i$ satisfies the same bound
without the validation term.
\end{theorem}

If $ng_b\ge\log(18K/\delta)$, the validation term in
\eqref{eq:lstd-misspecification} is at most
$\{4C_{\mathrm M}(G_Q+\sigma_Q)/\kappa_{\mathrm{LT}}(k^\star)\}
\sqrt{\log(18K/\delta)/(ng_b)}$.
A multiplicative best-candidate bound additionally requires
$\|Q_{k^\star}-\mathcal B_\pi Q_{k^\star}\|_{2,\nu}
\le L_\nu\|Q_{k^\star}-Q^\pi\|_{2,\nu}$.
Under this condition, the first two terms become
\[
\left\{1+\frac{2L_\nu}{\kappa_{\mathrm{LT}}(k^\star)}\right\}
\min_i\|Q_i-Q^\pi\|_{2,\nu}.
\]
One-step concentrability with
$C_{\nu,\pi}=\|\dd(\nu P_\pi)/\dd\nu\|_\infty<\infty$ is sufficient,
with $L_\nu=1+\gamma\sqrt{C_{\nu,\pi}}$, by Jensen's inequality and
$Q_{k^\star}-\mathcal B_\pi Q_{k^\star}
=(I-\gamma P_\pi)(Q_{k^\star}-Q^\pi)$.

\subsection{Moment identification and validation deviations}

Throughout the proofs below, write $k=k^\star$.

\begin{lemma}[LSTD moment identification]
\label{lem:lstd-identification}
Under Conditions~\ref{cond:lstd-candidate-envelope}
and~\ref{cond:lstd-invertibility}, for every $i\ne k$,
\begin{equation}
\|Q_i-Q_k\|_{2,\nu}
\le\frac{1}{\kappa_{\mathrm{LT}}(k)B_Q}
\max_{\ell\in\{i,k\}}|m_{i\ell}-m_{k\ell}|.
\label{eq:lstd-identification}
\end{equation}
\end{lemma}

\begin{proof}
Since $Q_j-\gamma P_\pi Q_j=\mu_j+r\in L^1(\nu)$ and the candidates
are bounded, all matrix and moment entries below are finite.
If $\|Q_i-Q_k\|_{2,\nu}=0$, the inequality holds directly.
Otherwise, the difference is covered by Condition~\ref{cond:lstd-invertibility}.
Write $B=B_Q$ and $s=s_{\mathrm{LT}}(k)$. For $v=(1,-1)^\top$,
linearity gives
\[
A_{ik}v
=\mathbb E_\nu[\phi_{ik}(I-\gamma P_\pi)(Q_i-Q_k)]
=\begin{pmatrix}
m_{ii}-m_{ki}\\
m_{ik}-m_{kk}
\end{pmatrix}.
\]
The candidate bound and $\|A_{ik}v\|_2\ge s\|v\|_2$ imply
\[
\|Q_i-Q_k\|_{2,\nu}
\le\sqrt2 B\|v\|_2
\le\frac{\sqrt2 B}{s}\|A_{ik}v\|_2
\le\frac{2B}{s}\|A_{ik}v\|_\infty.
\]
Substituting \eqref{eq:lstd-conditioning} proves
\eqref{eq:lstd-identification}.
\end{proof}

\begin{lemma}[Validation moment deviations]
\label{lem:lstd-deviations}
Under the sampling and TD conditions of
Theorem~\ref{thm:lstd-misspecification} and
Condition~\ref{cond:lstd-candidate-envelope}, define
\[
\mathcal P_k
:=\{(k,j):j\in[K]\}
\cup\{(i,i),(i,k):i\ne k\}.
\]
For every $\delta\in(0,1)$, conditional on training, with probability
at least $1-\delta$,
\begin{equation}
\max_{(i,j)\in\mathcal P_k}|\widehat m_{ij}-m_{ij}|
\le\tau_{n,\delta},
\qquad
\tau_{n,\delta}
:=C_{\mathrm M}B_Q(G_Q+\sigma_Q)
\{r_{n,\delta}+r_{n,\delta}^2\}.
\label{eq:lstd-deviations}
\end{equation}
\end{lemma}

\begin{proof}
For each $(i,j)\in\mathcal P_k$, apply
Lemma~\ref{lem:trajectory-bernstein}(ii) with multiplier $Q_j$, conditional
mean $\mu_i$, centered TD residual $\varepsilon_i-\mu_i(X)$, and bounds
$B=B_Q$, $D=G_Q$, $\sigma=\sigma_Q$, and $v=B_Q$.
The validation sampling setup ensures the required conditional centering
and exponential moment bounds given the pre-transition history.
With $u=\log(18K/\delta)$, each deviation is bounded by
$\tau_{n,\delta}$ outside an event of probability $6e^{-u}$.
Since $|\mathcal P_k|=3K-2$, a union bound proves
\eqref{eq:lstd-deviations}.
\end{proof}

\subsection{Proof of the oracle inequality}

\begin{proof}[Proof of Theorem~\ref{thm:lstd-misspecification}]
Condition on training and work on the event in
Lemma~\ref{lem:lstd-deviations}. Write
$\widehat i=\widehat i_{\mathrm{LT}}$,
$\kappa=\kappa_{\mathrm{LT}}(k)$, and $\tau=\tau_{n,\delta}$.
If $\widehat i=k$, the result holds directly. Otherwise,
$(\widehat i,\ell)\in\mathcal P_k$ for
$\ell\in\{\widehat i,k\}$, so
\[
|m_{\widehat i\ell}|
\le\widehat T_{\widehat i}+\tau
\le\widehat T_k+\tau
\le T_k+2\tau,
\qquad
|m_{k\ell}|\le T_k.
\]
Lemma~\ref{lem:lstd-identification}, applied to the pair $(\widehat i,k)$, gives
\[
\|Q_{\widehat i}-Q_k\|_{2,\nu}
\le\frac{2(T_k+\tau)}{\kappa B_Q}
\le\frac{2}{\kappa}\|\mu_k\|_{2,\nu}
+\frac{2\tau}{\kappa B_Q},
\]
where the last inequality follows from
$T_k=\max_j|\langle Q_j,\mu_k\rangle_\nu|
\le B_Q\|\mu_k\|_{2,\nu}$ by Cauchy--Schwarz.
Adding $\|Q_k-Q^\pi\|_{2,\nu}$ and substituting
\eqref{eq:lstd-deviations} proves \eqref{eq:lstd-misspecification}.
For the population selector, use $T_{\widetilde i_{\mathrm{LT}}}\le T_k$
in the same comparison with $\tau=0$.
\end{proof}

\section{Normalized LSTD-Tournament: guarantees and limitations}
\label{app:lstd-normalized}
\label{app:lstd-empirical-normalization}

\subsection{Population oracle inequality}

The experimental variant of \citet[Appendix~E.2]{liu2025modelselection}
uses each candidate and its pairwise differences as critics, divided by
their standard deviations. For $f\in L^2(\nu)$, write
$\sigma_\nu(f):=\|f-\mathbb E_\nu f\|_{2,\nu}$ and retain
$h_{ij}=Q_i-Q_j$. With positive standard deviations, define
\begin{equation}
T_i^{\mathrm{nd}}
:=\max\left\{
\frac{|\langle Q_i,\mu_i\rangle_\nu|}{\sigma_\nu(Q_i)},
\max_{j\ne i}
\frac{|\langle h_{ij},\mu_i\rangle_\nu|}{\sigma_\nu(h_{ij})}
\right\},
\qquad
\widetilde i_{\mathrm{LT,nd}}\in\argmin_{i\in[K]}T_i^{\mathrm{nd}}.
\label{eq:lstd-normalized-population-selector}
\end{equation}
For a comparator $k$, set
\begin{equation}
C_k:=\max\left\{
\frac{\|Q_k\|_{2,\nu}}{\sigma_\nu(Q_k)},
\max_{j\ne k}
\frac{\|h_{kj}\|_{2,\nu}}{\sigma_\nu(h_{kj})}
\right\}.
\label{eq:lstd-normalized-critic-norms}
\end{equation}

\begin{theorem}[Population oracle inequality for normalized LSTD-Tournament]
\label{thm:lstd-normalized-population}
Suppose Condition~\ref{cond:q-unweighted-domain} holds, all standard
deviations in \eqref{eq:lstd-normalized-population-selector} are positive,
and the inequality in Condition~\ref{cond:q-pairwise-coercivity} holds
for every $i\in[K]$. Then every population minimizer satisfies
\begin{equation}
\|Q_{\widetilde i_{\mathrm{LT,nd}}}-Q^\pi\|_{2,\nu}
\le
\|Q_{k^\star}-Q^\pi\|_{2,\nu}
+\frac{1+C_{k^\star}}{\kappa_\nu}
\|Q_{k^\star}-\mathcal B_\pi Q_{k^\star}\|_{2,\nu}.
\label{eq:lstd-normalized-population-oracle}
\end{equation}
\end{theorem}

\subsection{Failure of empirical standard-deviation normalization}

For the empirical selector, replace $\langle f,\mu_i\rangle_\nu$ by
$\mathbb P_{n,Q}[f(X)\varepsilon_i]$ and $\sigma_\nu(f)$ by
\[
\widehat\sigma_n(f)^2
:=\mathbb P_{n,Q}[(f-\mathbb P_{n,Q}f)^2].
\]
Denote the resulting scores by $\widehat T_i^{\mathrm{nd}}$ and choose
$\widehat i_{\mathrm{LT,nd}}\in\argmin_i\widehat T_i^{\mathrm{nd}}$
when all required empirical standard deviations are positive.

\begin{proposition}[Counterexample for empirical standard-deviation normalization]
\label{prop:lstd-normalized-empirical-failure}
For every $n\ge8$, there exist a single-action MDP with $\gamma=0$ and
two candidates, all rewards and candidates taking values in $[0,3]$,
satisfying Conditions~\ref{cond:validation-spectral-gap},
\ref{cond:q-pairwise-coercivity}, \ref{cond:q-bounded-signal},
\ref{cond:q-subgaussian-td}, and~\ref{cond:q-unweighted-domain} with
$g_b=\kappa_\nu=1$, $H_Q=G_Q=3$, and $\sigma_Q=1$, such that
$k^\star=1$, $C_1\le\sqrt5$, and
\[
\|Q_1-Q^\pi\|_{2,\nu}=\frac1{2n^2},
\qquad
\|Q_2-Q^\pi\|_{2,\nu}=\frac{\sqrt5}{2}.
\]
With probability at least
\[
1-\frac1n-2e^{-n/8}-2^{1-n},
\]
all empirical denominators are positive and the unique empirical
minimizer is $\widehat i_{\mathrm{LT,nd}}=2$.
\end{proposition}

Consequently, these conditions alone do not yield a uniform oracle
inequality with a fixed coefficient on the best candidate's Bellman
residual and a validation remainder tending to zero. The sample can
underestimate the best candidate's standard deviation and inflate its
normalized moment.

\input{sections/lstd_empirical_normalization.tex}

\subsection{Proofs}
\label{app:lstd-empirical-normalization-proofs}

\subsubsection{Population oracle inequality}

\begin{proof}[Proof of Theorem~\ref{thm:lstd-normalized-population}]
Write $k=k^\star$, $i=\widetilde i_{\mathrm{LT,nd}}$, and
$b=\|\mu_k\|_{2,\nu}$. Cauchy--Schwarz and
\eqref{eq:lstd-normalized-critic-norms} give $T_k^{\mathrm{nd}}\le C_kb$.
If $i=k$, the result holds directly. Otherwise, set
$h=h_{ik}$ and $d=\|h\|_{2,\nu}>0$. Since
$\mu_i-\mu_k=(I-\gamma P_\pi)h$, coercivity gives
\begin{align*}
\kappa_\nu d^2
&\le |\langle h,\mu_i-\mu_k\rangle_\nu|\\
&\le \sigma_\nu(h)T_i^{\mathrm{nd}}+db\\
&\le d\{T_k^{\mathrm{nd}}+b\}
\le (1+C_k)bd,
\end{align*}
where the third line uses $\sigma_\nu(h)\le d$ and the optimality of $i$.
Dividing by $\kappa_\nu d$ and applying the triangle inequality proves
\eqref{eq:lstd-normalized-population-oracle}.
\end{proof}

\subsubsection{Failure of empirical standard-deviation normalization}

\begin{proof}[Proof of Proposition~\ref{prop:lstd-normalized-empirical-failure}]
Fix $n\ge8$ and put $p=n^{-2}$ and $\epsilon=p^3$.
Let $F,G$ be independent random variables taking values $-1$ and $1$
with equal probability, and let $Z\sim\operatorname{Bernoulli}(p)$ be
independent of them. Take the state to be $(F,G,Z)$, with a single action
and transitions drawing a fresh state independently from their joint law
$\nu$. Set $\gamma=0$ and define the reward and candidates by
\[
R=Q^\pi:=\frac p2+\epsilon G+Z,
\qquad
Q_1:=Q^\pi+\frac p2,
\qquad
Q_2:=Q^\pi+1+\frac F2.
\]
The validation states are independent, so $g_b=1$. All rewards and
candidates lie in $[0,3]$. The TD residuals equal their conditional means,
$\mu_1=p/2$ and $\mu_2=1+F/2$, and $(I-\gamma P_\pi)h=h$ for every $h$.
Thus the stated conditions hold with the displayed constants, including
comparisons with zero. The two error norms follow directly from these
residuals. Moreover,
\[
\sigma_\nu(Q_1)^2=p(1-p)+\epsilon^2,
\qquad
h_{12}=\frac p2-1-\frac F2,
\qquad
\sigma_\nu(h_{12})^2=\frac14.
\]
Consequently,
\[
C_1^2
=\max\left\{
1+\frac{4p^2}{p(1-p)+\epsilon^2},
1+4(1-p/2)^2
\right\}\le5.
\]
The remaining population variance is
$\sigma_\nu(Q_2)^2=p(1-p)+\epsilon^2+1/4>0$.

Write $X_t=(F_t,G_t,Z_t)$ and consider the event
\[
E_n:=\left\{
Z_t=0\text{ for all }t\in[n],\quad
|\mathbb P_{n,Q}F|\le\frac12,\quad
\widehat\sigma_n(G)>0
\right\}.
\]
The probability of observing any $Z_t=1$ is at most $np$.
Hoeffding's inequality bounds the probability of
$|\mathbb P_{n,Q}F|>1/2$ by $2e^{-n/8}$, and
$\Pr\{\widehat\sigma_n(G)=0\}=2^{1-n}$. Hence
\[
\Pr(E_n)\ge1-np-2e^{-n/8}-2^{1-n}.
\]
On $E_n$, $Q_1(X_t)=p+\epsilon G_t$ and
$0<\widehat\sigma_n(Q_1)=\epsilon\widehat\sigma_n(G)\le\epsilon$.
The moment with critic $Q_1$ therefore gives
\[
\widehat T_1^{\mathrm{nd}}
\ge
\frac{(p/2)\mathbb P_{n,Q}Q_1}{\widehat\sigma_n(Q_1)}
\ge\frac{p(p-\epsilon)}{2\epsilon}
\ge\frac1{4p}>12.
\]
Also $\widehat\sigma_n(F)\ge\sqrt3/2$ on $E_n$.
The reverse triangle inequality for the empirical centered $L^2$ norm gives
\[
\widehat\sigma_n(Q_2)
\ge\frac12\widehat\sigma_n(F)
-\epsilon\widehat\sigma_n(G)
\ge\frac{\sqrt3}{4}-\epsilon>\frac14,
\qquad
\widehat\sigma_n(h_{21})
=\frac12\widehat\sigma_n(F)>\frac14.
\]
On this event, $|Q_2|\le2$, $|h_{21}|\le2$, and
$|\varepsilon_2|=|\mu_2|\le3/2$. Thus
$\widehat T_2^{\mathrm{nd}}\le12<\widehat T_1^{\mathrm{nd}}$.
All empirical denominators are positive on $E_n$, and candidate $2$ is
the unique minimizer. Substituting $p=n^{-2}$ into the error norms and
the probability bound proves the proposition.
\end{proof}

\input{sections/lstd_empirical_normalization_proofs.tex}

\section{Adjoint LSTD-Tournament for occupancy-ratio selection}
\label{app:adjoint-lstd-tournament}

\subsection{Algorithm and oracle guarantee}

Adjoint LSTD-Tournament selects among occupancy-ratio candidates
$w_1,\ldots,w_M$ fitted independently of the validation data.
Let $\mathbb P_n$ average over validation transitions and
$\mathbb P_m^0$ over initial draws from $d_0$.
For each ordered pair, compute
\begin{align}
\widehat m_{ij}
&:=\mathbb P_n\!\left[
w_i(X)\{w_j(X)-\gamma(\pi w_j)(S')\}
\right]-(1-\gamma)\mathbb P_m^0[w_j],
\nonumber\\
\widehat T_i
&:=\max_{j\in[M]}|\widehat m_{ij}|,
\qquad
\widehat i_{\mathrm{ALT}}\in\argmin_{i\in[M]}\widehat T_i.
\label{eq:adj-lstd-selector}
\end{align}
Ties may be resolved by candidate order. If initial moments are known
exactly, replace $\mathbb P_m^0[w_j]$ by $\mathbb E_{d_0}[w_j]$.
The selector uses the candidates themselves as critics, absolute moments,
and no critic penalty. Let $m_{ij}$ and $T_i=\max_j|m_{ij}|$ denote the
population counterparts.

The oracle inequality below concerns this unnormalized selector.
In the experiments, we use the normalized-difference version, with critics
$w_i$ and $w_i-w_j$, dividing each moment by the critic's empirical
standard deviation.

To relate this criterion to LSTD, take
$\phi_{ij}:=(w_i,w_j)^\top$ and define
\[
\widehat A_{ij}
:=\mathbb P_n\!\left[
\phi_{ij}(X)\{\phi_{ij}(X)-\gamma(\pi\phi_{ij})(S')\}^\top
\right],
\qquad
\widehat c_{ij}:=(1-\gamma)\mathbb P_m^0[\phi_{ij}].
\]
For $w_\theta=\phi_{ij}^\top\theta$, the adjoint moment vector is
$\widehat A_{ij}^\top\theta-\widehat c_{ij}$.
Evaluating it at the two coordinate vectors recovers the corresponding moments
in \eqref{eq:adj-lstd-selector}.
Solving the moment equations instead gives linear minimax weight
learning \citep[Example~2]{ueharaEtAl2020MWLMQL}.
The tournament evaluates these moments at the fitted candidates.

Retain the validation setup of Section~\ref{sec:offline-data-setup} and
Condition~\ref{cond:validation-spectral-gap}. For the initial moments, use
Condition~\ref{cond:w-initial-sample} or exact expectations under $d_0$.
Let $M\ge2$ and let
$k^\star\in\argmin_{i\in[M]}\|w_i-w^\pi\|_{2,\nu}$ index a best candidate.
Impose the following conditions:
\begin{enumerate}[label=\textbf{(ALT\arabic*)}, ref=ALT\arabic*]
\item \label{cond:adj-lstd-envelope}
\textit{Bounded ratio candidates.}
There is a finite $\overline w>0$ such that $0\le w_i\le\overline w$
almost surely under each of $\nu$, $d_0$, and $\nu P_\pi$,
for every $i\in[M]$.

\item \label{cond:adj-lstd-domain}
\textit{Second moments under the offline distribution.}
The target ratio satisfies $w^\pi\in L^2(\nu)$. The adjoint update
$\mathcal B_\pi^\ast w_{k^\star}$ exists as a $\nu$-density, and its residual
$\rho_{k^\star}:=w_{k^\star}-\mathcal B_\pi^\ast w_{k^\star}$ belongs to $L^2(\nu)$.

\item \label{cond:adj-lstd-invertibility}
\textit{Adjoint LSTD moments along oracle-comparison directions.}
For $\phi_{i k^\star}=(w_i,w_{k^\star})^\top$, let
\[
A_{i k^\star}
:=\mathbb E_\nu\!\left[
\phi_{i k^\star}(X)
\{\phi_{i k^\star}(X)-\gamma(P_\pi\phi_{i k^\star})(X)\}^\top
\right].
\]
There is a constant $s_{\mathrm{ALT}}(k^\star)>0$ such that
\begin{equation}
\sum_{\ell\in\{i,k^\star\}}
\left\langle w_i-w_{k^\star},(I-\gamma P_\pi)w_\ell\right\rangle_\nu^2
\ge 2s_{\mathrm{ALT}}(k^\star)^2,
\qquad
\kappa_{\mathrm{ALT}}(k^\star):=\frac{s_{\mathrm{ALT}}(k^\star)}{2\overline w^2}.
\label{eq:adj-lstd-conditioning}
\end{equation}
The inequality is required for every $i$ with
$\|w_i-w_{k^\star}\|_{2,\nu}>0$.
\end{enumerate}
Condition~\ref{cond:adj-lstd-envelope} bounds the critics at successors
and initial draws.
The full singular-value bound
$\min_{i\ne k^\star}\sigma_{\min}(A_{i k^\star})
\ge s_{\mathrm{ALT}}(k^\star)>0$ is sufficient for
Condition~\ref{cond:adj-lstd-invertibility}: its moment sum is
$\|A_{i k^\star}^\top v\|_2^2$ for $v=(1,-1)^\top$, with $\|v\|_2^2=2$.
The adjoint matrices $A_{i k^\star}^\top$ have the same singular values as
$A_{i k^\star}$. Condition~\ref{cond:adj-lstd-invertibility} allows the
pairwise matrices themselves to be singular.

\begin{theorem}[Adjoint LSTD-Tournament oracle inequality]
\label{thm:adj-lstd-oracle}
Suppose Condition~\ref{cond:validation-spectral-gap} and
Conditions~\ref{cond:adj-lstd-envelope}--\ref{cond:adj-lstd-invertibility}
hold. Use Condition~\ref{cond:w-initial-sample} or exact $d_0$ moments.
For every $\delta\in(0,1)$, set
\[
u_\delta:=\log(20M/\delta),
\qquad
r_T:=\sqrt{\frac{u_\delta}{ng_b}},
\qquad
r_0:=\sqrt{\frac{u_\delta}{m}},
\]
with $r_0=0$ for exact initial moments. Conditional on training, with
probability at least $1-\delta$, the selector in
\eqref{eq:adj-lstd-selector} satisfies
\begin{align}
\|w_{\widehat i_{\mathrm{ALT}}}-w^\pi\|_{2,\nu}
\le{}&
\|w_{k^\star}-w^\pi\|_{2,\nu}
+\frac{2}{\kappa_{\mathrm{ALT}}(k^\star)}
\|w_{k^\star}-\mathcal B_\pi^\ast w_{k^\star}\|_{2,\nu}
\nonumber\\
&+\frac{2C_{\mathrm M}\overline w}{\kappa_{\mathrm{ALT}}(k^\star)}
(r_T+r_T^2)
+\frac{\sqrt2(1-\gamma)}{\kappa_{\mathrm{ALT}}(k^\star)}r_0,
\label{eq:adj-lstd-oracle}
\end{align}
where $C_{\mathrm M}$ is the constant in
Lemma~\ref{lem:trajectory-bernstein}. Every population minimizer
$\widetilde i_{\mathrm{ALT}}\in\argmin_iT_i$ satisfies the same bound
without the two validation terms.
\end{theorem}

For $ng_b\ge u_\delta$, $r_T+r_T^2\le2r_T$. If, in addition,
$\|\rho_{k^\star}\|_{2,\nu}\le
L_\nu^\ast\|w_{k^\star}-w^\pi\|_{2,\nu}$,
the first two terms in \eqref{eq:adj-lstd-oracle} become
\[
\left\{1+\frac{2L_\nu^\ast}{\kappa_{\mathrm{ALT}}(k^\star)}\right\}
\min_i\|w_i-w^\pi\|_{2,\nu}.
\]
For example, finite one-step concentrability $C_{\nu,\pi}$ gives
$\|P_\pi^\ast\|_{L^2(\nu)\to L^2(\nu)}\le\sqrt{C_{\nu,\pi}}$ by
duality, so $L_\nu^\ast=1+\gamma\sqrt{C_{\nu,\pi}}$ is sufficient.

\subsection{Adjoint moment identification}

Throughout the proofs below, write $k=k^\star$.

\begin{lemma}[Adjoint LSTD moment identification]
\label{lem:adj-lstd-identification}
Under Conditions~\ref{cond:adj-lstd-envelope}
and~\ref{cond:adj-lstd-invertibility}, for every $i\ne k$,
\begin{equation}
\|w_i-w_k\|_{2,\nu}
\le\frac{1}{\kappa_{\mathrm{ALT}}(k)\overline w}
\max_{\ell\in\{i,k\}}|m_{i\ell}-m_{k\ell}|.
\label{eq:adj-lstd-identification}
\end{equation}
\end{lemma}

\begin{proof}
If $\|w_i-w_k\|_{2,\nu}=0$, the inequality holds directly.
Otherwise, the difference is covered by Condition~\ref{cond:adj-lstd-invertibility}.
Write $B=\overline w$, $s=s_{\mathrm{ALT}}(k)$, and $v=(1,-1)^\top$.
The initial-distribution term cancels in each moment difference, giving
\[
A_{ik}^\top v
=\mathbb E_\nu\!\left[
\{\phi_{ik}-\gamma P_\pi\phi_{ik}\}(w_i-w_k)
\right]
=\begin{pmatrix}
m_{ii}-m_{ki}\\
m_{ik}-m_{kk}
\end{pmatrix}.
\]
The candidate bound and $\|A_{ik}^\top v\|_2\ge s\|v\|_2$ imply
\[
\|w_i-w_k\|_{2,\nu}
\le\sqrt2 B\|v\|_2
\le\frac{\sqrt2 B}{s}\|A_{ik}^\top v\|_2
\le\frac{2B}{s}\|A_{ik}^\top v\|_\infty.
\]
Substituting \eqref{eq:adj-lstd-conditioning} proves
\eqref{eq:adj-lstd-identification}.
\end{proof}

\subsection{Validation moment deviations}

\begin{lemma}[Adjoint validation moment deviations]
\label{lem:adj-lstd-deviations}
Suppose Conditions~\ref{cond:validation-spectral-gap}
and~\ref{cond:adj-lstd-envelope} hold. Use
Condition~\ref{cond:w-initial-sample} or exact $d_0$ moments.
Define
\[
\mathcal P_k
:=\{(k,j):j\in[M]\}
\cup\{(i,i),(i,k):i\ne k\}.
\]
With $r_T,r_0$ as in Theorem~\ref{thm:adj-lstd-oracle}, conditional on
training, with probability at least $1-\delta$,
\begin{equation}
\max_{(i,j)\in\mathcal P_k}|\widehat m_{ij}-m_{ij}|
\le\tau,
\qquad
\tau:=C_{\mathrm M}\overline w^2(r_T+r_T^2)
+\frac{(1-\gamma)\overline w}{\sqrt2}r_0.
\label{eq:adj-lstd-deviations}
\end{equation}
\end{lemma}

\begin{proof}
Set $B=\overline w$ and
$Y_{ij,t}:=w_i(X_t)\{w_j(X_t)-\gamma(\pi w_j)(S_t')\}$.
Condition~\ref{cond:adj-lstd-envelope} implies
$0\le(\pi w_j)(S_t')\le B$ almost surely: the candidate bound holds
$\nu P_\pi$-almost surely, and integration over the target action
preserves it. Hence $|Y_{ij,t}|\le B^2$
and $\mathbb E Y_{ij,t}^2\le B^4$.
Apply Lemma~\ref{lem:trajectory-bernstein}(i), using the validation
filtration in Appendix~\ref{app:trajectory-concentration}.
For each $(i,j)\in\mathcal P_k$, outside an event of probability
$6e^{-u_\delta}$,
\[
|\mathbb P_nY_{ij}-\mathbb E Y_{ij}|
\le C_{\mathrm M}B^2(r_T+r_T^2).
\]
For the initial sample, $0\le w_j\le B$ also holds $d_0$-almost surely.
Hoeffding's inequality gives, outside an event of probability
$2e^{-u_\delta}$ for each $j$,
\[
|\mathbb P_m^0[w_j]-\mathbb E_{d_0}[w_j]|
\le B\sqrt{\frac{u_\delta}{2m}}=\frac{B}{\sqrt2}r_0.
\]
There are $3M-2$ transition moments and $M$ initial moments.
Their total failure probability is at most
$\{6(3M-2)+2M\}e^{-u_\delta}\le\delta$.
Combining the two deviations proves \eqref{eq:adj-lstd-deviations}.
For exact initial moments, the initial deviation is zero.
\end{proof}

\subsection{Proof of the oracle inequality}

\begin{proof}[Proof of Theorem~\ref{thm:adj-lstd-oracle}]
Condition on training and work on the event in
Lemma~\ref{lem:adj-lstd-deviations}. Write
$\widehat i=\widehat i_{\mathrm{ALT}}$ and
$\kappa=\kappa_{\mathrm{ALT}}(k)$.
If $\widehat i=k$, the result holds directly.
Otherwise, for each $\ell\in\{\widehat i,k\}$,
$(\widehat i,\ell)\in\mathcal P_k$ and empirical optimality gives
\[
|m_{\widehat i\ell}|
\le\widehat T_{\widehat i}+\tau
\le\widehat T_k+\tau
\le T_k+2\tau,
\qquad
|m_{k\ell}|\le T_k.
\]
By Lemma~\ref{lem:adj-lstd-identification}, applied to $(\widehat i,k)$,
\[
\|w_{\widehat i}-w_k\|_{2,\nu}
\le\frac{2(T_k+\tau)}{\kappa \overline w}.
\]
The adjoint moment identity \eqref{eq:w-adjoint-moment} and
Condition~\ref{cond:adj-lstd-domain} give
$m_{kj}=\langle\rho_k,w_j\rangle_\nu$.
Cauchy--Schwarz therefore yields
$T_k\le \overline w\|\rho_k\|_{2,\nu}$, so
\[
\|w_{\widehat i}-w_k\|_{2,\nu}
\le\frac{2}{\kappa}\|\rho_k\|_{2,\nu}
+\frac{2\tau}{\kappa \overline w}.
\]
Adding $\|w_k-w^\pi\|_{2,\nu}$ and substituting
\eqref{eq:adj-lstd-deviations} proves \eqref{eq:adj-lstd-oracle}.
For the population selector, repeat the comparison using
$T_{\widetilde i_{\mathrm{ALT}}}\le T_k$ and $\tau=0$.
\end{proof}

\section{Additional related work}
\label{app:related-work}

Section~\ref{sec:related-work-main} compares PBV with existing Bellman
validation methods and LSTD-Tournament. This appendix discusses OPE model
selection, occupancy-ratio estimation, supervised aggregation, and
statistical inverse problems.

\textbf{Model selection for $Q$-functions and occupancy ratios.}
Existing OPE model-selection methods have focused primarily on $Q$-functions,
replacing the unavailable supervised loss with structured Bellman criteria.
BErMin estimates Bellman error with complexity penalties; BVFT and BVFT-PE use
projected Bellman errors over candidate-dependent state aggregations; and ModBE
uses relative square-loss tests for nested model classes
\citep{farahmand2011model,xie2021batch,zhang2021towards,lee2022oracle}.
Supervised Bellman validation instead regresses the one-step TD target on
the current state and action separately for each candidate. Duality-based
Residual Estimation learns residual critics and selects among critic
configurations on an additional split
\citep{zitovsky2023revisiting,miyaguchi2026duality}. PBV instead constructs
critics directly from the candidates, avoiding candidate-dependent projection,
nestedness, and auxiliary critic fitting.

\paragraph{Model-selection targets in OPE.}
Model selection arises at several levels in off-policy evaluation. One line of
work selects or combines scalar value estimates, including MAGIC, SWITCH,
shrinkage estimators, SLOPE, proxy-task and cross-validation methods, and OPERA
\citep{thomasBrunskill2016DataEfficientOPE,wang2017optimal,su2020shrinkage,
su2020adaptive,udagawa2023policy,cief2025cross,nie2024opera}. A second line
uses OPE to rank policies or tune complete offline-RL pipelines
\citep{irpan2019off,paine2020hyperparameter,tang2021model,
yang2022offline,nie2022data}. \citet{zhang2021towards} primarily study offline
policy selection but also use BVFT-PE to select among candidate $Q^\pi$
estimates for a fixed policy. We instead study nuisance selection for a fixed target policy.
From finite classes of candidates fitted independently of the validation
data, we select the $Q$-function or occupancy-ratio estimate used for OPE. This differs from
selecting scalar value estimates or policies because neither nuisance is
observed, so their target losses are not directly available.

\paragraph{Occupancy-ratio estimation and selection.}
Marginalized importance sampling, DICE methods, and minimax weight learning
estimate discounted occupancy ratios from adjoint Bellman equations, typically
using prescribed ratio and critic classes
\citep{xieEtAl2019MarginalizedIS,nachumEtAl2019DualDICE,
zhangEtAl2020GenDICE,ueharaEtAl2020MWLMQL,ueharaEtAl2021FiniteSampleMinimax}.
FORE is an occupancy-ratio analogue of
FQE that estimates the ratio through fitted adjoint Bellman recursion without
requiring Bellman completeness \citep{van2026fitted}, while FORC is a related
finite-horizon method that does require Bellman completeness
\citep{huangChenJiang2023DensityFeatures,huangJiang2024OccupancyPG}. These
methods construct an occupancy-ratio estimate within a specified function
class. A-PBV instead selects post hoc among already-fitted candidates using
pairwise log-ratio critics and Hellinger-penalized adjoint Bellman scores.

\paragraph{Supervised model selection and aggregation.}
In supervised prediction, held-out outcomes make the risk of a fixed candidate
directly estimable. Cross-validation uses this risk to select algorithms or
tuning parameters, while stacking and Super Learner combine candidates using
cross-validated predictions
\citep{stone1974CrossValidatory,wolpert1992Stacked,
van2003unified,van2007super}. Model-selection aggregation instead seeks to
match the best element of a finite dictionary; under squared loss,
Q-aggregation attains sharp oracle inequalities with logarithmic remainders
\citep{daiRigolletZhang2012QAggregation,lan2024causal}.
These supervised criteria are unavailable here because neither $Q^\pi$ nor
$w^\pi$ is observed, squared temporal-difference error is not a valid target
loss, and offline-distribution Bellman moments need not control target-policy
error. PBV instead constructs directly estimable pairwise criteria and selects a
candidate from the fitted class rather than forming an aggregate.

\paragraph{Model selection in linear inverse problems.}
Related model-selection challenges arise in statistical linear inverse problems,
where instability of the inverse operator motivates operator-dependent risk
estimates and penalties
\citep{cavalierEtAl2002OracleInverse,abramovichEtAl2018IllPosedSelection}.
For example, Regularized DeepIV uses an out-of-sample Tikhonov criterion to
obtain oracle guarantees under misspecification in nonparametric instrumental-variable
regression \citep{liEtAl2024RegularizedDeepIV}. In general, small forward error
need not control structural error without regularization and source or link
conditions. Here, the Bellman evaluation operator is a $\sqrt{\gamma}$-contraction
in $L^2(d_{\pi,\gamma})$, while the adjoint Bellman update contracts
generalized KL divergence by a factor of $\gamma$. These contractions convert pairwise Bellman
moments into target-error bounds without a Tikhonov bias--variance choice or
inverse-problem source conditions. IW-PBV requires coverage and occupancy-ratio
accuracy, A-PBV requires bounded ratio candidates and bounded pairwise log ratios,
and unweighted PBV requires a coercivity condition under the offline
distribution.

\end{document}

%% file: sections/experiments_defaults.tex
\FloatBarrier
\section{Experiments}
\label{sec:experiments}

\input{figures/pbv_default_main_results.tex}

\begin{figure}[t]
\centering
\includegraphics[width=.99\linewidth]{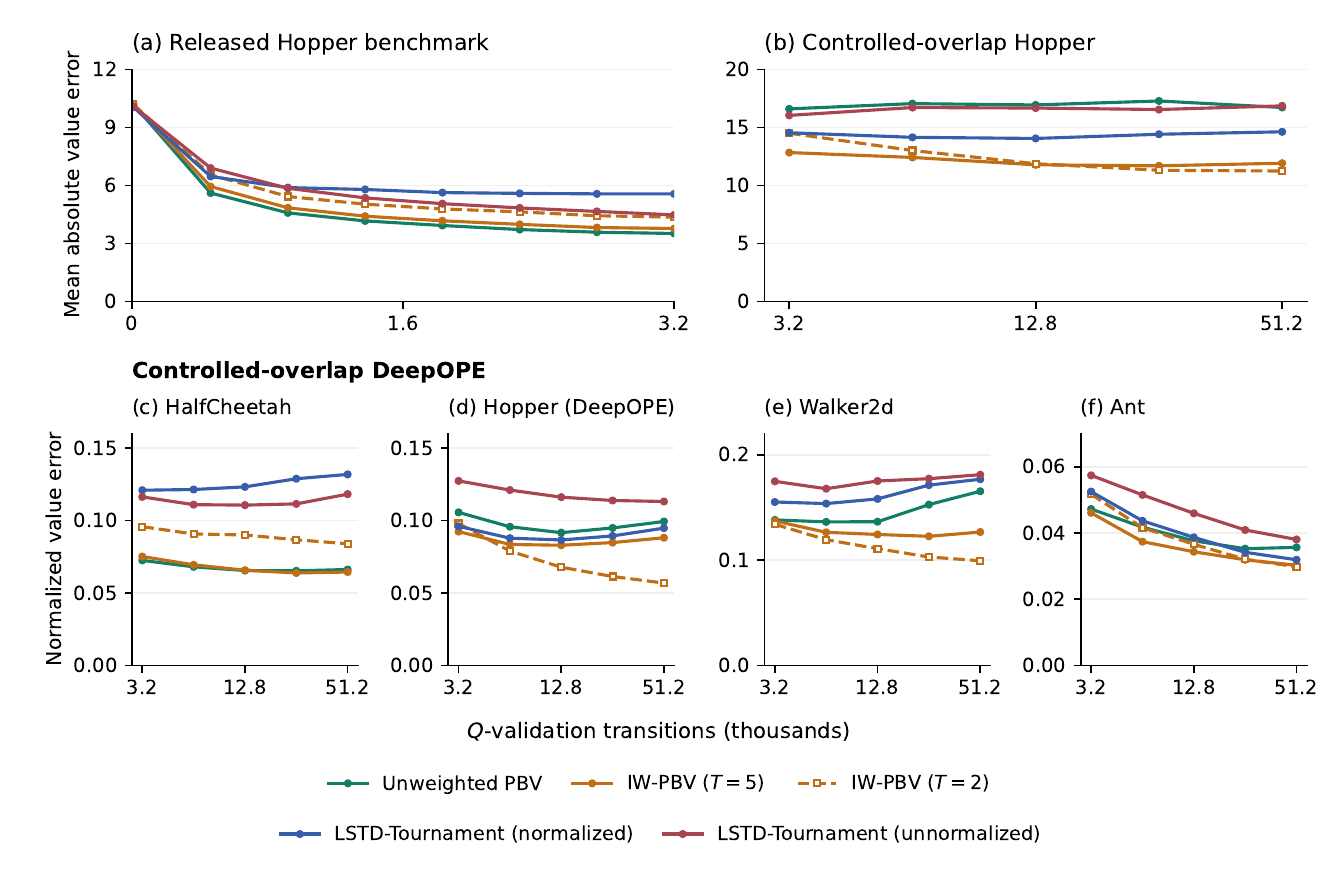}
\caption{Mean absolute policy-value error as the $Q$-validation sample size increases.
(a) Released model-free Hopper benchmark: 120 cases with the original 15 candidates.
(b) Controlled-overlap Hopper: 60 cases.
(c)--(f) Controlled-overlap DeepOPE: eleven target policies in each of four environments.
Panels (b)--(f) use 60 FQE candidates.
DeepOPE errors are divided by the range of reference policy values in each environment;
the two Hopper benchmark panels use original return units.
Solid IW-PBV curves use $T=5$; dashed curves use $T=2$.
Candidates and selected occupancy ratios are held fixed; methods share validation draws.
Full DeepOPE results on D4RL datasets are reported in Appendix~\ref{app:full-deepope}.}
\label{fig:pbv-default-validation}
\end{figure}

We evaluate unweighted PBV and IW-PBV on two benchmarks: the released Hopper
benchmark of \citet{liu2025modelselection}, and four MuJoCo environments using
DeepOPE target policies and D4RL offline datasets
\citep{fu2021benchmarks,fuEtAl2020D4RL}.
To assess target-occupancy weighting under controlled overlap, we also collect
logging data by sampling one benchmark policy uniformly at the start of each
trajectory and following it until termination. Each component policy is
evaluated in turn as the target, so the expected fraction of trajectories
following the target policy is $1/10$ for Hopper and $1/11$ for DeepOPE.

On Hopper, we use the released offline transitions, candidate predictions,
and reference values. These include 60 model-free settings on the main gravity
and noise grids, 60 under poor coverage, and 36 model-based settings.
The model-free and model-based candidate classes contain 15 and five
candidates, respectively. For DeepOPE, four environments, five dataset
qualities, and eleven target policies give 220 settings.
We fit 60 neural FQE candidates per setting, varying network architecture,
learning rate, and training duration.
The controlled-overlap studies contain 60 Hopper and 44 DeepOPE settings,
with 60 FQE candidates per setting.

For IW-PBV, we fit candidate occupancy ratios using fitted occupancy-ratio
evaluation (FORE; \citealp{van2026fitted}) and select among them with A-PBV
on independent data. Before selection, we discard candidates with training
relative ESS below 5\% and cap the remaining predictions at 50.
Following \citet{sinha2022experience}, we temper the selected ratio with
power $1/T$, using $T=5$ and also reporting $T=2$.
We compare with the unnormalized and normalized-difference
variants of LSTD-Tournament and with squared TD error, using identical
candidates and validation draws. We report mean absolute policy-value error
and initial-state value RMSE, with DeepOPE errors normalized by each
environment's range of reference policy values \citep{fu2021benchmarks}.
Appendix~\ref{app:experiments} gives the evaluation details and additional results.

On the released Hopper benchmark, unweighted PBV has lower mean absolute
policy-value error and initial-state value RMSE than both LSTD-Tournament
variants in all three benchmark groups (Table~\ref{tab:pbv-main-results}).
Under controlled overlap, IW-PBV with $T=5$ has lower aggregate mean absolute
policy-value error than unweighted PBV and both LSTD-Tournament variants
at every validation sample size shown (Figure~\ref{fig:pbv-default-validation}).
Comparing the two temperatures, $T=2$ gives lower mean value error at larger
sample sizes in controlled Hopper
and in the DeepOPE Hopper and Walker2d environments; $T=5$ gives lower
error in HalfCheetah. On full DeepOPE, unnormalized LSTD-Tournament has
slightly lower aggregate errors, but the differences among the PBV and
LSTD-Tournament variants are small relative to the overall error
(Appendix~\ref{app:full-deepope}).

We study occupancy-ratio selection in Appendix~\ref{app:occupancy-experiments}.
In the controlled Hopper ratio experiment, A-PBV has lower aggregate
Hellinger distance, ratio RMSE, generalized KL, and reward error than
Adjoint LSTD-Tournament across three candidate libraries.
However, using A-PBV rather than Adjoint LSTD-Tournament to select the occupancy
ratio does not consistently improve downstream policy-value estimation
with IW-PBV, with results varying across benchmarks.
Compared with untempered weighting, tempering substantially reduces errors
on the main model-free and model-based Hopper grids; its effect on DeepOPE
depends on the benchmark and validation sample size.

\FloatBarrier

%% file: figures/pbv_default_main_results.tex
\begin{table}[!htbp]
\centering
\small
\setlength{\tabcolsep}{3pt}
\renewcommand{\arraystretch}{1.08}
\caption{Selection at \(n_Q=3{,}200\): mean absolute policy-value error (VE) and initial-state value RMSE. MF, poor MF, and MB are the main, poor-coverage, and model-based Hopper grids, with their released 15-, 15-, and five-candidate classes. DeepOPE uses 60 FQE candidates. Hopper errors are in original return units; DeepOPE errors are normalized by the range of reference policy values in each environment. Cases have equal weight; RMSE takes the square root after pooling squared errors. Bold marks the smallest mean.}
\label{tab:pbv-main-results}
\begin{tabular*}{\linewidth}{@{\extracolsep{\fill}}lrrrrrrrr@{}}
\toprule
& \multicolumn{2}{c}{MF (60)} & \multicolumn{2}{c}{Poor MF (60)} & \multicolumn{2}{c}{MB (36)} & \multicolumn{2}{c}{DeepOPE (220)} \\
\cmidrule(lr){2-3}\cmidrule(lr){4-5}\cmidrule(lr){6-7}\cmidrule(lr){8-9}
Selector & VE & RMSE & VE & RMSE & VE & RMSE & VE & RMSE \\
\midrule
Unweighted PBV & \textbf{1.78} & \textbf{4.22} & 5.26 & 10.28 & \textbf{1.25} & \textbf{3.09} & 0.599 & 0.704 \\
IW-PBV ($T=5$) & 2.31 & 5.04 & \textbf{5.25} & \textbf{10.24} & 1.61 & 3.76 & 0.598 & 0.705 \\
LSTD-Tournament (normalized) & 3.09 & 7.45 & 8.04 & 14.28 & 1.36 & 3.43 & 0.597 & 0.702 \\
LSTD-Tournament (unnormalized) & 2.46 & 5.90 & 6.50 & 12.20 & 1.89 & 4.34 & \textbf{0.587} & \textbf{0.691} \\
Squared TD error & 6.75 & 14.40 & 12.02 & 20.25 & 2.56 & 5.35 & 0.733 & 0.826 \\
\bottomrule
\end{tabular*}
\end{table}

%% file: sections/q_fitted_weight_theory.tex
\subsection{PBV under approximate fitted-weight coercivity}
\label{sec:q-theory:fitted-weight}

We extend the unweighted PBV guarantees to the fitted measure, allowing
an additive defect in the coercivity condition.
For a nonnegative fitted weight, write
\[
\widehat\nu(\dd x):=\widehat w(x)\nu(\dd x),
\qquad
\|f\|_{2,\widehat\nu}^2:=\mathbb E_\nu[\widehat w(X)f(X)^2],
\qquad
\mu_i:=\mathbb E(\varepsilon_i\mid X)=Q_i-\mathcal B_\pi Q_i.
\]
The measure $\widehat\nu$ need not be normalized.
Let $k^\star_{\widehat\nu}$ be the smallest index minimizing
$\|Q_i-Q^\pi\|_{2,\widehat\nu}$ over $i\in[K]$.
In addition to Condition~\ref{cond:validation-spectral-gap}, we impose
the following analogues of the unweighted conditions.

\begin{enumerate}[label=\textbf{(B\arabic*)}, ref=B\arabic*, start=6]
\item \label{cond:q-fitted-coercive-weights}
\textit{Independent bounded fitted weights.}
The $Q$-validation fold is independent of all data used to construct
$\widehat w$ and the candidates. There is a finite constant $W_Q>0$ such
that $0\le\widehat w\le W_Q$, $\nu$-almost surely.
Conditional on these fitting data, the validation process follows the
setup of Section~\ref{sec:offline-data-setup} and satisfies
Condition~\ref{cond:validation-spectral-gap}.

\item \label{cond:q-fitted-approximate-coercivity}
\textit{Approximate coercivity along oracle-comparison directions.}
There are constants $\kappa_{\widehat\nu}\in(0,1]$ and
$\varepsilon_{\mathrm{coer}}\ge0$ such that
\[
\left|\left\langle h_{i k^\star_{\widehat\nu}},
(I-\gamma P_\pi)h_{i k^\star_{\widehat\nu}}
\right\rangle_{\widehat\nu}\right|
\ge\kappa_{\widehat\nu}
\|h_{i k^\star_{\widehat\nu}}\|_{2,\widehat\nu}^2
-\varepsilon_{\mathrm{coer}}^2
\quad\text{for every }i\in\mathcal J_Q.
\]

\item \label{cond:q-fitted-coercive-signal}
\textit{Bounded candidate differences and conditional TD means.}
There are finite constants $H_Q,G_Q$ such that, $\widehat\nu$-almost surely,
\[
\max_{i,j\in\mathcal J_Q}|h_{ij}(X)|\le H_Q,
\qquad
\max_{i\in\mathcal J_Q}|\mu_i(X)|\le G_Q.
\]

\item \label{cond:q-fitted-coercive-noise}
\textit{Conditional subexponential TD noise.}
There is a finite constant $\sigma_Q>0$ such that,
$\widehat\nu$-almost surely,
\[
\max_{i\in\mathcal J_Q}
\mathbb E\!\left[\left.
\exp\left\{\frac{|\varepsilon_i-\mu_i(X)|}{\sigma_Q}\right\}
\right|X\right]\le2.
\]

\item \label{cond:q-fitted-coercive-domain}
\textit{Second moments under the fitted measure.}
The true action-value function, candidates, rewards, and TD residuals satisfy
\[
\|Q^\pi\|_{2,\widehat\nu}^2
+\max_{i\in[K]}\|Q_i\|_{2,\widehat\nu}^2
+\mathbb E_\nu[\widehat w(X)R^2]
+\max_{i\in[K]}\mathbb E_\nu[\widehat w(X)\varepsilon_i^2]
<\infty.
\]
\end{enumerate}

These are fitted-measure analogues of
Conditions~\ref{cond:q-pairwise-coercivity},
\ref{cond:q-bounded-signal}, \ref{cond:q-subgaussian-td}, and
\ref{cond:q-unweighted-domain}, with
Condition~\ref{cond:q-fitted-approximate-coercivity} allowing an additive
coercivity defect. Setting $\varepsilon_{\mathrm{coer}}=0$ recovers
exact coercivity.
The validation observations are still sampled under $\nu$, so the
spectral gap remains that of the behavior process. The weight envelope
controls the additional multiplier in the empirical moments.

\paragraph{Population oracle inequality.}
Define the fitted-measure population scores and selector by
\[
\mathcal M_{ij}^{\mathrm{fit}}
:=|\langle h_{ij},\mu_i\rangle_{\widehat\nu}|
-\frac14|\langle h_{ij},(I-\gamma P_\pi)h_{ij}\rangle_{\widehat\nu}|,
\qquad
\widetilde i_Q^{\mathrm{fit}}
\in\argmin_{i\in[K]}\max_{j\in\mathcal J_Q}\mathcal M_{ij}^{\mathrm{fit}}.
\]
Here $\widehat{\mathcal M}_{ij}^{\mathrm{TA}}(\widehat w)$ denotes the
weighted empirical score in Algorithm~\ref{alg:q-pairwise-selection},
and $\widehat i_Q$ is its minimizer.

\begin{theorem}[Population TA-PBV under fitted-weight coercivity]
\label{thm:q-fitted-coercive-population}
Suppose Conditions~\ref{cond:q-fitted-coercive-weights},
\ref{cond:q-fitted-approximate-coercivity}, and~\ref{cond:q-fitted-coercive-domain}
hold with $\varepsilon_{\mathrm{coer}}=0$.
Then every population minimizer satisfies
\[
\|Q_{\widetilde i_Q^{\mathrm{fit}}}-Q^\pi\|_{2,\widehat\nu}
\le\|Q_{k^\star_{\widehat\nu}}-Q^\pi\|_{2,\widehat\nu}
+\frac{2}{\kappa_{\widehat\nu}}
\|Q_{k^\star_{\widehat\nu}}-\mathcal B_\pi Q_{k^\star_{\widehat\nu}}\|_{2,\widehat\nu}.
\]
The same bound holds for any comparator $k\in[K]$ for which
Condition~\ref{cond:q-fitted-approximate-coercivity} holds with $k$ in place of
$k^\star_{\widehat\nu}$.
\end{theorem}

\begin{theorem}[Finite-sample PBV under approximate coercivity]
\label{thm:q-fitted-approximate-finite-sample}
Suppose Condition~\ref{cond:validation-spectral-gap} and
Conditions~\ref{cond:q-fitted-coercive-weights}--\ref{cond:q-fitted-coercive-domain}
hold. There is a finite constant $C_{Q,\mathrm{fit}}>0$, depending only on
$H_Q,G_Q,\sigma_Q$, and $W_Q$, such that, for every $\delta\in(0,1)$,
conditional on the fitting data, with probability at least $1-\delta$,
\begin{align}
\|Q_{\widehat i_Q}-Q^\pi\|_{2,\widehat\nu}
\le{}&\|Q_{k^\star_{\widehat\nu}}-Q^\pi\|_{2,\widehat\nu}
+\frac{2}{\kappa_{\widehat\nu}}
\|Q_{k^\star_{\widehat\nu}}-\mathcal B_\pi Q_{k^\star_{\widehat\nu}}\|_{2,\widehat\nu}
\notag\\
&+\frac{2\varepsilon_{\mathrm{coer}}}{\sqrt{3\kappa_{\widehat\nu}}}
+\frac{C_{Q,\mathrm{fit}}}{\kappa_{\widehat\nu}}
\sqrt{\frac{\log(12K/\delta)}{ng_b}}.
\label{eq:q-fitted-approximate-finite-sample}
\end{align}
More generally, the same bound holds for any comparator $k$ fixed
conditional on the fitting data for which
Condition~\ref{cond:q-fitted-approximate-coercivity} holds with $k$ in place of
$k^\star_{\widehat\nu}$.
\end{theorem}

Proofs and an explicit choice of $C_{Q,\mathrm{fit}}$ are given in
Appendix~\ref{app:q-fitted-coercive-proofs}.
Taking $\varepsilon_{\mathrm{coer}}=0$ gives the exact fitted-weight
coercivity bound. If
$\|\mu_{k^\star_{\widehat\nu}}\|_{2,\widehat\nu}
\le L_{\widehat\nu}\|Q_{k^\star_{\widehat\nu}}-Q^\pi\|_{2,\widehat\nu}$,
then \eqref{eq:q-fitted-approximate-finite-sample} becomes
\begin{align*}
\|Q_{\widehat i_Q}-Q^\pi\|_{2,\widehat\nu}
\le{}&\left(1+\frac{2L_{\widehat\nu}}{\kappa_{\widehat\nu}}\right)
\min_{i\in[K]}\|Q_i-Q^\pi\|_{2,\widehat\nu}
\\
&+\frac{2\varepsilon_{\mathrm{coer}}}{\sqrt{3\kappa_{\widehat\nu}}}
+\frac{C_{Q,\mathrm{fit}}}{\kappa_{\widehat\nu}}
\sqrt{\frac{\log(12K/\delta)}{ng_b}}.
\end{align*}
One-step concentrability under the fitted measure is sufficient:
if $\widehat\nu P_\pi\ll\widehat\nu$ and
$C_{\widehat\nu,\pi}
:=\|\dd(\widehat\nu P_\pi)/\dd\widehat\nu\|_\infty<\infty$,
one may take $L_{\widehat\nu}=1+\gamma\sqrt{C_{\widehat\nu,\pi}}$.
If additionally $\gamma\sqrt{C_{\widehat\nu,\pi}}<1$,
Condition~\ref{cond:q-fitted-approximate-coercivity} holds with
$\kappa_{\widehat\nu}=1-\gamma\sqrt{C_{\widehat\nu,\pi}}$ and
$\varepsilon_{\mathrm{coer}}=0$.
For fixed envelope constants and coercivity bounded away from zero,
the fitted-measure error vanishes when the oracle approximation error,
its mean Bellman residual, and $\varepsilon_{\mathrm{coer}}$ vanish and
$\log K=o(ng_b)$. Taking $\widehat w\equiv1$ and
$\varepsilon_{\mathrm{coer}}=0$ recovers the unweighted results.

\begin{lemma}[Approximate coercivity for accurate weights]
\label{lem:q-fitted-approximate-coercivity}
Suppose Condition~\ref{cond:q-bounded-signal} holds and
$\widehat w\ge0$ satisfies $H_\nu(\widehat w,w^\pi)<\infty$.
Then, for every $i,j\in\mathcal J_Q$,
\begin{align*}
\left\langle h_{ij},(I-\gamma P_\pi)h_{ij}
\right\rangle_{\widehat w\nu}
\ge{}&
\frac{1-\sqrt\gamma}{2}
\|h_{ij}\|_{2,\widehat w\nu}^2\\
&-
\frac{\{(1+\sqrt\gamma)H_Q+2G_Q\}^2}
{2(1-\sqrt\gamma)}
H_\nu(\widehat w,w^\pi)^2.
\end{align*}
\end{lemma}

The lemma verifies Condition~\ref{cond:q-fitted-approximate-coercivity} with
\[
\kappa_{\widehat\nu}=\frac{1-\sqrt\gamma}{2},
\qquad
\varepsilon_{\mathrm{coer}}
=\frac{(1+\sqrt\gamma)H_Q+2G_Q}{\sqrt{2(1-\sqrt\gamma)}}
H_\nu(\widehat w,w^\pi).
\]
The additional term in
Theorem~\ref{thm:q-fitted-approximate-finite-sample} is then
\[
\frac{2\{(1+\sqrt\gamma)H_Q+2G_Q\}}{\sqrt3(1-\sqrt\gamma)}
H_\nu(\widehat w,w^\pi).
\]
The proof of the main-text TA-PBV guarantee
(Theorem~\ref{thm:sequential-ta-pbv}) uses the same weight-perturbation
argument through the refined score bounds in
Appendix~\ref{app:q-finite-sample-proof}. Those bounds give the sharper
Hellinger term $H_\nu(\widehat w,w^\pi)/\sqrt{1-\sqrt\gamma}$ in
Theorem~\ref{thm:q-finite-sample-oracle}, which measures error under
$d_{\pi,\gamma}$.
Exact coercivity follows if this error is sufficiently small relative
to the fitted norm of each comparison direction. For example, if
\[
\{(1+\sqrt\gamma)H_Q+2G_Q\}
H_\nu(\widehat w,w^\pi)
\le
\frac{1-\sqrt\gamma}{\sqrt2}
\|h\|_{2,\widehat w\nu}
\]
for every oracle-comparison direction $h$ with nonzero fitted norm,
then Condition~\ref{cond:q-fitted-approximate-coercivity} holds with
$\kappa_{\widehat\nu}=(1-\sqrt\gamma)/4$ and
$\varepsilon_{\mathrm{coer}}=0$.

\begin{proof}[Proof of Lemma~\ref{lem:q-fitted-approximate-coercivity}]
Fix $h=h_{ij}$. Condition~\ref{cond:q-bounded-signal} gives
$\|h\|_{\infty,\nu}\le H_Q$ and, since
$\gamma P_\pi h=h-(\mu_i-\mu_j)$,
$\|\gamma P_\pi h\|_{\infty,\nu}\le H_Q+2G_Q$.
The triangle inequality for the square-root weights and
Lemma~\ref{lem:q-occupancy-operator} give
\begin{align*}
\|\gamma P_\pi h\|_{2,\widehat w\nu}
&\le
\sqrt\gamma\,\|h\|_{2,d_{\pi,\gamma}}
+(H_Q+2G_Q)H_\nu(\widehat w,w^\pi)\\
&\le
\sqrt\gamma\,\|h\|_{2,\widehat w\nu}
+\{(1+\sqrt\gamma)H_Q+2G_Q\}
H_\nu(\widehat w,w^\pi).
\end{align*}
Cauchy--Schwarz therefore yields
\begin{align*}
\langle h,(I-\gamma P_\pi)h\rangle_{\widehat w\nu}
\ge{}&
(1-\sqrt\gamma)\|h\|_{2,\widehat w\nu}^2\\
&-
\{(1+\sqrt\gamma)H_Q+2G_Q\}
H_\nu(\widehat w,w^\pi)\|h\|_{2,\widehat w\nu}.
\end{align*}
Applying Young's inequality to the final term proves the claim.
\end{proof}

\paragraph{Weights selected on the shared validation fold.}
The independence requirement can be replaced by a finite-library argument,
as in Algorithm~\ref{alg:q-pairwise-selection}.
Suppose $w_1,\ldots,w_M$ are fixed conditional on training, and each satisfies
Conditions~\ref{cond:q-fitted-coercive-weights}--\ref{cond:q-fitted-coercive-domain}
with common envelope constants and the corresponding measure
$\nu_a:=w_a\nu$, oracle $k^\star_{\nu_a}$, coercivity constant
$\kappa_{\nu_a}$, and defect $\varepsilon_{\mathrm{coer},a}$.
Then, with conditional probability at least $1-\delta$,
\eqref{eq:q-fitted-approximate-finite-sample} holds simultaneously for all
$M$ weights with $\log(12MK/\delta)$ in place of $\log(12K/\delta)$.
It therefore holds for $\widehat\nu=w_{\widehat i_w}\nu$ even when
$\widehat i_w$ uses the same validation fold. The bound compares with
the best $Q$-candidate under that selected measure.

%% file: sections/experiments_details.tex
\section{\texorpdfstring{\(Q\)}{Q}-function selection: experimental details and additional results}
\label{app:experiments}

This appendix gives the protocols and additional results for the benchmark
comparisons and controlled-overlap studies in Section~\ref{sec:experiments}.
Occupancy-ratio fitting and its direct evaluation are described
separately in Appendix~\ref{app:occupancy-experiments}.

\subsection{Benchmark designs and candidate classes}

\paragraph{Released LSTD-Tournament benchmark.}
The four model-free Hopper panels vary gravity or transition noise under
the original and poor-coverage logging conditions. Each panel contains ten
target-policy checkpoints evaluated under three environment settings,
giving 120 cases. The model-based gravity and noise panels contain six
targets under three settings each, giving 36 cases. We use the released
offline transitions, candidate predictions, and policy-value caches.
Model-free classes contain 15 simulator-induced $Q$-functions; model-based
classes contain five candidate models, each inducing a $Q$-function and value
estimate. Simulator variation controls candidate errors without training
FQE; the subgrid experiments vary candidate misspecification.
We use new paired bootstrap draws.
We compare both LSTD-Tournament variants and squared TD error with our selectors.

\paragraph{DeepOPE/D4RL.}
We use all eleven target policies for each of HalfCheetah, Hopper, Walker2d,
and Ant, with random, medium, medium-replay, medium-expert, and expert D4RL-v2
datasets. The resulting $4\times5\times11=220$ cases each use 60 FQE candidates.
Four SiLU architectures---two hidden layers of width 64, 128, or 256,
and one hidden layer of width 64---are crossed with learning rates
$10^{-4}$, $3\times10^{-4}$, and $10^{-3}$.
Five checkpoints, after 50, 100, 200, 400, and 800 fitted-Q iterations,
give 60 candidates. Each iteration uses 20 AdamW updates with batch size 512,
weight decay $10^{-4}$, and Huber loss.
Inputs are standardized, and the target network is updated after each
iteration. Fits use $\gamma=0.99$ without early stopping. Training and
validation splits preserve complete trajectories. Each nominal block
contains 51,200 transitions. For HalfCheetah and Walker2d medium-replay,
the available data cannot supply two disjoint blocks of this size: we first
form disjoint trajectory pools and complete each block by resampling whole
trajectories within its own pool, retaining at least 90\% distinct source
transitions. For stochastic targets, $Q$-selection predictions average eight
next-action draws; reset-value evaluation uses 32 action draws.

\paragraph{Controlled-overlap DeepOPE.}
For each DeepOPE environment, the logging policy independently draws one
of the eleven benchmark policies uniformly at the start of each trajectory
and follows it until termination. Every component is evaluated as a target,
giving 44 environment--target cases. Five independent ratio-data replicates
are used per case, while the fitted $Q$-class and its validation pool remain
fixed. Ratio training and validation each use at least 51,200 transitions,
with at least three trajectories from every policy component in each split.
These samples are independent of the $Q$-validation sample.
The mixture is uniform over trajectory starts; pooling complete trajectories
need not give equal transition mass to each component.

\subsection{Selection criteria}

We use unweighted PBV and IW-PBV as defined in Section~\ref{sec:unweightedpbv}
and Algorithm~\ref{alg:q-pairwise-selection}. IW-PBV's $T=5$ FORE weights
are held fixed during $Q$-function selection
(Appendix~\ref{app:occupancy-experiments}).
We also report $T=2$ in Figure~\ref{fig:pbv-default-validation}, keeping the selected ratios fixed.

For LSTD-Tournament, write
$\varepsilon_i=Q_i(X)-R-\gamma C(\pi Q_i)(S')$, where $C$ is the
continuation indicator. The unnormalized variant minimizes
\[
\max_{j\in[M]}|\mathbb P_n[Q_j\varepsilon_i]|.
\]
The normalized-difference variant minimizes
\[
\max\left\{
 \frac{|\mathbb P_n[Q_i\varepsilon_i]|}{\widehat\sigma_n(Q_i)},\;
 \max_{j\ne i}
 \frac{|\mathbb P_n[(Q_i-Q_j)\varepsilon_i]|}
      {\widehat\sigma_n(Q_i-Q_j)}
\right\},\qquad
\widehat\sigma_n(f)^2=\mathbb P_n[(f-\mathbb P_n f)^2].
\]
The variance uses divisor $n$. A zero standard deviation for any required
critic makes the criterion undefined; ties are broken by candidate order.
Squared TD error minimizes $\mathbb P_n[\varepsilon_i^2]$.
Both PBV criteria use absolute empirical moments and absolute empirical
Bellman energies. The zero function is included as a comparator and cannot
be selected. All selectors use identical candidate predictions and validation
draws. DeepOPE IW-PBV treats scores within $10^{-12}$ of the minimum as ties;
Hopper IW-PBV uses exact minima.

\subsection{Evaluation metrics}

Absolute policy-value error is
$|\widehat V-V^\pi|$, where $\widehat V$ is the selected candidate's
estimated policy value. For learned candidates, the estimated reset value
averages $\widehat Q(S_0,A_0)$ over 51,200 reset-state draws, integrating target
actions as above. Reference values come from the released Hopper caches
and, for DeepOPE, from 1,024 target-policy trajectories of horizon at most
1,000.

Following \citet{fu2021benchmarks}, we normalize true and estimated DeepOPE
policy values using the minimum and maximum reference policy values across each
environment's eleven target policies. Absolute errors are thus divided by
this range, and initial-state squared errors by its square before pooling.
The reference policy values use the same discount, $\gamma=0.99$, as FQE.
Each environment's range is shared across dataset qualities, full and
controlled-overlap studies, methods, validation sample sizes, and candidate counts.
Hopper benchmark errors use original return units. The per-environment
DeepOPE curves use the same normalization as the aggregate results.
Validation resamples, ratio-data replicates, and candidate orders are
averaged within each case before cases receive equal weight.

\paragraph{Initial-state value RMSE.}
Let $\widehat v(s)=(\pi\widehat Q)(s)$ and let $v_H^\pi(s)$ be the expected
discounted return from $s$ over horizon $H$, with zero rewards after
termination. We use $H=1{,}024$ for Hopper and $H=1{,}000$ for DeepOPE.
For each of $N_0=1{,}024$ reset states $S_{0j}$, draw two independent initial
actions $A_{0j}^{(k)}\sim\pi(\cdot\mid S_{0j})$ and two conditionally
independent target-policy continuations $G_j^{(k)}$, $k=1,2$.
The estimator
\[
\widehat{\mathcal E}_{v,0}(\widehat Q)
=\frac{1}{N_0}\sum_{j=1}^{N_0}
\prod_{k=1}^{2}
\{\widehat Q(S_{0j},A_{0j}^{(k)})-G_j^{(k)}\}
\]
is unbiased for the mean squared state-value error of $\widehat v$ relative
to $v_H^\pi$ over the reset-state distribution, conditional on the fitted
candidate. Independent initial actions ensure that this targets
state-value error for stochastic policies. Reusing one initial action would
instead target action-value error. For the simulator-defined Hopper
candidates, each candidate prediction is replaced by a return in the
candidate environment from the same reset state. Candidate and reference
environments share random numbers within each branch; the two branches
remain independent. Candidate simulations are reused only after checking
identical candidate and target-policy parameters. DeepOPE references are
shared across dataset qualities for each environment--target pair.
RMSE is the square root after pooling the estimated squared errors across
selections and cases, rather than the mean of case-level square roots.

\subsection{Validation draws and uncertainty}

The released Hopper and full DeepOPE comparisons use $n_Q=3{,}200$.
We also report controlled-overlap DeepOPE at $n_Q=51{,}200$.
Released Hopper uses 96 paired transition
bootstrap draws per case in both the model-free and model-based settings.
Its model-free sample-size curves use nested draws at
$n_Q\in\{8,464,920,1376,1832,2288,2744,3200\}$.
DeepOPE uses 96 complete-trajectory bootstrap draws at each nominal size.
The full and controlled-overlap DeepOPE grid is
$n_Q\in\{3{,}200,6{,}400,12{,}800,25{,}600,51{,}200\}$.
Whole-trajectory sampling can exceed the nominal size. Reference values and
evaluation losses are used only after selection; all methods use the same
validation draws.

Table~\ref{tab:pbv-paired-differences} reports 95\% intervals from 10,000
bootstrap draws over condition--truth groups for Hopper and
environment--dataset groups for DeepOPE, after averaging validation resamples,
ratio-data replicates, and candidate orders within cases. Methods remain
paired in every draw. Controlled-overlap Hopper uses its six
perturbation--truth groups.
Controlled-overlap DeepOPE instead resamples its four environments, averaging
the five ratio-data replicates within each target case first.
For RMSE differences, each resampled mean squared error is square-rooted
before taking the difference. The ranges of reference policy values remain fixed during
resampling. These intervals condition on the fitted candidate libraries and
saved Monte Carlo reference values.
Validation resamples are not independent training replications; the
controlled-overlap DeepOPE study has five independent ratio-data replicates
per target, whereas the original benchmarks have one fitted ratio library
per case.

\input{figures/pbv_default_paired.tex}
\FloatBarrier

\subsection{Candidate counts}
\label{app:candidate-counts}

The 15-candidate class contains the five checkpoints from each two-layer
architecture at learning rate $3\times10^{-4}$. The 30-, 45-, and 60-candidate
classes add complete training runs from the remaining architecture--rate
pairs, using 20 fixed nested orders shared by all methods and cases.
The fitted functions, target-action draws, validation resamples, and
occupancy weights are shared across candidate counts. Selections are fixed
before reference returns or candidate evaluation losses are accessed.

\input{figures/pbv_default_candidate_counts.tex}
\FloatBarrier

\subsection{Zero comparator}
\label{app:zero-comparator}

\begin{figure}[!htbp]
\centering
\includegraphics[width=.99\linewidth]{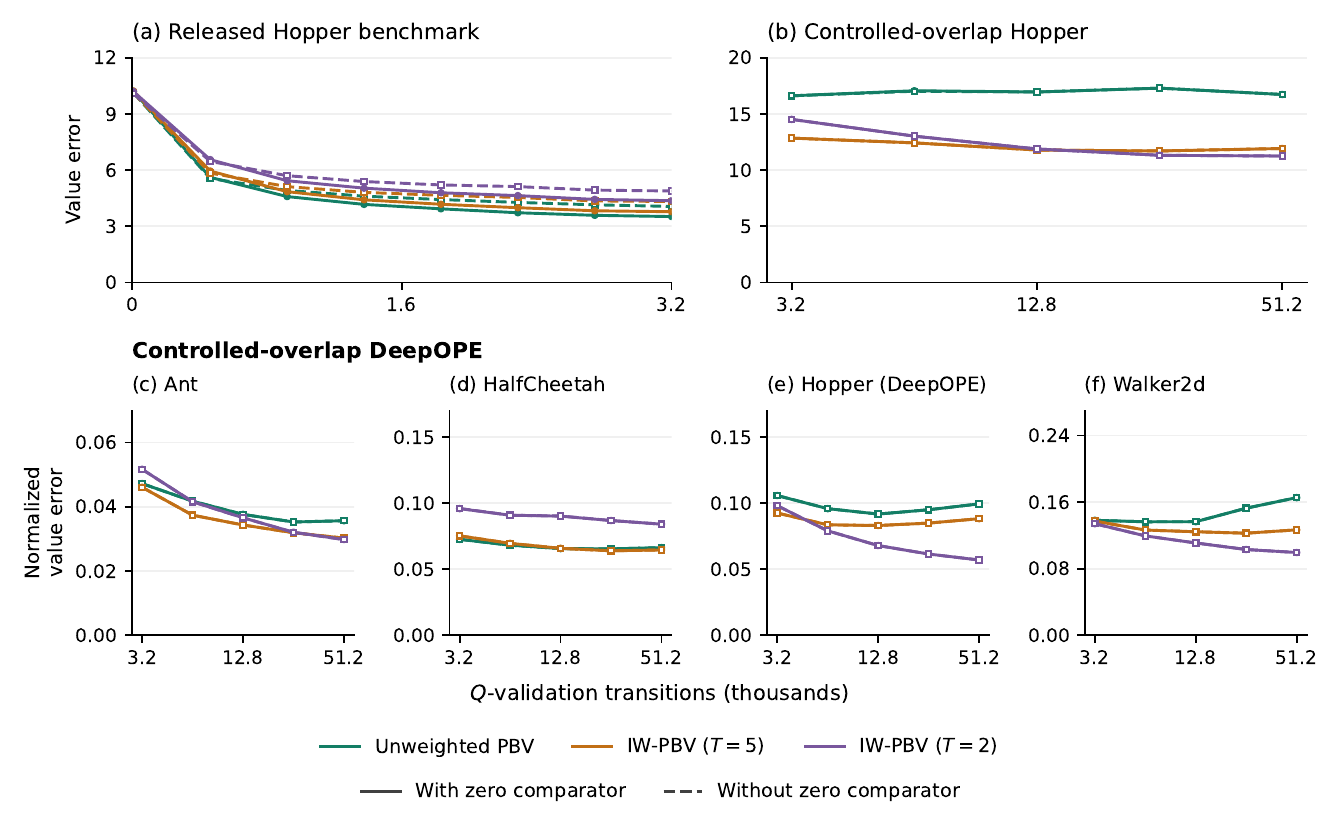}
\caption{Zero-comparator ablation for the six panels of
Figure~\ref{fig:pbv-default-validation}. Solid curves include zero as a
comparator; dashed curves omit it. Candidates, selected occupancy ratios,
validation draws, and tie rules are shared within each pair. The two Hopper
benchmark panels use original return units; DeepOPE errors use each
environment's fixed range of reference policy values.}
\label{fig:pbv-zero-comparator}
\end{figure}

We repeat the PBV and IW-PBV comparisons with and without the zero
comparator, holding fitted candidates, selected occupancy ratios, validation
draws, and tie rules fixed. Zero is never selectable.
Table~\ref{tab:pbv-zero-comparator} uses the benchmark groups and validation
sample size of Table~\ref{tab:pbv-main-results}; Figure~\ref{fig:pbv-zero-comparator}
uses the cases and sample sizes of Figure~\ref{fig:pbv-default-validation}.
Including zero substantially reduces both errors on the poor-coverage
model-free Hopper grid. Changes are small on the main model-free and
model-based Hopper grids and on full DeepOPE; the controlled-overlap curves
are nearly unchanged.

\input{figures/pbv_zero_comparator_table.tex}
\FloatBarrier

\subsection{Critic penalty}
\label{app:critic-penalty}

We ablate only the $Q$-selection penalty, changing its coefficient from
$1/4$ to zero. Fitted candidates, selected occupancy weights, validation
draws, candidate orders, and tie rules are fixed; zero remains a comparator
and is never selectable. A-PBV ratio selection is unchanged.
Table~\ref{tab:pbv-penalty-ablation} reports this exploratory matched
comparison at $n_Q=3{,}200$. Selections are frozen before evaluation losses
are accessed.

\input{figures/pbv_penalty_ablation.tex}

Removing the penalty increases mean absolute policy-value error on all
three released Hopper grids and on controlled-overlap DeepOPE for all
three selectors. It slightly reduces both errors on full DeepOPE and
improves unweighted PBV on controlled-overlap Hopper. On controlled-overlap
DeepOPE, removing the IW-PBV penalty increases value error while slightly
reducing initial-state value RMSE. Most released-Hopper paired intervals
contain zero.
\FloatBarrier

\subsection{Full DeepOPE on D4RL datasets}
\label{app:full-deepope}

Table~\ref{tab:pbv-full-deepope} reports all 220 D4RL environment--dataset--target
cases with 60 FQE candidates. Aggregate errors are similar across the PBV
and LSTD-Tournament variants, with slightly lower errors for unnormalized
LSTD-Tournament at both validation sample sizes. The mean value errors of the PBV and LSTD-Tournament variants are
close in HalfCheetah and Ant. In Hopper, unnormalized LSTD-Tournament has
lower mean value error than both temperatures of IW-PBV; in Walker2d,
$T=2$ gives lower mean value error than both LSTD-Tournament variants
at every displayed sample size.

\input{figures/pbv_full_deepope_results.tex}
\input{figures/pbv_default_deep_environment.tex}
\FloatBarrier

\input{sections/lstd_reproduction_figures.tex}

\subsection{Controlled-overlap Hopper}
\label{app:controlled-hopper}

The controlled-overlap Hopper study uses the same 60-candidate FQE construction
as DeepOPE, with independent training and validation samples of at least
51,200 transitions. The logging policy draws one of the ten benchmark policies
uniformly at the start of each trajectory and follows it until termination.
Ten targets in each of two perturbation grids and three true environments
give 60 cases. Thus, these candidates and data are distinct from the released
simulator-induced $Q$-classes in the LSTD-Tournament comparison.
Each case uses five fixed permutations of its 51,200-transition validation
pool, with nested subsamples; these are validation draws, not independent
FQE refits. Occupancy weights are held fixed across candidate counts.
Table~\ref{tab:pbv-controlled-hopper} reports both endpoints of the validation
grid. At $T=5$, IW-PBV has lower mean absolute policy-value error and
initial-state value RMSE than unweighted PBV and both LSTD-Tournament variants
at both sample sizes. The $T=5$ errors are lower than those at $T=2$ for
$n_Q=3{,}200$; this ordering reverses at $n_Q=51{,}200$.

\input{figures/pbv_default_controlled_hopper.tex}
\FloatBarrier

%% file: figures/pbv_default_paired.tex
\begin{table}[!htbp]
\centering
\small
\setlength{\tabcolsep}{3pt}
\renewcommand{\arraystretch}{1.08}
\caption{Paired differences at \(n_Q=3{,}200\), with 95\% cluster-bootstrap intervals; negative favors the first method. Released Hopper uses 15 model-free or five model-based candidates; all learned classes contain 60 candidates. Hopper errors use original return units; DeepOPE errors are normalized by the range of reference policy values in each environment. Resamples, ratio-data replicates, and candidate orders are averaged within cases before resampling condition--truth or environment--dataset groups; controlled DeepOPE resamples its four environments. Candidate fits and reference values are held fixed.}
\label{tab:pbv-paired-differences}
\begin{tabular*}{\linewidth}{@{\extracolsep{\fill}}llrr@{}}
\toprule
Benchmark & Contrast & Value error & Value RMSE \\
\midrule
Hopper MF & PBV $-$ LSTD (norm.) & -1.32 [-3.52, -0.05] & -3.23 [-5.88, -0.12] \\
Hopper MF & IW-PBV $-$ LSTD (norm.) & -0.79 [-2.30, +0.23] & -2.41 [-4.41, +0.23] \\
Hopper MF & PBV $-$ LSTD (unnorm.) & -0.69 [-1.15, -0.21] & -1.68 [-2.79, -0.31] \\
Hopper MF & IW-PBV $-$ LSTD (unnorm.) & -0.16 [-0.88, +0.64] & -0.86 [-1.37, -0.10] \\
Hopper poor & PBV $-$ LSTD (norm.) & -2.78 [-5.49, -0.29] & -3.99 [-5.82, -0.30] \\
Hopper poor & IW-PBV $-$ LSTD (norm.) & -2.79 [-5.56, -0.29] & -4.04 [-5.88, -0.30] \\
Hopper poor & PBV $-$ LSTD (unnorm.) & -1.24 [-3.60, +0.34] & -1.92 [-3.90, -0.32] \\
Hopper poor & IW-PBV $-$ LSTD (unnorm.) & -1.25 [-3.59, +0.36] & -1.97 [-3.89, -0.43] \\
Hopper MB & PBV $-$ LSTD (norm.) & -0.11 [-0.23, +0.00] & -0.34 [-0.51, -0.11] \\
Hopper MB & IW-PBV $-$ LSTD (norm.) & +0.24 [+0.03, +0.48] & +0.33 [-0.07, +0.71] \\
Hopper MB & PBV $-$ LSTD (unnorm.) & -0.64 [-1.40, +0.07] & -1.25 [-2.41, -0.27] \\
Hopper MB & IW-PBV $-$ LSTD (unnorm.) & -0.28 [-1.07, +0.46] & -0.58 [-1.68, +0.49] \\
Full DeepOPE & PBV $-$ LSTD (norm.) & +0.002 [-0.011, +0.018] & +0.002 [-0.015, +0.021] \\
Full DeepOPE & IW-PBV $-$ LSTD (norm.) & +0.001 [-0.018, +0.024] & +0.003 [-0.022, +0.030] \\
Full DeepOPE & PBV $-$ LSTD (unnorm.) & +0.012 [+0.000, +0.026] & +0.014 [+0.000, +0.028] \\
Full DeepOPE & IW-PBV $-$ LSTD (unnorm.) & +0.011 [-0.007, +0.032] & +0.014 [-0.007, +0.037] \\
Controlled Hopper & IW-PBV $-$ PBV & -3.77 [-11.16, +1.04] & -8.15 [-14.46, +0.15] \\
Controlled Hopper & IW-PBV $-$ LSTD (norm.) & -1.71 [-5.36, +1.47] & -4.28 [-7.29, +0.44] \\
Controlled Hopper & IW-PBV $-$ LSTD (unnorm.) & -3.20 [-8.88, +0.17] & -8.48 [-14.97, -0.65] \\
Controlled DeepOPE & IW-PBV $-$ PBV & -0.003 [-0.010, +0.002] & -0.001 [-0.012, +0.006] \\
Controlled DeepOPE & IW-PBV $-$ LSTD (norm.) & -0.018 [-0.036, -0.005] & -0.019 [-0.050, +0.002] \\
Controlled DeepOPE & IW-PBV $-$ LSTD (unnorm.) & -0.031 [-0.040, -0.018] & -0.055 [-0.075, -0.030] \\
\bottomrule
\end{tabular*}
\end{table}

%% file: figures/pbv_default_candidate_counts.tex
\begin{figure}[!htbp]
\centering
\includegraphics[width=.99\linewidth]{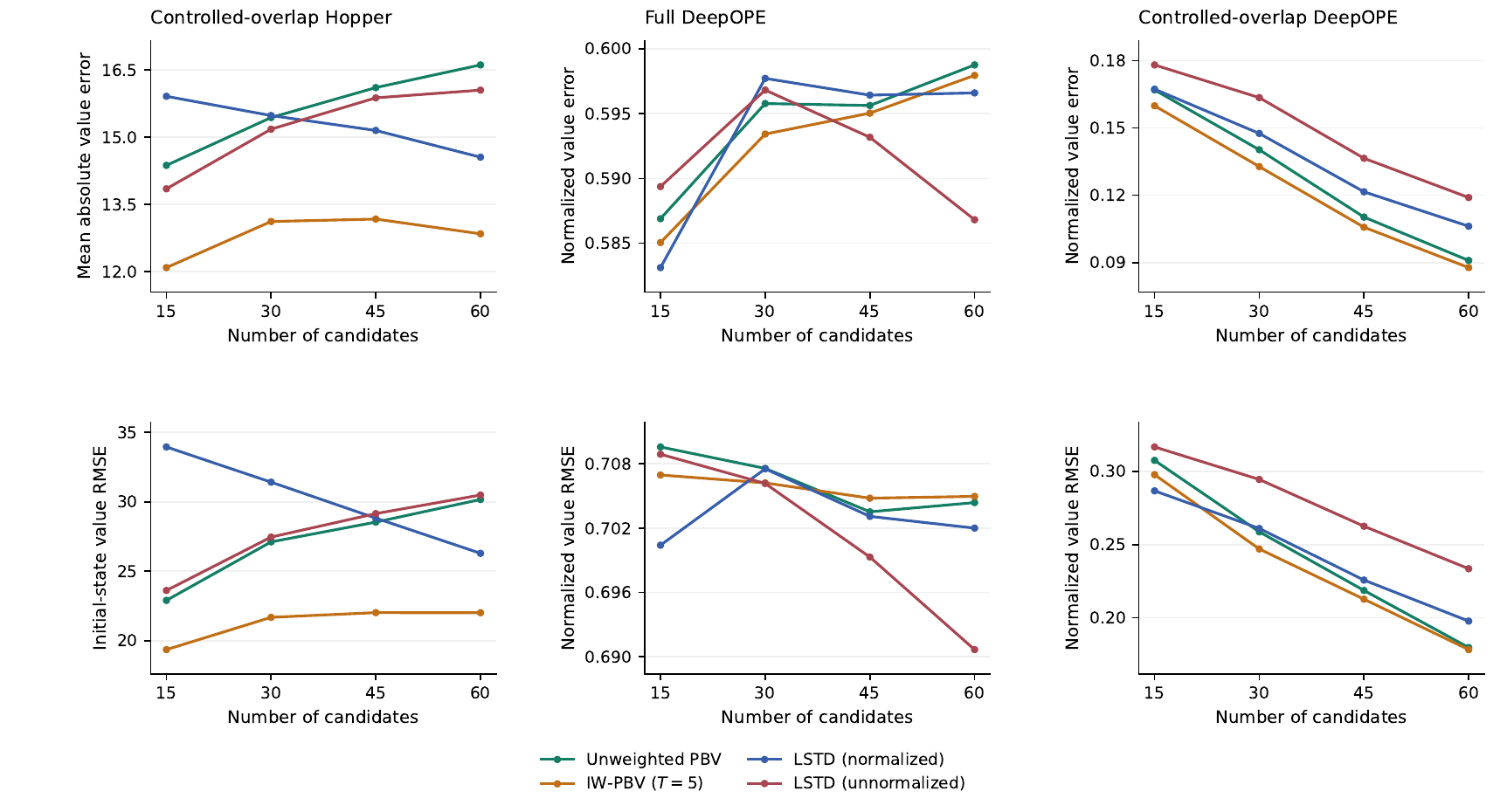}
\caption{Selection performance with 15, 30, 45, and 60 FQE candidates at
$n_Q=3{,}200$. Top: mean absolute policy-value error. Bottom: initial-state
value RMSE, with the square root taken after pooling squared errors.
DeepOPE errors are normalized by the range of reference policy values in
each environment; Hopper errors use original return units. Validation draws, ratio-data
replicates, and the 20 fixed nested candidate orders are averaged within
cases before averaging cases.}
\label{fig:pbv-candidate-counts}
\end{figure}

%% file: figures/pbv_zero_comparator_table.tex
\begin{table}[!htbp]
\centering\small
\setlength{\tabcolsep}{3pt}
\renewcommand{\arraystretch}{1.08}
\caption{Zero-comparator ablation for the comparisons in Table~\ref{tab:pbv-main-results}, at $n_Q=3{,}200$. VE is mean absolute policy-value error. Each pair uses identical candidates, occupancy weights, validation draws, and tie rules. Zero is never selectable. Units, normalization, case weights, and RMSE aggregation are as in Table~\ref{tab:pbv-main-results}.}
\label{tab:pbv-zero-comparator}
\begin{tabular*}{\linewidth}{@{\extracolsep{\fill}}llrrrrrrrr@{}}
\toprule
&& \multicolumn{2}{c}{MF (60)} & \multicolumn{2}{c}{Poor MF (60)} & \multicolumn{2}{c}{MB (36)} & \multicolumn{2}{c}{DeepOPE (220)} \\
\cmidrule(lr){3-4}\cmidrule(lr){5-6}\cmidrule(lr){7-8}\cmidrule(lr){9-10}
Selector & Zero & VE & RMSE & VE & RMSE & VE & RMSE & VE & RMSE \\
\midrule
Unweighted PBV & With & 1.78 & 4.22 & 5.26 & 10.28 & 1.25 & 3.09 & 0.599 & 0.704 \\
 & Without & 1.79 & 4.21 & 6.33 & 12.30 & 1.25 & 3.09 & 0.598 & 0.704 \\
\addlinespace
IW-PBV ($T=5$) & With & 2.31 & 5.04 & 5.25 & 10.24 & 1.61 & 3.76 & 0.598 & 0.705 \\
 & Without & 2.29 & 5.02 & 6.36 & 12.31 & 1.61 & 3.76 & 0.598 & 0.705 \\
\addlinespace
IW-PBV ($T=2$) & With & 3.47 & 6.87 & 5.27 & 10.27 & 2.07 & 4.52 & 0.587 & 0.695 \\
 & Without & 3.40 & 6.74 & 6.37 & 12.31 & 2.05 & 4.49 & 0.587 & 0.694 \\
\bottomrule
\end{tabular*}
\end{table}

%% file: figures/pbv_penalty_ablation.tex
% Source: output/penalty_ablation_20261031/{relative_changes,paired_contrasts}.csv
% Frozen exploratory comparison: penalty coefficients 1/4 and 0; n_Q=3200.
\begin{table}[!htbp]
\centering\small
\setlength{\tabcolsep}{3pt}
\renewcommand{\arraystretch}{1.08}
\caption{Critic-penalty ablation at $n_Q=3{,}200$. VE is mean absolute policy-value error; RMSE is initial-state value RMSE. Entries show with penalty / without penalty (percentage change upon removal); positive changes favor retaining the penalty. Bold percentages indicate paired 95\% cluster-bootstrap intervals excluding zero, using the groups and protocol described above. Hopper errors use original return units; DeepOPE errors use fixed per-environment reference ranges. Parentheses in the first column give case counts. Released Hopper uses 15 model-free or five model-based candidates; learned classes use 60.}
\label{tab:pbv-penalty-ablation}
\begin{tabular*}{\linewidth}{@{\extracolsep{\fill}}llrr@{}}
\toprule
Benchmark & Selector & VE: with / without & RMSE: with / without \\
\midrule
\multirow{3}{*}{Hopper MF (60)} & Unweighted PBV & 1.78 / 2.15 (+21.1\%) & 4.22 / 4.98 (\textbf{+17.9\%}) \\
 & IW-PBV ($T=5$) & 2.31 / 3.06 (\textbf{+32.8\%}) & 5.04 / 6.38 (\textbf{+26.7\%}) \\
 & IW-PBV ($T=2$) & 3.47 / 4.45 (\textbf{+28.2\%}) & 6.87 / 8.45 (\textbf{+22.9\%}) \\
\addlinespace
\multirow{3}{*}{Hopper poor MF (60)} & Unweighted PBV & 5.26 / 5.66 (+7.6\%) & 10.28 / 10.37 (+0.8\%) \\
 & IW-PBV ($T=5$) & 5.25 / 5.59 (+6.5\%) & 10.24 / 10.28 (+0.4\%) \\
 & IW-PBV ($T=2$) & 5.27 / 5.62 (+6.8\%) & 10.27 / 10.38 (+1.1\%) \\
\addlinespace
\multirow{3}{*}{Hopper MB (36)} & Unweighted PBV & 1.25 / 1.90 (+52.0\%) & 3.09 / 4.45 (\textbf{+44.0\%}) \\
 & IW-PBV ($T=5$) & 1.61 / 2.29 (+42.6\%) & 3.76 / 5.03 (\textbf{+33.6\%}) \\
 & IW-PBV ($T=2$) & 2.07 / 2.84 (+37.3\%) & 4.52 / 5.68 (\textbf{+25.7\%}) \\
\addlinespace
\multirow{3}{*}{Full DeepOPE (220)} & Unweighted PBV & 0.599 / 0.589 (\textbf{-1.6\%}) & 0.704 / 0.692 (\textbf{-1.7\%}) \\
 & IW-PBV ($T=5$) & 0.598 / 0.590 (\textbf{-1.4\%}) & 0.705 / 0.696 (\textbf{-1.3\%}) \\
 & IW-PBV ($T=2$) & 0.587 / 0.582 (\textbf{-0.9\%}) & 0.695 / 0.689 (\textbf{-0.8\%}) \\
\addlinespace
\multirow{3}{*}{Controlled Hopper (60)} & Unweighted PBV & 16.61 / 15.60 (\textbf{-6.1\%}) & 30.17 / 28.96 (-4.0\%) \\
 & IW-PBV ($T=5$) & 12.85 / 13.09 (+1.9\%) & 22.01 / 22.14 (+0.6\%) \\
 & IW-PBV ($T=2$) & 14.51 / 15.86 (+9.3\%) & 22.20 / 27.58 (+24.2\%) \\
\addlinespace
\multirow{3}{*}{Controlled DeepOPE (44)} & Unweighted PBV & 0.091 / 0.106 (\textbf{+16.8\%}) & 0.180 / 0.198 (\textbf{+10.2\%}) \\
 & IW-PBV ($T=5$) & 0.088 / 0.095 (\textbf{+8.8\%}) & 0.178 / 0.178 (-0.2\%) \\
 & IW-PBV ($T=2$) & 0.095 / 0.100 (+5.0\%) & 0.183 / 0.179 (-2.4\%) \\
\bottomrule
\end{tabular*}
\end{table}

%% file: figures/pbv_full_deepope_results.tex
\begin{table}[!htbp]
\centering
\small
\setlength{\tabcolsep}{4pt}
\renewcommand{\arraystretch}{1.08}
\caption{Full DeepOPE on 220 D4RL environment--dataset--target cases with 60 FQE candidates: mean absolute policy-value error (VE) and initial-state value RMSE. Errors are normalized by each environment's reference policy-value range. Cases receive equal weight; RMSE takes the square root after pooling squared errors. Bold marks the smallest mean.}
\label{tab:pbv-full-deepope}
\begin{tabular*}{\linewidth}{@{\extracolsep{\fill}}lrrrr@{}}
\toprule
& \multicolumn{2}{c}{$n_Q=3{,}200$} & \multicolumn{2}{c}{$n_Q=51{,}200$} \\
\cmidrule(lr){2-3}\cmidrule(lr){4-5}
Selector & VE & RMSE & VE & RMSE \\
\midrule
Unweighted PBV & 0.599 & 0.704 & 0.601 & 0.707 \\
IW-PBV ($T=5$) & 0.598 & 0.705 & 0.604 & 0.711 \\
IW-PBV ($T=2$) & 0.587 & 0.695 & 0.586 & 0.690 \\
LSTD-Tournament (normalized) & 0.597 & 0.702 & 0.593 & 0.697 \\
LSTD-Tournament (unnormalized) & \textbf{0.587} & \textbf{0.691} & \textbf{0.582} & \textbf{0.683} \\
Squared TD error & 0.733 & 0.826 & 0.744 & 0.834 \\
\bottomrule
\end{tabular*}
\end{table}

%% file: figures/pbv_default_deep_environment.tex
\begin{figure}[!htbp]
\centering
\includegraphics[width=.99\linewidth]{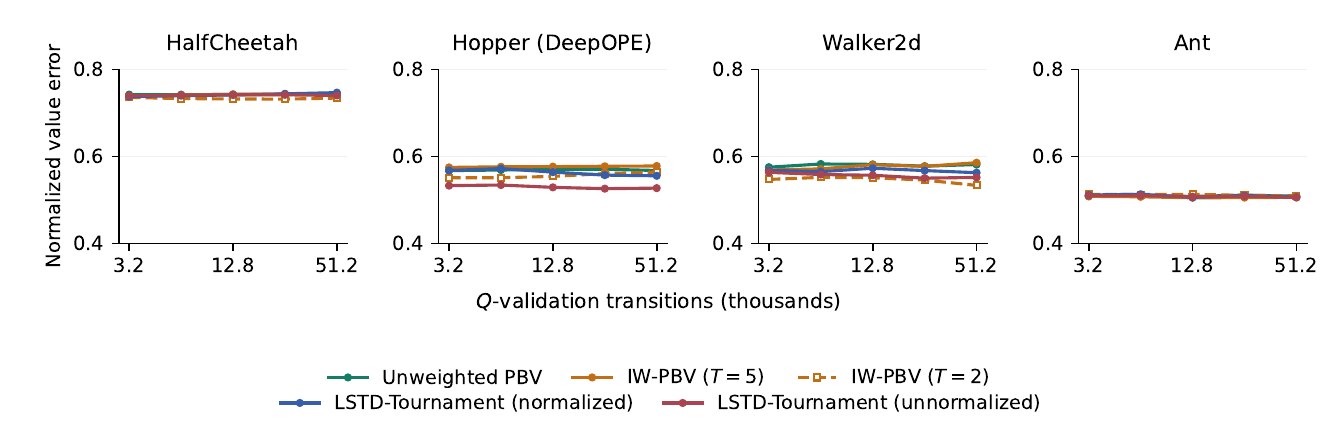}
\caption{Full DeepOPE on D4RL-v2 datasets, with 60 FQE candidates per case.
Each panel averages five dataset qualities and eleven target policies.
Errors are divided by each environment's reference policy-value range.
IW-PBV uses the fixed temperatures $T=2$ and $T=5$ with the same selected ratios.
All four vertical axes use the range $0.4$--$0.8$.
Hopper here uses DeepOPE policies and D4RL data, separately from the
released and controlled-overlap Hopper benchmarks.}
\label{fig:pbv-deep-environments}
\end{figure}

%% file: sections/lstd_reproduction_figures.tex
\subsection{Results on the released Hopper benchmark grids}
\label{app:lstd-grid-figures}

\begin{figure}[!htbp]
\centering
\includegraphics[width=.99\linewidth]{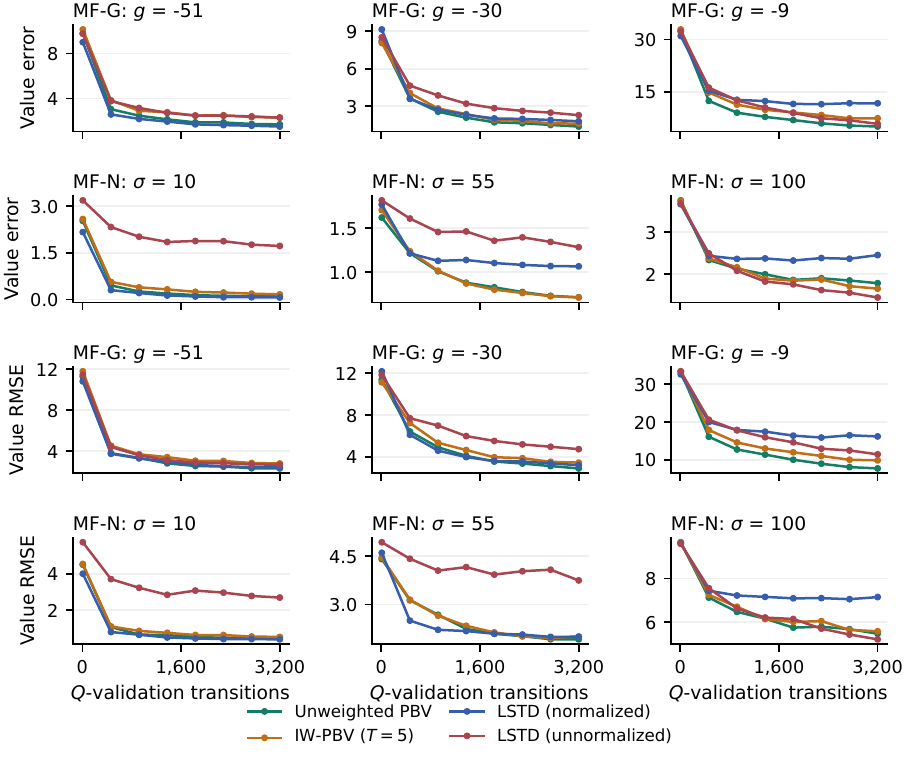}
\caption{Main model-free Hopper grids, corresponding to Figure~3 of
\citet{liu2025modelselection}. Columns correspond to the three true
environments; each panel averages ten target policies. The first two rows
show absolute value error for the gravity and noise grids, respectively;
the last two show initial-state value RMSE. Each point averages 96 paired
validation draws per target. IW-PBV uses independent data for
FORE fitting, with weights tempered by power $1/5$.}
\label{fig:lstd-model-free}
\end{figure}

\begin{figure}[!htbp]
\centering
\includegraphics[width=.99\linewidth]{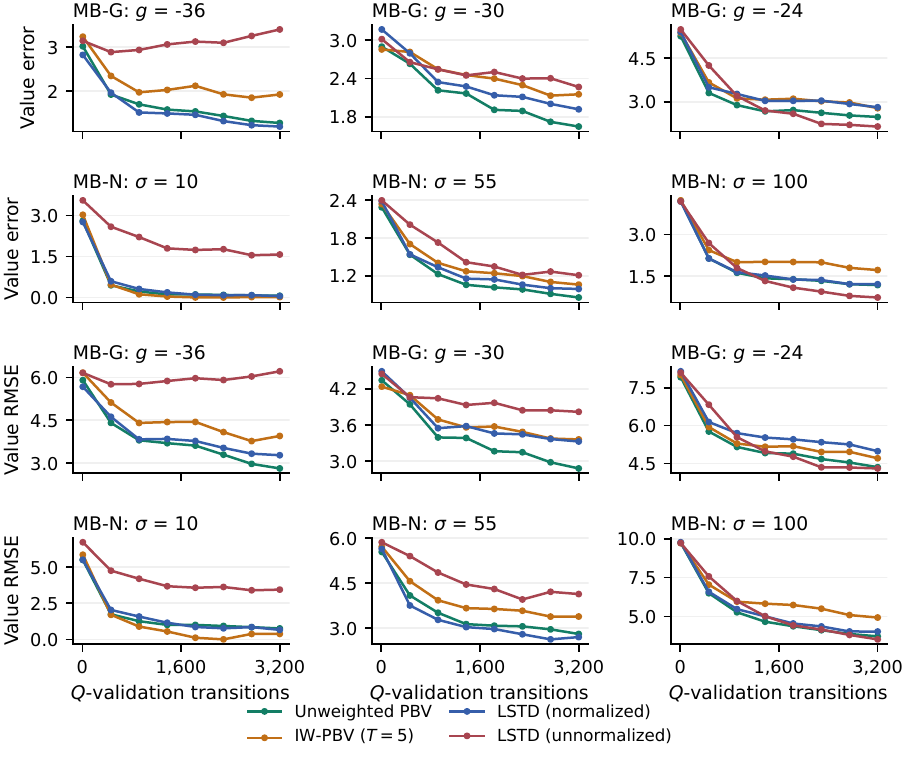}
\caption{Five-candidate model-based Hopper grids, corresponding to the
benchmark in Figure~4 of \citet{liu2025modelselection}. Columns correspond
to the three true environments; each panel averages six target policies.
Rows show gravity-grid value error, noise-grid value error, gravity-grid
initial-state value RMSE, and noise-grid initial-state value RMSE.
All four selectors use the same model-induced $Q$-functions and 96
paired validation draws per target; the comparison does not include the
original paper's additional model-based selectors.}
\label{fig:lstd-model-based}
\end{figure}

\begin{figure}[!htbp]
\centering
\includegraphics[width=.99\linewidth]{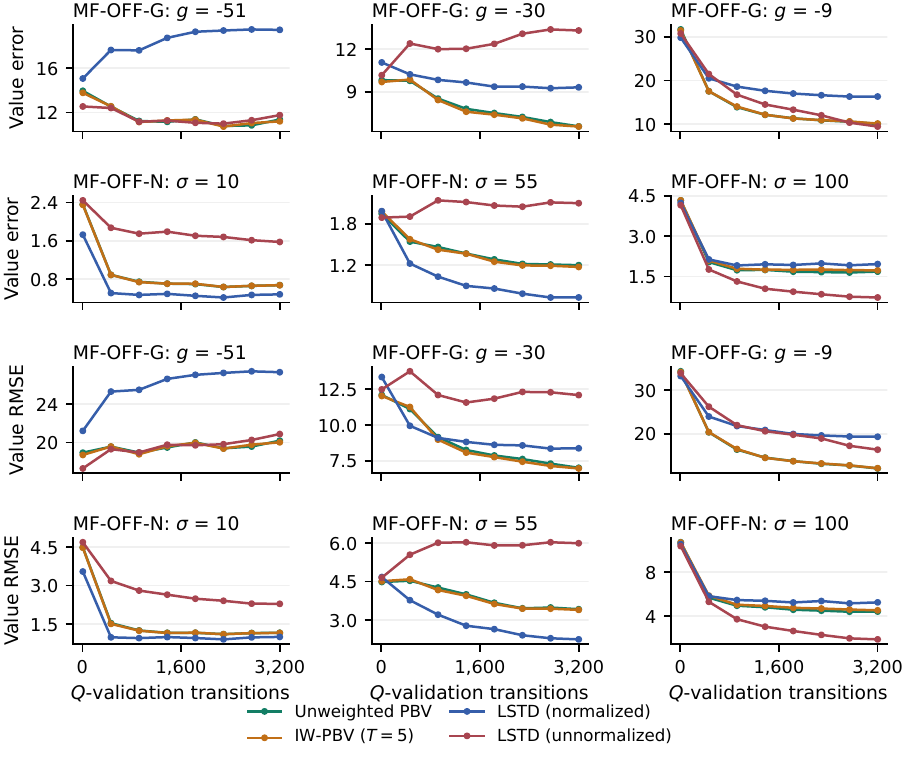}
\caption{Poor-coverage Hopper grids, corresponding to Figure~9 of
\citet{liu2025modelselection}. The layout and averaging match
Figure~\ref{fig:lstd-model-free}. Unweighted PBV and LSTD-Tournament use the
released logging data. IW-PBV additionally uses ratios fitted
from the reconstructed behavior sources described in
Appendix~\ref{app:occupancy-experiments}.}
\label{fig:lstd-poor-coverage}
\end{figure}

\begin{figure}[!htbp]
\centering
\includegraphics[width=.99\linewidth]{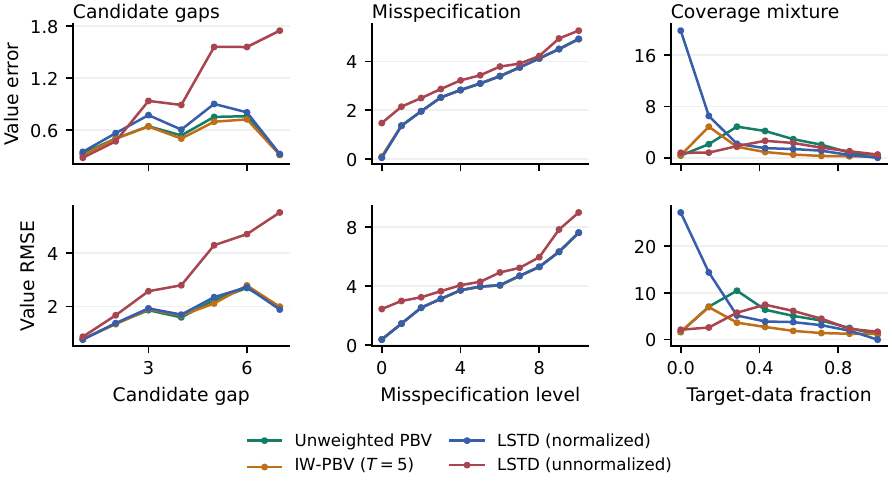}
\caption{Candidate gaps, misspecification, and coverage mixtures,
corresponding to the representative studies in Figure~5 of
\citet{liu2025modelselection}. Top: mean absolute value error.
Bottom: initial-state value RMSE. Unweighted PBV and LSTD-Tournament use the
released logging data; IW-PBV uses additional independent ratio
data, with deterministic reconstructed background actors in the coverage
panel.}
\label{fig:lstd-subgrids}
\end{figure}

Figures~\ref{fig:lstd-model-free}--\ref{fig:lstd-subgrids} use the
benchmark conditions of Figures~3--5 and~9 of
\citet{liu2025modelselection}, comparing unweighted PBV and IW-PBV with
both LSTD-Tournament variants.
The main model-free grids contain 60 cases; the poor-coverage counterparts
contain another 60. The five-candidate model-based grids contain 36 cases.
We show absolute value error and initial-state value RMSE on the original
return scale. The plots report our paired reevaluation of the released candidates.
Each curve first averages validation resamples within each target and then
averages across targets; RMSE takes the square root after averaging
squared errors.
The fitted candidates and released Monte Carlo reference values remain fixed.
Paired uncertainty for the aggregate comparisons is reported in
Table~\ref{tab:pbv-paired-differences}.
Among the two LSTD-Tournament variants, the normalized variant has lower
aggregate errors on the model-based Hopper grids,
whereas unnormalized LSTD-Tournament has lower errors on the main and
poor-coverage model-free grids (Table~\ref{tab:pbv-main-results}).

\FloatBarrier

\FloatBarrier

\FloatBarrier

The subgrid experiments reuse the released candidate predictions.
For the gap experiments, the true noise-grid model is indexed by 7,
and the candidates have indices $\{7-k,7,7+k\}$, $k=1,\ldots,7$.
For misspecification, the true model is indexed by 0 and the candidate
set is $\{k,\ldots,k+4\}$, $k=0,\ldots,10$.
These panels average ten targets and 32 paired resamples per subset,
with $n_Q=2{,}048$ for gaps and $3{,}200$ for misspecification.
The coverage panel uses the gravity-grid model indexed by 0, target policy 8,
eight target-data fractions, and 256 paired resamples at $n_Q=3{,}200$.
The choice of policy 8 follows the released implementation and Table~1
of \citet{liu2025modelselection}.
For IW-PBV, fresh auxiliary background and target-policy rollouts
are mixed in the same proportions. The background uses the reconstructed
deterministic actor, without added action noise.
Ratio training and validation each use 3,200 transitions, with the same
56-candidate FORE class, A-PBV selection, and $1/5$ tempering.
The original $Q$-validation draws are retained.
Because the original background checkpoints are unavailable, IW-PBV
uses additional data from reconstructed policies;
Appendix~\ref{app:occupancy-experiments} details this limitation.

\FloatBarrier

%% file: figures/pbv_default_controlled_hopper.tex
\begin{table}[!htbp]
\centering
\small
\setlength{\tabcolsep}{3pt}
\renewcommand{\arraystretch}{1.08}
\caption{Controlled-overlap Hopper with 60 FQE candidates: mean absolute policy-value error and initial-state value RMSE in original return units. Each entry averages 60 environment--target cases after averaging validation draws and candidate orders within cases. RMSE takes the square root after pooling squared errors.}
\label{tab:pbv-controlled-hopper}
\begin{tabular*}{\linewidth}{@{\extracolsep{\fill}}lrrrr@{}}
\toprule
& \multicolumn{2}{c}{\(n_Q=3{,}200\)} & \multicolumn{2}{c}{\(n_Q=51{,}200\)} \\
\cmidrule(lr){2-3}\cmidrule(lr){4-5}
Selector & Value error & Value RMSE & Value error & Value RMSE \\
\midrule
Unweighted PBV & 16.61 & 30.17 & 16.73 & 30.98 \\
IW-PBV ($T=5$) & 12.85 & 22.01 & 11.92 & 21.35 \\
IW-PBV ($T=2$) & 14.51 & 22.20 & 11.25 & 19.31 \\
LSTD-Tournament (normalized) & 14.55 & 26.29 & 14.64 & 25.81 \\
LSTD-Tournament (unnormalized) & 16.05 & 30.49 & 16.87 & 32.23 \\
Squared TD error & 99.01 & 116.71 & 98.16 & 116.66 \\
\bottomrule
\end{tabular*}
\end{table}

%% file: sections/occupancy_experiments.tex
\section{Occupancy-ratio selection experiments}
\label{app:occupancy-experiments}

We compare A-PBV with our adjoint LSTD extension on direct occupancy-ratio
errors and on the resulting IW-PBV selections. The direct comparison uses
FORE, DualDICE, and their combined candidate library; the downstream comparison
uses five benchmark groups. A-PBV gives lower aggregate direct ratio
errors across all three libraries, while downstream performance varies across
benchmarks. We also examine tempering; the main experiments report the
fixed temperatures $T=2$ and $T=5$.

\subsection{Ratio fitting, screening, and normalization}
\label{app:occupancy-fitting}

\paragraph{Benchmark FORE libraries.}
The benchmark ratio libraries contain neural FORE estimators
\citep{van2026fitted} fitted with AdamW, learning rates
$3\times10^{-4}$ and $10^{-3}$, weight decay $10^{-4}$, and checkpoints
after 25, 50, 100, 200, 400, 800, and 1,600 updates.
DeepOPE uses one or two hidden layers of width 64, giving 28 candidates.
The Hopper extension also includes two-layer networks of widths 32 and
128, giving 56 candidates. Ratio training and validation each use 3,200
transitions for Hopper, 25,600 for full DeepOPE, and at least 51,200
for controlled-overlap DeepOPE. The latter uses five independent data
replicates per environment--target pair.
In the released 15-candidate Hopper comparisons, ratio predictions outside
the known target-action support are set to the prediction floor.
The model-based and coverage-mixture extensions use unprojected predictions.

\paragraph{Direct ratio libraries.}
FORE uses architectures $(64)$, $(64,64)$, $(32,32)$, and $(128,128)$,
with the learning rates and checkpoints above. DualDICE uses architectures
$(64)$, $(64,64)$, $(128,128)$, and $(256,256)$; multipliers $0.3$ and $1$
on its default learning rates $10^{-4}$ and $10^{-3}$ for the two networks;
and checkpoints at 100, 250, 500, 1,000, 2,500, 5,000, and 10,000 updates.
We use the unchanged continuous-action update from the official Google
implementation. FORE and DualDICE use minibatches of 1,024 and 256,
respectively.

\paragraph{Screening and normalization.}
The screen follows the fitted estimator's training normalization.
Let $u_j$ denote its predictions after flooring the raw output at $10^{-6}$
and applying the training normalization.
Using the ratio-training empirical measure $\mathbb P_{n,w}$ and the
continuation indicator $C$, define
\[
c_j(x)=\min\{u_j(x),50\},\qquad
v_j(x)=\frac{(1-\gamma)c_j(x)}
{\mathbb P_{n,w}[c_j(X)(1-\gamma C)]}.
\]
This rescaling enforces the empirical occupancy-mass identity on the
training sample. We retain candidate $j$ if
\[
\frac{(\mathbb P_{n,w}v_j)^2}{\mathbb P_{n,w}v_j^2}\ge 0.05,
\]
and use $\widetilde w_j(x)=\min\{v_j(x),50\}$ for selection and evaluation.
The capped candidates are not rescaled before ratio selection.
The same transformation applies
to current-state, next-target, reset, and $Q$-validation predictions.
For stochastic DeepOPE targets, ratio validation uses one common target-action
draw per next state and reset state, shared by all candidates.
The $Q$-selection action averages are described in
Appendix~\ref{app:experiments}.

For downstream weighting, we use
$w_T=\widetilde w^{1/T}/\mathbb P_{n,w}[\widetilde w^{1/T}]$ with $T=5$
\citep{sinha2022experience}. The normalizer is a positive common factor
and leaves IW-PBV selections unchanged.
The temperature comparison also includes $T=1$, which uses the
untempered ratio, and $T=2,3,4$.

\paragraph{Auxiliary data for Hopper.}
The unweighted Hopper comparisons reuse the released logging data.
IW-PBV additionally uses independently collected ratio-training
and validation samples. The poor-coverage behavior checkpoints are absent,
so these rollouts use policies reconstructed from the released state--action
pairs. For the coverage-mixture panel, we refit FORE using deterministic
reconstructed background actors without added action noise, retaining the
released $Q$-candidates and validation draws. Thus these weighted experiments
use an approximation to the original behavior policy.

\FloatBarrier

\subsection{Direct ratio errors}

We use a separate 18-case Hopper experiment with gravity and noise
perturbations, three perturbation levels, and three target policies.
Each case has disjoint training, validation, and evaluation samples of
10,000 transitions each. Three optimizer seeds $(17,19,23)$ use the same
data split. The logging law is the known equal mixture
\[
\nu=\tfrac13\sum_{\ell=1}^3\frac{d_{\pi_\ell,\gamma}}{m_\ell},
\qquad m_\ell=\int\dd d_{\pi_\ell,\gamma},
\]
where $d_{\pi_\ell,\gamma}$ is the preterminal discounted occupancy measure.
For a component target $\pi$, the true ratio is
$w^\pi=3m_\pi\Pr(\text{component}=\pi\mid X)$.
The component policies are deterministic: on a unique target action graph,
$w^\pi=3m_\pi$, and off that graph it is zero.
All 180,000 evaluation observations have identified graph membership at
action tolerance $10^{-5}$. At $10^{-4}$, two observations are ambiguous;
we retain error bounds for them. An independent sample of 1,024 complete
target trajectories per case is used to estimate $m_\pi$, with maximum
Monte Carlo standard error $0.00457$.

We compare three candidate libraries: 56 FORE candidates, 56 DualDICE
candidates \citep{nachumEtAl2019DualDICE}, and their 112-candidate union.
Each fitting family varies four architectures, two learning rates, and seven
checkpoints (Section~\ref{app:occupancy-fitting}). All predictions in this
experiment are unprojected.
The ESS screen removes 167 of 3,024 fitted FORE candidates and 4 of 3,024
DualDICE candidates, leaving 44--56 and 55--56 candidates per library,
respectively. Both selectors are defined in all 54 fitting runs for each
library. Our adjoint LSTD extension applies normalized-difference
LSTD-Tournament to the adjoint Bellman equation
(Appendix~\ref{app:adjoint-lstd-tournament}).

We report the unhalved Hellinger distance
$H_\nu(w,w^\pi)=\{\int(\sqrt w-\sqrt{w^\pi})^2\dd\nu\}^{1/2}$,
ratio RMSE $\{\int(w-w^\pi)^2\dd\nu\}^{1/2}$, generalized KL from the true
ratio to the candidate
\[
\int\{w^\pi\log(w^\pi/w)-w^\pi+w\}\dd\nu,
\]
and absolute importance-weighted reward error
$|\mathbb E_\nu[wr]-(1-\gamma)V^\pi|$.
Known component weights are used when averaging evaluation samples.
We average risks over optimizer seeds within cases, then over cases;
Hellinger distance and ratio RMSE take the square root after averaging.
Intervals use 10,000 paired cluster-bootstrap draws over the six
groups defined by perturbation type and level, holding the data splits
and estimated occupancy masses fixed.

\input{figures/pbv_default_ratio_results.tex}

\begin{figure}[!htbp]
\centering
\includegraphics[width=\linewidth]{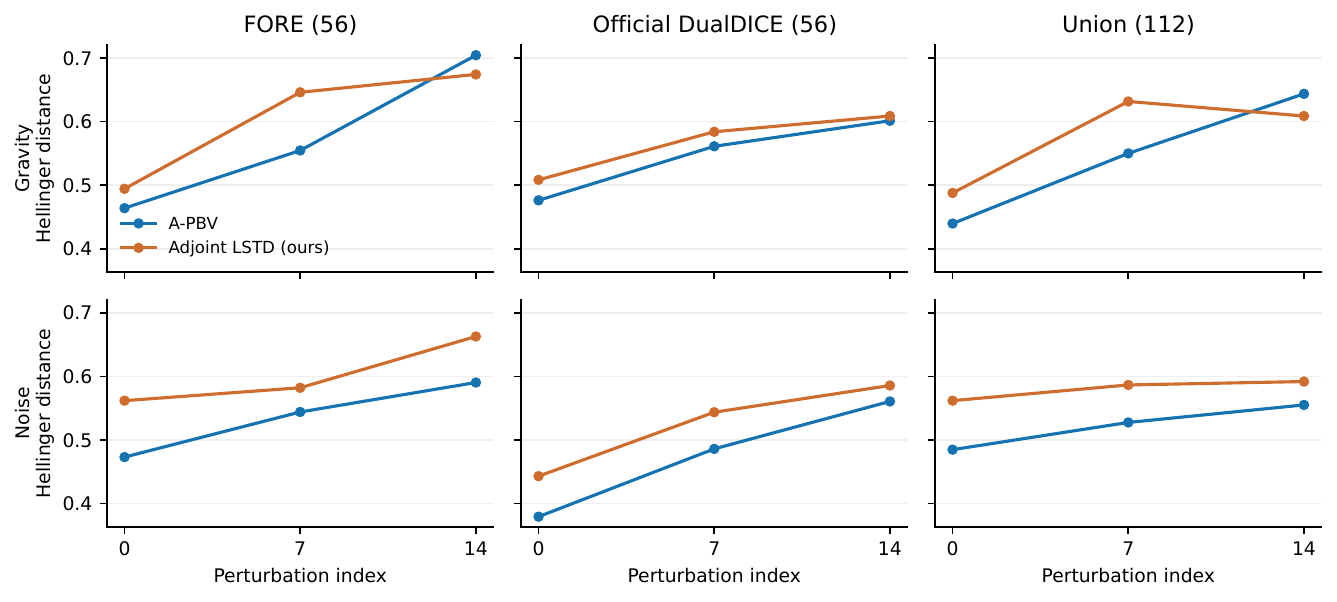}
\caption{Direct ratio selection under gravity and noise perturbations.
Each point averages squared Hellinger error across three target policies
and three optimizer seeds before taking the square root. Both selectors
use the same candidate screen and cap.}
\label{fig:pbv-ratio-conditions}
\end{figure}

A-PBV has lower aggregate error on all four metrics in each library
(Table~\ref{tab:pbv-ratio-selection}). Its Hellinger differences relative
to adjoint LSTD are $-0.046$ (95\% interval $[-0.079,-0.008]$) for FORE,
$-0.032$ ($[-0.048,-0.019]$) for DualDICE, and $-0.043$
($[-0.071,-0.006]$) for their union.
DualDICE gives the lowest Hellinger distance, ratio RMSE, and reward error
under both selectors; FORE gives the lowest generalized KL.
Adding FORE candidates to DualDICE does not lower its aggregate errors.
Figure~\ref{fig:pbv-ratio-conditions} shows Hellinger error by perturbation.

The known-ratio construction uses a mixture of normalized discounted
occupancies. The released Hopper caches instead pool complete-trajectory
transitions, so we use the controlled experiment for direct ratio errors
and the benchmark data for downstream $Q$-selection errors.

\subsection{Downstream ratio selection and tempering}

We compare the two ratio selectors across five benchmark groups
at $n_Q=3{,}200$. The $Q$-selection rule is IW-PBV throughout; the fitted
$Q$-candidates, ratio data, ESS screen, cap, and validation draws are shared.
Only the ratio selector and temperature vary.
All 596 ratio selections are defined for both selectors.
These experiments use the benchmark FORE libraries described above.
DeepOPE uses the 60-candidate $Q$-libraries from the main experiments.
\input{figures/pbv_ratio_downstream.tex}

The two ratio selectors give similar aggregate errors on poor-coverage
Hopper and full DeepOPE at both temperatures
(Table~\ref{tab:pbv-ratio-downstream}). A-PBV has lower mean errors on
controlled DeepOPE, and Adjoint LSTD-Tournament on the main model-free and
model-based Hopper groups. The paired intervals for the DeepOPE and
poor-coverage comparisons include zero.
The direct ratio-error advantage of A-PBV therefore does not consistently
improve downstream errors.

\begin{figure}[!htbp]
\centering
\includegraphics[width=\linewidth]{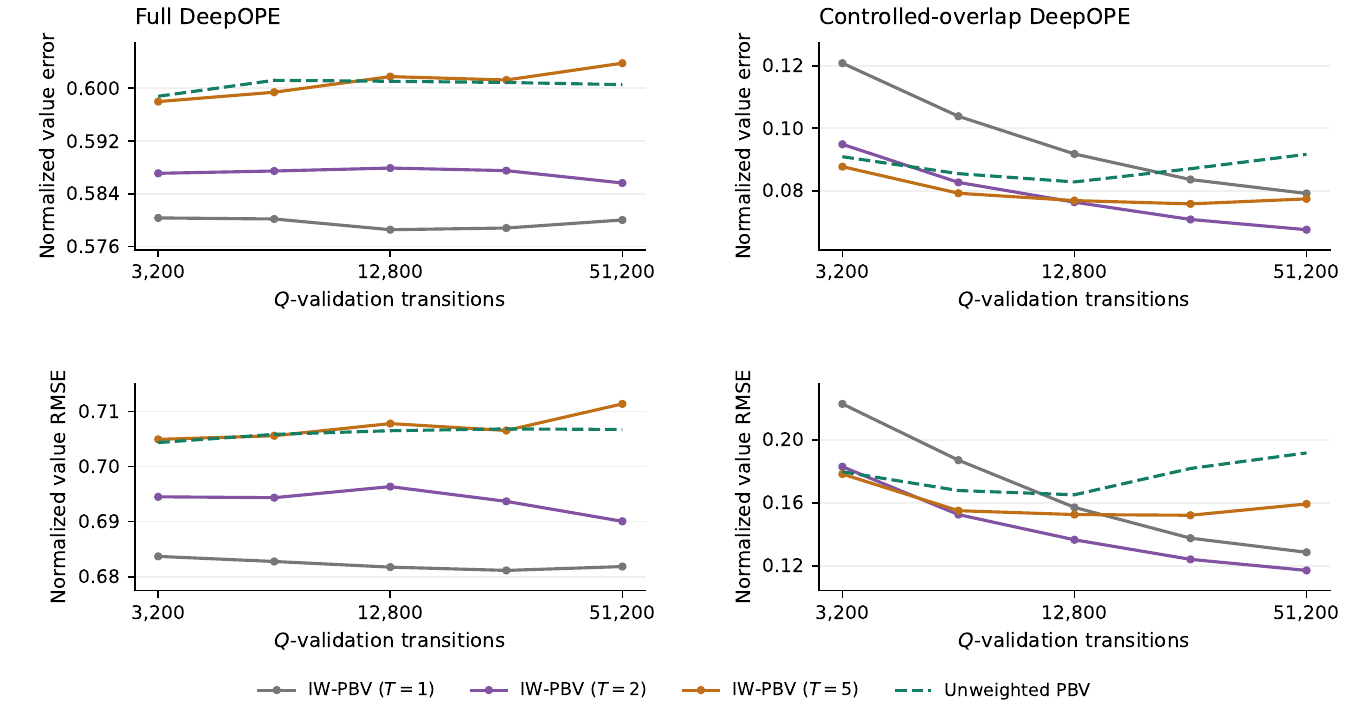}
\caption{Tempering and validation sample size with 60 FQE candidates per case.
Columns show full and controlled-overlap DeepOPE; rows show mean absolute
policy-value error and initial-state value RMSE, normalized by each
environment's reference policy-value range. Each temperature is fixed across
cases and sample sizes. A-PBV selects the ratio, with the same fitted candidates,
ratio data, and validation draws for every curve.}
\label{fig:pbv-ratio-temperature}
\end{figure}

Figure~\ref{fig:pbv-ratio-temperature} compares fixed temperatures $T=1,2,5$
across the DeepOPE validation sample sizes, holding fitted candidates and
A-PBV-selected ratios fixed. Untempered weighting ($T=1$) gives slightly
lower aggregate errors than $T=2$ and $T=5$ on full DeepOPE at every
displayed sample size. Among the three plotted temperatures, $T=5$ gives lower
errors under controlled overlap at $n_Q=3{,}200$, whereas $T=2$ gives the
lowest mean value error and RMSE at
$n_Q=25{,}600$ and $51{,}200$. At the latter sample size, $T=2$ lowers these
errors by 13\% and 27\% relative to $T=5$.

At $n_Q=3{,}200$, however, $T=5$ has the lowest aggregate errors on all
three Hopper groups. Averaging percentage changes equally over the five
benchmark groups, untempered A-PBV-selected weights increase mean value
error by $49.7\%$ (95\% cluster-bootstrap interval $[35.3\%,66.3\%]$) and
initial-state value RMSE by $35.1\%$ ($[21.2\%,47.5\%]$) relative to $T=5$.
The corresponding increases with Adjoint LSTD-Tournament are $28.4\%$ and
$20.4\%$. The bootstrap resamples environment/truth groups for Hopper and
environments for DeepOPE, with shared draws across related benchmark groups.

\paragraph{Independent validation sample.}
We repeated $Q$-function selection for the 44 controlled-overlap DeepOPE
cases on independently collected mixture trajectories, using a nominal
$51{,}200$ validation transitions per environment. The fitted candidates,
selected ratios, temperatures $T=2,5$, and evaluation reference values were
held fixed. All selections on the new sample were saved before evaluation.
Table~\ref{tab:pbv-fresh-validation} reports the resulting errors.
Relative to unweighted PBV, the percentage changes with $T=2$ are
$-5.9\%$ for value error (95\% interval $[-25.5\%,45.3\%]$) and
$-8.0\%$ for initial-state value RMSE ($[-26.4\%,86.4\%]$).
The paired bootstrap resamples the four environments; negative changes
favor IW-PBV. The aggregate gains are smaller than on the original
validation pool.

\input{figures/pbv_fresh_validation_results.tex}

\FloatBarrier

%% file: figures/pbv_default_ratio_results.tex
\begin{table}[!htbp]
\centering
\small
\setlength{\tabcolsep}{3pt}
\renewcommand{\arraystretch}{1.12}
\caption{Direct ratio errors on 18 controlled Hopper cases, with three optimizer seeds per case. Library sizes are before the ESS screen. Hellinger distance and ratio RMSE take square roots after averaging squared errors. GKL is generalized KL from the true ratio to the estimate; reward error is $|\widehat{\mathbb E}_{\nu}[wr]-(1-\gamma)V^\pi|$. Both selectors use the same ESS screen and cap.}
\label{tab:pbv-ratio-selection}
\begin{tabular*}{\linewidth}{@{\extracolsep{\fill}}llrrrr@{}}
\toprule
Library & Selector & Hellinger & Ratio RMSE & GKL & Reward error \\
\midrule
FORE (56) & A-PBV & 0.561 & 1.473 & 0.685 & 0.313 \\
FORE (56) & Adjoint LSTD & 0.607 & 1.498 & 0.816 & 0.359 \\
DualDICE (56) & A-PBV & 0.516 & 0.684 & 0.808 & 0.267 \\
DualDICE (56) & Adjoint LSTD & 0.548 & 0.729 & 1.019 & 0.278 \\
Union (112) & A-PBV & 0.537 & 1.083 & 0.835 & 0.411 \\
Union (112) & Adjoint LSTD & 0.580 & 1.232 & 1.149 & 0.433 \\
\bottomrule
\end{tabular*}
\end{table}

%% file: figures/pbv_ratio_downstream.tex
\begin{table}[!htbp]
\centering
\small
\setlength{\tabcolsep}{2.5pt}
\renewcommand{\arraystretch}{1.12}
\caption{Downstream IW-PBV errors at $n_Q=3{,}200$, with tempered ($T=5$) and untempered ($T=1$) occupancy weights. A / L gives the errors when A-PBV or Adjoint LSTD-Tournament selects the ratio. Differences are A minus L, with 95\% cluster-bootstrap intervals below. DeepOPE uses 60 FQE candidates and policy-range-normalized errors; Hopper uses its released candidate classes and original return units. Cases receive equal weight after averaging data replicates; RMSE takes the square root after pooling squared errors.}
\label{tab:pbv-ratio-downstream}
\begin{tabular*}{\linewidth}{@{\extracolsep{\fill}}lcrrrr@{}}
\toprule
& & \multicolumn{2}{c}{Value error} & \multicolumn{2}{c}{Initial-state value RMSE} \\
\cmidrule(lr){3-4}\cmidrule(lr){5-6}
Benchmark & $T$ & A / L & Difference & A / L & Difference \\
\midrule
Hopper MF & 5 & 2.31 / 1.90 & \shortstack{+0.41\\[-2pt]{\scriptsize [+0.02, +1.10]}} & 5.04 / 4.42 & \shortstack{+0.61\\[-2pt]{\scriptsize [-0.06, +1.32]}} \\
Hopper MF & 1 & 5.50 / 3.32 & \shortstack{+2.18\\[-2pt]{\scriptsize [+0.31, +5.42]}} & 10.20 / 6.94 & \shortstack{+3.26\\[-2pt]{\scriptsize [+0.36, +5.85]}} \\
Poor-coverage MF & 5 & 5.25 / 5.26 & \shortstack{-0.01\\[-2pt]{\scriptsize [-0.07, +0.08]}} & 10.24 / 10.23 & \shortstack{+0.01\\[-2pt]{\scriptsize [-0.12, +0.09]}} \\
Poor-coverage MF & 1 & 5.43 / 5.40 & \shortstack{+0.04\\[-2pt]{\scriptsize [-0.81, +0.92]}} & 10.52 / 10.13 & \shortstack{+0.39\\[-2pt]{\scriptsize [-1.10, +1.65]}} \\
Hopper MB & 5 & 1.61 / 1.50 & \shortstack{+0.11\\[-2pt]{\scriptsize [-0.05, +0.29]}} & 3.76 / 3.50 & \shortstack{+0.26\\[-2pt]{\scriptsize [-0.10, +0.71]}} \\
Hopper MB & 1 & 2.76 / 1.92 & \shortstack{+0.84\\[-2pt]{\scriptsize [+0.42, +1.26]}} & 5.57 / 4.27 & \shortstack{+1.31\\[-2pt]{\scriptsize [+0.24, +2.51]}} \\
Full DeepOPE & 5 & 0.598 / 0.597 & \shortstack{+0.001\\[-2pt]{\scriptsize [-0.004, +0.007]}} & 0.705 / 0.705 & \shortstack{+0.000\\[-2pt]{\scriptsize [-0.004, +0.005]}} \\
Full DeepOPE & 1 & 0.580 / 0.583 & \shortstack{-0.002\\[-2pt]{\scriptsize [-0.011, +0.006]}} & 0.684 / 0.686 & \shortstack{-0.002\\[-2pt]{\scriptsize [-0.011, +0.005]}} \\
Controlled DeepOPE & 5 & 0.088 / 0.090 & \shortstack{-0.002\\[-2pt]{\scriptsize [-0.010, +0.002]}} & 0.178 / 0.188 & \shortstack{-0.009\\[-2pt]{\scriptsize [-0.033, +0.005]}} \\
Controlled DeepOPE & 1 & 0.121 / 0.125 & \shortstack{-0.004\\[-2pt]{\scriptsize [-0.013, +0.002]}} & 0.223 / 0.238 & \shortstack{-0.015\\[-2pt]{\scriptsize [-0.035, +0.003]}} \\
\bottomrule
\end{tabular*}
\end{table}

%% file: figures/pbv_fresh_validation_results.tex
\begin{table}[!htbp]
\centering
\small
\setlength{\tabcolsep}{5pt}
\caption{Controlled-overlap DeepOPE selection on an independent $51{,}200$-transition validation pool per environment, using the same 60 fitted FQE candidates and selected ratios. All 44 target cases are retained. Errors use the original per-environment policy-value ranges and reference values; RMSE takes the square root after pooling squared errors.}
\label{tab:pbv-fresh-validation}
\begin{tabular}{lrr}
\toprule
Selector & Value error & Initial-state value RMSE \\
\midrule
Unweighted PBV & 0.0817 & 0.1571 \\
IW-PBV ($T=5$) & 0.0788 & 0.1524 \\
IW-PBV ($T=2$) & 0.0769 & 0.1445 \\
LSTD-Tournament (normalized) & 0.1025 & 0.1790 \\
LSTD-Tournament (unnormalized) & 0.0977 & 0.1728 \\
\bottomrule
\end{tabular}
\end{table}

%% file: sections/gaussian_kernel_experiment.tex
\section{On-policy examples for LSTD-Tournament}
\label{app:gaussian-kernel-experiment}

The three examples in Figure~\ref{fig:gaussian-normalization-mechanisms}
study amplification of moment noise, error in estimating critic standard
deviations, and near-collinearity between candidates.

\subsection{Noise amplification}

\paragraph{Data-generating process.}
We use a continuing Markov decision process with one action and state space
$[-1,1]^5$. The initial state and each subsequent state are independent
uniform draws from this cube. Rewards are $R=0.1+\xi$, where
$\xi\sim N(0,0.2^2)$ is independent of the states and other rewards.
With discount $\gamma=0.9$, the true value function is $Q^\pi(s)=1$ and
the policy value is $V^\pi=1$. The validation distribution is the stationary
distribution of the target policy, so the experiment is on-policy.

\paragraph{Candidates.}
The fixed candidate class is $\mathcal Q=\{Q_h:h\in\mathcal H\}$, where
\[
Q_h(s)=1+\exp\!\left(-\frac{\|s\|^2}{2h^2}\right).
\]
The 200 bandwidths comprise 100 logarithmically spaced values in
$[10^{-6},0.1)$ and 100 in $[0.1,10]$. More precisely,
\[
\mathcal H=
\{10^{-6+5j/100}:j=0,\ldots,99\}
\cup\{10^{-1+2j/99}:j=0,\ldots,99\}.
\]
All candidates take values between one and two.

\paragraph{Selectors.}
For validation observations $(S_t,R_t,S_t')_{t=1}^n$, write
$\mathbb P_n$ for the empirical average and define
\[
\widehat m_h(f)
=\mathbb P_n\!\left[f(S)\{Q_h(S)-R-\gamma Q_h(S')\}\right],
\qquad
\widehat{\mathcal E}(f)
=\mathbb P_n\!\left[f(S)\{f(S)-\gamma f(S')\}\right].
\]
Unweighted PBV minimizes
\[
\widehat T_h^{\mathrm{PBV}}
=\max_{g\in\mathcal Q\cup\{0\}}
\left\{|\widehat m_h(Q_h-g)|
-\tfrac14|\widehat{\mathcal E}(Q_h-g)|\right\}.
\]
Zero is an additional comparator; the selected function belongs to
$\mathcal Q$. Unnormalized LSTD-Tournament minimizes
\[
\widehat T_h^{\mathrm{LT}}
=\max_{g\in\mathcal Q}|\widehat m_h(g)|.
\]
Normalized LSTD-Tournament uses the candidate and difference critics
of the published implementation and minimizes
\[
\widehat T_h^{\mathrm{LT,nd}}
=\max\left\{
\frac{|\widehat m_h(Q_h)|}{\widehat\sigma_n(Q_h)},\;
\max_{g\in\mathcal Q\setminus\{Q_h\}}
\frac{|\widehat m_h(Q_h-g)|}{\widehat\sigma_n(Q_h-g)}
\right\},
\]
where $\widehat\sigma_n(f)^2
=\mathbb P_n[(f-\mathbb P_nf)^2]$.
The $L^2$-normalized tournament uses the same candidate and difference
critics, replacing each $\widehat\sigma_n(f)$ by
$\{\mathbb P_n f^2\}^{1/2}$.
We evaluate normalized scores using rescaled Gaussian bumps and logarithms to avoid numerical underflow.
All four selectors use the same observations. Candidate order is randomized
once per repetition and shared across methods to resolve ties.

\begin{figure}[!htbp]
\centering
\includegraphics[width=.99\linewidth]{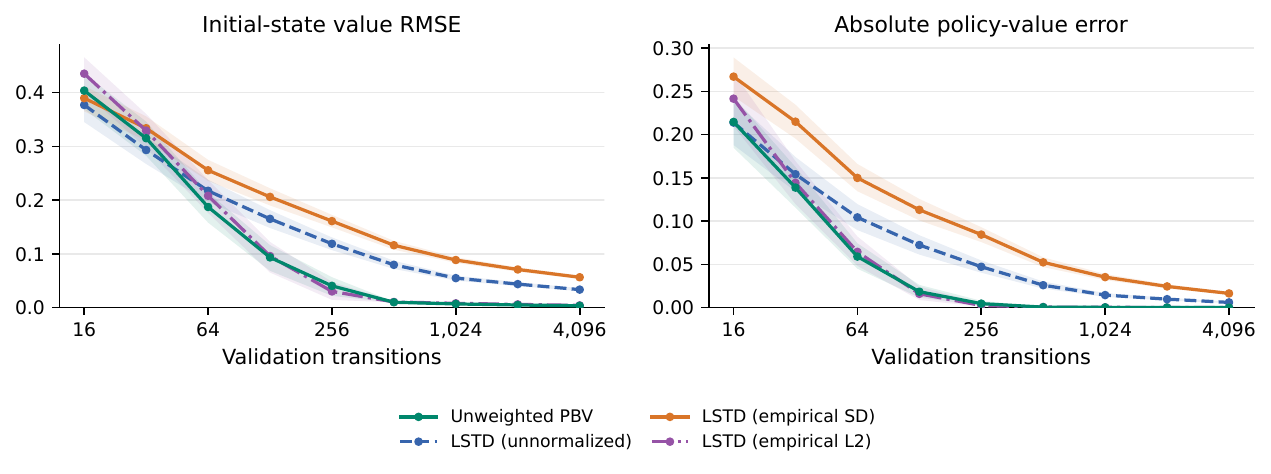}
\caption{On-policy selection among 200 Gaussian-kernel candidates in five
dimensions, with $Q^\pi(s)=V^\pi=1$. Initial-state value RMSE is shown on
the left and absolute policy-value error on the right. Curves summarize
500 repetitions; bands give 95\% bootstrap intervals. The $L^2$ variant
changes only the normalizing factors in the SD-normalized tournament.}
\label{fig:gaussian-kernel-selection}
\end{figure}

\paragraph{Evaluation.}
We generate 500 independent validation datasets of 4,096 transitions and use
nested prefixes of sizes
$n\in\{16,32,64,128,256,512,1024,2048,4096\}$.
Errors are evaluated exactly under the uniform initial-state distribution.
For $g_h=Q_h-1$ and $p\in\{1,2\}$,
\[
m_p(h):=\mathbb E[g_h(S)^p]
=\left\{\sqrt{\frac{\pi}{2}}\frac{h}{\sqrt p}
\operatorname{erf}\!\left(\frac{\sqrt p}{\sqrt2h}\right)\right\}^{5}.
\]
The absolute policy-value error of $Q_h$ is $m_1(h)$, and its initial-state
mean squared value error is $m_2(h)$. Figure~\ref{fig:gaussian-kernel-selection}
reports the square root of the average $m_2$ across selected candidates
and the average $m_1$ in its left and right panels, respectively.
The 95\% intervals use 5,000 bootstrap resamples of the independent
datasets, preserving pairing across selectors and validation sizes.
Each bootstrap RMSE is computed after averaging squared errors.

\FloatBarrier

\subsection{Standard-deviation estimation error}

\paragraph{Data-generating process and candidates.}
We use one action, discount $\gamma=0$, and independent states
$S=(X,U,V)$ with independent coordinates uniform on $[-1,1]$.
For $h=10^{-4}$, define
\[
k_h(x)=\exp\{-x^2/(2h^2)\},\qquad
p=\mathbb E[k_h(X)]
=\sqrt{\frac\pi2}\,h\operatorname{erf}\!\left(\frac1{\sqrt2h}\right),
\qquad \epsilon=p^3.
\]
The deterministic reward and true value function are
\[
R(S)=Q^\pi(S)=\frac p2+\epsilon U+k_h(X).
\]
The 200 fixed candidates share this spiky component and differ through
their approximation errors:
\[
Q_1=Q^\pi+\frac p2,\qquad
Q_{j+1}=Q^\pi+k_{b_j}(V),\qquad j=1,\ldots,199,
\]
where $b_j=\tfrac14\,8^{(j-1)/198}$ gives 199 logarithmically spaced
bandwidths from $0.25$ to $2$. The Gaussian perturbations have no
constant shift. Their TD residuals are $p/2$ and $k_{b_j}(V)$, so there
is no conditional TD noise. The background term $\epsilon U$ makes empirical critic
standard deviations positive almost surely.

\paragraph{Selectors and population standard deviations.}
We apply the same unweighted PBV, unnormalized LSTD, and empirical-SD-normalized
LSTD criteria to this library. The fourth selector is an oracle
control: it retains exactly the empirical moments and critics of the
SD-normalized tournament but uses population standard deviations.
For independent uniform $Z\in[-1,1]$, define
\[
\begin{gathered}
M(b)=\mathbb E[k_b(Z)]
=\sqrt{\frac\pi2}\,b\operatorname{erf}\!\left(\frac1{\sqrt2b}\right),\\
C(b,c)=M\!\left((b^{-2}+c^{-2})^{-1/2}\right)-M(b)M(c).
\end{gathered}
\]
Thus $v_h:=\operatorname{Var}_\nu(Q^\pi)
=M(h/\sqrt2)-p^2+\epsilon^2/3$,
$\sigma_\nu(Q_1)^2=v_h$, and
$\operatorname{Cov}_\nu(Q_{i+1},Q_{j+1})=v_h+C(b_i,b_j)$.
Since $Q_1-Q_{j+1}=p/2-k_{b_j}(V)$, its variance is $C(b_j,b_j)$;
the remaining difference variances follow from the displayed covariance.

\paragraph{Evaluation.}
We use 500 independent datasets, the same nested validation sizes as
above, a common randomized candidate order across methods, and 5,000
paired bootstrap resamples. The exact initial-state mean squared errors
are $p^2/4$ for $Q_1$ and $M(b_j/\sqrt2)$ for $Q_{j+1}$; the absolute
policy-value errors are $p/2$ and $M(b_j)$, respectively.

\begin{figure}[!htbp]
\centering
\includegraphics[width=.99\linewidth]{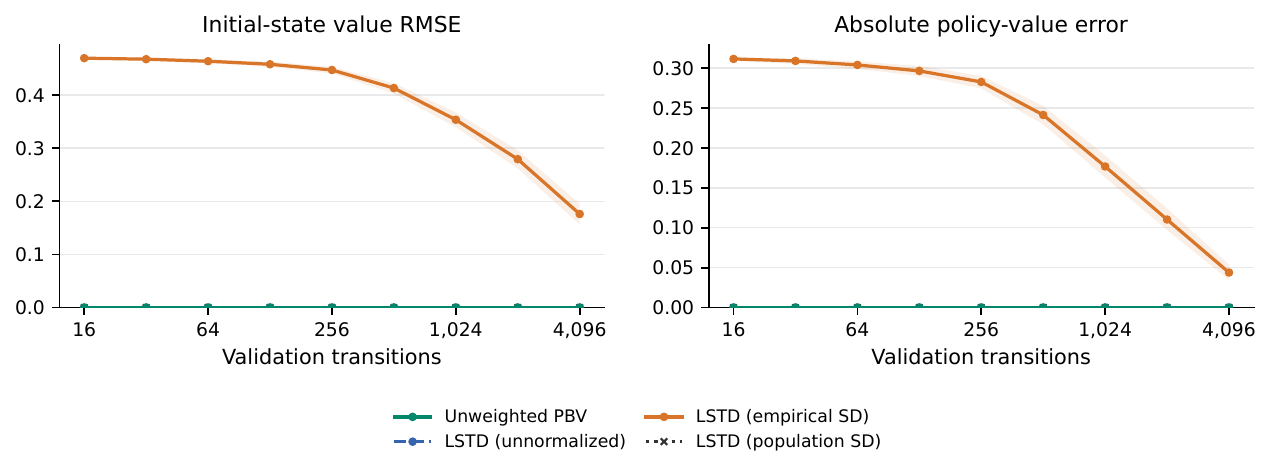}
\caption{On-policy selection among 200 candidates with a Gaussian-spike
truth: initial-state value RMSE (left) and absolute policy-value error
(right). Curves summarize 500 repetitions, with 95\% bootstrap bands.
PBV, unnormalized LSTD, and oracle-SD LSTD select the same candidate,
so their curves overlap.}
\label{fig:gaussian-spiky-truth}
\end{figure}
\FloatBarrier

\subsection{Near-collinearity}

\paragraph{Data-generating process and candidates.}
We use one action, discount $\gamma=0$, and independent states
$S=(F,U)$, where $F$ and $U$ are independent uniform draws from
$[-\sqrt3,\sqrt3]$. Write $G=1+U$ and set $A=10^4$.
Rewards are $R=AF+\xi$, with independent
$\xi\sim N(0,0.2^2)$, so $Q^\pi(S)=AF$ and $V^\pi=0$.
The 200 fixed candidates are
\[
Q_t(S)=AF+tG,\qquad t\in\mathcal T.
\]
We form $\mathcal T$ from 200 equally spaced values in $[-1,1]$,
replacing the positive value closest to zero by zero.
All candidates share the large component $AF$, while
$Q_t-Q_s=(t-s)G$. Their population standard deviations are
$\sigma_\nu(Q_t)=\sqrt{A^2+t^2}$ and
$\sigma_\nu(Q_t-Q_s)=|t-s|$ for $t\ne s$.
The centered candidate span has spikiness measure $6$, and the noise
amplification factors are at most $\sqrt2$.

\paragraph{Selectors and evaluation.}
We apply unweighted PBV and the unnormalized and empirical-SD-normalized
LSTD-Tournament criteria defined above. All three use the same observations
and randomized candidate order.
We use 500 independent datasets, the same nested validation sizes, and
5,000 paired bootstrap resamples as in the other examples.
For $Q_t$, the exact initial-state mean squared value error is $2t^2$,
and the absolute policy-value error is $|t|$.

\begin{figure}[!htbp]
\centering
\includegraphics[width=.99\linewidth]{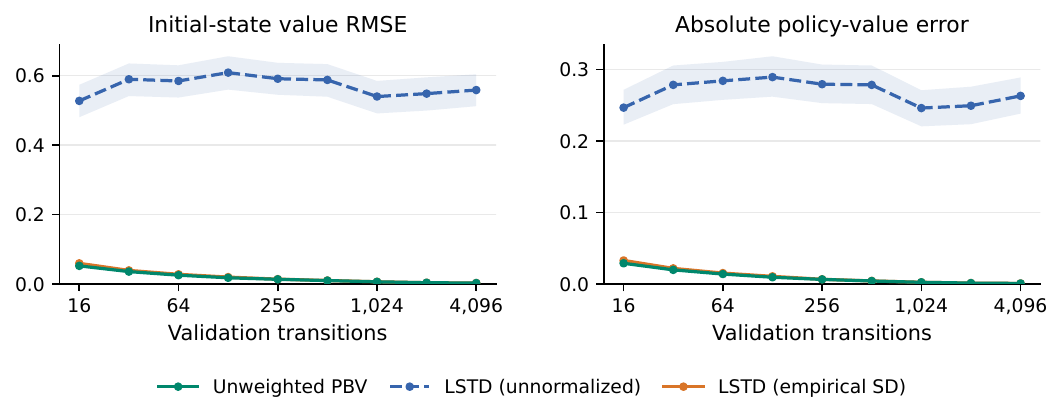}
\caption{On-policy selection among 200 nearly collinear candidates:
initial-state value RMSE (left) and absolute policy-value error (right).
Curves summarize 500 repetitions, with 95\% bootstrap bands.}
\label{fig:collinear-tournament}
\end{figure}
\FloatBarrier

%% file: sections/q_fitted_weight_proofs.tex
\subsection{PBV under approximate fitted-weight coercivity}
\label{app:q-fitted-coercive-proofs}

Condition throughout on the fitting data, so that $\widehat w$,
$\widehat\nu$, and the candidate functions are fixed. Write
$\mathcal B_{\widehat\nu}(f,g)
:=\langle f,(I-\gamma P_\pi)g\rangle_{\widehat\nu}$ and
$T_i^{\mathrm{fit}}:=\max_{j\in\mathcal J_Q}\mathcal M_{ij}^{\mathrm{fit}}$.

\subsubsection{Population score bounds}

\begin{proof}[Proof of Theorem~\ref{thm:q-fitted-coercive-population}]
With $\varepsilon_{\mathrm{coer}}=0$, fix a comparator $k$ satisfying
Condition~\ref{cond:q-fitted-approximate-coercivity},
and set $b_k:=\|\mu_k\|_{2,\widehat\nu}$.
Condition~\ref{cond:q-fitted-coercive-domain} makes the score moments
finite by Cauchy--Schwarz. Since
$\mu_i=\mu_k+(I-\gamma P_\pi)h_{ik}$, the reverse triangle inequality gives
\begin{align}
\mathcal M_{ik}^{\mathrm{fit}}
&\ge\frac34|\mathcal B_{\widehat\nu}(h_{ik},h_{ik})|
-|\langle h_{ik},\mu_k\rangle_{\widehat\nu}|
\notag\\
&\ge\frac{3\kappa_{\widehat\nu}}4\|h_{ik}\|_{2,\widehat\nu}^2
-b_k\|h_{ik}\|_{2,\widehat\nu}.
\label{eq:app-q-fitted-score-lower}
\end{align}
For the comparator's row, $h_{kj}=-h_{jk}$ and the same coercivity
inequality imply, for every $j\in\mathcal J_Q$,
\begin{equation}
\mathcal M_{kj}^{\mathrm{fit}}
\le b_k\|h_{kj}\|_{2,\widehat\nu}
-\frac{\kappa_{\widehat\nu}}4\|h_{kj}\|_{2,\widehat\nu}^2,
\qquad
T_k^{\mathrm{fit}}\le\frac{b_k^2}{\kappa_{\widehat\nu}}.
\label{eq:app-q-fitted-score-upper}
\end{equation}
For $u:=\|h_{\widetilde i_Q^{\mathrm{fit}},k}\|_{2,\widehat\nu}$,
population optimality and
\eqref{eq:app-q-fitted-score-lower}--\eqref{eq:app-q-fitted-score-upper}
give $3\kappa_{\widehat\nu}u^2/4-b_ku\le b_k^2/\kappa_{\widehat\nu}$.
The nonnegative root is $2b_k/\kappa_{\widehat\nu}$.
Thus $u\le2b_k/\kappa_{\widehat\nu}$, and the triangle inequality gives
the claimed population bound.
\end{proof}

\subsubsection{Concentration of the fitted-weight scores}

\begin{lemma}[Localized fitted-weight score deviation]
\label{lem:q-fitted-coercive-validation}
Suppose Condition~\ref{cond:validation-spectral-gap} and
Conditions~\ref{cond:q-fitted-coercive-weights},
\ref{cond:q-fitted-coercive-signal}, and~\ref{cond:q-fitted-coercive-noise}
hold. There is a universal constant $C_0>0$ such that, for every
$\delta\in(0,1)$ and nonempty collection
$\mathcal P\subseteq[K]\times\mathcal J_Q$ fixed conditional on the
fitting data, with conditional probability at least $1-\delta$,
simultaneously for all $(i,j)\in\mathcal P$,
\begin{equation}
\left|\widehat{\mathcal M}_{ij}^{\mathrm{TA}}(\widehat w)
-\mathcal M_{ij}^{\mathrm{fit}}\right|
\le C_0(G_Q+\sigma_Q)
\left\{\sqrt{W_Q}\|h_{ij}\|_{2,\widehat\nu}r
+W_QH_Qr^2\right\},
\qquad
r:=\sqrt{\frac{\log(6|\mathcal P|/\delta)}{ng_b}}.
\label{eq:app-q-fitted-localized-validation}
\end{equation}
\end{lemma}

\begin{proof}
Let $\mathbb P_{n,Q}$ average over the $Q$-validation transitions.
Put $a_{ij}:=\widehat w h_{ij}$ and
$\xi_i:=\varepsilon_i-\mu_i(X)$. The multiplier satisfies
\[
|a_{ij}|\le W_QH_Q,
\qquad
\mathbb E_\nu(a_{ij}^2)
\le W_Q\mathbb E_\nu[\widehat w h_{ij}^2]
=W_Q\|h_{ij}\|_{2,\widehat\nu}^2.
\]
The signal and noise bounds hold on $\{\widehat w>0\}$; outside this set
all multiplier products are zero. The validation setup therefore permits
Lemma~\ref{lem:trajectory-bernstein}(ii) with multiplier $a_{ij}$,
$D=G_Q$, and $\sigma=\sigma_Q$. It gives a deviation bound for
$\mathbb P_{n,Q}[a_{ij}\varepsilon_i]$ from
$\mathbb E_\nu[a_{ij}\varepsilon_i]$.
For the penalty, use
$a_{ij}(\varepsilon_i-\varepsilon_j)$.
Its conditional mean factor is bounded by $2G_Q$, and conditional
Cauchy--Schwarz gives, on $\{\widehat w>0\}$,
\[
\mathbb E\!\left[\left.
e^{|\xi_i-\xi_j|/(2\sigma_Q)}\right|X\right]
\le
\left\{
\mathbb E[e^{|\xi_i|/\sigma_Q}\mid X]
\mathbb E[e^{|\xi_j|/\sigma_Q}\mid X]
\right\}^{1/2}\le2.
\]
Apply the same lemma with $D=2G_Q$ and $\sigma=2\sigma_Q$.
For either moment, the resulting deviation is bounded by a universal
constant times
\[
(G_Q+\sigma_Q)
\left\{\sqrt{W_Q}\|h_{ij}\|_{2,\widehat\nu}
\sqrt{\frac{u}{ng_b}}+W_QH_Q\frac{u}{ng_b}\right\}
\]
outside an event of probability $6e^{-u}$.
Take $u=\log(12|\mathcal P|/\delta)$ and a union bound over both moments
and all pairs in $\mathcal P$. The inequality
$\big||x|-|y|\big|\le|x-y|$ bounds the score deviation by the moment
deviation plus one quarter of the penalty-moment deviation.
Increasing the universal constant to replace
$\log(12|\mathcal P|/\delta)$ by $\log(6|\mathcal P|/\delta)$ proves
\eqref{eq:app-q-fitted-localized-validation}.
\end{proof}

\subsubsection{Finite-sample oracle inequality}

With $C_0$ as in Lemma~\ref{lem:q-fitted-coercive-validation}, take
\[
C_{Q,\mathrm{fit}}
:=\sqrt{W_Q}\left\{
2C_0(G_Q+\sigma_Q)
+\frac{2\sqrt6}{3}\sqrt{C_0H_Q(G_Q+\sigma_Q)}
\right\}.
\]

\begin{proof}[Proof of Theorem~\ref{thm:q-fitted-approximate-finite-sample}]
Fix a comparator $k$ satisfying
Condition~\ref{cond:q-fitted-approximate-coercivity}.
Apply Lemma~\ref{lem:q-fitted-coercive-validation} to
\[
\mathcal P_k:=\{(i,k):i\in[K]\}
\cup\{(k,j):j\in\mathcal J_Q\}.
\]
This collection is fixed conditional on the fitting data and has at most
$2K$ elements. With conditional probability at least $1-\delta$,
the localized deviation bound therefore holds with
\[
r=\sqrt{\frac{\log(12K/\delta)}{ng_b}},
\qquad
\eta=C_0\sqrt{W_Q}(G_Q+\sigma_Q)r,
\qquad
v=C_0W_QH_Q(G_Q+\sigma_Q)r^2.
\]
On this event, set $u:=\|h_{\widehat i_Q,k}\|_{2,\widehat\nu}$ and
$b_k:=\|\mu_k\|_{2,\widehat\nu}$.
The reverse triangle inequality and
Condition~\ref{cond:q-fitted-approximate-coercivity} give
\[
\mathcal M_{ik}^{\mathrm{fit}}
\ge\frac{3\kappa_{\widehat\nu}}4\|h_{ik}\|_{2,\widehat\nu}^2
-b_k\|h_{ik}\|_{2,\widehat\nu}
-\frac34\varepsilon_{\mathrm{coer}}^2.
\]
Since $h_{kj}=-h_{jk}$, the comparator's scores satisfy
\[
\mathcal M_{kj}^{\mathrm{fit}}
\le b_k\|h_{kj}\|_{2,\widehat\nu}
-\frac{\kappa_{\widehat\nu}}4\|h_{kj}\|_{2,\widehat\nu}^2
+\frac14\varepsilon_{\mathrm{coer}}^2.
\]
Adding the localized deviations and maximizing the comparator's
quadratic upper bound gives
\begin{align*}
\max_j\widehat{\mathcal M}_{\widehat i_Q,j}^{\mathrm{TA}}(\widehat w)
&\ge\frac{3\kappa_{\widehat\nu}}4u^2-(b_k+\eta)u
-\frac34\varepsilon_{\mathrm{coer}}^2-v,\\
\max_j\widehat{\mathcal M}_{kj}^{\mathrm{TA}}(\widehat w)
&\le\frac{(b_k+\eta)^2}{\kappa_{\widehat\nu}}
+\frac14\varepsilon_{\mathrm{coer}}^2+v.
\end{align*}
Empirical optimality therefore implies
\[
\frac{3\kappa_{\widehat\nu}}4u^2-(b_k+\eta)u
-\frac{(b_k+\eta)^2}{\kappa_{\widehat\nu}}
-\varepsilon_{\mathrm{coer}}^2-2v\le0.
\]
Solving the quadratic and using $\sqrt{a+b}\le\sqrt a+\sqrt b$ yields
\[
u\le\frac{2(b_k+\eta)}{\kappa_{\widehat\nu}}
+\frac{2\varepsilon_{\mathrm{coer}}}{\sqrt{3\kappa_{\widehat\nu}}}
+\frac{2\sqrt6}{3}\sqrt{\frac v{\kappa_{\widehat\nu}}}.
\]
The triangle inequality, $\kappa_{\widehat\nu}\le1$, and the definitions
of $\eta,v,C_{Q,\mathrm{fit}}$ give
\[
\|Q_{\widehat i_Q}-Q^\pi\|_{2,\widehat\nu}
\le\|Q_k-Q^\pi\|_{2,\widehat\nu}
+\frac{2b_k}{\kappa_{\widehat\nu}}
+\frac{2\varepsilon_{\mathrm{coer}}}{\sqrt{3\kappa_{\widehat\nu}}}
+\frac{C_{Q,\mathrm{fit}}}{\kappa_{\widehat\nu}}r.
\]
Since $b_k=\|Q_k-\mathcal B_\pi Q_k\|_{2,\widehat\nu}$, taking
$k=k^\star_{\widehat\nu}$ proves
\eqref{eq:q-fitted-approximate-finite-sample}.
For the shared-fold extension, apply the theorem separately to each
training-fixed $w_a$ with failure probability $\delta/M$ and take a union
bound. The resulting event controls all $M$ weighted selectors, so it
also controls the selector using any validation-selected weight.
\end{proof}

For the stated concentrability implications, conditional Jensen's inequality
gives $\|P_\pi f\|_{2,\widehat\nu}
\le\sqrt{C_{\widehat\nu,\pi}}\|f\|_{2,\widehat\nu}$.
The Bellman equation and the triangle inequality give the bound with
$L_{\widehat\nu}=1+\gamma\sqrt{C_{\widehat\nu,\pi}}$.
Also,
$\langle f,(I-\gamma P_\pi)f\rangle_{\widehat\nu}
\ge(1-\gamma\sqrt{C_{\widehat\nu,\pi}})\|f\|_{2,\widehat\nu}^2$,
which proves the sufficient condition for fitted-weight coercivity.

%% file: sections/lstd_empirical_normalization.tex
\subsection{Finite-sample oracle inequality}

Proposition~\ref{prop:lstd-normalized-empirical-failure} shows that
bounded critics can have poorly estimated standard deviations when their
variation is concentrated on rare states. To quantify this concentration,
define
\[
\mathcal F_{\mathrm{nd}}
:=\{Q_i:i\in[K]\}\cup\{h_{ij}:i,j\in[K],\ i\ne j\},
\qquad
\Lambda_{\mathrm{nd}}
:=\max_{f\in\mathcal F_{\mathrm{nd}}}
\frac{\|f-\mathbb E_\nu f\|_\infty^2}{\sigma_\nu(f)^2}.
\]
For a bounded finite candidate class with positive critic standard
deviations, $1\le\Lambda_{\mathrm{nd}}<\infty$, but it can be arbitrarily
large.

Retain $C_{k^\star}$ from \eqref{eq:lstd-normalized-critic-norms} and
$\kappa_{\mathrm{LT}}(k^\star)$ from \eqref{eq:lstd-conditioning}, and set
\[
r_{n,\delta}:=\sqrt{\frac{\log(18K^2/\delta)}{ng_b}},
\qquad
b_{k^\star}:=\|Q_{k^\star}-\mathcal B_\pi Q_{k^\star}\|_{2,\nu}.
\]
Define the empirical selector for every validation sample by setting
$\widehat T_i^{\mathrm{nd}}=+\infty$ whenever a required empirical
standard deviation in row $i$ is zero, and resolve ties by candidate order.

\begin{theorem}[Empirically normalized LSTD-Tournament]
\label{thm:lstd-normalized-finite-sample}
Condition on training and retain the validation setup of
Section~\ref{sec:offline-data-setup} and
Condition~\ref{cond:validation-spectral-gap}. Assume
Conditions~\ref{cond:q-unweighted-domain},
\ref{cond:lstd-candidate-envelope}, and~\ref{cond:lstd-invertibility}, the
conditional TD mean bound $G_Q$ in
Condition~\ref{cond:q-bounded-signal}, and the noise bound in
Condition~\ref{cond:q-subgaussian-td}, both for $i\in[K]$.
Suppose $\sigma_\nu(f)>0$ for every $f\in\mathcal F_{\mathrm{nd}}$.
There is a universal constant $c>0$ such that, for every $n\ge1$ and
$\delta\in(0,1)$, with conditional probability at least $1-\delta$,
every empirical minimizer satisfies
\begin{equation}
\begin{aligned}
\|Q_{\widehat i_{\mathrm{LT,nd}}}-Q^\pi\|_{2,\nu}
\le{}&\|Q_{k^\star}-Q^\pi\|_{2,\nu}
+\frac{1+4C_{k^\star}}{\kappa_{\mathrm{LT}}(k^\star)}b_{k^\star}\\
&+\frac{cC_{k^\star}(G_Q+\sigma_Q)}{\kappa_{\mathrm{LT}}(k^\star)}
r_{n,\delta}
+cB_Q\Lambda_{\mathrm{nd}}r_{n,\delta}^2.
\end{aligned}
\label{eq:lstd-normalized-finite-sample}
\end{equation}
\end{theorem}

The factor $C_{k^\star}$ measures amplification from dividing uncentered
moments by critic standard deviations, as in the population bound.
The $\Lambda_{\mathrm{nd}}$ term accounts for estimating the critic
standard deviations: uniform relative concentration holds once $ng_b$
exceeds a universal constant times
$\Lambda_{\mathrm{nd}}\log(18K^2/\delta)$.
For a fixed candidate class and validation law, the two validation terms
have orders $n^{-1/2}$ and $n^{-1}$, respectively.
Under one-step concentrability,
$b_{k^\star}\le(1+\gamma\sqrt{C_{\nu,\pi}})
\|Q_{k^\star}-Q^\pi\|_{2,\nu}$, so
\eqref{eq:lstd-normalized-finite-sample} gives a multiplicative oracle
inequality relative to the best candidate's $L^2(\nu)$ error.

The centered candidate span gives a geometric upper bound on
$\Lambda_{\mathrm{nd}}$. If $\psi_1,\ldots,\psi_d$ is an
$L^2(\nu)$-orthonormal basis of
$\mathcal V_K:=\operatorname{span}\{Q_i-\mathbb E_\nu Q_i:i\in[K]\}$,
then Cauchy--Schwarz gives
\[
\Lambda_{\mathrm{nd}}
\le\Lambda_K:=\left\|\sum_{j=1}^d\psi_j^2\right\|_\infty<\infty.
\]

%% file: sections/lstd_empirical_normalization_proofs.tex
\subsubsection{Finite-sample oracle inequality}

\begin{proof}[Proof of Theorem~\ref{thm:lstd-normalized-finite-sample}]
Condition on training. Write $k=k^\star$, $r=r_{n,\delta}$,
$D_Q=G_Q+\sigma_Q$, $B=B_Q$,
$\kappa=\kappa_{\mathrm{LT}}(k)$, and
$\Lambda=\Lambda_{\mathrm{nd}}$.
Let $C_0$ be a sufficiently large universal constant.
If $\Lambda r^2>1/C_0$, boundedness gives
\[
\|Q_{\widehat i_{\mathrm{LT,nd}}}-Q_k\|_{2,\nu}
\le2B\le2C_0B\Lambda r^2.
\]
The triangle inequality proves
\eqref{eq:lstd-normalized-finite-sample} in this case when $c\ge2C_0$.

Suppose now $\Lambda r^2\le1/C_0$.
For a critic $f\in\mathcal F_{\mathrm{nd}}$, set
\[
g_f:=f-\mathbb E_\nu f,
\qquad \sigma_f:=\sigma_\nu(f),
\qquad c_f:=\frac{\|f\|_{2,\nu}}{\sigma_f},
\qquad R_f:=\frac{\|g_f\|_\infty}{\sigma_f}.
\]
Then $c_f\ge1$, $1\le R_f\le\sqrt\Lambda$, and
$\|f\|_\infty\le\sigma_f(c_f+R_f)$.
Let $\varepsilon_i=Q_i(X)-R-\gamma(\pi Q_i)(S')$ and
$\mu_i=Q_i-\mathcal B_\pi Q_i$. For a critic in row $i$, write
$\widehat m_{if}=\mathbb P_{n,Q}[f(X)\varepsilon_i]$ and
$m_{if}=\langle f,\mu_i\rangle_\nu$.

Apply Lemma~\ref{lem:trajectory-bernstein}(i) to $g_f$ and $g_f^2$,
and part~(ii) to $f(X)\varepsilon_i$, with
$u=\log(18K^2/\delta)$. The bound for $g_f^2$ uses
$\mathbb E_\nu g_f^4\le\sigma_f^4R_f^2$.
The MDP sampling law and the TD mean and noise conditions verify the
conditional-law and moment assumptions in part~(ii).
There are $K^2$ row--critic pairs: $(i,Q_i)$ and $(i,h_{ij})$, $j\ne i$.
A union bound gives an event of conditional probability at least
$1-\delta$ on which, simultaneously for these pairs,
\begin{equation}
\begin{aligned}
|\mathbb P_{n,Q}g_f|
&\le C\sigma_f(r+R_fr^2),\\
|\mathbb P_{n,Q}g_f^2-\sigma_f^2|
&\le C\sigma_f^2(R_fr+R_f^2r^2),\\
|\widehat m_{if}-m_{if}|
&\le CD_Q\sigma_f\{c_fr+(c_f+R_f)r^2\}.
\end{aligned}
\label{eq:lstd-normalized-concentration}
\end{equation}
Work on this event. Since
$\widehat\sigma_n(f)^2
=\mathbb P_{n,Q}g_f^2-(\mathbb P_{n,Q}g_f)^2$,
\eqref{eq:lstd-normalized-concentration} and
$R_fr\le C_0^{-1/2}$ imply
\[
\left|\frac{\widehat\sigma_n(f)}{\sigma_f}-1\right|
\le CR_fr\le\frac12
\]
after increasing $C_0$.
All empirical denominators are therefore positive. Decomposing the
error in the normalized moment gives
\begin{align*}
\left|\frac{\widehat m_{if}}{\widehat\sigma_n(f)}
-\frac{m_{if}}{\sigma_f}\right|
&\le\frac{|\widehat m_{if}-m_{if}|}{\widehat\sigma_n(f)}
+\frac{|m_{if}|}{\sigma_f}
\frac{|\widehat\sigma_n(f)-\sigma_f|}{\widehat\sigma_n(f)}\\
&\le Cc_fD_Qr+C\frac{|m_{if}|}{\sigma_f}R_fr,
\end{align*}
where the last inequality uses $c_f,R_f\ge1$ and
$R_fr\le C_0^{-1/2}$.
For every critic in row $k$,
$c_f\le C_k$ and $|m_{kf}|/\sigma_f\le C_kb_k$.
Consequently,
\begin{equation}
\widehat T_k^{\mathrm{nd}}
\le C_kb_k+CC_k(D_Q+b_k\sqrt\Lambda)r.
\label{eq:lstd-normalized-comparator-score}
\end{equation}

Write $i=\widehat i_{\mathrm{LT,nd}}$.
If $i=k$, the claimed bound holds directly. Otherwise, put $h=h_{ik}$.
For $f\in\{Q_i,h\}$, the candidate bound gives
$\|f\|_{2,\nu},\sigma_f\le2B$.
Thus \eqref{eq:lstd-normalized-concentration} implies
\[
|\widehat m_{if}-m_{if}|
\le CBD_Q\{r+(1+\sqrt\Lambda)r^2\}
\le CBD_Qr.
\]
Since $Q_k=Q_i-h$, the two critics in row $i$ recover the moments
against both $Q_i$ and $Q_k$:
\begin{align*}
|\mathbb P_{n,Q}[Q_i(X)\varepsilon_i]|
&\le\widehat\sigma_n(Q_i)\widehat T_i^{\mathrm{nd}},\\
|\mathbb P_{n,Q}[Q_k(X)\varepsilon_i]|
&\le\{\widehat\sigma_n(Q_i)+\widehat\sigma_n(h)\}
\widehat T_i^{\mathrm{nd}}.
\end{align*}
The deterministic bounds
$\widehat\sigma_n(Q_i)\le B$ and
$\widehat\sigma_n(h)\le2B$, together with empirical optimality, yield
\[
\max_{\ell\in\{i,k\}}|m_{i\ell}|
\le3B\widehat T_k^{\mathrm{nd}}+CBD_Qr.
\]
Also $|m_{k\ell}|\le Bb_k$ for $\ell\in\{i,k\}$.
Lemma~\ref{lem:lstd-identification} therefore gives
\begin{align*}
\|Q_i-Q_k\|_{2,\nu}
&\le\frac{3\widehat T_k^{\mathrm{nd}}+b_k+CD_Qr}{\kappa}\\
&\le\frac{1+3C_k}{\kappa}b_k
+\frac{CC_k}{\kappa}(D_Q+b_k\sqrt\Lambda)r,
\end{align*}
where the second inequality uses
\eqref{eq:lstd-normalized-comparator-score} and $C_k\ge1$.
Increasing $C_0$ absorbs the term
$CC_kb_k\sqrt\Lambda r/\kappa$ into $C_kb_k/\kappa$, so
\[
\|Q_i-Q_k\|_{2,\nu}
\le\frac{1+4C_k}{\kappa}b_k
+\frac{CC_kD_Q}{\kappa}r.
\]
Choose $c\ge\max\{C,2C_0\}$ and apply the triangle inequality.
This proves \eqref{eq:lstd-normalized-finite-sample} in both cases.
\end{proof}